\documentclass[10pt,twocolumn,letterpaper]{article}
\usepackage[T1]{fontenc}

\usepackage[pagenumbers]{iccv} 

\usepackage{xargs}

\usepackage{multicol}
\usepackage{longtable}
\usepackage{rotating}
\usepackage{multirow}
\usepackage{nicefrac}
\usepackage{amsthm}
\usepackage{xargs}[2008/03/08]
\usepackage{pifont}

\usepackage{thmtools}

\newtheorem{definition}{Definition}

\newcommand{\mysizefortwo}{0.48\linewidth}

\usepackage{xcolor}
\usepackage{wasysym}

\newcommand\ptn{\aPerformance(\eventTN)}
\newcommand\pfp{\aPerformance(\eventFP)}
\newcommand\pfn{\aPerformance(\eventFN)}
\newcommand\ptp{\aPerformance(\eventTP)}

\definecolor{iccvblue}{rgb}{0.21,0.49,0.74}
\usepackage[pagebackref,breaklinks,colorlinks,allcolors=iccvblue]{hyperref}

\def\paperID{*****} 
\def\confName{ICCV}
\def\confYear{2025}

\title{A Hitchhiker's Guide to Understanding Performances of Two-Class Classifiers}

\author{Ana\"is Halin\thanks{Equal contributions.}, S\'ebastien Pi\'erard\printfnsymbol{1}, Anthony Cioppa, and Marc Van Droogenbroeck\\
Montefiore Institute, University of Li\`ege, Li\`ege, Belgium\\
{\tt\small \{Anais.Halin,S.Pierard,Anthony.Cioppa,M.VanDroogenbroeck\}@uliege.be}
}

\newcommand{\paperA}{paper~A~\cite{Pierard2024Foundations-arxiv}\xspace}
\newcommand{\paperB}{paper~B~\cite{Pierard2024TheTile-arxiv}\xspace}

\newcommand{\PaperA}{Paper~A~\cite{Pierard2024Foundations-arxiv}\xspace}
\newcommand{\PaperB}{Paper~B~\cite{Pierard2024TheTile-arxiv}\xspace}

\global\long\def\sampleSpace{\Omega}%
\global\long\def\aSample{\omega}%
\global\long\def\eventSpace{\Sigma}%
\global\long\def\measurableSpace{(\sampleSpace,\eventSpace)}%
\global\long\def\aPerformance{P}%
\global\long\def\allPerformances{\mathbb{\aPerformance}_{\measurableSpace}}%
\global\long\def\aScore{X}%
\global\long\def\allScores{\mathbb{\aScore}_{\measurableSpace}}%
\newcommandx\domainOfScore[1][usedefault, addprefix=\global, 1=\aScore]{\mathrm{dom}(#1)}%
\global\long\def\randVarGroundtruthClass{Y}%
\global\long\def\randVarPredictedClass{\hat{Y}}%
\global\long\def\classNeg{c_-}%
\global\long\def\classPos{c_+}%
\global\long\def\sampleTN{tn}%
\global\long\def\sampleFP{fp}%
\global\long\def\sampleFN{fn}%
\global\long\def\sampleTP{tp}%
\global\long\def\eventTN{\{\sampleTN\}}%
\global\long\def\eventFP{\{\sampleFP\}}%
\global\long\def\eventFN{\{\sampleFN\}}%
\global\long\def\eventTP{\{\sampleTP\}}%
\global\long\def\scoreAccuracy{A}%
\global\long\def\scoreTNR{TNR}%
\global\long\def\scoreFPR{FPR}%
\global\long\def\scoreTPR{TPR}%
\global\long\def\scoreNPV{NPV}%
\global\long\def\scorePPV{PPV}%
\newcommandx\scoreFBeta[1][usedefault, addprefix=\global, 1=\beta]{F_{#1}}%
\global\long\def\priorpos{\pi_+}%
\global\long\def\priorneg{\pi_-}%
\global\long\def\entitiesToRank{\mathbb{E}}%
\global\long\def\anEntity{\epsilon}%
\global\long\def\randVarImportance{I}%

\newcommandx\rankingScore[1][usedefault, addprefix=\global, 1=\randVarImportance]{R_{#1}}%

\global\long\def\relWorseOrEquivalent{\lesssim}%
\global\long\def\rank{\mathrm{rank}_\entitiesToRank}%
\global\long\def\ordering{\relWorseOrEquivalent}%
\global\long\def\LScityscapes{\ding{171}}
\global\long\def\LSade{\ding{170}}
\global\long\def\LSvoc{\ding{169}}
\global\long\def\LScoco{\ding{168}}

\global\long\def\realNumbers{\mathbb{R}}%
\newcommand{\indep}{\perp \!\!\! \perp}

\global\long\def\cityscapes{\LScityscapes{}~Cityscapes}
\global\long\def\ade{\LSade{}~ADE20K}
\global\long\def\voc{\LSvoc{}~Pascal VOC 2012}
\global\long\def\coco{\LScoco{}~COCO-Stuff 164k}

\newcommand{\tile}{Tile\xspace}
\newcommand{\tiles}{Tiles\xspace}
\newcommand{\valueTile}{Value Tile\xspace}
\newcommand{\valueTiles}{Value Tiles\xspace}
\newcommand{\baselineTile}{Baseline Value Tile\xspace}

\newcommand{\SOTATile}{State-of-the-Art Value Tile\xspace}

\newcommand{\noSkillTile}{No-Skill Tile\xspace}

\newcommand{\skillTile}{Relative-Skill Tile\xspace}

\newcommand{\correlationTile}{Correlation Tile\xspace}
\newcommand{\correlationTiles}{Correlation Tiles\xspace}
\newcommand{\rankingTile}{Ranking Tile\xspace}
\newcommand{\rankingTiles}{Ranking Tiles\xspace}
\newcommand{\entityTile}{Entity Tile\xspace}
\newcommand{\entityTiles}{Entity Tiles\xspace}

\global\long\def\allNonSkilledPerformances{\mathbb{\aPerformance}^{\randVarGroundtruthClass\indep\randVarPredictedClass}_{\measurableSpace}}%

\global\long\def\allPriorFixedPerformances{\mathbb{\aPerformance}^{\priorpos}_{\measurableSpace}}%

\newcommand{\comma}{\,,}
\newcommand{\point}{\,.}

\newcommandx\unconditionalProbabilisticScore[1]{\aScore_{#1}^{U}}%
\newcommandx\conditionalProbabilisticScore[2]{\aScore_{#1 \vert #2}^{C}}%
\renewcommand{\paperA}{\cite{Pierard2025Foundations}\xspace}
\renewcommand{\paperB}{\cite{Pierard2024TheTile-arxiv}\xspace}

\renewcommand{\PaperA}{\citet{Pierard2025Foundations}\xspace}
\renewcommand{\PaperB}{\citet{Pierard2024TheTile-arxiv}\xspace}

\newcommand{\scenarioA}{\includegraphics[height=\fontcharht\font`\B]{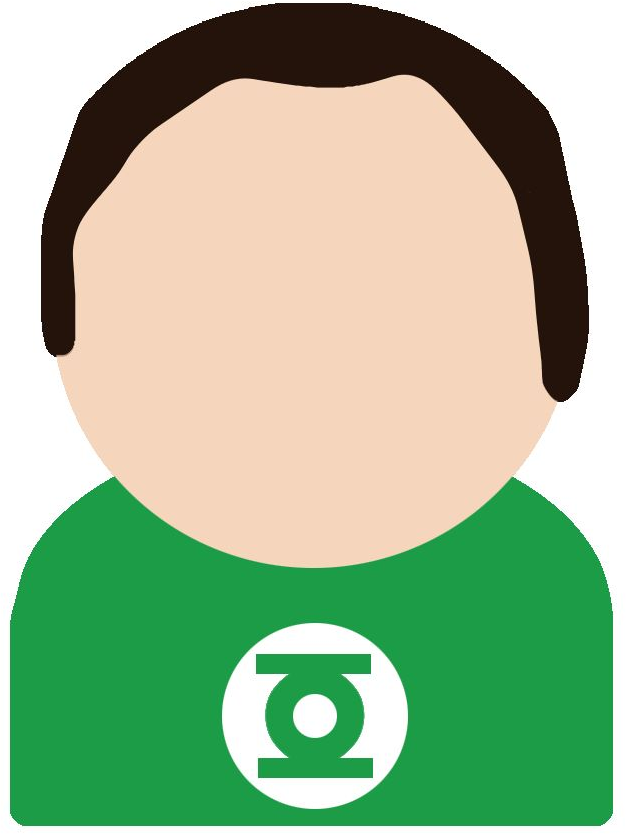}}

\newcommand{\scenarioB}{\includegraphics[height=\fontcharht\font`\B]{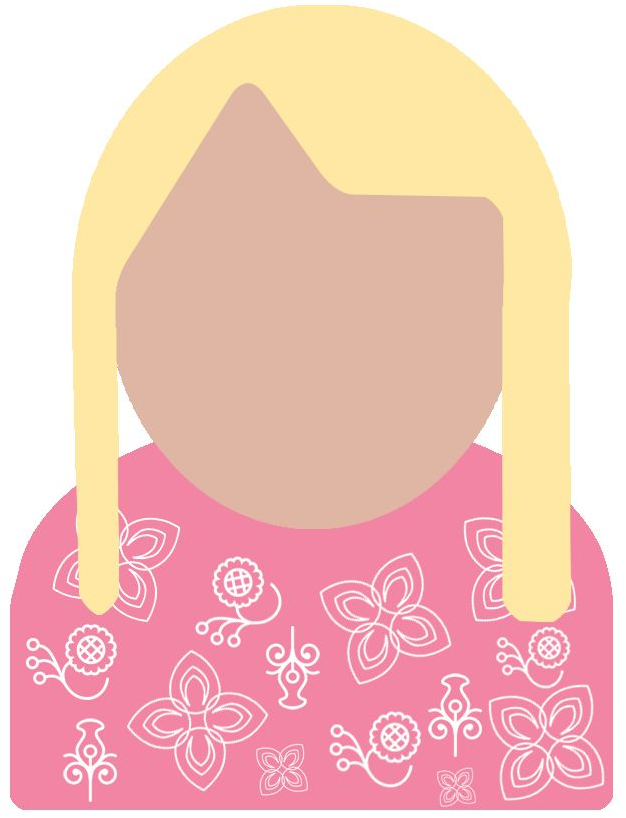}}

\newcommand{\scenarioC}{\includegraphics[height=\fontcharht\font`\B]{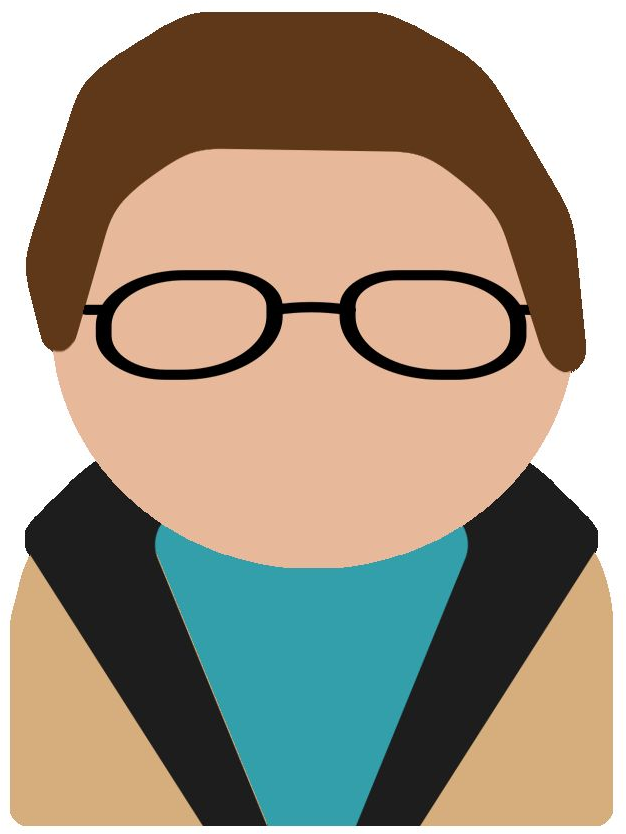}}

\newcommand{\scenarioD}{\includegraphics[height=\fontcharht\font`\B]{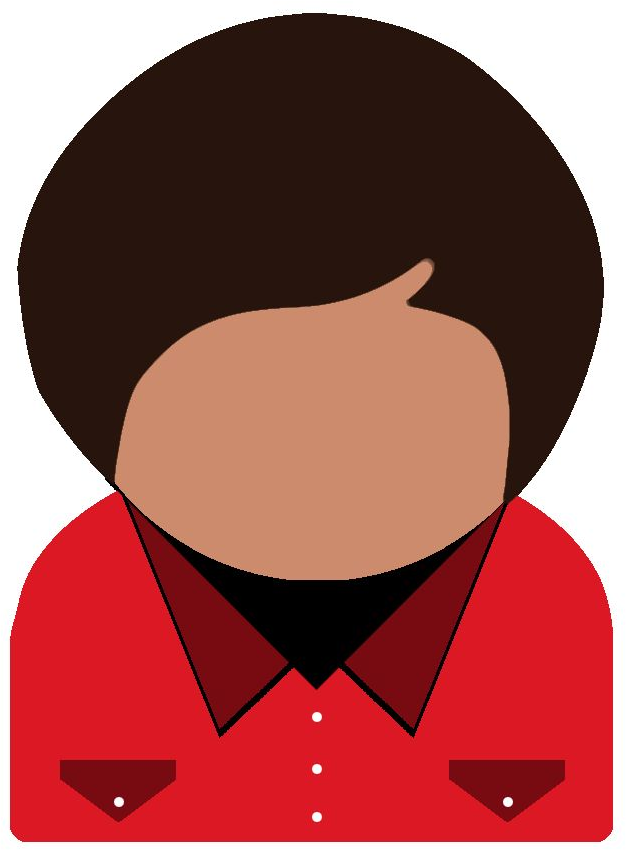}}

\makeatletter
\newcommand{\printfnsymbol}[1]{%
  \textsuperscript{\@fnsymbol{#1}}%
}
\makeatother

\begin{document}

\newcommand{\mysection}[1]{\vspace{2pt}\noindent\textbf{#1}}

\maketitle
\begin{abstract}

Properly understanding the performances of classifiers is essential in various scenarios. However, the literature often relies only on one or two standard scores to compare classifiers, which fails to capture the nuances of application-specific requirements. 
The Tile is a recently introduced visualization tool organizing an infinity of ranking scores into a 2D map.
Thanks to the \tile{}, it is now possible to compare classifiers efficiently, displaying all possible application-specific preferences instead of having to rely on a pair of scores. 
This hitchhiker's guide to understanding the performances of two-class classifiers presents four scenarios showcasing different user profiles: a theoretical analyst, a method designer, a benchmarker, and an application developer. We introduce several interpretative flavors adapted to the user's needs by mapping different values on the \tile{}. We illustrate this guide by ranking and analyzing the performances of $74$ state-of-the-art semantic segmentation models through the perspective of the four scenarios. 
Through these user profiles, we demonstrate that the \tile{} effectively captures the behavior of classifiers in a single visualization, while accommodating an infinite number of ranking scores. Code for mapping the different \tile{} flavors is available in supplementary material.


\end{abstract}

\begin{figure}
    \centering
    \includegraphics[width=0.95\linewidth]{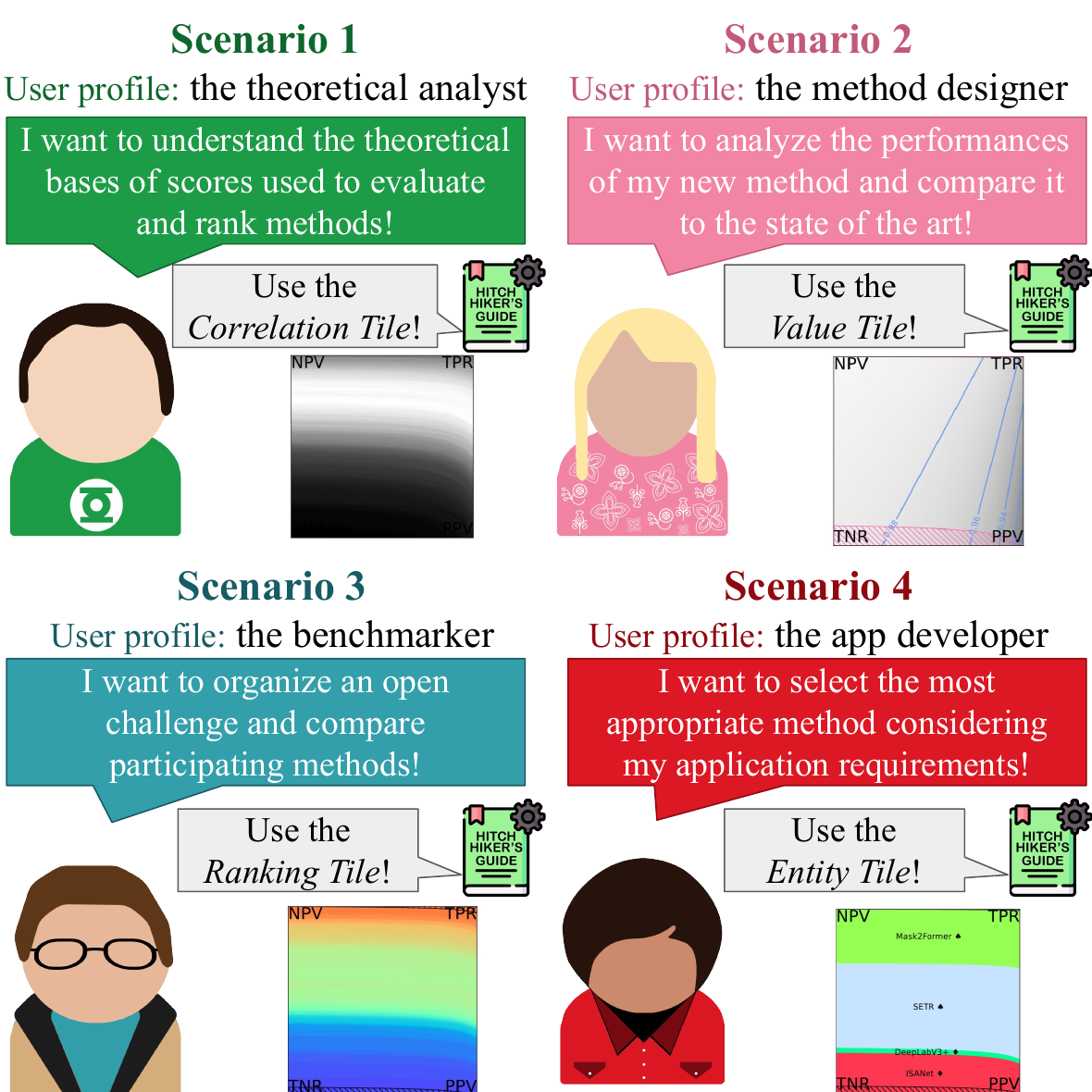}
    \caption{\textbf{Our hitchhiker's guide.} This hitchhiker's guide to understanding performances of two-class classifiers addresses four scenarios, answering specific requests from four user profiles: (1)~\emph{the theoretical analyst}, who is interested in understanding the theoretical relationship between different scores typically used for evaluating or ranking methods, (2)~\emph{the method designer}, who would like to analyze the performances of his/her new method and compare it to others, (3)~\emph{the benchmarker}, who organizes challenges for the scientific community and would like to know how to rank participating methods, and finally (4)~\emph{the application developer}, who wants to select the most appropriate method for his/her application. This guide provides specific tools and explains how to interpret them for each of those four scenarios.}
    \label{fig:graphical_abstract}
\end{figure}

\section{Introduction}
\label{sec:intro}
As humans, performance and ranking are widespread in all aspects of our lives.
For instance, at school, teachers evaluate tests using a score which reflects the performance of students. 
In some disciplines such as calculus, evaluation is straightforward as there are only two possible cases: either the answer is correct or wrong. The score can then be calculated as the ratio of correct answers to the total number of questions. Likewise, in most team sports, team A beats team B if they score more points. Even for reviewing papers, area chairs use scores provided by the reviewers to assess if a paper should be accepted or rejected~\cite{Liu2022integrating}. 

However, not all evaluations are well-defined. For instance, when grocery shopping, consumers may choose product A over product B looking at different characteristics such as the price, the amount of sugar, or the packaging. In this case, the choice is based on several, sometimes contradictory scores. The question is, therefore, which score should the choice be based on?
Similar questions arise in the field of machine learning: 
How can we determine if a newly designed classifier outperforms existing ones? Which score(s) should we use to analyze its performance? 
How do we decide which score to consider for ranking the classifiers? 
Answers to these questions are not trivial, while potentially having a big impact on the development of a whole field of research.

In this paper, we propose a step-by-step hitchhiker's guide to help compare, analyze, and rank two-class classifiers. 
Throughout this guide, we study four scenarios, answering specific requests that four common user profiles may have, and provide visual tools adapted to their needs, as illustrated in \cref{fig:graphical_abstract}.
To do so, we rely on the infinite and parametric family of scores satisfying an axiomatic definition of the performance-based rankings introduced in~\paperA and the tool called the \emph{\tile{}} introduced in~\paperB. 
More precisely, we leverage the newly introduced \tile{} tool to visualize various elements useful for understanding the performances of two-class classifiers from different perspective and considering an infinite number of scores.
As such, we propose several ways to construct, use, and interpret different \tile{} flavors in the four scenarios, offering practical guidance to different user profiles with varied objectives, such as theoretically analyzing scores, developing new methods, benchmarking challenges, or designing applications.

We provide Python code 
that takes a list of two-class classification performances 
and generates a report (see supplementary material) from which most of the figures in this guide are issued. This report is intended for the theoretical analyst \scenarioA, the method designer \scenarioB, the benchmarker \scenarioC, and the application developer \scenarioD. It contains a recap of the compared performances (tables and plots), all flavors described in this guide (\correlationTiles, \valueTiles, \baselineTile, \SOTATile, \rankingTiles, and \entityTile) as well as others (\noSkillTile, \skillTile), and gives several recommendations for the models to select (see scenario 4). Besides generating a report, our code also provides implementations of algorithms introduced in \cite{Pierard2024TheTile-arxiv}.

\mysection{Contributions}. We summarize our contributions as follows.
\textbf{(i)} We provide the first hitchhiker's guide to understanding the performances of two-class classifiers, anchored in rigorous theoretical foundations.
\textbf{(ii)} Throughout the guide, we answer the specific needs of four common user profiles via four scenarios by detailing which tool they should use, how they are constructed, and how they should interpret the results.
\textbf{(iii)} We illustrate our guide for the computer vision community with an analysis and ranking of $74$ state-of-the-art semantic segmentation models.

\section{Related Work}

For the analysis of performance measures, which is the focus of our guide, two strategies have emerged. 

The first strategy consists in providing a series of performance scores that highlight the benefits and trade-offs in designing a classifier. 
\citet{Tharwat2021Classification} analyzed a series of scores for classification tasks that can be used alone or in combination. 
For example, the weighted harmonic mean between the recall $\scoreTPR$ and the precision $\scorePPV$, denoted by $\scoreFBeta$, or simply by $\scoreFBeta[1]$ when weights are equal, is advocated by many authors. A score similar to $\scoreFBeta[1]$ is Jaccard's coefficient. 
Yet, it is known that Jaccard's coefficient and the $\scoreFBeta[1]$ score lead to the same ranking of classifiers, although Jaccard is probabilistic, unlike $\scoreFBeta[1]$~\cite{Flach2003TheGeometry,Powers2011Evaluation}. 
However, the problem with the first strategy is that even though we can use some scores for analyzing the performances of classifiers and ranking them, it is not clear which ones to use.

The second strategy makes use of so-called evaluation spaces, such as the \emph{Receiver Operating Characteristic} (ROC)~\cite{Nahm2022Receiver} or \emph{Precision-Recall} (PR) spaces, that combine two scores.
In~\cite{Fawcett2004Roc,Fawcett2006AnIntroduction}, Fawcett explained how to use ROC graphs and avoid interpretation pitfalls. 
For instance, the area under the ROC curve (AUC-ROC) is commonly used as a performance indicator, as it has statistical significance~\cite{Bamber1975TheArea}.  
Although several authors raised concerns about ROC-derived scores in decision-making~\cite{Briggs2008TheSkill,Carrington2023Deep}, recent works such as the one on contingency spaces~\cite{Ahmadzadeh2023Contingency} still build upon ROC graphs. 
More recently, the PR space has become the de facto replacement for the ROC space in the presence of imbalance~\cite{HughesOliver2018Population}, ignoring the fact that there is a bijection between these two spaces~\cite{Davis2006TheRelationship}. Moreover, the PR space suffers from the presence of an unachievable region~\cite{Boyd2012Unachievable}, which is often overlooked in practice. 
The primary drawback of ROC- or PR-based curves analyzes is that, according to \citet{Menon2013Statistical}, the AUC-ROC and AUC-PR apply not to a classifier but to a scoring function, and that a scoring function yields a family of classifiers obtained for different thresholds. In other words, these curves consider a parametric family of classifiers rather than a specific classifier, and they are therefore inadequate to select classifiers when it comes to comparing unique instances from different families. 
\begin{figure}[t!]
\begin{centering}
\includegraphics[width=0.8\linewidth]{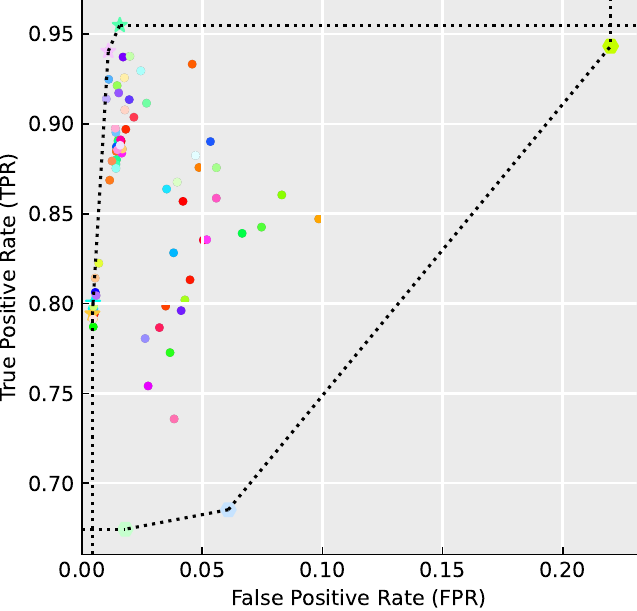}
\includegraphics[width=\linewidth]{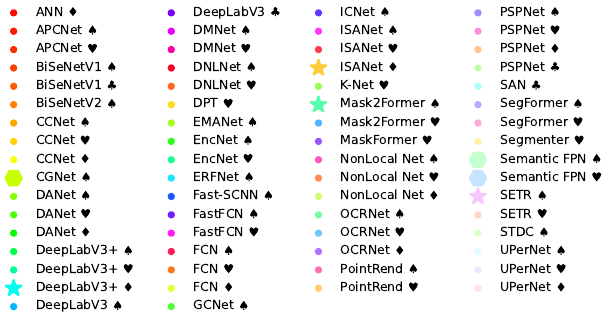}
\par\end{centering}
\caption{
\textbf{How can we rank all these classification performances using the ROC space?} The classical ROC space does not provide an answer at a glance. The $74$ performances shown here (for a positive prior equal to $0{.}124$) serve as the showcase for our illustration in this guide. The dash lines correspond to the supremum and infimum of all achievable performances using a combination of the classifiers.
\label{fig:roc}
}
\end{figure}
As an example of these drawbacks, we show the performance of $74$ two-class classifiers in the ROC space in \cref{fig:roc}. As can be seen, it is extremely challenging to uniquely determine which classifier is the best or the worst. 
Moreover, it is even harder to rank all $74$ classifiers based only on this ROC space.

While these strategies provide a good insight from multiple perspectives, they only offer a scattered interpretation due to the inherent partial redundancy between scores, and ultimately only offer an incomplete look on the performances and ranking. In a recent attempt to formalize the notion of ranking, \citet{Nguyen2023HowTrustworthy} proposed to impose three properties for ranking: (1)~reliability,  
(2)~meaningfulness (evaluated by humans), and (3)~mathematical consistency.
\PaperA presents an alternative formal axiomatic definition of performance-based rankings, grounded in order theory and anchored within a probabilistic framework. The axioms ensure the stability of rankings, \ie , that if multiple entities, in our case two-class classifiers, are ranked, adding or removing an entity does not affect the relative order of the previously present entities.
Additionally, it introduces ranking scores that satisfy these axioms, parameterized by a random variable $\randVarImportance$, called \emph{importance}, which allows for consideration of application-specific preferences. 
\PaperB proposes a spatial organization of the ranking scores in a 2D square map, called the \tile{}, for the particular case of two-class classification. It also studies properties of the \tile{} from a theoretical perspective. 
In this work, we show how both the axiomatic framework and the \tile can be used in practice and propose a hitchhiker's guide to use the \tile{} in various ways through four scenarios and user profiles.

\section{Hitchhiker’s Guide}\label{sec:RankingTile}

This section presents our hitchhiker's guide to understanding performances of two-class classifiers. We begin by providing the recommended basic theoretical knowledge for using the guide, including a description of the terminology and notations, the general \tile{} tool, and the illustrative context used throughout. Next, we detail four scenarios, each aligned with the needs of a specific user profile. For each scenario, we highlight the relevant flavors of the \tile{}, and explain how to interpret the results.

\subsection{Recommended Basic Theoretical Knowledge}

\mysection{Terminology and notations.}
To be consistent with~\paperA and \paperB, we use the same terminology and notations. Hence, an \emph{entity} $\anEntity$ is a two-class classifier and the set of all entities of interest is denoted by $\entitiesToRank$. \emph{Performances}, denoted by $\aPerformance$, are probability measures, and the performance $\aPerformance_\anEntity$ of an entity $\anEntity$ is the evaluation of this entity.
\emph{Scores} are functions associating a real value to performances, that is, 
$\aScore:\domainOfScore\rightarrow\realNumbers:\aPerformance\mapsto\aScore(\aPerformance)$,
with the domain of the score, $\domainOfScore$, included in the set of all probability measures, $\allPerformances$, on the measurable space $\measurableSpace$, where $\sampleSpace$ is the sample space (\ie, the set of outcomes) and $\eventSpace$ is the event space (\ie, a $\sigma$-algebra on $\sampleSpace$).

\mysection{Construction and interpretation of the \tile.}
As explained in~\paperB, the \tile for two-class classification is obtained by organizing ranking scores on a 2D square map, where the two axes, $a$ and $b$, respectively represents the importance $\randVarImportance$ given to true positive~($\sampleTP$) compared to true negative~($\sampleTN$), and false negative~($\sampleFN$) compared to false positive~($\sampleFP$):
\begin{equation} \label{eq:a}
a=\randVarImportance(\sampleTP)=1-\randVarImportance(\sampleTN)\comma
\end{equation}
\begin{equation} \label{eq:b}
b=\randVarImportance(\sampleFN)=1-\randVarImportance(\sampleFP)\point
\end{equation}
The importance value can be arbitrarily chosen to reflect application-specific preferences. 
In the particular case of two-class classification, the ranking scores are given by
\begin{equation} \label{eq:ranking-scores}
\resizebox{.9\hsize}{!}{$
    \rankingScore(\aPerformance)=\frac{
        \randVarImportance(\sampleTN)\aPerformance(\eventTN)+\randVarImportance(\sampleTP)\aPerformance(\eventTP)
    }{
        \randVarImportance(\sampleTN)\aPerformance(\eventTN)+\randVarImportance(\sampleFP)\aPerformance(\eventFP)+\randVarImportance(\sampleFN)\aPerformance(\eventFN)+\randVarImportance(\sampleTP)\aPerformance(\eventTP)
    }
    \comma
    $}
\end{equation}
with $\aPerformance(\eventTN)$ (resp. $\aPerformance(\eventFP)$, $\aPerformance(\eventFN)$, and $\aPerformance(\eventTP)$) corresponding to the probability of the event $\eventTN$ (resp. $\eventFP$, $\eventFN$, and $\eventTP$). The \tile{} is then defined as follows:
\begin{definition}
    The \tile{} for two-class classification is the mapping 
    $$[0,1]^2\rightarrow\allScores : (a,b)\mapsto\rankingScore[\randVarImportance]$$ 
    with $\allScores$ denoting all possible scores on $\measurableSpace$ and $\randVarImportance(\sampleTN)=1-a$, $\randVarImportance(\sampleFP)=1-b$, $\randVarImportance(\sampleFN)=b$, $\randVarImportance(\sampleTP)=a$.
\end{definition}

\begin{figure}
\begin{centering}
\includegraphics[width=\linewidth]{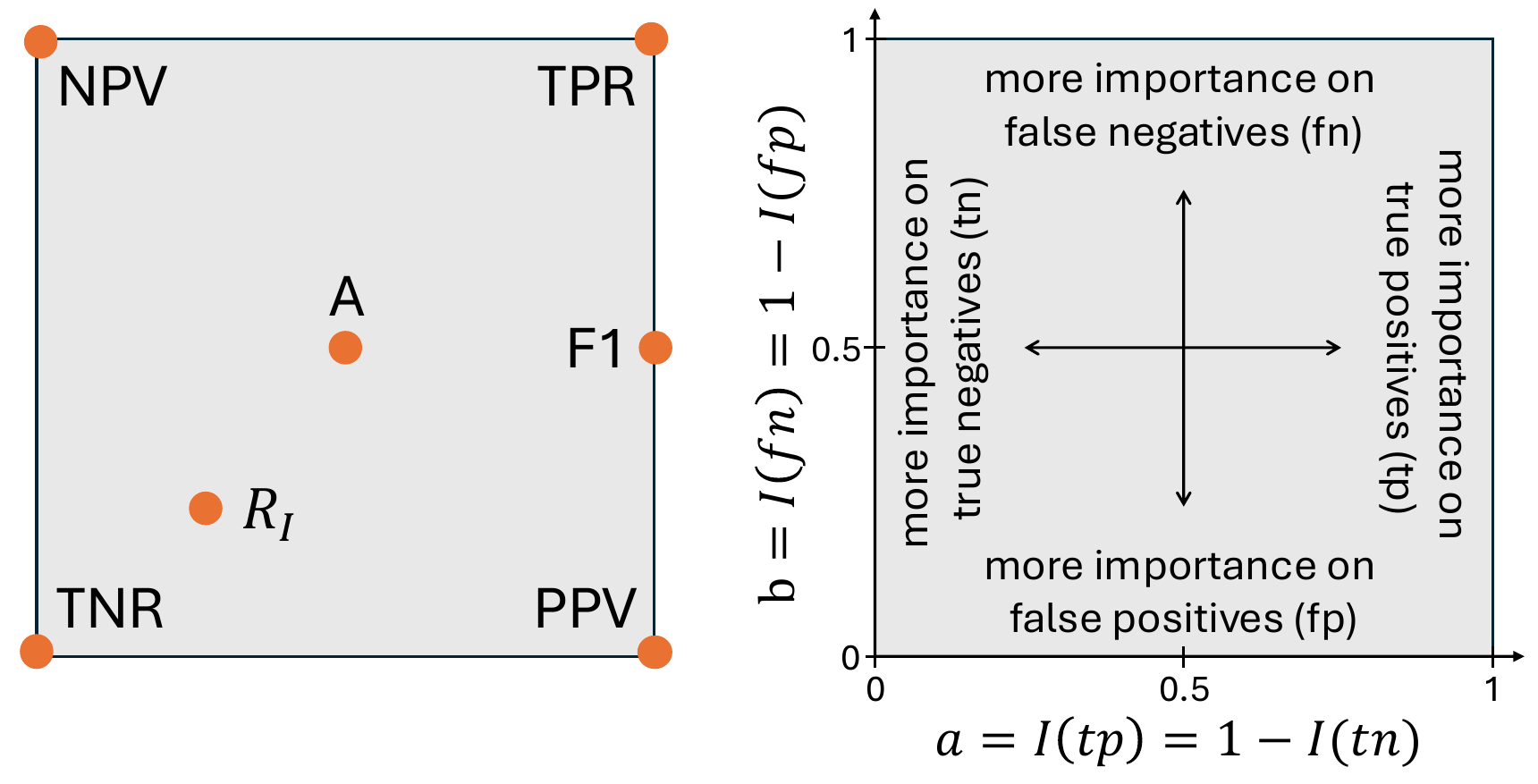}
\par\end{centering}
\caption{\textbf{\tile{} with ranking scores for two-class classification.} 
Each point of the \tile{} corresponds to a ranking score $\rankingScore[\randVarImportance]$ which can be computed using \cref{eq:ranking-scores}. The \tile therefore contains an infinity of ranking scores, including the popular Accuracy ($\scoreAccuracy$), Negative Predictive Value ($\scoreNPV$), True Positive Rate ($\scoreTPR$), Positive Predictive Value ($\scorePPV$), True Negative Rate ($\scoreTNR$), and $\scoreFBeta[1]$ scores. 
The location of a ranking score $\rankingScore[\randVarImportance]$ in the \tile{} is defined by the values of $a$ and $b$, which indicate the importance given to true positives ($\sampleTP$), true negatives ($\sampleTN$), false negatives ($\sampleFN$), and false positives ($\sampleFP$). 
In all following figures, the \tile flavors adopt the same layout as the one on the left hand side. For ease of reading, the names of the scores at the four corners are systematically written.  
\label{fig:tile}}
\end{figure}

The layout of the \tile{} is shown in \cref{fig:tile}.
By construction, the top-right corner of the \tile{} gives maximum importance values to both $\sampleTP$ and $\sampleFN$ (\ie, $\randVarImportance(\sampleTP)=\randVarImportance(\sampleFN)=1$). 
The ranking score in this corner is thus the True Positive Rate ($\scoreTPR$), also known as recall or sensitivity:
\begin{equation} \label{eq:TPR}
 \resizebox{.9\hsize}{!}{$\scoreTPR = \frac{\aPerformance(\eventTP)}{\aPerformance(\eventTP) + \aPerformance(\eventFN)} = P(\randVarGroundtruthClass=\randVarPredictedClass|\randVarGroundtruthClass=\classPos)\comma$}
\end{equation}
where $\randVarGroundtruthClass$ is the ground truth, $\randVarPredictedClass$ the prediction, and $\classPos$ the positive class.
Conversely, the bottom-left corner gives minimum importance values to both $\sampleTP$ and $\sampleFN$ (\ie, $\randVarImportance(\sampleTP)=\randVarImportance(\sampleFN)=0$), leading to the ranking score named the True Negative Rate ($\scoreTNR$), also known as specificity: 
\begin{equation} \label{eq:TNR}
 \resizebox{.9\hsize}{!}{$\scoreTNR = \frac{\aPerformance(\eventTN)}{\aPerformance(\eventTN) + \aPerformance(\eventFP)} = P(\randVarGroundtruthClass=\randVarPredictedClass|\randVarGroundtruthClass=\classNeg)\comma$}
\end{equation}
where $\classNeg$ is the negative class.
Similarly, the two other corners correspond to the Negative Predictive Value ($\scoreNPV$) and the Positive Predictive Value ($\scorePPV$), also known as precision:
\begin{equation} \label{eq:NPV}
 \resizebox{.9\hsize}{!}{$\scoreNPV = \frac{\aPerformance(\eventTN)}{\aPerformance(\eventTN) + \aPerformance(\eventFN)} = P(\randVarGroundtruthClass=\randVarPredictedClass|\randVarPredictedClass=\classNeg)\comma$}
\end{equation}
\begin{equation} \label{eq:PPV}
 \resizebox{.9\hsize}{!}{$\scorePPV = \frac{\aPerformance(\eventTP)}{\aPerformance(\eventTP) + \aPerformance(\eventFP)} = P(\randVarGroundtruthClass=\randVarPredictedClass|\randVarPredictedClass=\classPos)\point$}
\end{equation}
The score in the middle of the \tile{}, \ie, giving the same importance value to $\sampleTP$, $\sampleTN$, $\sampleFN$, and $\sampleFP$ (\ie, $\randVarImportance(\sampleTP)=\randVarImportance(\sampleTN)=\randVarImportance(\sampleFN)=\randVarImportance(\sampleFP)=0.5$) is the popular accuracy $\scoreAccuracy$, also defined as $P(\randVarGroundtruthClass=\randVarPredictedClass)$:
\begin{equation} \label{eq:Accuracy}
\resizebox{.9\hsize}{!}{$
 \scoreAccuracy = \frac{0.5\,\aPerformance(\eventTP) + 0.5\,\aPerformance(\eventTN)}{0.5\,\aPerformance(\eventTP) + 0.5\,\aPerformance(\eventFN) + 0.5\,\aPerformance(\eventFP)+ 0.5\,\aPerformance(\eventTN)} \point
 $}
\end{equation}

Finally, $\scoreFBeta$ scores, with $\beta=\sqrt{\nicefrac{b}{1-b}}$, are on the right-hand side that joins $\scoreTPR$ to $\scorePPV$, $\scoreFBeta[1]$ being located in the center of this side following:
\begin{equation} \label{eq:FBeta1}
 \scoreFBeta[1] = \frac{\aPerformance(\eventTP)}{\aPerformance(\eventTP)  + 0.5\,\aPerformance(\eventFP) + 0.5\,\aPerformance(\eventFN)}\point
\end{equation}

As shown in this section, the \tile{} organizes an infinity of ranking scores into a 2D map through the concept of importance. Each point on the \tile{} corresponds to a score that assigns varying importance values to $\sampleTP$, $\sampleTN$, $\sampleFP$, and $\sampleFN$. 
This makes the \tile{} a useful visual tool for displaying the values these scores take for a given entity. Through our scenarios, we explain how to map different values on this tile for different user profiles.

\begin{table*}[t!]
    \centering
    \caption{List of the models (columns) sorted 
    \wrt the date of publication (from 2015 to 2023) from the MMSegmentation toolbox~\cite{Mmsegmentation2020} trained on one or multiple datasets (rows). We respectively have $31$ models trained on the \cityscapes{}~\cite{Cordts2016The} dataset, $27$ on the \ade{}~\cite{Zhou2017Scene} dataset, $12$ on the \voc{}~\cite{Everingham2010PascalVOC} dataset, and $4$ on the \coco{}~\cite{Caesar2018COCO} dataset, totaling $74$ models.}
    \resizebox{\textwidth}{!}{
    \begin{tabular}{l|c|c|c|c|c|c|c|c|c|c|c|c|c|c|c|c|c|c|c|c|c|c|c|c|c|c|c|c|c|c|c|c|c|c|c|c}
        & \rotatebox{90}{FCN~\cite{Long2015Fully,Shelhamer2017Fully}} & \rotatebox{90}{DeepLabV3~\cite{Chen2017Rethinking-arxiv}} & \rotatebox{90}{PSPNet~\cite{Zhao2017Pyramid}} & \rotatebox{90}{ERFNet~\cite{Romera2018ERFNet}} & \rotatebox{90}{DeepLabV3+~\cite{Chen2018EncoderDecoder}} & \rotatebox{90}{EncNet~\cite{Zhang2018Context}} & \rotatebox{90}{NonLocal Net~\cite{Wang2018Non}} & \rotatebox{90}{BiSeNetV1~\cite{Yu2018Bisenet}} & \rotatebox{90}{ICNet~\cite{Zhao2018ICNet}} & \rotatebox{90}{PSANet~\cite{Zhao2018PSANet}} & \rotatebox{90}{UPerNet~\cite{Xiao2018Unified}} & \rotatebox{90}{FastFCN~\cite{Wu2019FastFCN-arxiv}} & \rotatebox{90}{Fast-SCNN~\cite{Poudel2019FastSCNN-arxiv}} & \rotatebox{90}{ISANet~\cite{Huang2019Interlaced-arxiv}} & \rotatebox{90}{APCNet~\cite{He2019Adaptive}} & \rotatebox{90}{DANet~\cite{Fu2019Dual}} & \rotatebox{90}{Semantic FPN~\cite{Kirillov2019Panoptic}} & \rotatebox{90}{ANN~\cite{Zhu2019Asymmetric}} & \rotatebox{90}{CCNet~\cite{Huang2019CCNet}} & \rotatebox{90}{DMNet~\cite{He2019Dynamic} } & \rotatebox{90}{EMANet~\cite{Li2019Expectation}} & \rotatebox{90}{GCNet~\cite{Cao2019GCNet}} & \rotatebox{90}{PointRend~\cite{Kirillov2020PointRend}} & \rotatebox{90}{DNLNet~\cite{Yin2020Disentangled}} & \rotatebox{90}{OCRNet~\cite{Yuan2020Object}} & \rotatebox{90}{CGNet~\cite{Wu2021CGNet}} & \rotatebox{90}{SETR~\cite{Zheng2021Rethinking}} & \rotatebox{90}{BiSeNetV2~\cite{Yu2021BiSeNet}} & \rotatebox{90}{STDC~\cite{Fan2021Rethinking}} & \rotatebox{90}{DPT~\cite{Ranftl2021Vision}} & \rotatebox{90}{Segmenter~\cite{Strudel2021Segmenter}} & \rotatebox{90}{K-Net~\cite{Zhang2021Knet}} & \rotatebox{90}{MaskFormer~\cite{Cheng2021Per}} & \rotatebox{90}{SegFormer~\cite{Xie2021SegFormer}} & \rotatebox{90}{Mask2Former~\cite{Cheng2022Masked}} & \rotatebox{90}{SAN~\cite{Xu2023Side}}   \\
        \hline
        \hline
        \cityscapes{}~\cite{Cordts2016The}                   & \checkmark & \checkmark & \checkmark & \checkmark & \checkmark & \checkmark & \checkmark & \checkmark & \checkmark & \checkmark & \checkmark & \checkmark & \checkmark & \checkmark & \checkmark & \checkmark & \checkmark & \checkmark & \checkmark & \checkmark & \checkmark & \checkmark & \checkmark & \checkmark & \checkmark & \checkmark & \checkmark & \checkmark & \checkmark & & & & & \checkmark & \checkmark & \\
        \ade{}~\cite{Zhou2017Scene}                       & \checkmark & \checkmark & \checkmark & & \checkmark & \checkmark & \checkmark & & & \checkmark & \checkmark & \checkmark & & \checkmark & \checkmark & \checkmark & \checkmark & \checkmark & \checkmark & \checkmark & & \checkmark & \checkmark & \checkmark & \checkmark & & \checkmark & & & \checkmark & \checkmark & \checkmark & \checkmark & \checkmark & \checkmark & \\
        \voc{}~\cite{Everingham2010PascalVOC}    & \checkmark & \checkmark & \checkmark & & \checkmark & & \checkmark & & & & \checkmark & & & \checkmark & & \checkmark & & \checkmark & \checkmark & & & \checkmark & & & \checkmark & & & & & & & & & & & \\
        \coco{}~\cite{Caesar2018COCO}             & & \checkmark & \checkmark & & & & & \checkmark & & & & & & & & & & & & & & & & & & & & & & & & & & & & \checkmark \\
    \end{tabular}}
    \vspace{-0.4mm}
    \label{tab:models-mmsegmentation}
\end{table*}

\mysection{Description of our illustration.} In the following, we illustrate all our \tile flavors on a same use case related to computer vision for autonomous driving. We demonstrate that the applicability of the \tile and its flavors is far from being limited to two-class classification problems. Specifically, we compare and rank $74$ state-of-the-art semantic segmentation models from the MMSegmentation toolbox~\cite{Mmsegmentation2020}. What makes this use case challenging is that the sets of predicted classes differ from one model to another. This is because the models have been trained on several datasets with heterogeneous labels. As shown in \cref{tab:models-mmsegmentation}, these include $31$ models trained on \cityscapes{}~\cite{Cordts2016The}, $27$ on \ade{}~\cite{Zhou2017Scene}, $12$ on \voc{}~\cite{Everingham2010PascalVOC}, and $4$ on \coco{}~\cite{Caesar2018COCO}. Motivated by the preponderant role played by the foreground-background differentiation in autonomous driving, we decided to group the semantic labels into a negative class, corresponding to background objects (\eg, road, sidewalk, or building), and a positive class, corresponding to foreground, potentially moving, objects (\eg, person, rider, or car). The grouping has been performed for both the learning sets and BDD100K~\cite{Yu2020BDD100K}, that we selected as testing set (see supplementary material for a detailed explanation on how it was performed). By doing so, the performances of the $74$ models become directly comparable.

\subsection{Four Scenarios of our Hitchhiker’s Guide}

This section provides a step-by-step guide for using the \tile{} to rank two-class classifiers (\ie, entities), by illustrating its various potential uses through four scenarios.
Each scenario corresponds to one user profile among (1)~the theoretical analyst \scenarioA, who is interested in understanding the theoretical relationship between different scores typically used for evaluating or ranking methods, (2)~the method designer \scenarioB, who
needs to analyze the performances of his/her new method and compare it to others, (3)~the benchmarker \scenarioC, who organizes challenges for the scientific community and has to rank participating methods, and finally (4)~the application developer \scenarioD, who should be able to select the most appropriate method for his/her application.
In the final scenario, we also discuss various strategies for selecting an entity based on the \tiles{}.

\subsubsection*{Scenario 1: The Theoretical Analyst \scenarioA}

The theoretical analyst seeks to understand the foundational principles behind the scores used to evaluate and rank classifiers. Unlike other users, his/her focus is not on a specific application but rather on the theoretical relationships between various ranking scores. For this user, it is crucial to explore how different scores compare to one another, and how they correlate in terms of both value and ranking. The analyst aims to ensure that each selected score provides unique, non-redundant information, thereby enriching the evaluation process. Therefore, he/she requires a tool that can effectively illustrate the relationships between scores, helping discern whether the chosen scores are complementary or overlapping in the information they convey. To address these needs, the next section introduces the \correlationTile{}, specifically designed for this type of analysis.

\mysection{\correlationTile{}.}
The \correlationTile{} displays the correlation, using a linear (Pearson's $r$) or a rank (Spearman's $\rho$) correlation coefficient, between a score $\aScore$, typically one used as a reference in a research field of interest, and the canonical ranking scores $\rankingScore$ across the \tile{}. The \correlationTile{} is defined as:
\begin{definition}
    The \correlationTile{} is the mapping
    $$corr_f : [0,1]^2\rightarrow[-1,1] : (a,b)\mapsto f(\aScore,\rankingScore)\comma$$
    where f is a correlation coefficient.
\end{definition}

\begin{figure}
\begin{centering}
\hfill
\includegraphics[height=3.1cm]{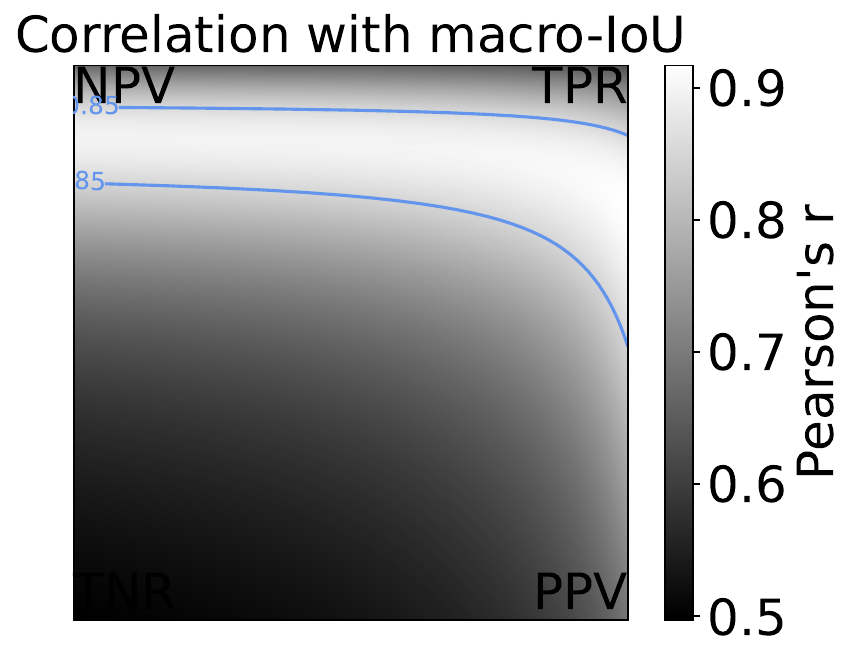}
\hfill
\hfill
\includegraphics[height=3.1cm]{images/results/tile_correlation_macro_iou_spearmanr}
\hfill
\end{centering}
\caption{\textbf{\correlationTile{} for linear (Pearson's $r$, left) and rank (Spearman's $\rho$, right) correlation coefficients.} These tiles show the linear and rank correlations between the canonical ranking scores and the macro-averaged IoU score that is usually taken as reference in the field of semantic segmentation, computed for the $31$ models trained on \cityscapes{}. The blue lines delineate the area where the correlation coefficients are $\ge 0.85$. 
\label{fig:correlation-tile}}
\end{figure}

\Cref{fig:correlation-tile} illustrates this correlation for the mean intersection over union (mIoU), a score commonly used to benchmark semantic segmentation models. A strong correlation between the mIoU and the \tile{} scores is observed in a horizontal band near the top.
Hence, selecting a score in this band will lead to similar conclusions than the ones provided by the mIoU. However, selecting a score outside this band will most probably allow understanding the performances of semantic segmentation algorithms under a fresh angle. This \correlationTile{} is therefore a powerful tool for understanding how to select complementary scores.

\subsubsection*{Scenario 2: The Method Designer \scenarioB}
 
The method designer focuses on evaluating the performance of his/her newly developed method, comparing it against state-of-the-art and other baseline methods. This user is particularly interested in understanding how a method performs across different importances given to true positives ($\sampleTP$), false positives ($\sampleFP$), false negatives ($\sampleFN$), and true negatives ($\sampleTN$). 
In cases of a parametric method, the designer also seeks insights into the optimal hyper-parameter settings that maximize performance.
Hereafter, we show how the \valueTile{}, \baselineTile{}, and \SOTATile{} can help the method designer understand the strengths and weaknesses of a method, providing insights needed to fine-tune performance and optimize hyper-parameters.

\mysection{\valueTile{}.}\label{subsec:ValueTile}
The \valueTile{} is a map displaying the value of each ranking score $\rankingScore$ across the \tile for a given entity and is defined as: 
\begin{definition}
    For an entity $\anEntity$, the \emph{\valueTile{}} is the mapping 
    $$V_\anEntity : [0,1]^2\rightarrow[0,1] : (a,b)\mapsto\rankingScore[\randVarImportance](\aPerformance_\anEntity)\point $$
\end{definition}

In practice, several methods can be used to compute the \emph{\valueTile{}}, which are discussed in supplementary material.

\Cref{fig:Value-Tile-best-methods} shows the \valueTile{} for the $4$ entities ranked first among the $74$ state-of-the-art semantic segmentation models (see \cref{fig:ranks}). This \tile can be used to get in one look where the entity performs best, and therefore where improvements should come from the method designer.

\begin{figure}
\begin{centering}
\includegraphics[width=\mysizefortwo]{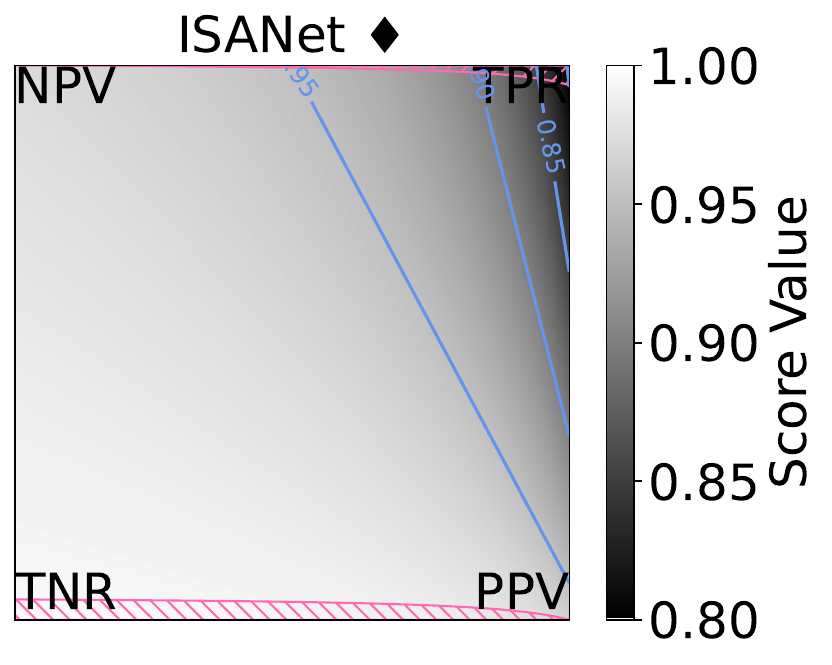}
\hfill
\includegraphics[width=\mysizefortwo]{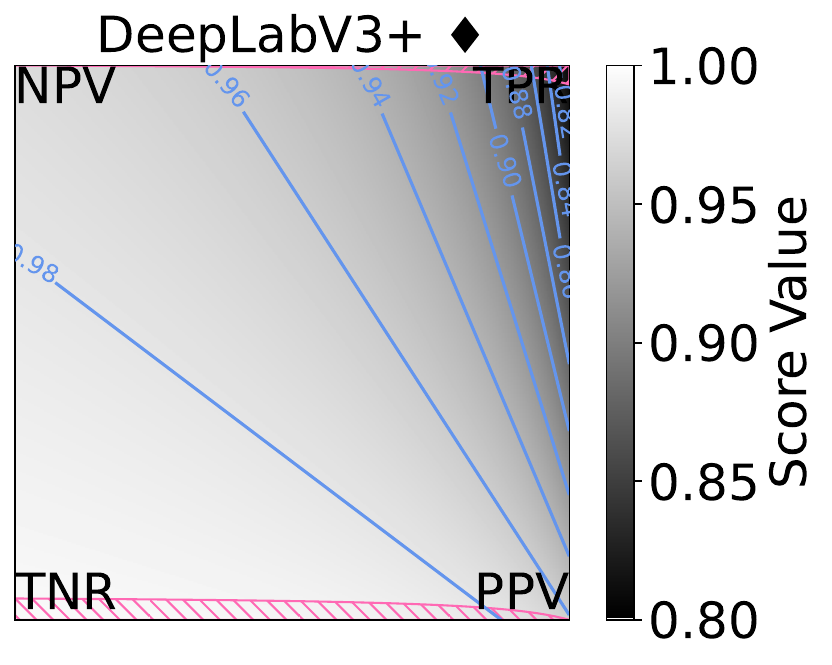}\\
\includegraphics[width=\mysizefortwo]{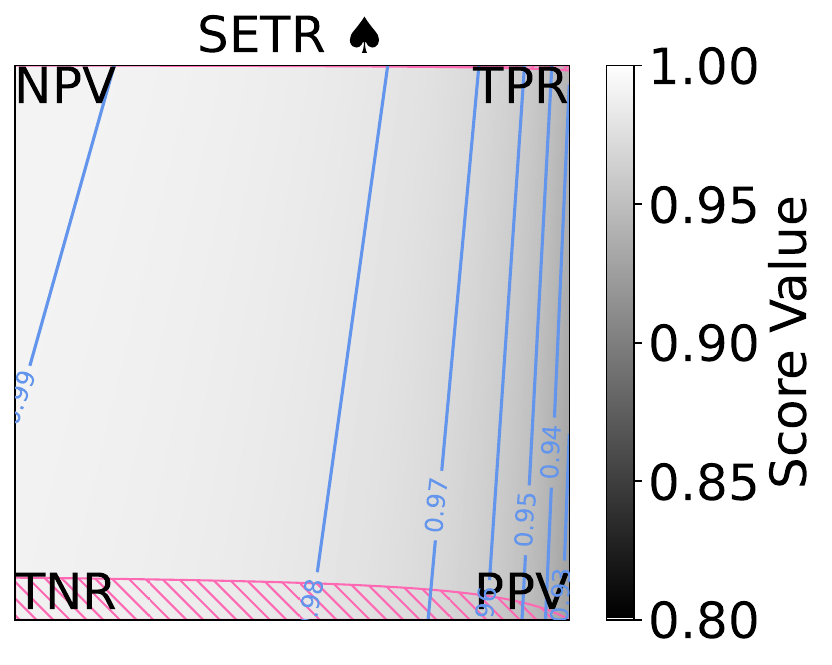}
\hfill
\includegraphics[width=\mysizefortwo]{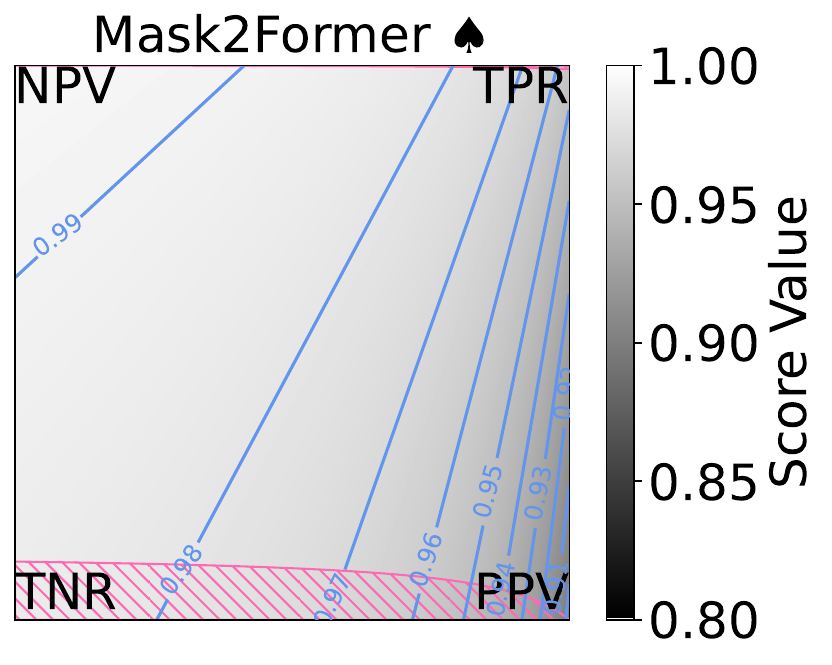}
\end{centering}
\caption{\textbf{\valueTile{} for the $4$ entities ranked first (see \cref{fig:ranks}).} The \valueTile{} shows the values of the canonical ranking scores. Blue lines are iso-value lines, \ie, lines on which all scores have the same value. Hatched areas indicate regions where non-skilled performances (\ie, when the ground truth $\randVarGroundtruthClass$ and the prediction $\randVarPredictedClass$ are independent)  surpass that of the entity. \label{fig:Value-Tile-best-methods}}
\end{figure}

The \valueTile shows a negative, zero, or positive horizontal gradient $\frac{\partial \rankingScore}{\partial a}$ when, respectively, $\ptp$ is less than, equal to or greater than $\ptn$. Similarly, the vertical gradient $\frac{\partial \rankingScore}{\partial b}$ is negative, zero, or positive when, respectively, $\pfn$ is less than, equal to, or greater than $\pfp$. As we have $\ptp < \ptn$ for all the models in our particular illustrative case, it is enough to look at the direction of the iso-value lines in order to analyze the bias (in the sense of \citet{Byrt1993Bias}): a model is biased towards the negative class (see, \eg, ISANet \LSvoc{} and DeepLabV3+ \LSvoc{} in \cref{fig:Value-Tile-best-methods}), unbiased, or biased towards the positive class (see, \eg, SETR \LScityscapes{} and Mask2Former \LScityscapes{} in \cref{fig:Value-Tile-best-methods}) when, respectively, the iso-value lines are tilted to the left, vertical, or tilted to the right.

\mysection{\baselineTile{} and \SOTATile{}.}\label{subsubsec:BaselineValueTile}
The \baselineTile{} is a map displaying, for a given set of entities, the minimum value for each ranking score $\rankingScore$ across the \tile, while the \SOTATile{} is a map displaying the maximum value for each ranking score $\rankingScore$. They are respectively defined as follows.
\begin{definition}
    For a  given set $\entitiesToRank$ of entities, the \baselineTile{} is the mapping 
    $$
    [0,1]^2\rightarrow[0,1] : (a,b)\mapsto\min_{\anEntity\in\entitiesToRank} \rankingScore(\aPerformance_\anEntity)\point$$
\end{definition}
\begin{definition}
    For a  given set $\entitiesToRank$ of entities, the \SOTATile{} is the mapping 
    $$
    [0,1]^2\rightarrow[0,1] : (a,b)\mapsto\max_{\anEntity\in\entitiesToRank} \rankingScore(\aPerformance_\anEntity)\point$$
\end{definition}

In other words, for a given set $\entitiesToRank$ of entities, the \baselineTile{} gives the value of each score corresponding to the entity that is ranked last in each point of the \tile, while the \SOTATile{} gives the value of the score corresponding to the entity that is ranked first.

\begin{figure}
\begin{centering}
\includegraphics[height=3.1cm]{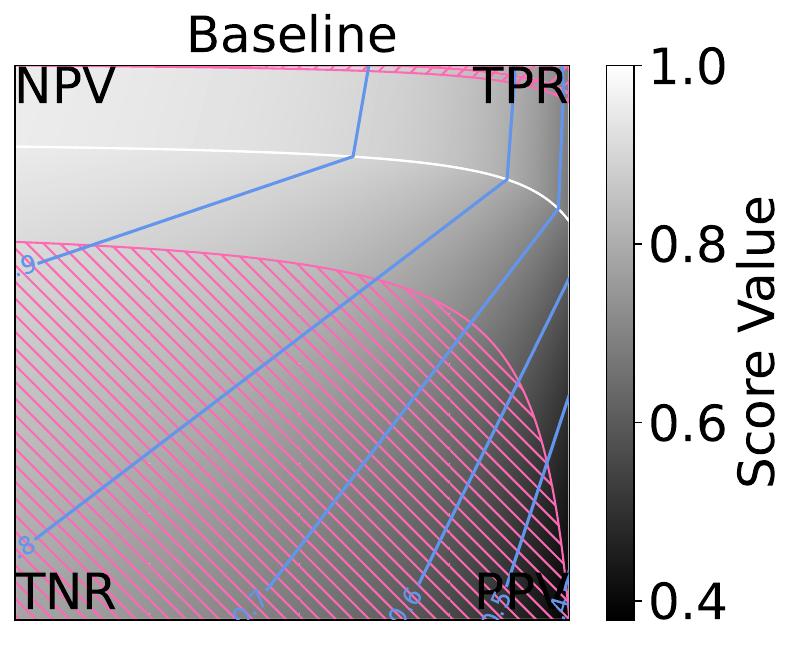}
\hfill
\includegraphics[height=3.1cm]{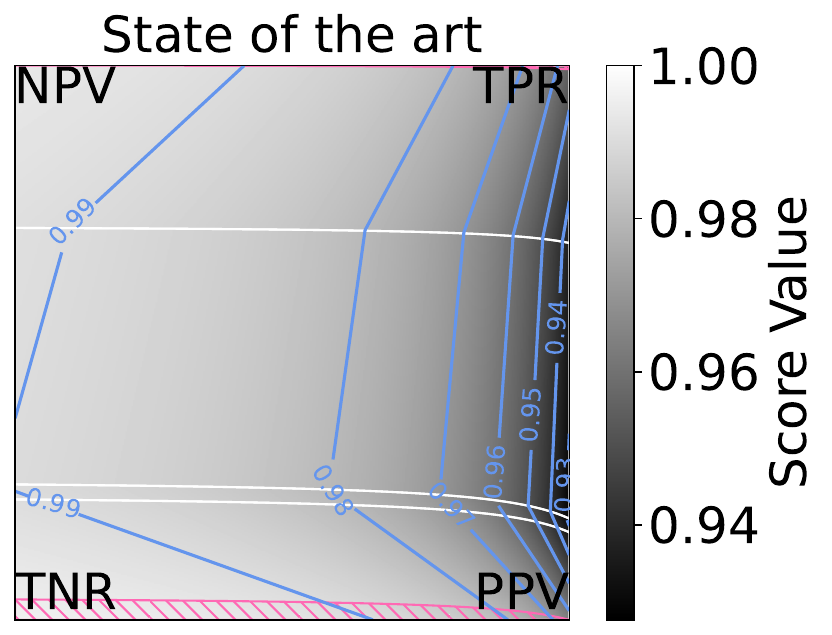}
\end{centering}
\caption{\textbf{\baselineTile{} (left) and \SOTATile{} (right).} These two tiles describe the current benchmark in semantic segmentation. The \baselineTile{} and the \SOTATile{} provide the \emph{infimum} and the \emph{supremum}, respectively, of the canonical ranking scores for the $74$ semantic segmentation models. In other words, they give the value of the scores for the entities ranked last and first, respectively. The blue lines are iso-value lines, \ie, lines on which all scores have the same value. The white lines mark the boundaries of these entities.} 
\label{fig:baseline-SOTA-tiles}
\end{figure}

The left-hand side of \cref{fig:baseline-SOTA-tiles} illustrates the \baselineTile{} for the $74$ semantic segmentation models, showing the lowest canonical ranking score values in each point among all compared entities, while the right-hand side illustrates the \SOTATile{} for the $74$ semantic segmentation models, showing the highest canonical ranking score values in each point among all compared entities. The hatched areas indicate that, if the classifiers always predicting the positive class, $\classPos$, or the negative class, $\classNeg$, were added to the set of entities to rank, they would occupy the top rank within these areas. These tiles are therefore interesting for the method designer when he/she compares them with the \valueTile{} of his/her entity. One can then easily compare where a method is close or above the \SOTATile{} and where it is close or below the \baselineTile{}. This indicates the advantages and drawbacks of a new method compared with previous works. Furthermore, these tiles may only consider a parametric family of the method, showing where different hyper-parameters perform well or bad, guiding the designer's choice.

\subsubsection*{Scenario 3: The Benchmarker \scenarioC}

The benchmarker is focused on comparing methods from the literature to identify which ones outperform others. This user profile aims to establish a clear ranking among different methods, ensuring that their assessment is fair and consistent. Additionally, the benchmarker may be interested in organizing open challenges, where the goal is to evaluate and rank participating methods to determine the top performer. In this context, it becomes crucial to accurately determine a winner by ranking the different entries based on their performance.
To meet these needs, the benchmarker requires a tool that can generate reliable rankings.

\mysection{\rankingTile{}.}\label{subsec:RankingTile}
The \rankingTile{} maps, for a given entity $\anEntity$, the rank of that entity across the tile.
Following \paperA, for a given set of entities $\entitiesToRank$, rankings are based on an ordering $\ordering_{\rankingScore[\randVarImportance]}$ of the performances induced by the ranking scores $\rankingScore$. 
The \rankingTile{} is defined as:

\begin{definition}
    The \rankingTile{} is the mapping
    $$[0,1]^2\rightarrow[1,|\entitiesToRank|] : (a,b)\mapsto\rank(\anEntity)\comma$$
    where $\rank(\anEntity)$ is computed according to $\ordering_{\rankingScore[\randVarImportance]}$.
\end{definition}

Therefore, the \rankingTile{} can be obtained by ordering the performances for each ranking score, \ie, ordering in each point $(a,b)$ the values of the ranking score for the \valueTile{} of each entity of the set. 

\begin{figure}
\begin{centering}
\includegraphics[width=\mysizefortwo]{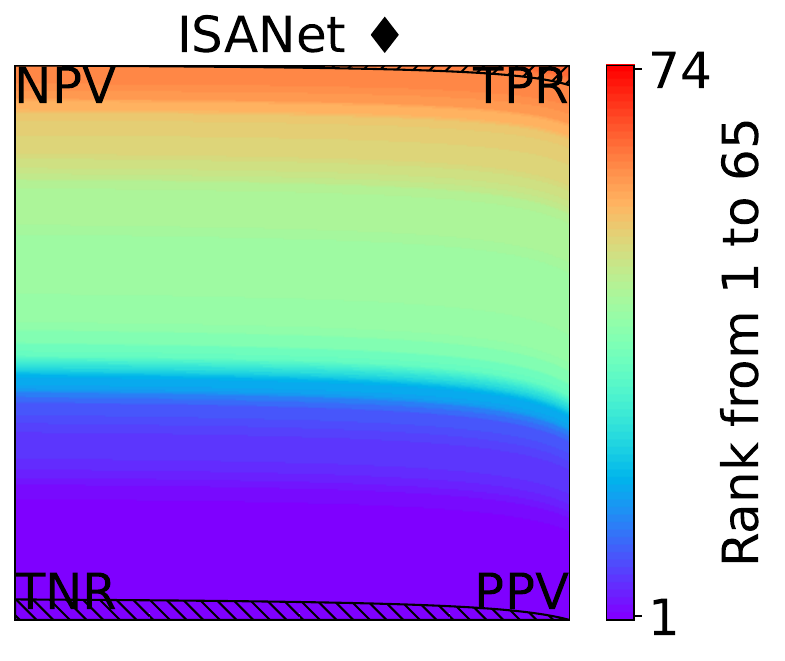} 
\hfill
\includegraphics[width=\mysizefortwo]{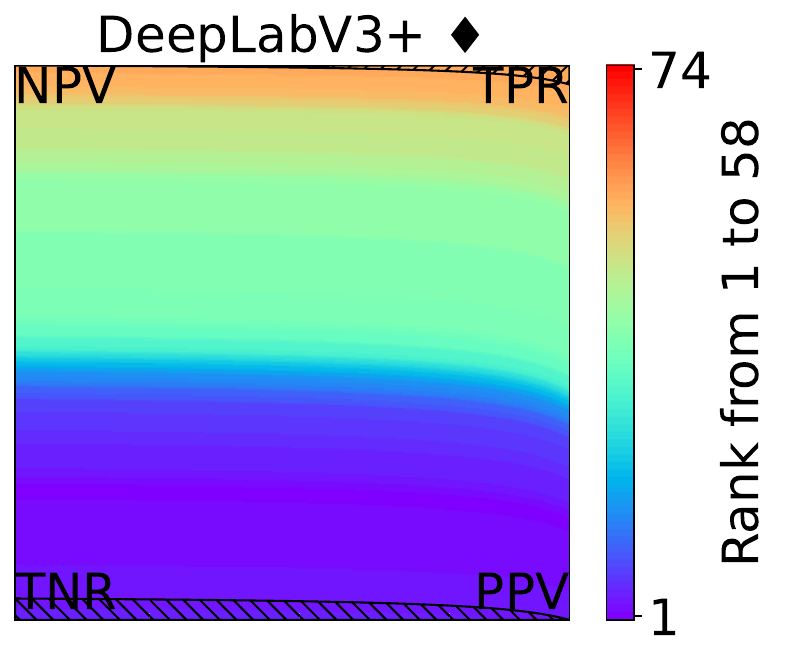}\\
\includegraphics[width=\mysizefortwo]{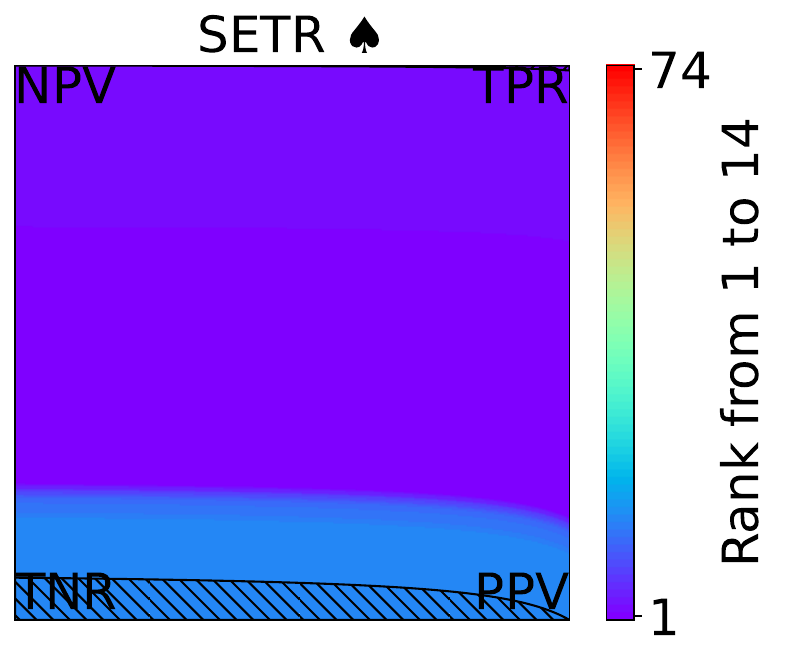}
\hfill
\includegraphics[width=\mysizefortwo]{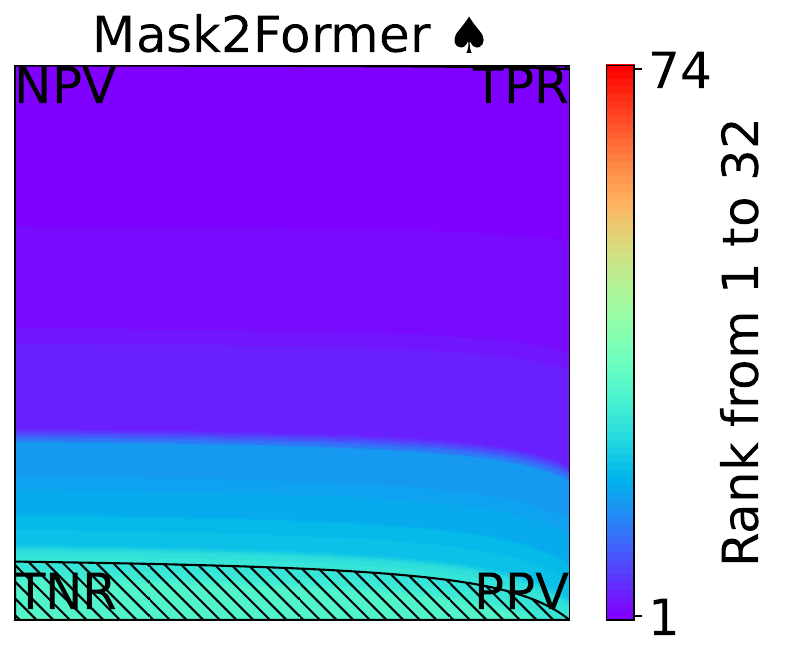}
\end{centering}
\caption{\textbf{\rankingTile{} for the $4$ entities that are ranked first, somewhere on the \tile{}.} The \rankingTile{} shows the ranks of a given entity across the tile.
Hatched areas highlight regions of the tile where no-skilled performances surpass that of the entity.
Remarkably, we directly see that (1)~ISANet~\LSvoc{} and DeepLabV3+~\LSvoc{} have poor rankings in the upper part of the tile, while (2)~SETR~\LScityscapes{} has the best overall performance and remains rather stable on the \tile, ranging from rank $1$ to $14$.
\label{fig:Ranking-Tile-best-methods}}
\end{figure}

\Cref{fig:Ranking-Tile-best-methods} shows the \rankingTile{} for the $4$ entities ranked first among the $74$ state-of-the-art semantic segmentation models (see \cref{fig:ranks}).
The \valueTile{} (\cref{fig:Value-Tile-best-methods}) and the \rankingTile{} (\cref{fig:Ranking-Tile-best-methods}) provide, at a glance, an overview of an entity's performance across the entire tile.

The \rankingTile{} is therefore a great solution to help the benchmarker identify the best-performing methods and appropriately determine the winners in open challenges, as it provides a structured way to know in which case a method has a particular rank.

\subsubsection*{Scenario 4: The Application Developer \scenarioD}

The application developer is interested in selecting the best method that aligns with the specific requirements of his/her application. This developer may already know the importance of true positives, false positives, false negatives, and true negatives relevant to a particular use case. The goal is therefore to choose a method that best meets these predefined criteria. For this purpose, the application developer needs a tool that allows him/her to efficiently identify the most suitable method based on specific priorities. 

\mysection{\entityTile{}.}\label{subsec:EntityTile}
The \entityTile{} maps entities that are at a given rank across the tile and is defined as follows.
\begin{definition}
    The \entityTile{}, for a given rank $r\in[1,|\entitiesToRank|]$, is the mapping
    $$[0,1]^2\rightarrow\entitiesToRank : (a,b)\mapsto\anEntity_r\comma $$
    where $\anEntity_r$ is the entity ranked $r$-th, according to $\ordering_{\rankingScore[\randVarImportance]}$.
\end{definition}

\begin{figure}
\begin{centering}
\includegraphics[width=0.49\linewidth]{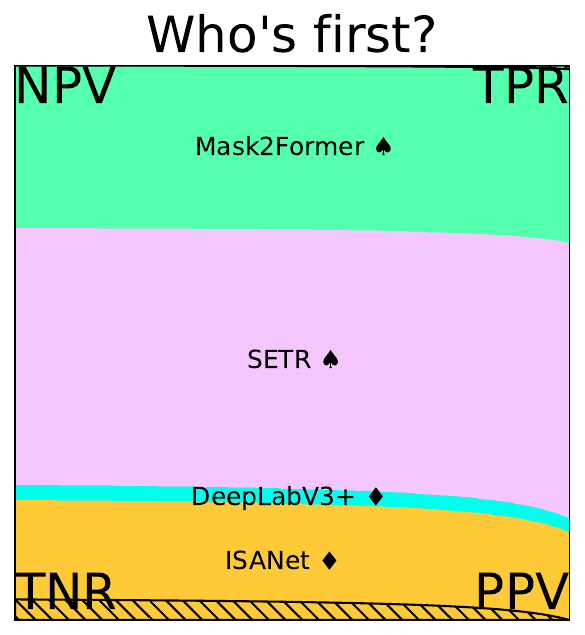}
\hfill
\includegraphics[width=0.49\linewidth]{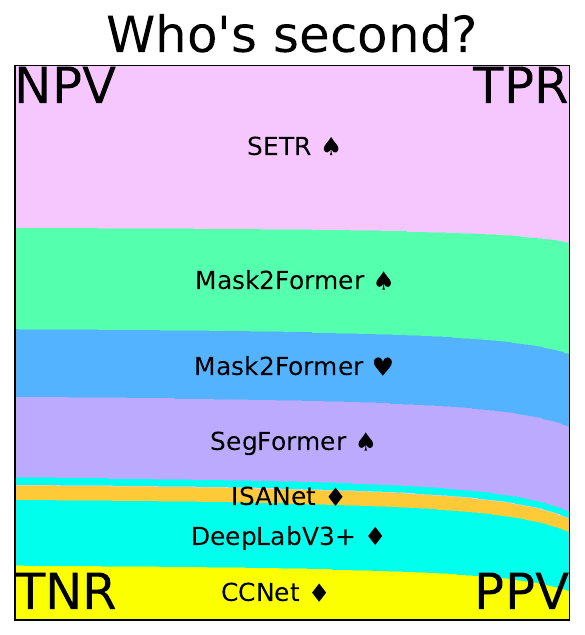}
\includegraphics[width=0.49\linewidth]{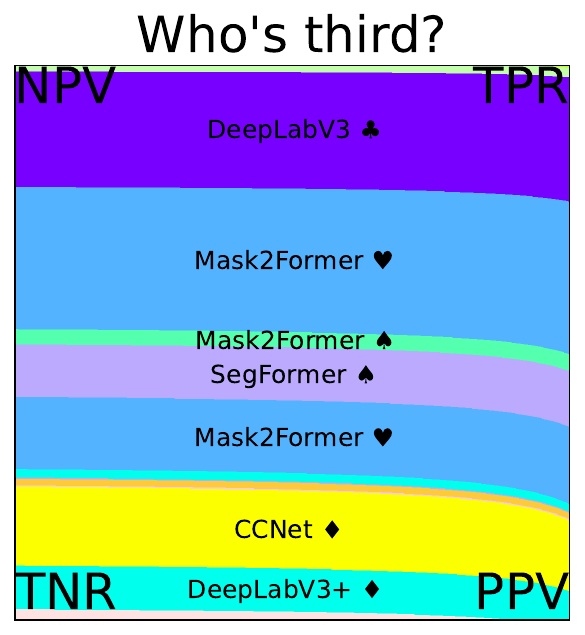}
\hfill
\includegraphics[width=0.49\linewidth]{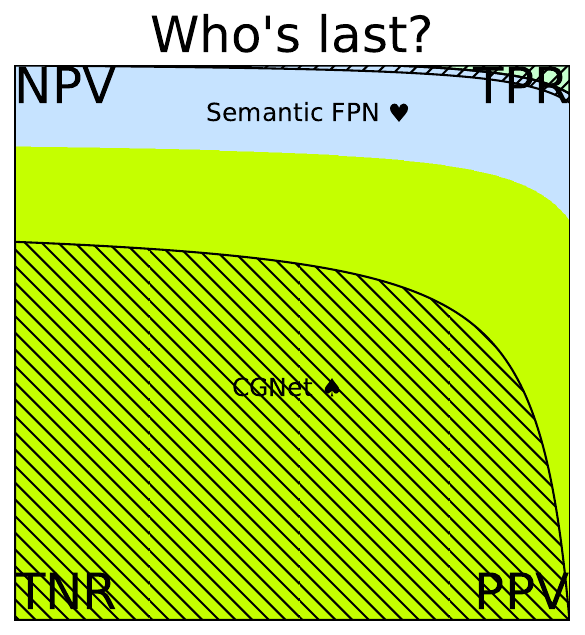}
\end{centering}
\caption{\textbf{\entityTile{} showing the entities that are ranked at the first, second, third, and last positions.}
Hatched areas in the ``who's first?'' and ``who's last?'' tiles highlight regions where no-skilled performances surpass that of the entities in the current map. \label{fig:ranks}}
\end{figure}

The \entityTile{} thus shows all entities that are at a given rank and is constructed similarly to the \rankingTile{}. 
In the special case of $r=1$, the \entityTile{} shows the best entities among $\entitiesToRank$, while for $r=|\entitiesToRank|$, it shows the worst entities among $\entitiesToRank$. 
As illustrated in \cref{fig:ranks} for the semantic segmentation models, the first rank is shared by $4$ entities, namely Mask2Former and SETR, both trained on \cityscapes{}, and DeepLabV3+ and ISANet, both trained on \voc{}. 
These $4$ entities also appear in the second and third ranks in other areas of the \rankingTiles{}, \ie, for other canonical ranking scores.
Interestingly, the entities ranked first are the ones on the upper dashed broken line of \cref{fig:roc} in the ROC space, and the ones ranked last are on the lower dashed broken line. This is a consequence of properties studied in \paperB.

The application developer may choose an entity based on those tiles depending on $3$ cases, based on the knowledge of importance values: 
\noindent (1) In the first case, the importance values are \emph{known}. Hence, the application developer simply selects the method ranked number 1 at the corresponding place on the \tile. This is typically how most benchmarks rank methods, \ie, based on a single score.
\noindent (2) In the second case, the importance values are \emph{unknown but can be determined} by analyzing the community practices. For example, in semantic segmentation, the ``mean intersection over union'' (mIoU) is often considered to be a good criterion. Although this score is not part of the \tile{}, the \correlationTile{} provides a direct comparison between the orderings induced by all ranking scores and the macro-averaged IoU. The application developer can then simply select the method ranked first in the area, where Spearman's $\rho$ is the highest in the tile. 
\noindent (3) In the last case, there is \emph{no information} on the importance values. 
One selection mechanism consists in minimizing the maximum rank over the \rankingTile{} and, in case of ex aequo, minimizing the average rank. 

Therefore, the \entityTile{} enables the application developer to select the best method by visualizing how different options perform relative to the importance values, whether they are known or not. This allows to make informed decisions tailored to specific application requirements.

\section{Discussion}

\mysection{Bands.}
Unlike the \valueTiles{}, all Correlation, Ranking and Entity \tiles illustrating this guide exhibit roughly horizontal bands. They can be explained as follows. All performances such that $\ptn\propto\ptp$ are perfectly rank-correlated with all ranking scores on any given horizontal line of the \tile. In our illustration, the ratio $\frac{\ptp}{\ptn}$ is roughly constant ($\mu=0.12$, $\sigma=0.01$), which explains the roughly horizontal bands. Similarly, all performances such that $\pfp\propto\pfn$ are perfectly rank-correlated with all ranking scores on any given vertical line of the \tile. This covers the case, not illustrated in this guide, in which all compared performances are unbiased, in the sense of \citet{Byrt1993Bias}, \ie $\aPerformance(\randVarGroundtruthClass)=\aPerformance(\randVarPredictedClass$).

\mysection{Multi-class classification extension.} 
The universal theoretical foundations for performance-based ranking~\cite{Pierard2025Foundations}, on top of which the \tile was build~\cite{Pierard2024TheTile-arxiv}, are indeed applicable to multi-class problems. With $n$ classes, specifying the relative importance given to the samples $\aSample$ such that $\randVarPredictedClass(\aSample)=\randVarGroundtruthClass(\aSample)$ (as done on the horizontal axis of the \tile) requires $n-1$ dimensions, and specifying the relative importance given to the samples $\aSample$ such that $\randVarPredictedClass(\aSample)\ne\randVarGroundtruthClass(\aSample)$ (as done on the vertical axis of the \tile) requires $n^2-n-1$ dimensions. A "ranking space" similar to the \tile, but for $n$ classes, would therefore require $n^2-2$ dimensions. This can be handled algorithmically, but it cannot be easily visualized. A typical alternative is to convert multi-class problems into two-class problems by micro- or macro-averaging~\cite{Sokolova2009ASystematic}. 
While micro- and macro-averaged \valueTiles{} can be constructed and used to derive \entityTiles{}, this approach is not necessarily advisable as it raises serious concerns. On the one hand, all micro-averaged scores are only functions of the multi-class accuracy, which leads to a loss of information, and, on the other hand, macro-averaging comes with interpretability issues~\cite{Pierard2020Summarizing}. We hope that this will motivate further studies on the topic.

\mysection{Choice of a positive class.} 
Once the labels have been grouped into two classes, there remains to choose which is the positive one, in order to comply with the classical convention. However, the opposite choice leads to identical \tiles, up to a central symmetry~\paperB. In particular, the list of entities (classifiers) identified as being the best (ranked first) does not depend on which class is chosen as positive.

\section{Conclusion}

In this paper, we introduced a hitchhiker’s guide to understanding the performance of two-class classifiers. We organized our guide into four distinct scenarios, each corresponding to a specific user profile, covering the theoretical analyst \scenarioA, the method designer \scenarioB, the benchmarker \scenarioC, and the application developer \scenarioD. For each scenario, we provided examples of challenges each user may encounter when comparing classifier performances and identified the most suitable tools for their needs. Central to our approach is the idea of displaying different types of information, called \emph{flavors}, on the \tile{}. 
By leveraging different mappings, we demonstrated the versatility of the \tile{} and its various flavors to address the diverse requirements of these user profiles. This guide offers a flexible and robust framework that enables users to effectively evaluate, rank, and interpret the performance of two-class classifiers, making it a valuable resource for researchers, practitioners, and developers alike.

\clearpage
 \mysection{Acknowledments.}
 The work by A. Halin and S. Pi{\'e}rard was supported by the Walloon Region (Service Public de Wallonie Recherche, Belgium) under grant n°2010235 (ARIAC by \href{https://www.digitalwallonia.be/en/}{DIGITALWALLONIA4.AI}). 
 A. Cioppa is 
 funded by the \href{https://www.frs-fnrs.be}{F.R.S.-FNRS}.


\clearpage 

\appendix
\section{Supplementary Material} \label{sec:supplementary}

We provide in this supplementary material (1)~the list of symbols used in this paper (\Cref{SM-subsec:Symbols}), (2)~a description of the software in the Python Jupyter Notebook (\cref{SM-subsec:Software}), (3)~a definition of the \skillTile{} (\cref{SM-subsec:SkillTile}), (4)~more details about the illustration used in the paper (\cref{SM-subsec:Example}), (5)~an analysis of the influence of a learning set (\cref{SM-subsec:LS}), (6)~an analysis of the behavior of scores, using the \correlationTile{}, explaining the patterns observed in the various tiles presented in the paper (\cref{SM-subsec:Scores}), and finally (7)~the comprehensive report generated by the Python Jupyter Notebook for our illustration (\cref{SM-subsec:all-results}).


\subsection{List of Symbols}\label{SM-subsec:Symbols}

\subsubsection*{Symbols used for the probability theory}
\begin{itemize}
    \item $\sampleSpace$: the sample space (universe)
    \item $\eventSpace$: the event space (a $\sigma$-algebra on $\sampleSpace$, \eg $2^\sampleSpace$)
    \item $\measurableSpace$: the measurable space
\end{itemize}

\subsubsection*{Symbols used for our probabilistic framework for performances}
\begin{itemize}
    \item $\aPerformance$: a performance, \ie, a probability measure
    \item $\aPerformance_\anEntity$: the performance of an entity $\anEntity$
    \item $\allPerformances$: all performances on $\measurableSpace$
    \item $\aScore$: a score
    \item $\allScores$: all scores on $\measurableSpace$
    \item $\domainOfScore$: the domain of the score $\aScore$
\end{itemize}

\subsubsection*{Symbols used for two-class classifications}
\begin{itemize}
    \item $\randVarGroundtruthClass$: the random variable for the ground truth
    \item $\randVarPredictedClass$: the random variable for the prediction
    \item $\classNeg$: the negative class
    \item $\classPos$: the positive class
    \item $\sampleTN$: the sample \emph{true negative}
    \item $\sampleFP$: the sample \emph{false positive}
    \item $\sampleFN$: the sample \emph{false negative}
    \item $\sampleTP$: the sample \emph{true positive}
    \item $\scoreAccuracy$: the score \emph{accuracy}
    \item $\scoreTNR$: the score \emph{true negative rate}
    \item $\scoreTPR$: the score \emph{true positive rate}
    \item $\scoreFPR$: the score \emph{false positive rate}
    \item $\scoreNPV$: the score \emph{negative predictive value}
    \item $\scorePPV$: the score \emph{positive predictive value}
    \item $\scoreFBeta$: the F-scores
    \item $\priorpos$: the score \emph{prior of the positive class}
    \item $\priorneg$: the score \emph{prior of the negative class}
    \item ROC: Receiver Operating Characteristic
    \item PR: Precision-Recall
    \item AUC-ROC: Area Under The ROC curve
    \item AUC-PR: Area Under The PR curve
\end{itemize}

\subsubsection*{Symbols used for the performance-based ranking of entities}
\begin{itemize}
    \item $\rank$: the \emph{ranking} function, relative to the set of entities~$\entitiesToRank$
    \item $\entitiesToRank$: the set of entities to rank
    \item $\anEntity$: an entity (\ie, an element of $\entitiesToRank$)
    \item $\randVarImportance$: the random variable \emph{Importance}
    \item $\rankingScore$: the \emph{ranking score} parameterized by the importance $\randVarImportance$
    \item $\ordering_{\rankingScore}$: the ordering induced by the ranking score $\rankingScore$
    \item $a$: the parameter specifying the relative importance given to the incorrect outcomes (\ie, $\sampleTP$ and $\sampleTN$), it corresponds to the horizontal axis of the \tile{}
    \item $b$: the parameter specifying the relative importance given to the correct outcomes (\ie, $\sampleFN$ and $\sampleFP$), it corresponds to the vertical axis of the \tile{}
\end{itemize}

\subsubsection*{Symbols used for our illustration}
\begin{itemize}
    \item \LScityscapes{}: the learning set \emph{Cityscapes}
    \item \LSade{}: the learning set \emph{ADE20K}
    \item \LSvoc{}: the learning set \emph{Pascal VOC 2012}
    \item \LScoco{}: the learning set \emph{COCO-Stuff 164k}
\end{itemize}

\subsubsection*{Other symbols}
\begin{itemize}
    \item $\realNumbers$: the real numbers
    \item mIoU: mean intersection over union
    \item $r$: the linear correlation coefficient of Pearson
    \item $\rho$: the rank correlation coefficient of Spearman
    \item $\tau$: the rank correlation coefficient of Kendall
\end{itemize}

\subsection{Software Description}\label{SM-subsec:Software}

The \emph{Python Jupyter Notebook} generates a comprehensive report related to the illustration of this paper, featuring all the tiles presented in the paper and more. It takes as input a list of two-class classification performances (\ie, the values of $\aPerformance(\{\aSample\})$, for all $\aSample \in \{\sampleTN, \sampleFP, \sampleFN, \sampleTP\}$).
The various tiles are obtained by integrating two types of information. The first involves point-specific data, such as the value of a ranking score, a rank, an entity, or a correlation value, as detailed in this paper. This is achieved by discretizing the \tile{} using a grid size parameter along the two axis, $a$ and $b$, and mapping the respective data, accordingly to the ranking scores in the \tile{}, to a point $(a,b)$.
The second type, relevant when priors are fixed (\ie, when all compared performances have the same class priors), pertains to area-based information. The algorithm used to identify these areas of interest, described in \paperB, determines the areas (and their respective boundaries) within the \tile{} where an entity holds a specific rank (typically, the first or the last) without requiring a discretization of the \tile{}. The hatched areas on the various tiles provided in the paper (\eg, the \valueTile{} or \rankingTile{}) are drawn using this algorithm, as well as the white boundary lines on the \baselineTile{} and the \SOTATile{}.

\subsubsection{Three Methods to Compute \valueTiles}

In practice, several methods can be used to compute the \emph{\valueTile{}}, depending on the availability of data.

\begin{enumerate}
    \item \emph{Direct computation:} If the values of $\aPerformance(\eventFP)$, $\aPerformance(\eventFN)$, $\aPerformance(\eventTP)$, and $\aPerformance(\eventTN)$ are known, the first method consists in computing the values of the ranking scores in each point of the \tile using \cref{eq:ranking-scores}.
    \item \emph{Interpolation:} Knowing the values taken by the scores $\scoreTPR$, $\scoreTNR$, $\scoreNPV$, and $\scorePPV$ at the four corners of the \tile{}, the values taken by the remaining scores can be determined by averaging, using $f$-means, vertically with $f:x\mapsto x^{-1}$ (\ie the harmonic mean) and horizontally with $f:x\mapsto (1-x)^{-1}$ \cite{Pierard2024TheTile-arxiv}.
    As one can see, the non-linearities differ between the vertical and horizontal axes.
    \item \emph{Equation system resolution:} The third method involves first, determining the values of $\aPerformance(\eventFP)$, $\aPerformance(\eventFN)$, $\aPerformance(\eventTP)$, and $\aPerformance(\eventTN)$ by solving a system of $4$ equations with $4$ unknowns, then, computing the values of the ranking scores using \cref{eq:ranking-scores}. Specifically, if the values of three ranking scores from the \valueTile{} are known, or if two ranking scores and the priors (for fixed priors) with $\priorneg = \aPerformance(\eventTN) + \aPerformance(\eventFP)$ and $\priorpos = \aPerformance(\eventFN) + \aPerformance(\eventTP)$, then, knowing that
    $
    \aPerformance(\eventFP) + \aPerformance(\eventFN) + \aPerformance(\eventTP) + \aPerformance(\eventTN) =1
    $ 
    makes the system complete.
\end{enumerate}

These three methods require discretizing the \tile{}. This can be implemented with a grid size parameter, which defines linearly spaced values for the two axes, $a$ and $b$. 
The provided code implements the direct computation method.

\subsection{\skillTile{}}\label{SM-subsec:SkillTile}

In this section, we define two new tiles, namely the \noSkillTile{} and the \skillTile{}. For this purpose, we first define two sets of performances.

We say that a performance $\aPerformance$ is \emph{no-skilled} when the ground truth $\randVarGroundtruthClass$ and the prediction $\randVarPredictedClass$ are independent. 
We denote the set of all non-skilled performances on $\measurableSpace$ by $\allNonSkilledPerformances$:
\begin{equation}
    \resizebox{.88\hsize}{!}{$\allNonSkilledPerformances=\left\{ \aPerformance \in \allPerformances : \aPerformance(\randVarGroundtruthClass,\randVarPredictedClass)=\aPerformance(\randVarGroundtruthClass) \aPerformance(\randVarPredictedClass) \right\}\point$}
\end{equation}
Therefore, the classifier always predicting the positive class $\classPos$ and the classifier always predicting the negative class $\classNeg$ have both no-skilled performances.

When the priors are the same for all compared performances, we can define the set of all performances at a given positive prior, $\priorpos$:
\begin{equation}
    \allPriorFixedPerformances=\left\{ \aPerformance \in \allPerformances : \aPerformance(\randVarGroundtruthClass=\classPos) = \priorpos \right\}\point
\end{equation}

\begin{definition}
    The \emph{\noSkillTile{}} is the mapping 
    $$noskill_\randVarImportance : [0,1]^2\rightarrow[0,1] : (a,b)\mapsto\max_{\aPerformance\in\allNonSkilledPerformances\cap\allPriorFixedPerformances} \rankingScore(\aPerformance)\point$$
\end{definition}

The \noSkillTile{} thus displays the value of the ranking scores in each point of the tile for the best-performing non-skilled performance. Note that, \eg, the hatched areas on \cref{fig:Value-Tile-best-methods} highlight regions where the \noSkillTile{} exhibits a higher value than the \valueTile{}.

\begin{definition}
    The \emph{\skillTile{}} is the mapping
    $$skill_\randVarImportance : [0,1]^2\rightarrow[0,1] : (a,b)\mapsto\frac{SOTA_\randVarImportance-noskill_\randVarImportance}{1-noskill_\randVarImportance}\comma$$
    where $SOTA_\randVarImportance$ corresponds to the values of the \SOTATile{}.
\end{definition}

\begin{figure}
\begin{centering}
\hfill
\includegraphics[height=3.1cm]{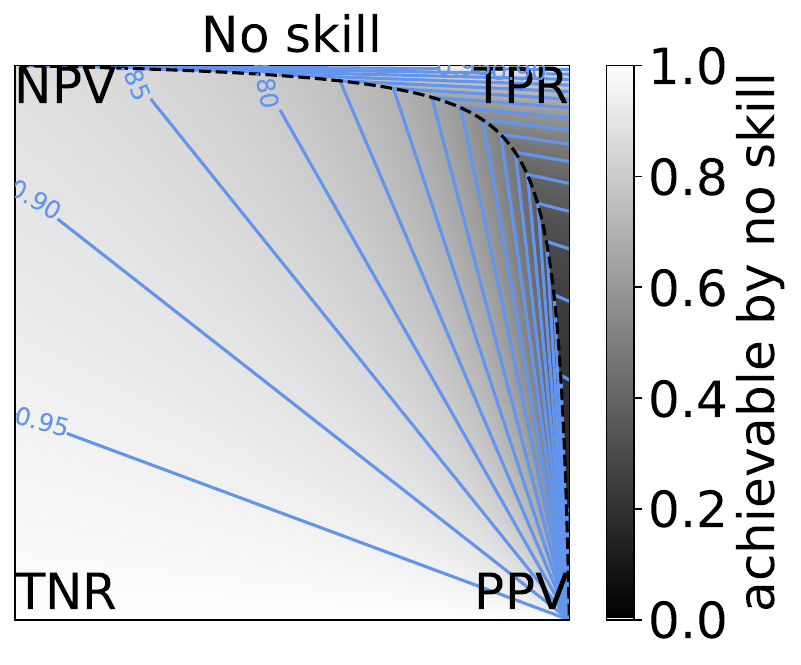}
\hfill
\hfill
\includegraphics[height=3.1cm]{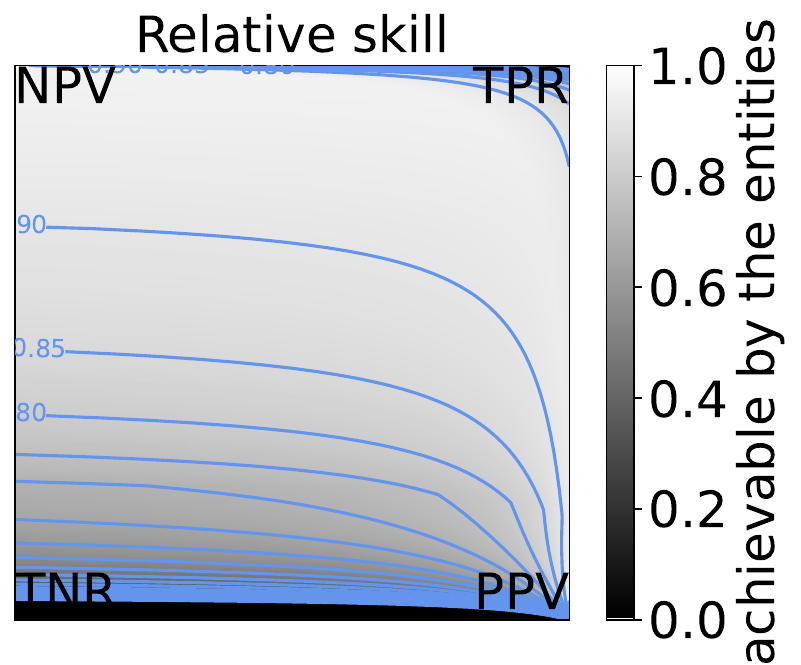}
\hfill
\end{centering}
\caption{\textbf{\noSkillTile{} (left) and \skillTile{} (right).} These tiles show, on the left, the score values of the best no-skilled performances and, on the right, the relative performance of the state of the art compared to no-skilled performances for the $74$ semantic segmentation models. The blue lines are iso-value lines, \ie, lines on which all scores have the same value. The dashed line indicates boundaries between different entities.
\label{fig:skill-tile}}
\end{figure}

\Cref{fig:skill-tile} illustrates the \noSkillTile{} and the \skillTile{}. Note that the code to obtain these tiles is also available in the Python Jupyter Notebook.

\subsection{More Details on the Illustration}\label{SM-subsec:Example}

Our illustration is for an arbitrarily chosen pixel-based two-class classification task derived from a pixel-based semantic segmentation task. The performances are those of $74$ models evaluated on $8{,}000$ images of the BDD100K dataset~\cite{Yu2020BDD100K}: this is our testing set. These models have been trained on the \cityscapes{}~\cite{Cordts2016The}, \ade{}~\cite{Zhou2017Scene}, \voc{}~\cite{Everingham2010PascalVOC}, or \coco{}~\cite{Caesar2018COCO} datasets: these are the learning sets. The set of semantic labels in BDD100K are identical to those in \cityscapes{}, but differ from the ones in \ade{}, \voc{}, and \coco{}.

\paragraph{Defining the two classes for the testing set.} We arbitrarily took the union of the first $11$ semantic labels of BDD100K as the negative class, corresponding to background objects (\eg, road, sidewalk, or building), and the union of the $8$ remaining ones as the positive class (person, rider, car, truck, bus, train, motorcycle, and bicycle).

\paragraph{Defining the two classes for the $4$ learning sets.} The semantic labels predicted by the models are those from the corresponding learning set. We therefore also had to map all these labels to our negative and positive classes. To this end, for each learning set, taking into account all the models learned on this learning set, we computed the proportion of pixels that are positive for each semantic label. The semantic label has been attributed to the positive class when this proportion is greater than the positive prior $\priorpos$ of BDD100K (about $0.1242$). 
                                     
\paragraph{} All the results presented in our paper, and also here-after, are specific to these arbitrary choices. In particular, we noticed that thresholding the posteriors at $0.5$ instead of $\priorpos$ leads to a significantly different two-class classification problem, for which the ranking of the models is different.

\subsection{Influence of a Learning Set}\label{SM-subsec:LS}
\begin{figure}
\begin{centering}
\includegraphics[width=\mysizefortwo]{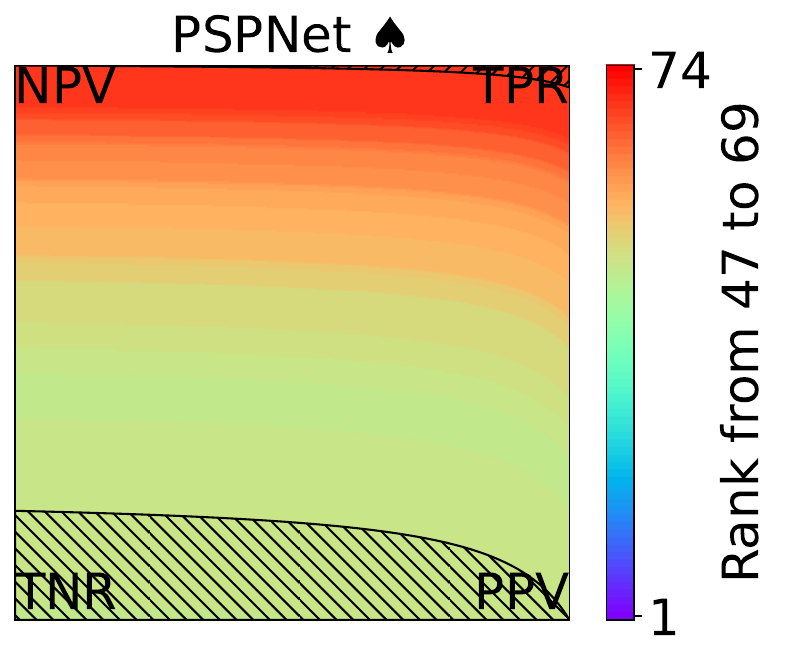}
\hfill
\includegraphics[width=\mysizefortwo]{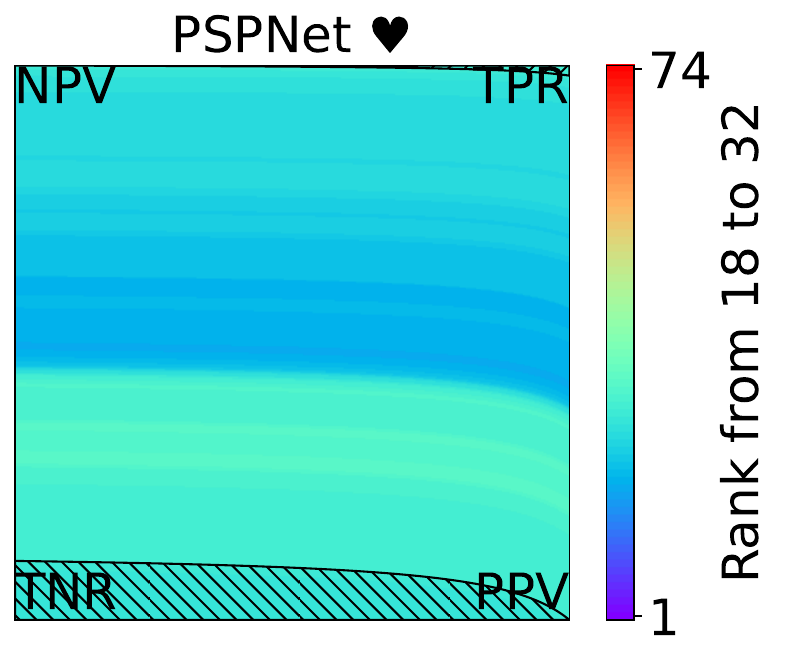}
\includegraphics[width=\mysizefortwo]{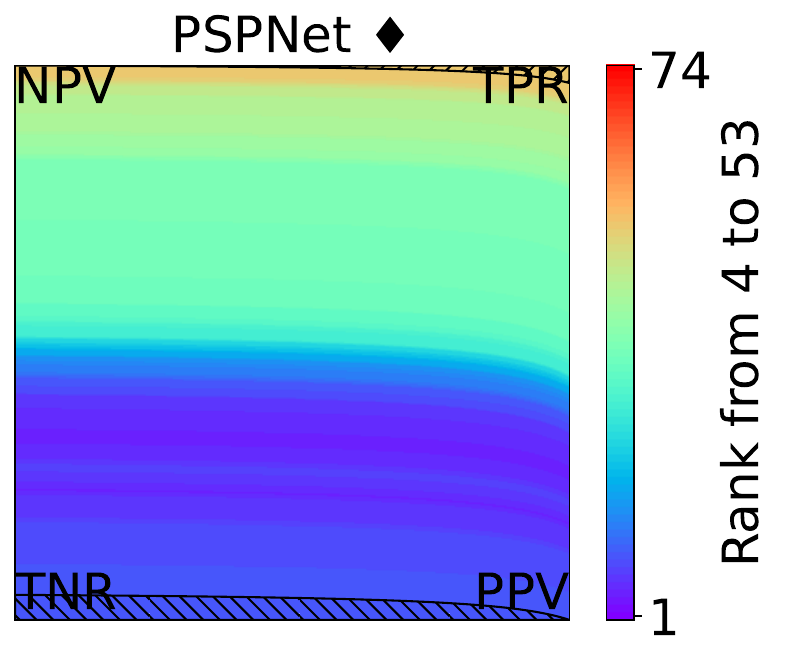}
\hfill
\includegraphics[width=\mysizefortwo]{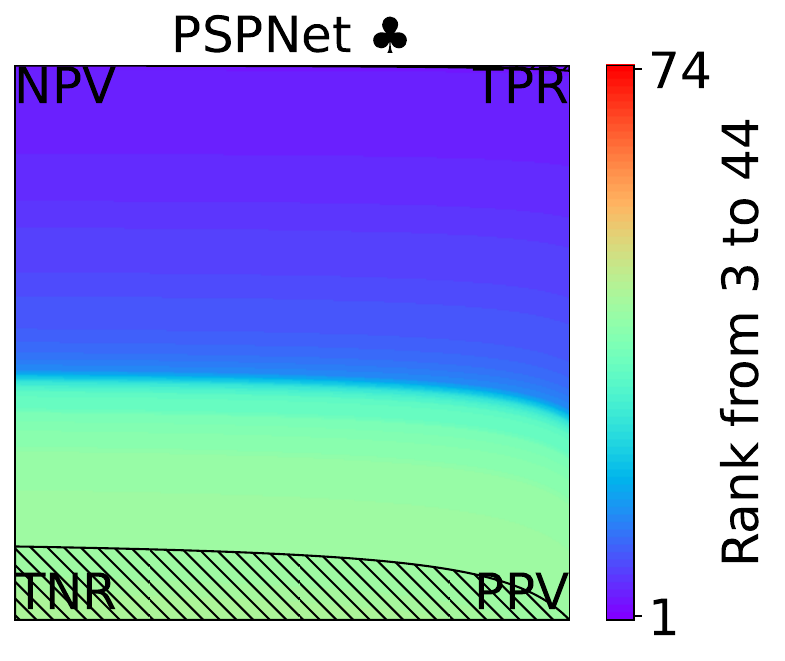}
\end{centering}
\caption{\textbf{\rankingTiles showing the rank of PSPNet trained on different learning datasets (from left to right: \cityscapes{}, \ade{}, \voc{}, and \coco{}).} We observe that the learning dataset strongly influences the ranking behavior.
\label{fig:influence-of-ls}}
\end{figure}

The influence of a learning set on the ranking of a semantic segmentation model can be seen by placing the \rankingTiles corresponding to that model trained on the different learning sets side by side. 
As shown in \cref{fig:influence-of-ls}, training PSPNet on \voc{} and \coco{} performs better than training it on \cityscapes{}. But the best stability over the tile is achieved by training PSPNet on \ade{}.

\subsection{Behavior of Scores}\label{SM-subsec:Scores}

\begin{figure}
\begin{centering}
\subfloat[True Negative Rate]{\begin{centering}
\includegraphics[width=0.3\linewidth]{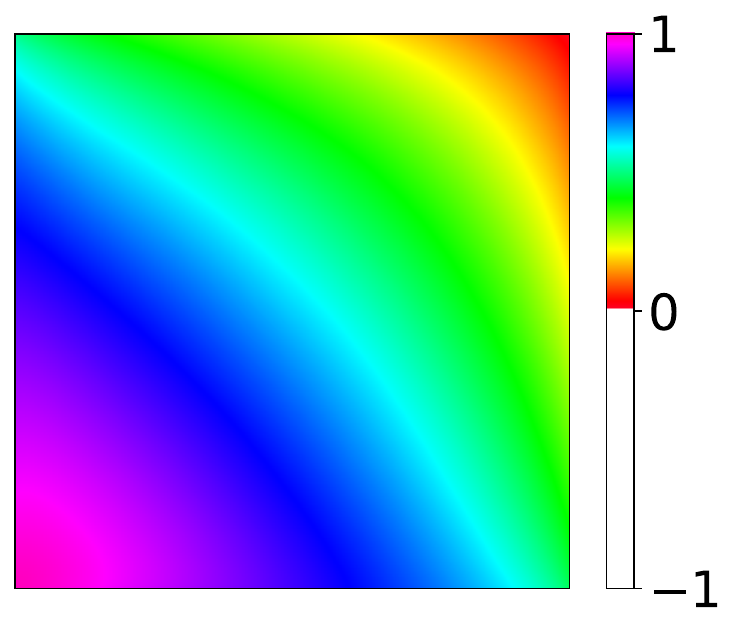}\includegraphics[width=0.3\linewidth]{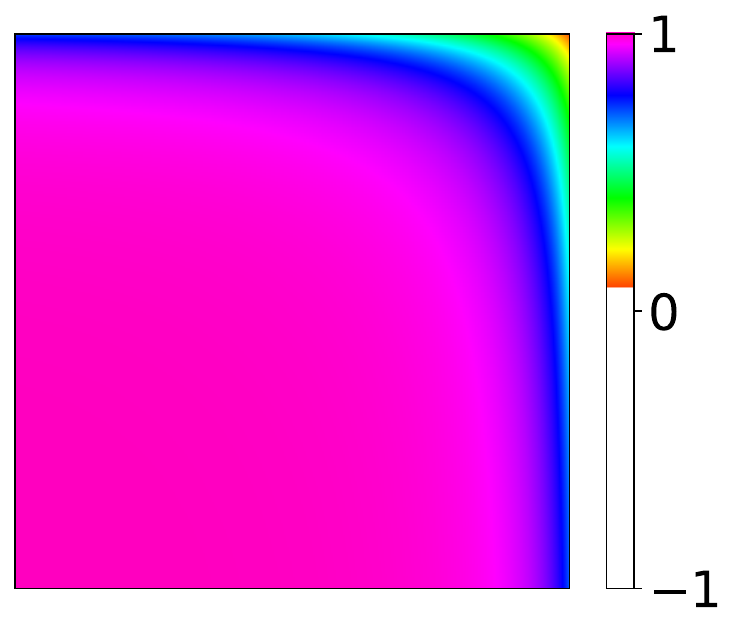}\includegraphics[width=0.3\linewidth]{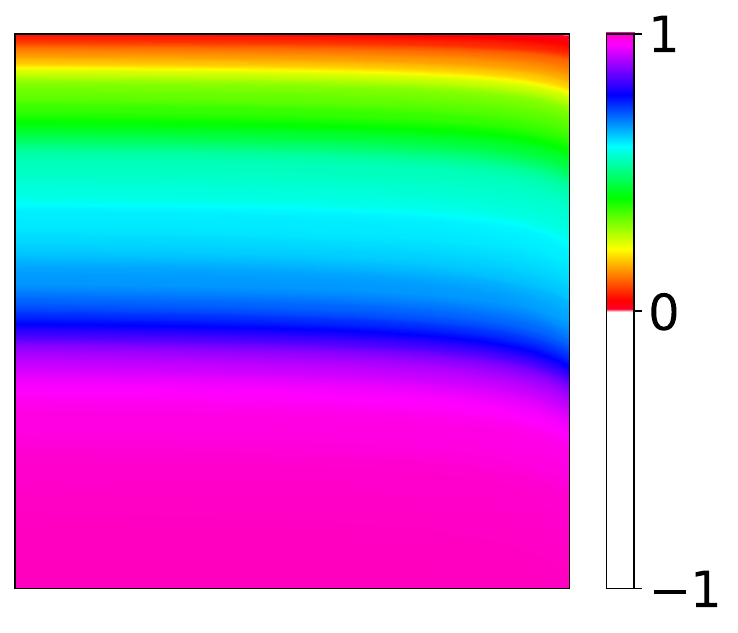}
\par\end{centering}
}
\par\end{centering}
\begin{centering}
\subfloat[True Positive Rate]{\begin{centering}
\includegraphics[width=0.3\linewidth]{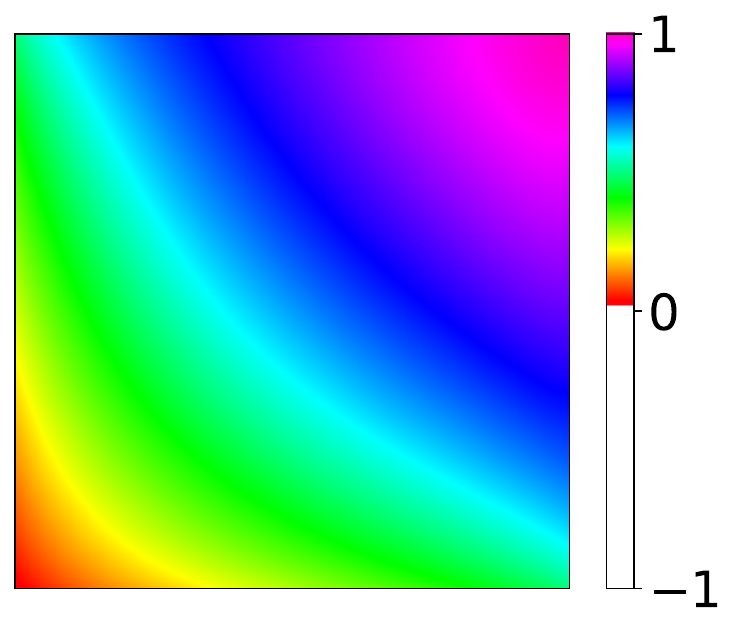}\includegraphics[width=0.3\linewidth]{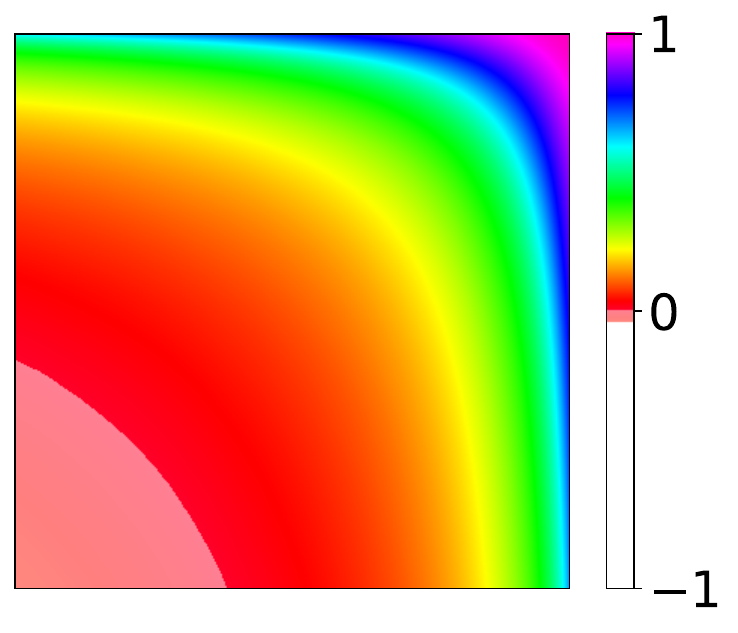}\includegraphics[width=0.3\linewidth]{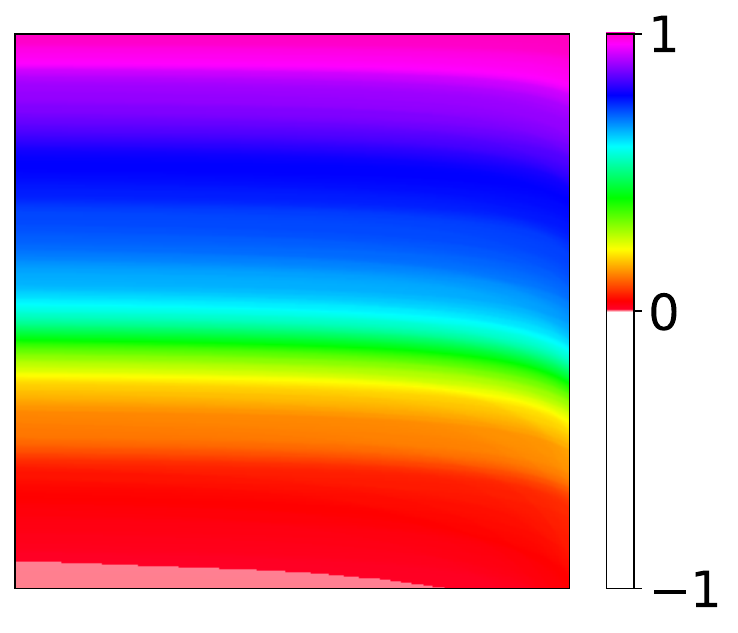}
\par\end{centering}
}
\par\end{centering}
\begin{centering}
\subfloat[Negative Predictive Value]{\begin{centering}
\includegraphics[width=0.3\linewidth]{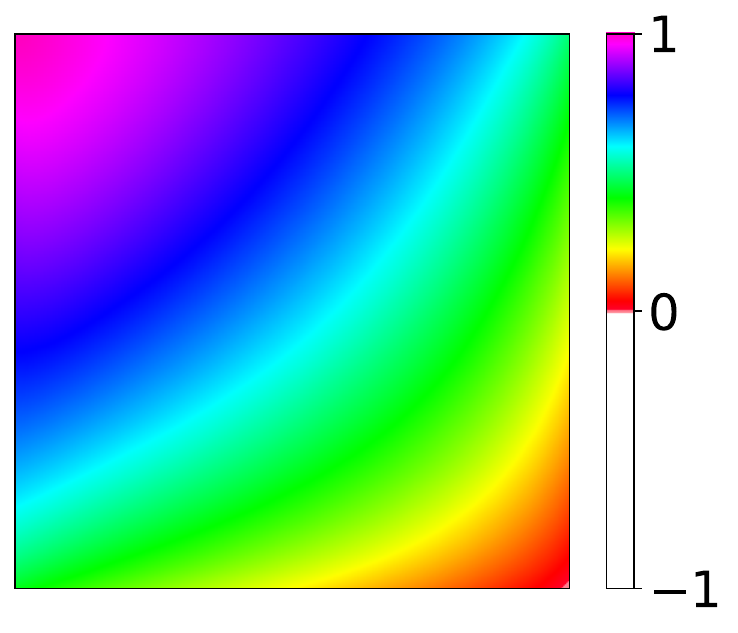}\includegraphics[width=0.3\linewidth]{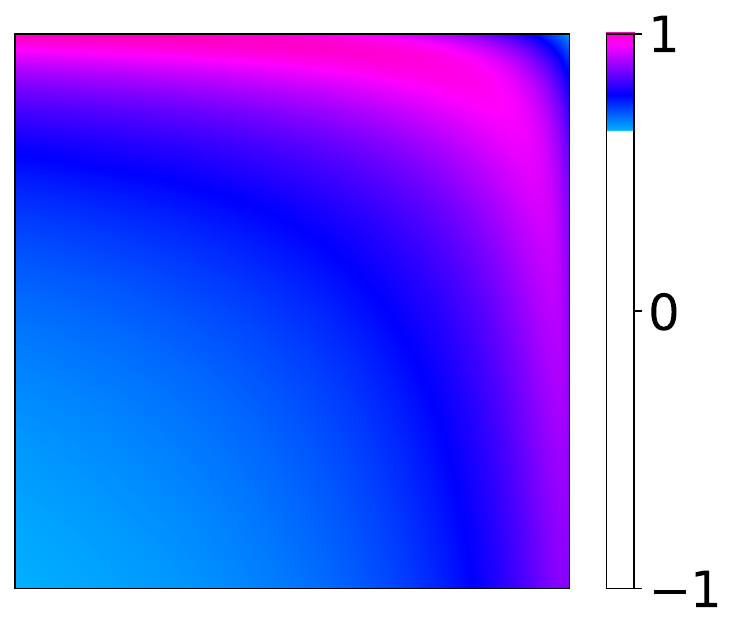}\includegraphics[width=0.3\linewidth]{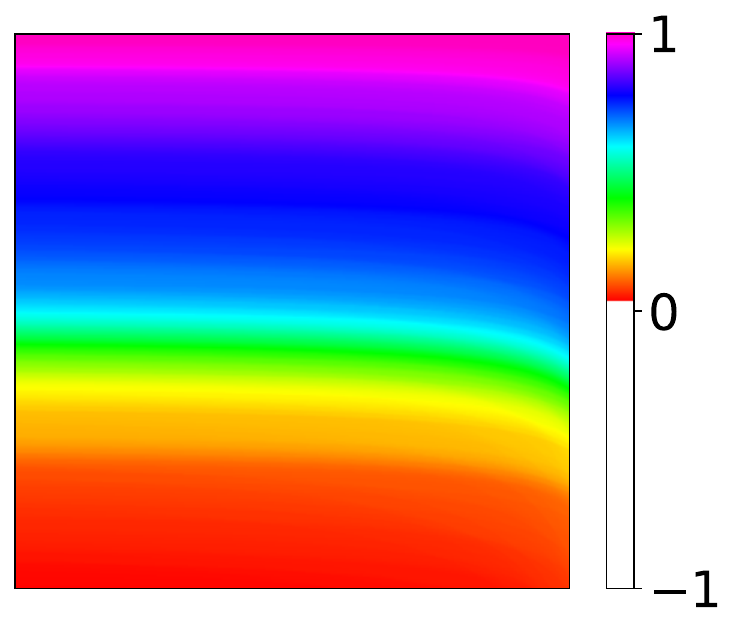}
\par\end{centering}
}
\par\end{centering}
\begin{centering}
\subfloat[Positive Predictive Value]{\begin{centering}
\includegraphics[width=0.3\linewidth]{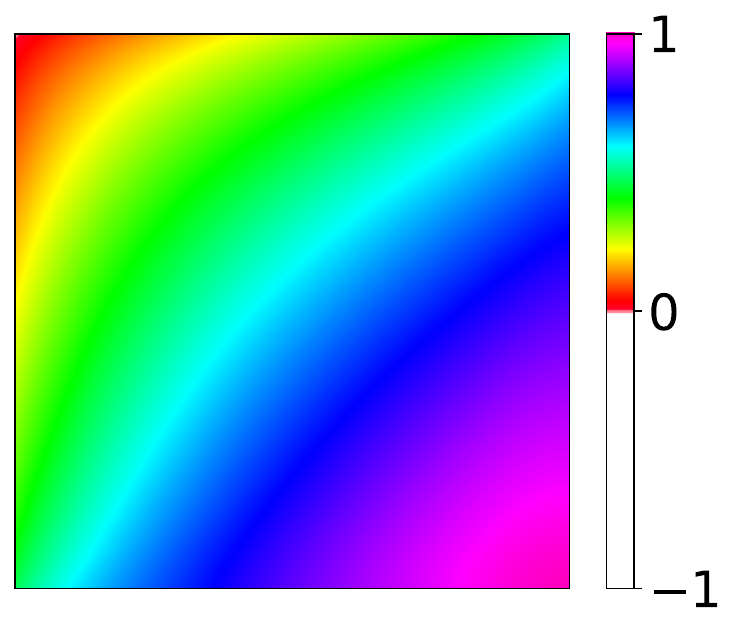}\includegraphics[width=0.3\linewidth]{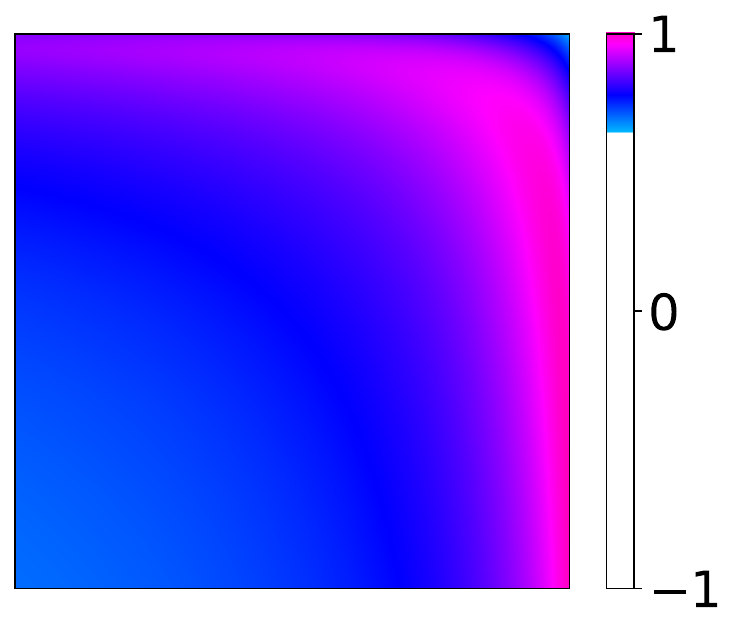}\includegraphics[width=0.3\linewidth]{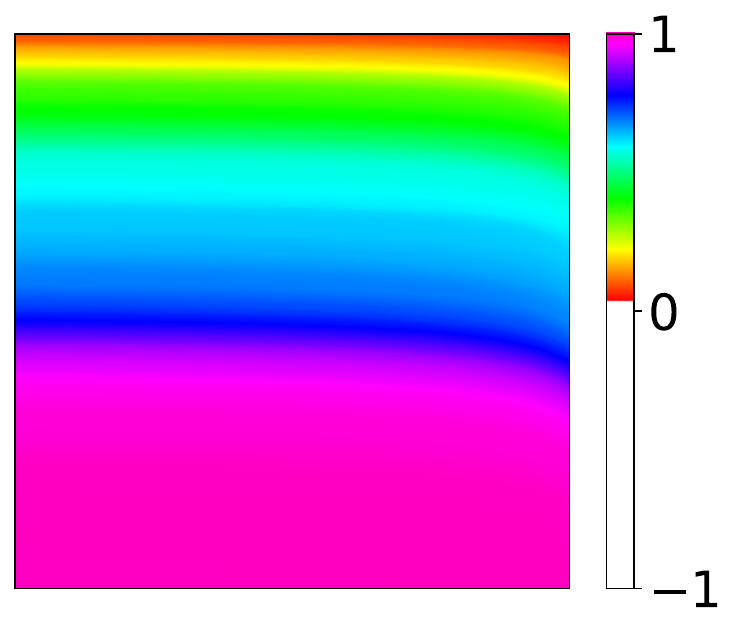}
\par\end{centering}
}
\par\end{centering}
\begin{centering}
\subfloat[Accuracy]{\begin{centering}
\includegraphics[width=0.3\linewidth]{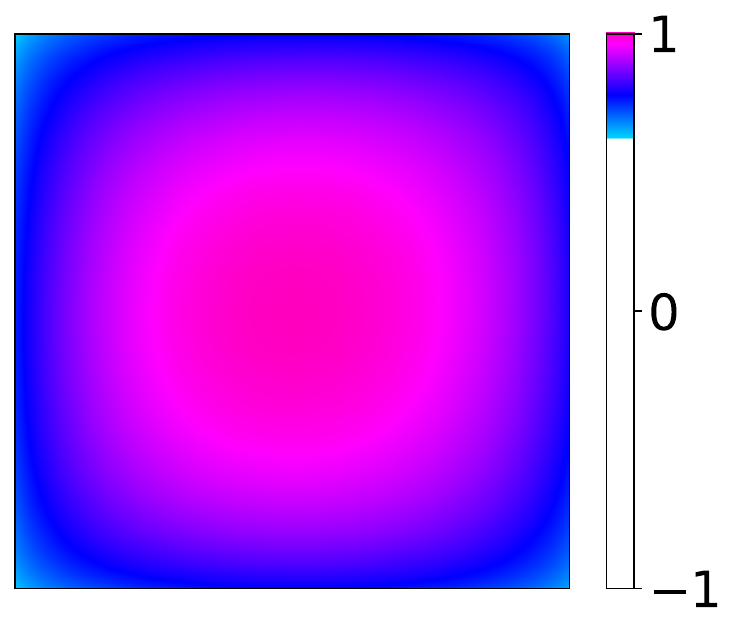}\includegraphics[width=0.3\linewidth]{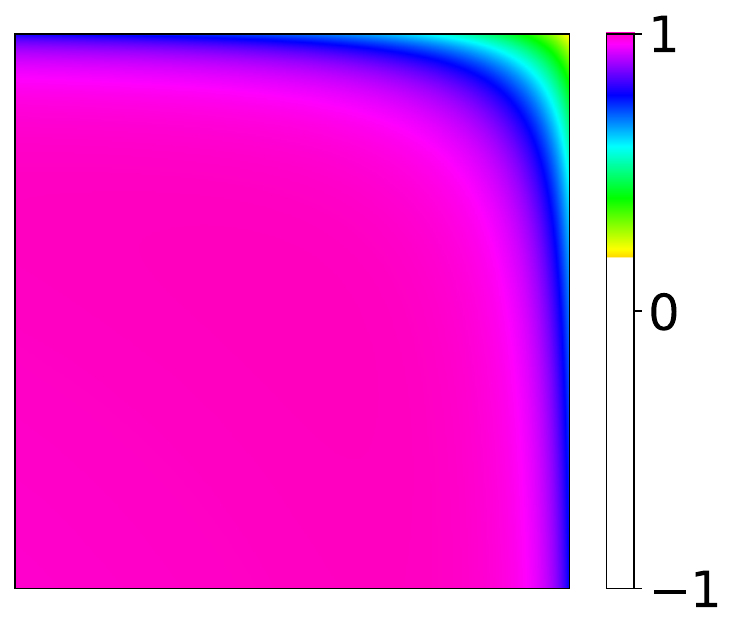}\includegraphics[width=0.3\linewidth]{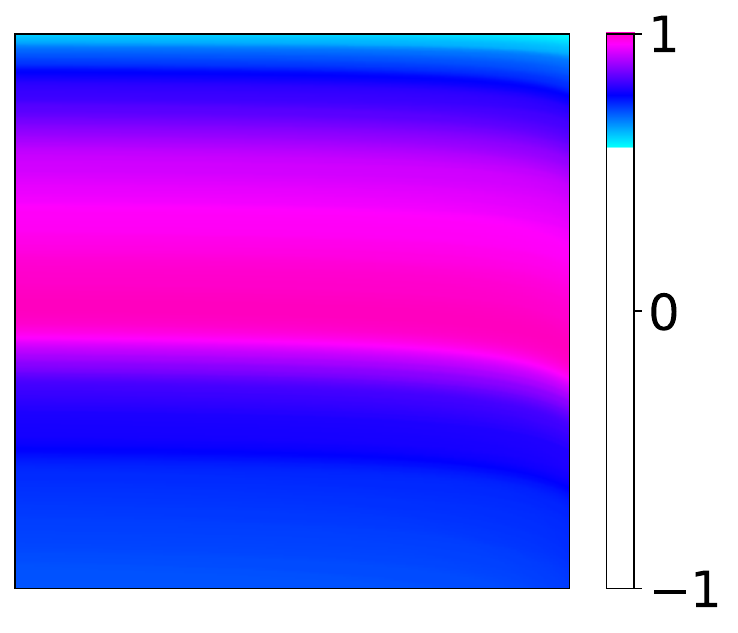}
\par\end{centering}
}
\par\end{centering}
\begin{centering}
\subfloat[F-one]{\begin{centering}
\includegraphics[width=0.3\linewidth]{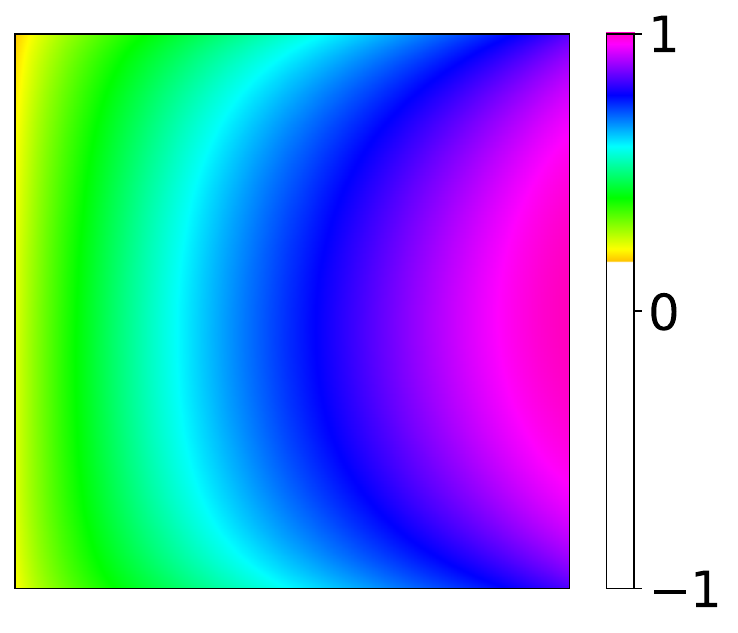}\includegraphics[width=0.3\linewidth]{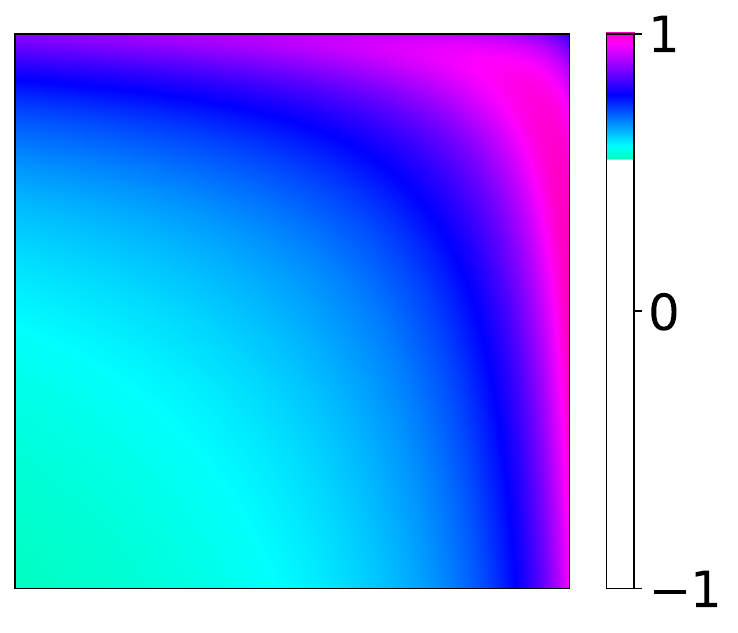}\includegraphics[width=0.3\linewidth]{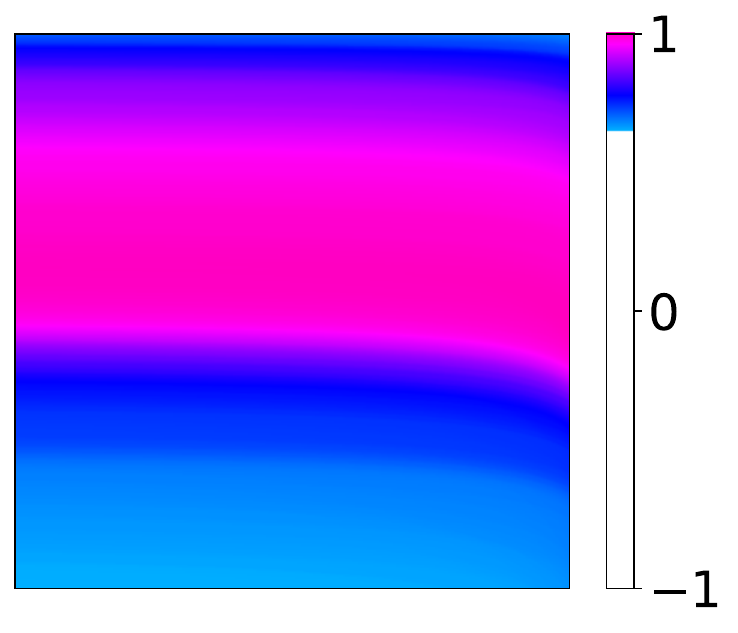}
\par\end{centering}
}
\par\end{centering}
\caption{\textbf{Behavior of $6$ scores for $3$ sets of performances.} The \correlationTile{} shows the estimated rank correlations (with the Spearman's $\rho$) between $6$ scores and all canonical ranking scores, for a uniform distribution over all possible performances (left), over all performances with a prior of the positive class equal to the one in our illustration ($\priorpos=0.124227$) (center), and over the $74$ performances compared in our illustration (right).\label{fig:behavior_of_scores}}

\end{figure}

The \correlationTile{} allows to depict the behavior of any score, showing the rank correlations between that score and all canonical ranking scores, for a given performance distribution.
On \cref{fig:behavior_of_scores}, we analyze the rank correlation, using Spearman correlation coefficient (Spearman's $\rho$), for $6$ scores belonging to the ranking scores (namely $\scoreTNR$, $\scoreTPR$, $\scoreNPV$, $\scorePPV$, $\scoreAccuracy$, and $\scoreFBeta[1]$) and $3$ distributions of performances (all $3$ distributions are uniform but are on different sets of performances). 
We analyze these scores (1)~for a uniform distribution over all performances, (2)~for a uniform distribution of performances having fixed priors corresponding to those of BDD100K, and (3)~for a uniform distribution of performances, where the set of performances is the one of our illustration.
Comparing the figures on the left with those on the center, we see the effect of having fixed priors with $\priorpos=0.1242$. The figures on the right show the effect of having fixed priors and an arbitrary choice of performances (the ones of the $74$ state-of-the-art semantic segmentation models). For the explanation of the quasi-horizontal bands, see \cref{tab:behavior_of_scores}.

\begin{table}[h]
    \centering
    \begin{tabular}{l|c|c|c}
         & Set 1 & Set 2 & Set 3 \\
        \hline
        \hline
         Num. perfs. (size)     & $1000$           & $1000$            & $74$  \\
         $\min(\theta)$             & $0.000628$      & $0.0000634$     & $0.0971$   \\
         $\max(\theta)$             & $1.57$          & $1.56$         & $0.170$    \\
         $\mu(\theta)$              & $0.806$        & $0.242$        & $0.125$    \\
         $\sigma(\theta)$           & $0.496$        & $0.281$        & $0.00997$ \\
    \end{tabular}
    \caption{\textbf{Comparison of the three sets of two-class classification performances used in \cref{fig:behavior_of_scores}:} (1) a uniform distribution over all possible performances, (2) over all performances with a prior of the positive class equal to the one in our illustration ($\priorpos=0.124227$), and (3) over the $74$ performances compared in our illustration. The score $\theta$ is given by $\theta=\arctan(\frac{\ptp}{\ptn})$. The set for which the standard deviation $\sigma(\theta)$ is minimal is the one for which $\ptn$ is the closer to be proportional to $\ptp$. It is the third one. This is why one can observe quasi horizontal bands in the \correlationTiles on the right hand side of \cref{fig:behavior_of_scores}.}
    \label{tab:behavior_of_scores}
\end{table}

\onecolumn
\subsection{Report Generated by the Python Jupyter Notebook for our Illustration}\label{SM-subsec:all-results}

%
%

\subsection*{The performances of the two-class classification entities}
\begin{longtable}[c]{|l|c|c|c|c|}
\hline 
entity & $\aPerformance(\eventTN)$ & $\aPerformance(\eventFP)$ & $\aPerformance(\eventFN)$ & $\aPerformance(\eventTP)$\tabularnewline
\hline 
\hline 
ANN \ding{171} & 0.8390 & 0.0368 & 0.0178 & 0.1064 \tabularnewline
ANN \ding{170} & 0.8615 & 0.0142 & 0.0136 & 0.1106 \tabularnewline
ANN \ding{169} & 0.8713 & 0.0045 & 0.0255 & 0.0987 \tabularnewline
APCNet \ding{171} & 0.8365 & 0.0393 & 0.0232 & 0.1010 \tabularnewline
APCNet \ding{170} & 0.8599 & 0.0159 & 0.0128 & 0.1114 \tabularnewline
BiSeNetV1 \ding{171} & 0.8453 & 0.0304 & 0.0250 & 0.0992 \tabularnewline
BiSeNetV1 \ding{168} & 0.8357 & 0.0401 & 0.0083 & 0.1159 \tabularnewline
BiSeNetV2 \ding{171} & 0.8332 & 0.0426 & 0.0154 & 0.1088 \tabularnewline
CCNet \ding{171} & 0.7896 & 0.0862 & 0.0190 & 0.1052 \tabularnewline
CCNet \ding{170} & 0.8621 & 0.0136 & 0.0142 & 0.1101 \tabularnewline
CCNet \ding{169} & 0.8719 & 0.0039 & 0.0253 & 0.0990 \tabularnewline
CGNet \ding{171} & 0.6831 & 0.1926 & 0.0070 & 0.1172 \tabularnewline
DANet \ding{171} & 0.8030 & 0.0728 & 0.0173 & 0.1069 \tabularnewline
DANet \ding{170} & 0.8626 & 0.0131 & 0.0136 & 0.1107 \tabularnewline
DANet \ding{169} & 0.8717 & 0.0040 & 0.0265 & 0.0978 \tabularnewline
DeepLabV3+ \ding{171} & 0.8175 & 0.0583 & 0.0200 & 0.1042 \tabularnewline
DeepLabV3+ \ding{170} & 0.8627 & 0.0131 & 0.0136 & 0.1106 \tabularnewline
DeepLabV3+ \ding{169} & 0.8718 & 0.0040 & 0.0248 & 0.0994 \tabularnewline
DeepLabV3 \ding{171} & 0.8424 & 0.0333 & 0.0213 & 0.1029 \tabularnewline
DeepLabV3 \ding{170} & 0.8633 & 0.0125 & 0.0140 & 0.1102 \tabularnewline
DeepLabV3 \ding{169} & 0.8710 & 0.0048 & 0.0241 & 0.1002 \tabularnewline
DeepLabV3 \ding{168} & 0.8609 & 0.0149 & 0.0078 & 0.1164 \tabularnewline
DMNet \ding{171} & 0.8517 & 0.0240 & 0.0305 & 0.0937 \tabularnewline
DMNet \ding{170} & 0.8618 & 0.0140 & 0.0135 & 0.1107 \tabularnewline
DNLNet \ding{171} & 0.8316 & 0.0441 & 0.0205 & 0.1038 \tabularnewline
DNLNet \ding{170} & 0.8633 & 0.0124 & 0.0143 & 0.1099 \tabularnewline
DPT \ding{170} & 0.8635 & 0.0123 & 0.0150 & 0.1092 \tabularnewline
EMANet \ding{171} & 0.8383 & 0.0375 & 0.0246 & 0.0996 \tabularnewline
EncNet \ding{171} & 0.8437 & 0.0320 & 0.0282 & 0.0960 \tabularnewline
EncNet \ding{170} & 0.8633 & 0.0125 & 0.0149 & 0.1093 \tabularnewline
ERFNet \ding{171} & 0.8450 & 0.0308 & 0.0169 & 0.1073 \tabularnewline
Fast-SCNN \ding{171} & 0.8290 & 0.0468 & 0.0136 & 0.1106 \tabularnewline
FastFCN \ding{171} & 0.8397 & 0.0361 & 0.0253 & 0.0989 \tabularnewline
FastFCN \ding{170} & 0.8613 & 0.0144 & 0.0144 & 0.1098 \tabularnewline
FCN \ding{171} & 0.8476 & 0.0282 & 0.0265 & 0.0977 \tabularnewline
FCN \ding{170} & 0.8657 & 0.0100 & 0.0163 & 0.1079 \tabularnewline
FCN \ding{169} & 0.8697 & 0.0061 & 0.0221 & 0.1022 \tabularnewline
GCNet \ding{171} & 0.8104 & 0.0654 & 0.0196 & 0.1047 \tabularnewline
GCNet \ding{170} & 0.8635 & 0.0123 & 0.0152 & 0.1090 \tabularnewline
GCNet \ding{169} & 0.8715 & 0.0043 & 0.0253 & 0.0989 \tabularnewline
ICNet \ding{171} & 0.8586 & 0.0172 & 0.0107 & 0.1135 \tabularnewline
ISANet \ding{171} & 0.8304 & 0.0454 & 0.0204 & 0.1038 \tabularnewline
ISANet \ding{170} & 0.8569 & 0.0189 & 0.0120 & 0.1123 \tabularnewline
ISANet \ding{169} & 0.8720 & 0.0038 & 0.0256 & 0.0987 \tabularnewline
K-Net \ding{170} & 0.8630 & 0.0127 & 0.0098 & 0.1144 \tabularnewline
Mask2Former \ding{171} & 0.8621 & 0.0137 & 0.0056 & 0.1186 \tabularnewline
Mask2Former \ding{170} & 0.8661 & 0.0097 & 0.0093 & 0.1149 \tabularnewline
MaskFormer \ding{170} & 0.8624 & 0.0133 & 0.0103 & 0.1139 \tabularnewline
NonLocal Net \ding{171} & 0.8269 & 0.0489 & 0.0176 & 0.1067 \tabularnewline
NonLocal Net \ding{170} & 0.8649 & 0.0109 & 0.0150 & 0.1092 \tabularnewline
NonLocal Net \ding{169} & 0.8716 & 0.0042 & 0.0250 & 0.0992 \tabularnewline
OCRNet \ding{171} & 0.8523 & 0.0235 & 0.0110 & 0.1132 \tabularnewline
OCRNet \ding{170} & 0.8635 & 0.0123 & 0.0130 & 0.1112 \tabularnewline
OCRNet \ding{169} & 0.8707 & 0.0051 & 0.0243 & 0.0999 \tabularnewline
PointRend \ding{171} & 0.8422 & 0.0335 & 0.0328 & 0.0914 \tabularnewline
PointRend \ding{170} & 0.8611 & 0.0147 & 0.0142 & 0.1101 \tabularnewline
PSANet \ding{171} & 0.8268 & 0.0489 & 0.0155 & 0.1088 \tabularnewline
PSANet \ding{170} & 0.8634 & 0.0123 & 0.0155 & 0.1087 \tabularnewline
PSPNet \ding{171} & 0.8528 & 0.0230 & 0.0273 & 0.0970 \tabularnewline
PSPNet \ding{170} & 0.8629 & 0.0129 & 0.0142 & 0.1100 \tabularnewline
PSPNet \ding{169} & 0.8710 & 0.0048 & 0.0231 & 0.1011 \tabularnewline
PSPNet \ding{168} & 0.8584 & 0.0174 & 0.0078 & 0.1165 \tabularnewline
SAN \ding{168} & 0.8544 & 0.0214 & 0.0088 & 0.1155 \tabularnewline
SegFormer \ding{171} & 0.8670 & 0.0088 & 0.0107 & 0.1135 \tabularnewline
SegFormer \ding{170} & 0.8637 & 0.0121 & 0.0127 & 0.1115 \tabularnewline
Segmenter \ding{170} & 0.8604 & 0.0154 & 0.0092 & 0.1150 \tabularnewline
Semantic FPN \ding{171} & 0.8600 & 0.0158 & 0.0404 & 0.0838 \tabularnewline
Semantic FPN \ding{170} & 0.8225 & 0.0533 & 0.0391 & 0.0851 \tabularnewline
SETR \ding{171} & 0.8663 & 0.0095 & 0.0074 & 0.1168 \tabularnewline
SETR \ding{170} & 0.8602 & 0.0155 & 0.0115 & 0.1128 \tabularnewline
STDC \ding{171} & 0.8411 & 0.0347 & 0.0164 & 0.1078 \tabularnewline
UPerNet \ding{171} & 0.8344 & 0.0413 & 0.0146 & 0.1096 \tabularnewline
UPerNet \ding{170} & 0.8619 & 0.0138 & 0.0139 & 0.1103 \tabularnewline
UPerNet \ding{169} & 0.8718 & 0.0039 & 0.0259 & 0.0983 \tabularnewline
\hline 
\end{longtable}
\begin{center}
\includegraphics[scale=0.8]{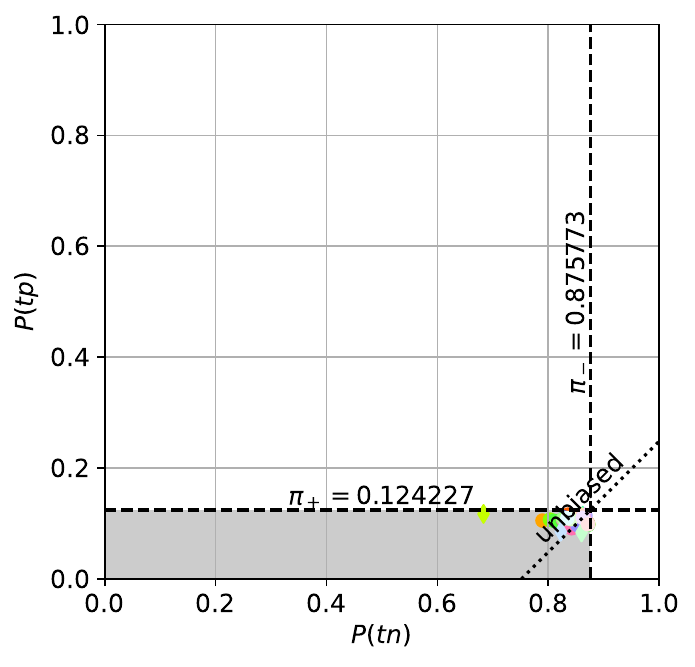}
\includegraphics[scale=0.8]{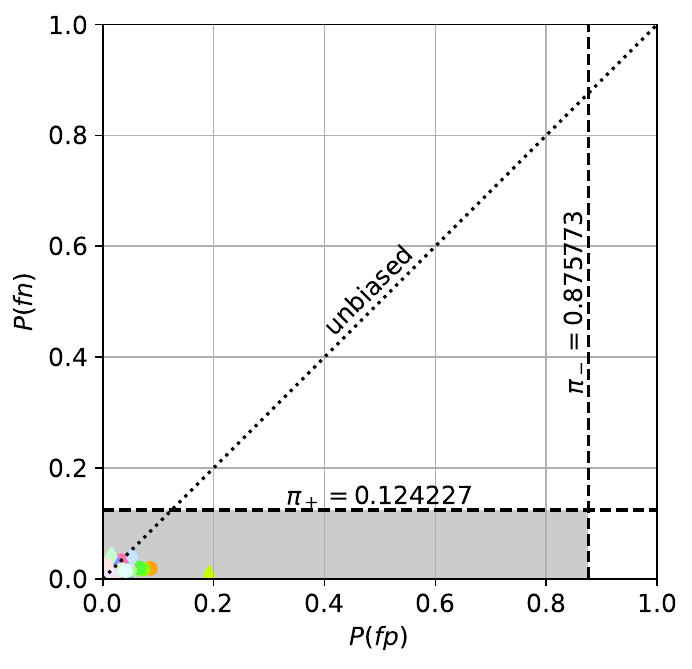}
\includegraphics[scale=0.8]{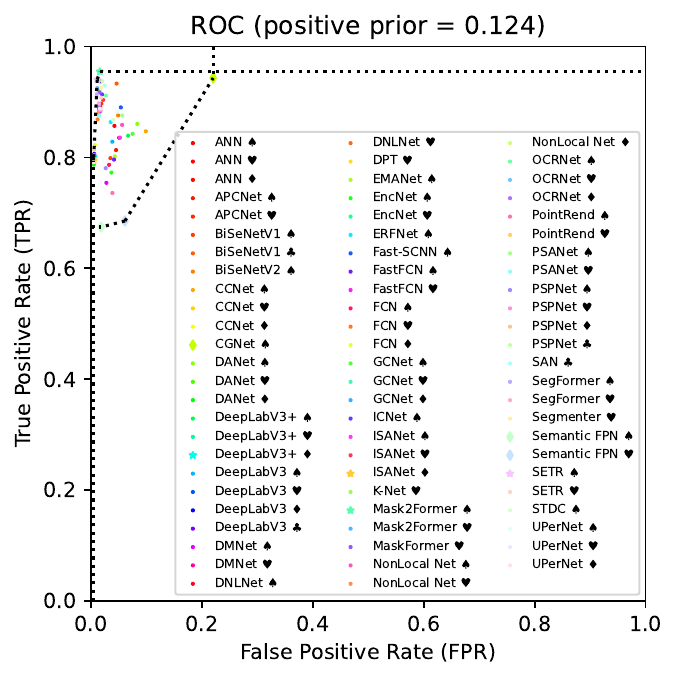}
\includegraphics[scale=0.8]{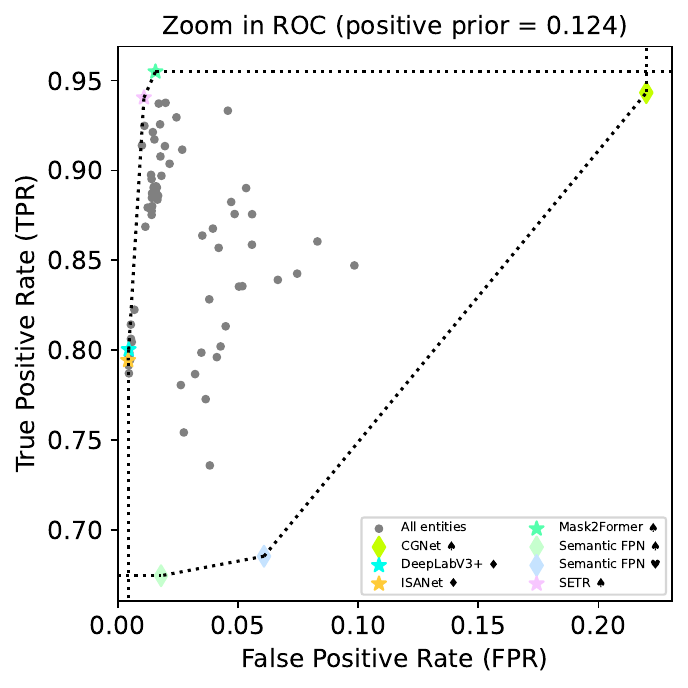}
\par\end{center}

\subsection*{Value Tile: Using the tile to show the canonical ranking scores values for each entity}
\begin{center}
\includegraphics[scale=0.4]{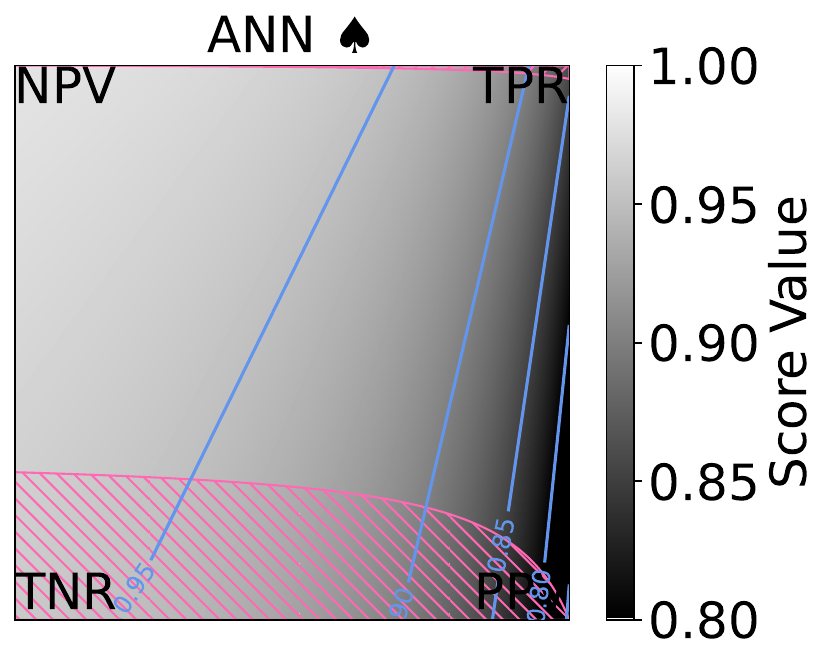}
\includegraphics[scale=0.4]{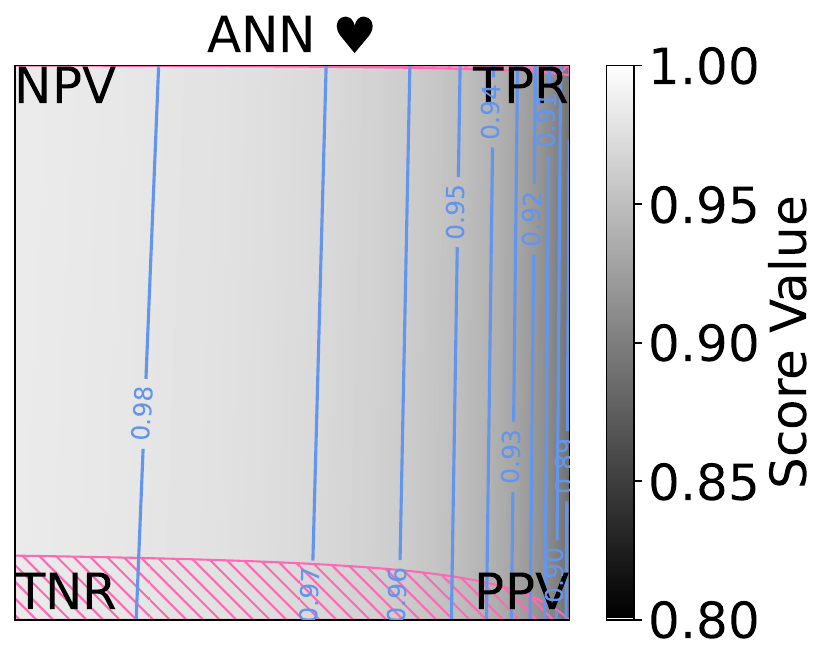}
\includegraphics[scale=0.4]{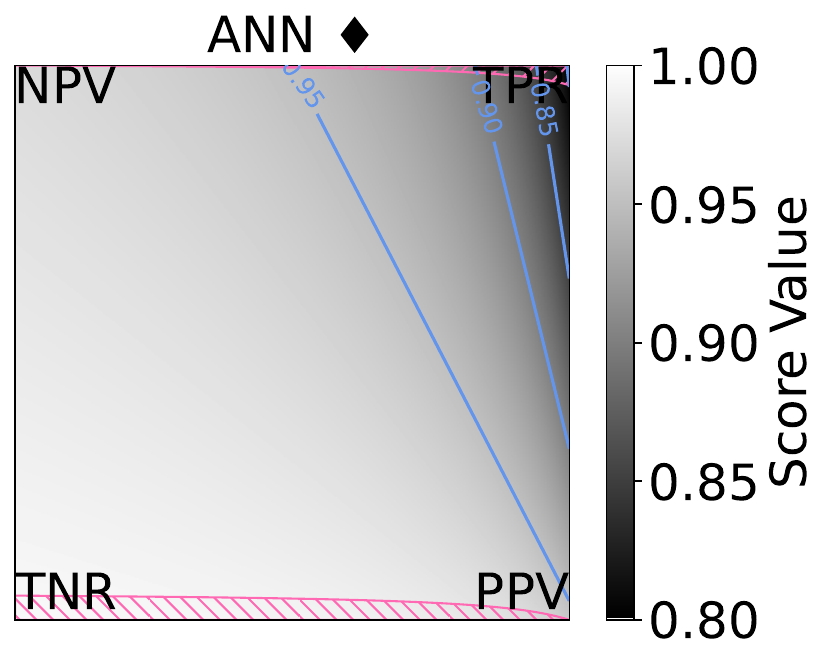}
\includegraphics[scale=0.4]{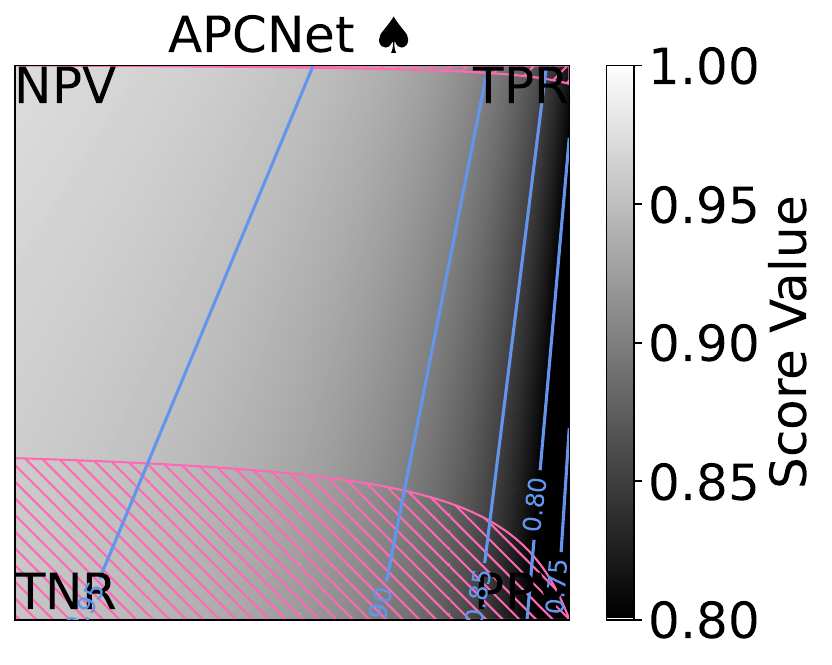}
\includegraphics[scale=0.4]{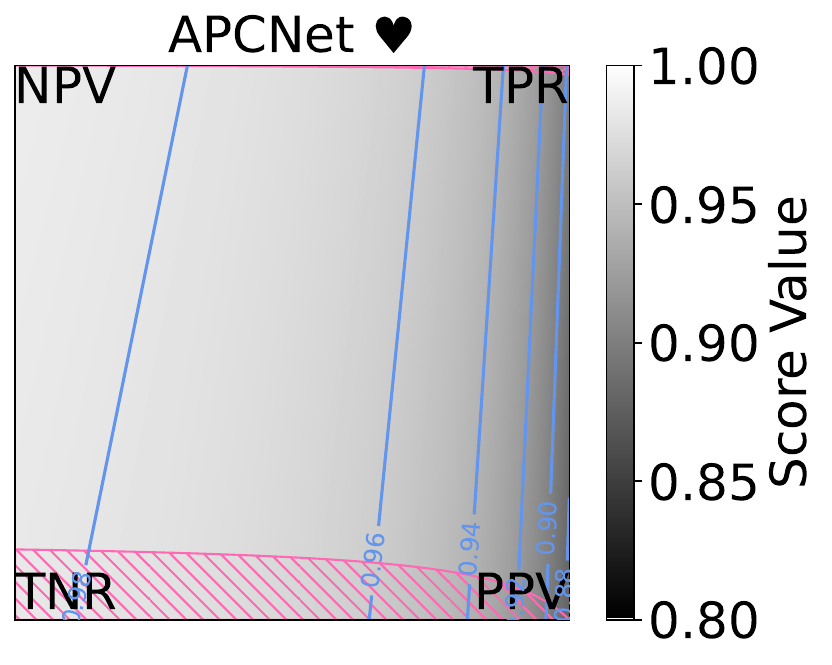}
\includegraphics[scale=0.4]{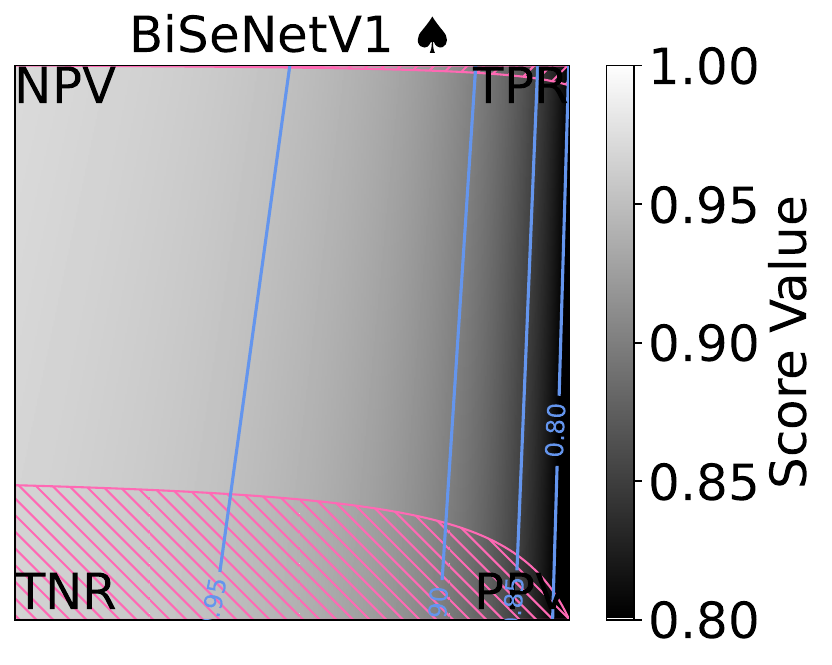}
\includegraphics[scale=0.4]{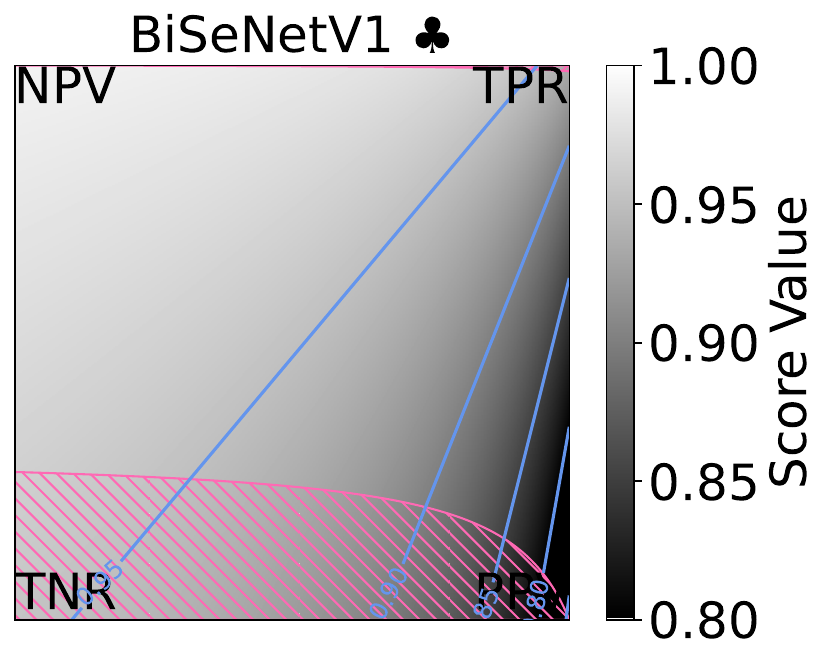}
\includegraphics[scale=0.4]{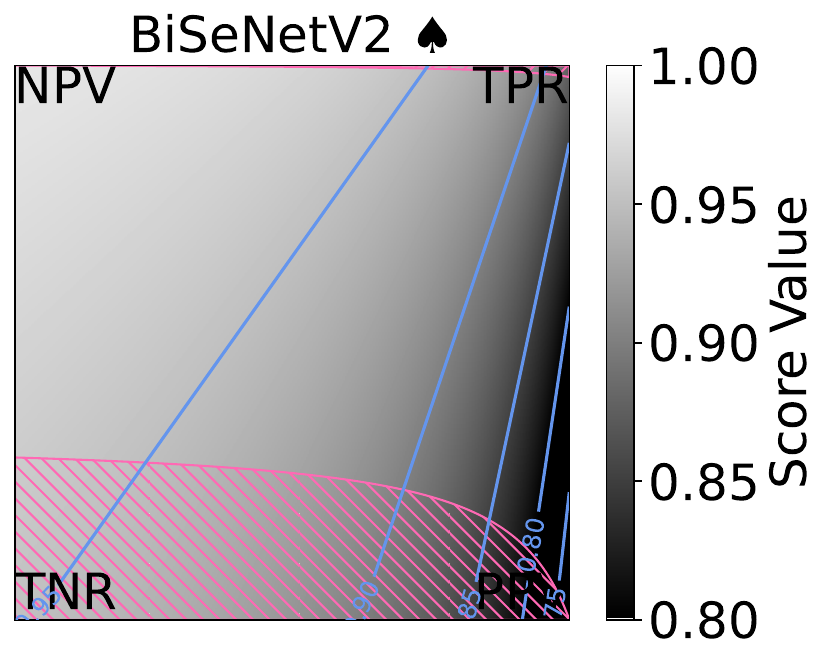}
\includegraphics[scale=0.4]{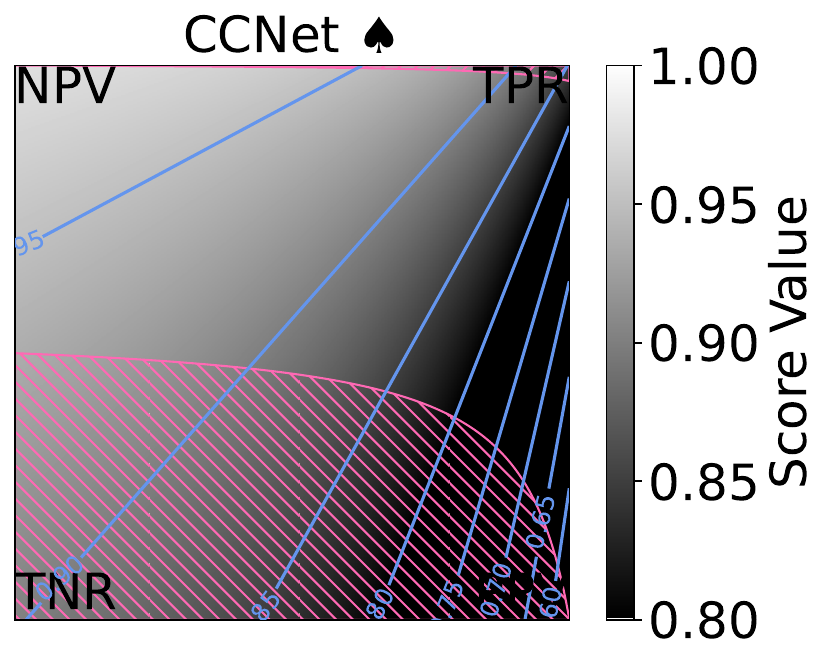}
\includegraphics[scale=0.4]{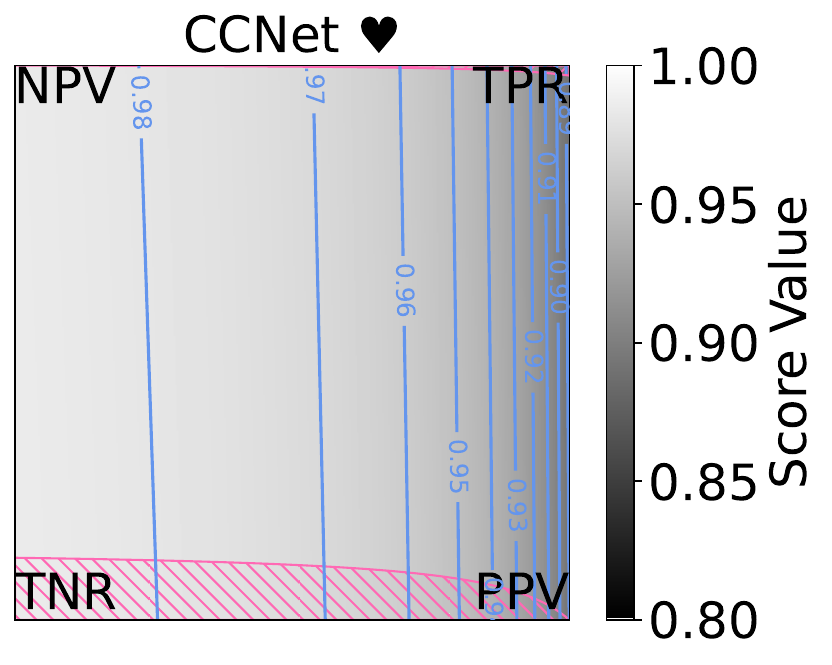}
\includegraphics[scale=0.4]{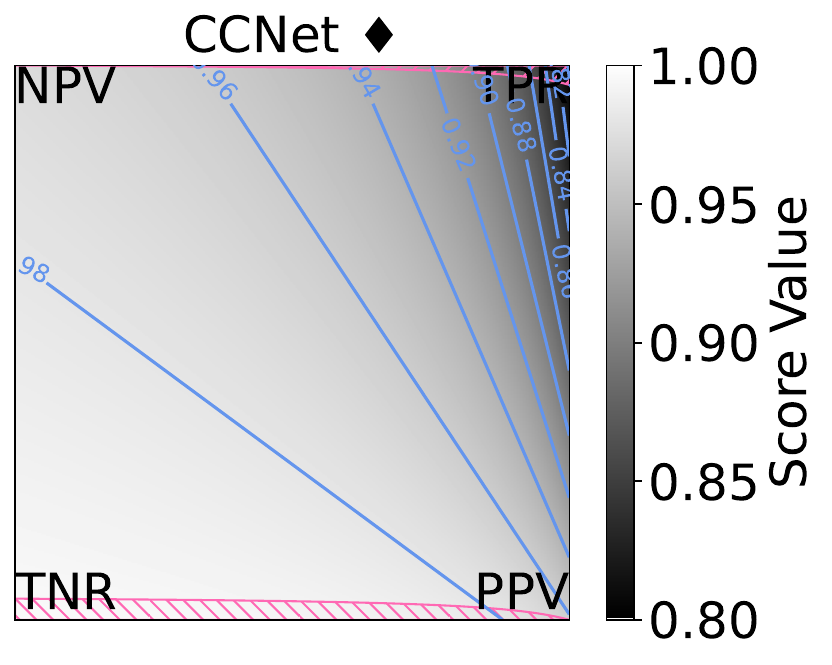}
\includegraphics[scale=0.4]{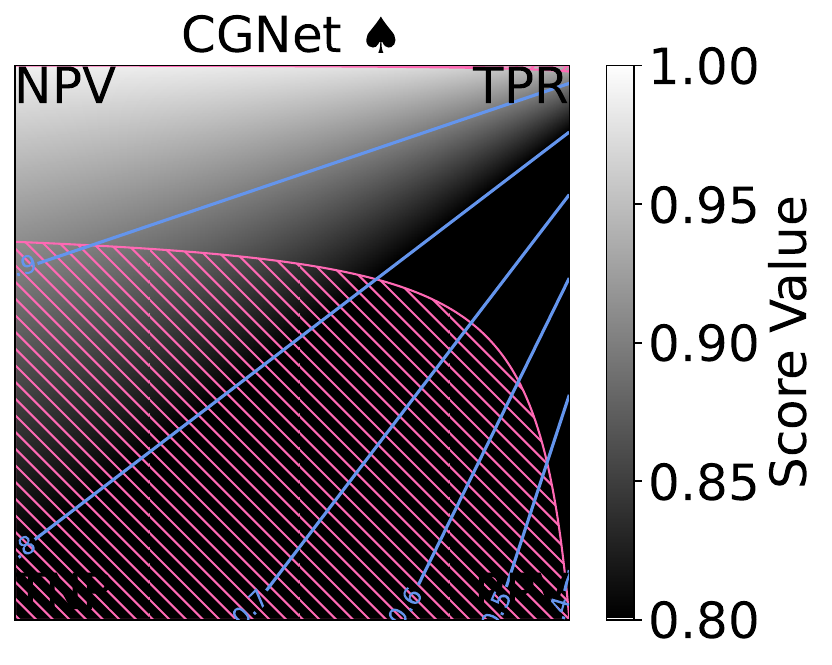}
\includegraphics[scale=0.4]{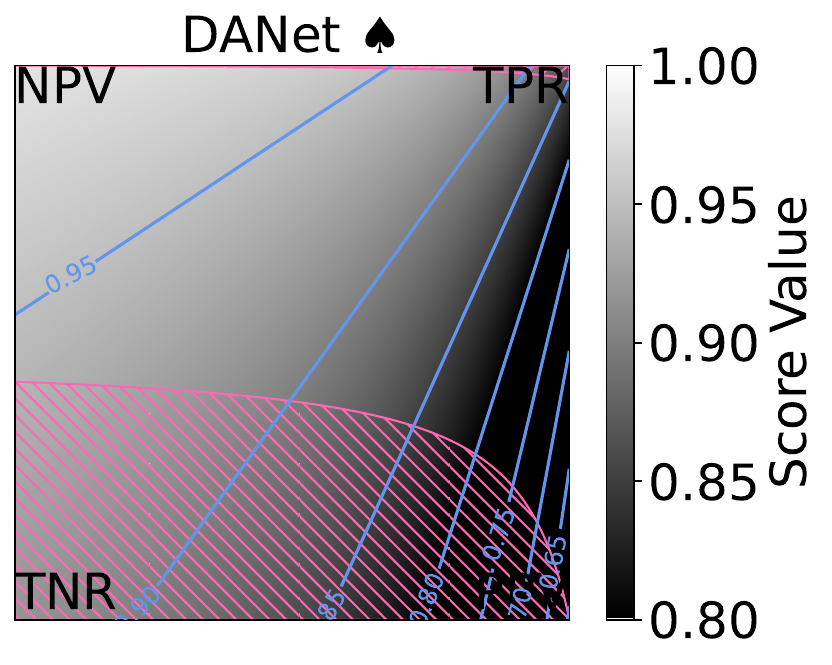}
\includegraphics[scale=0.4]{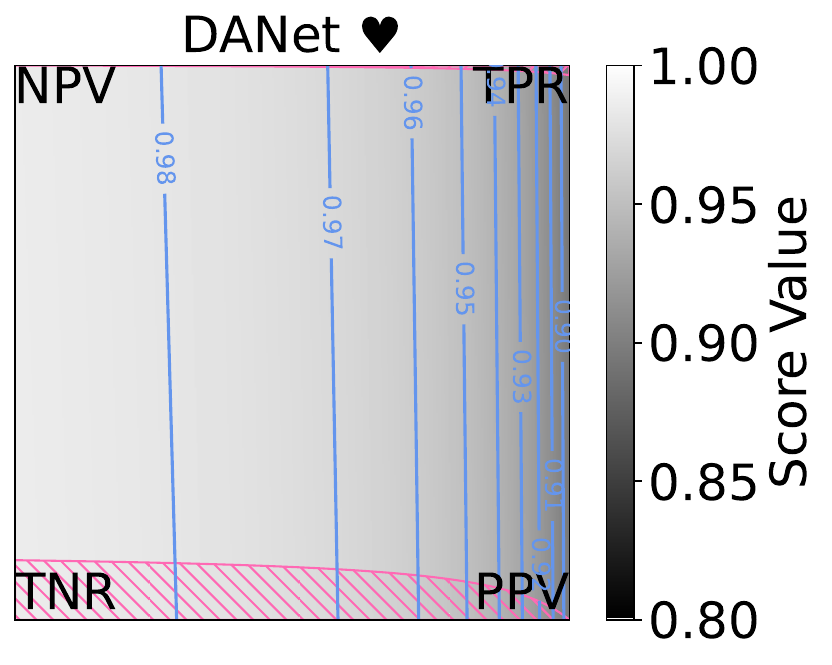}
\includegraphics[scale=0.4]{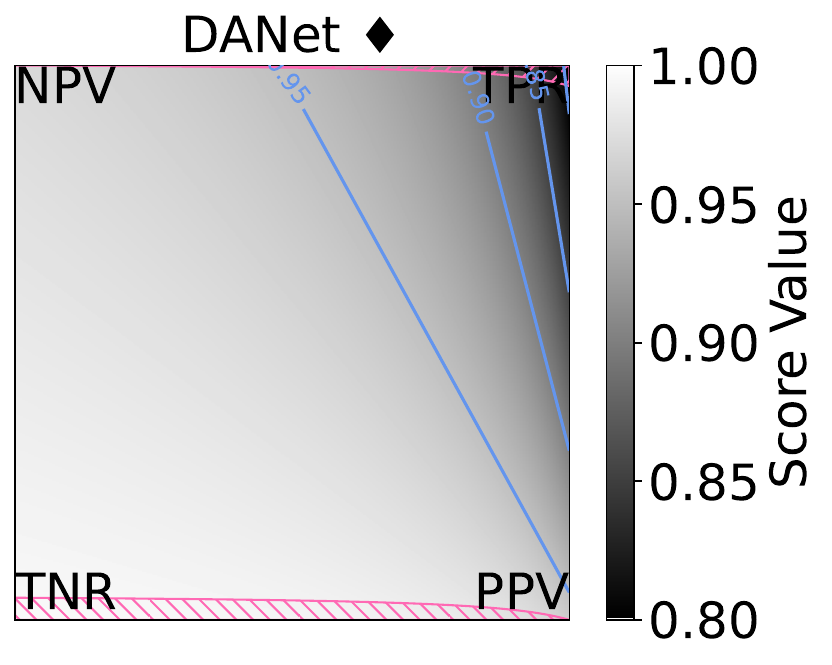}
\includegraphics[scale=0.4]{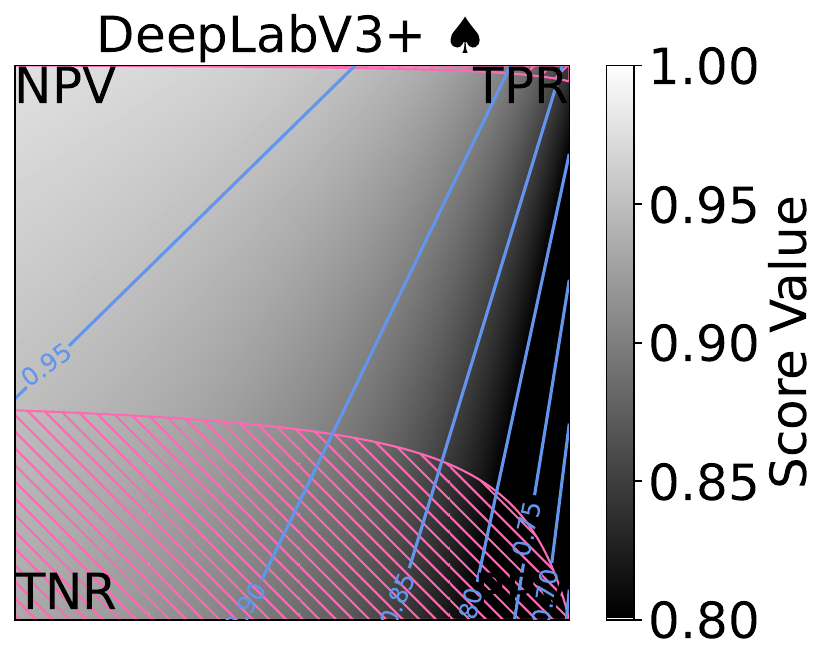}
\includegraphics[scale=0.4]{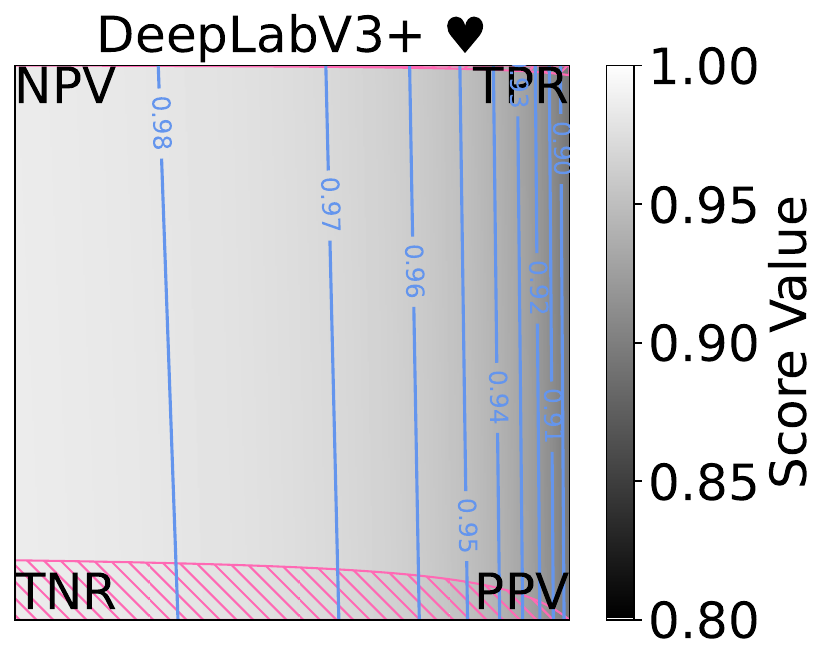}
\includegraphics[scale=0.4]{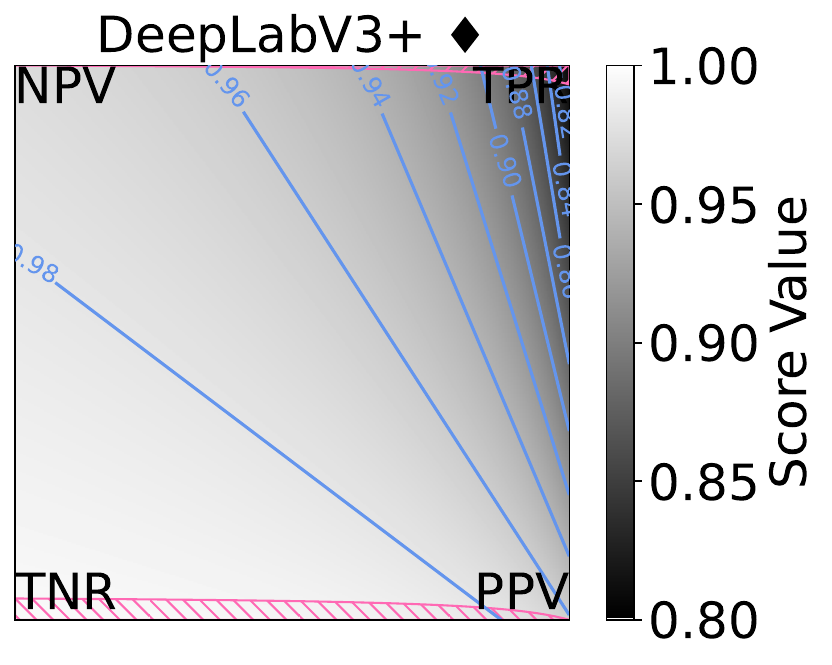}
\includegraphics[scale=0.4]{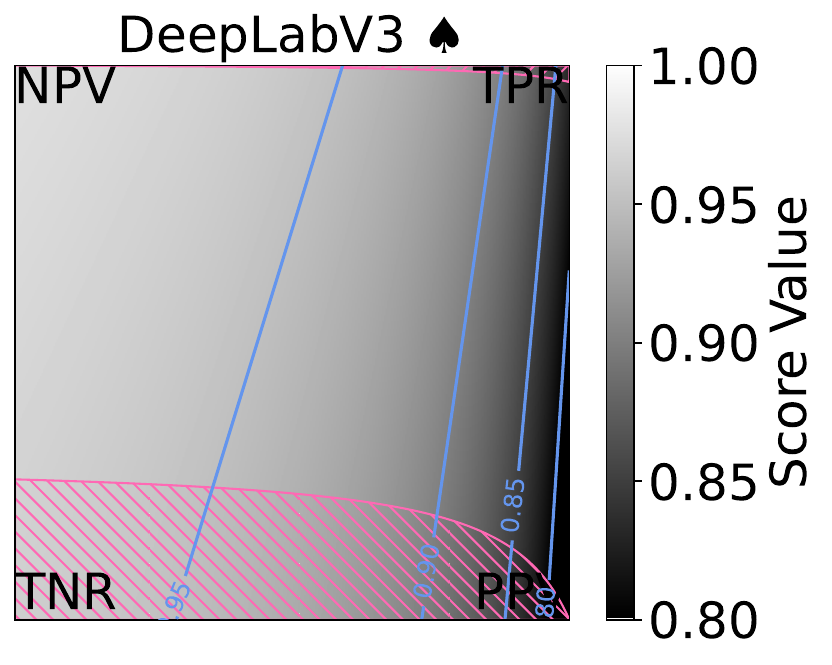}
\includegraphics[scale=0.4]{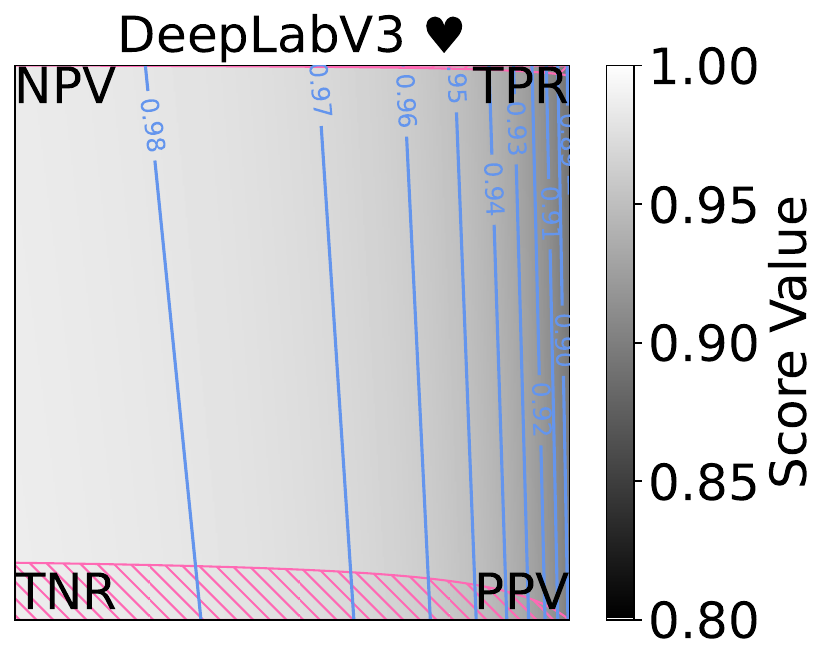}
\includegraphics[scale=0.4]{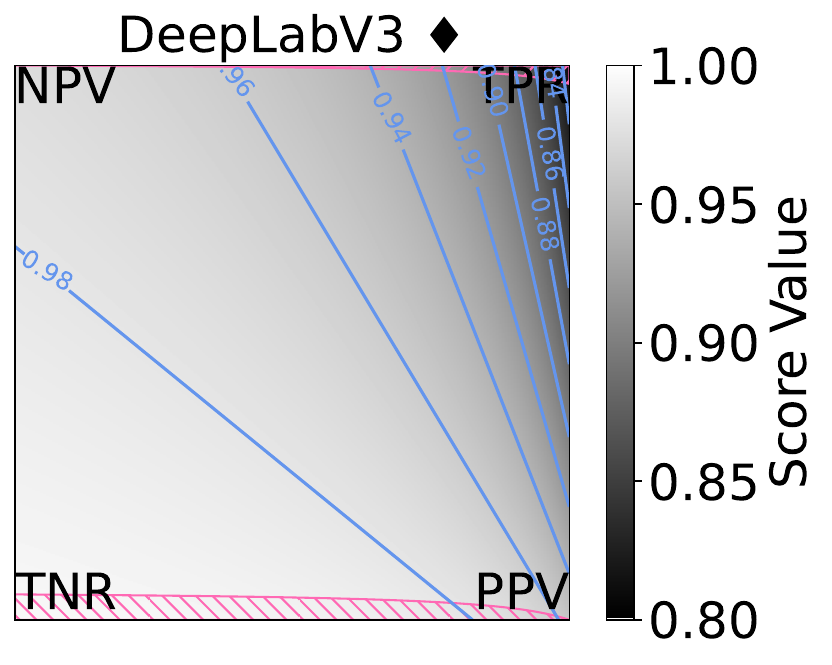}
\includegraphics[scale=0.4]{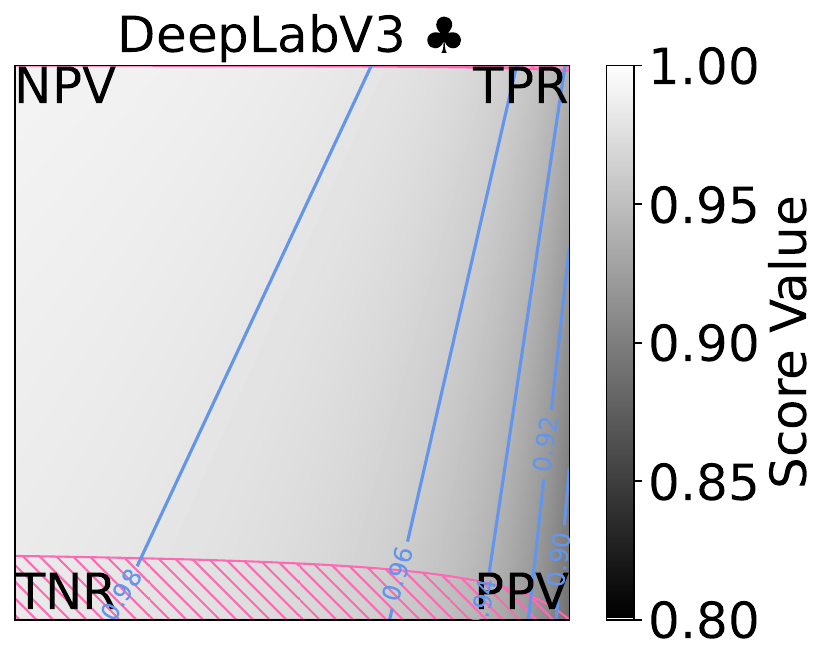}
\includegraphics[scale=0.4]{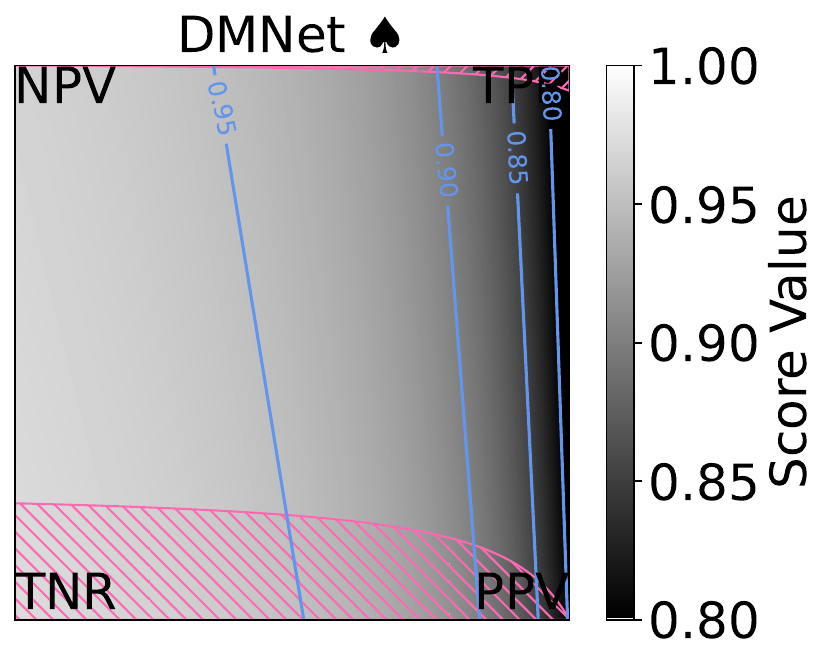}
\includegraphics[scale=0.4]{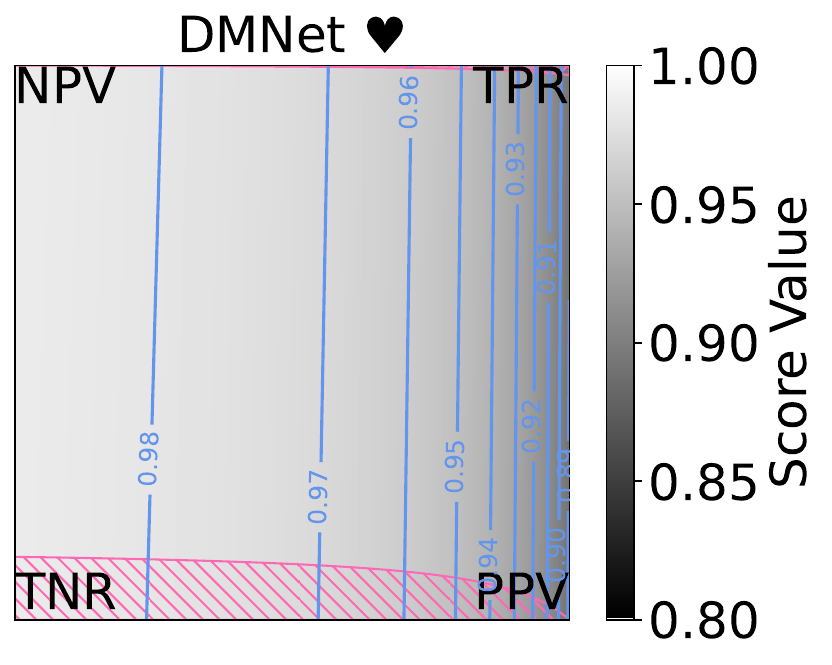}
\includegraphics[scale=0.4]{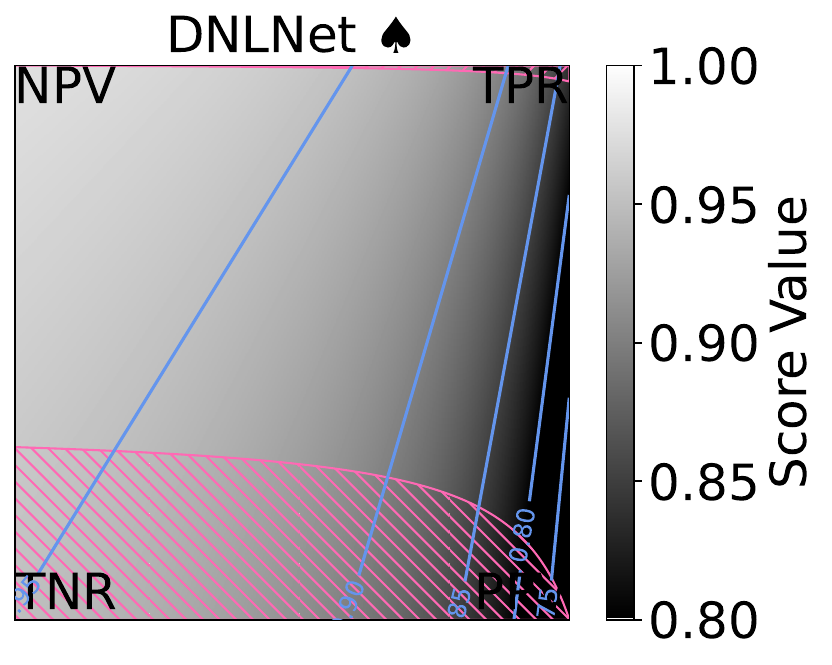}
\includegraphics[scale=0.4]{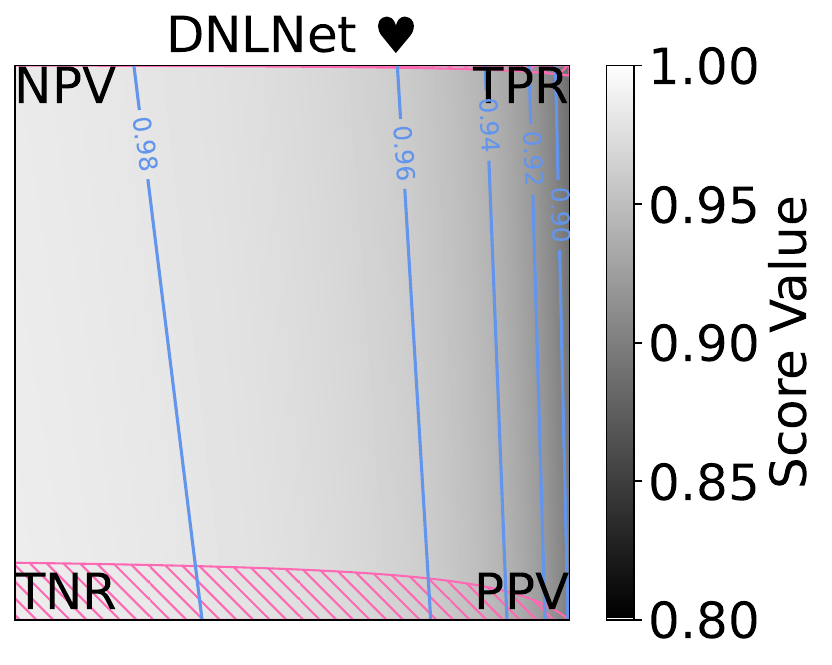}
\includegraphics[scale=0.4]{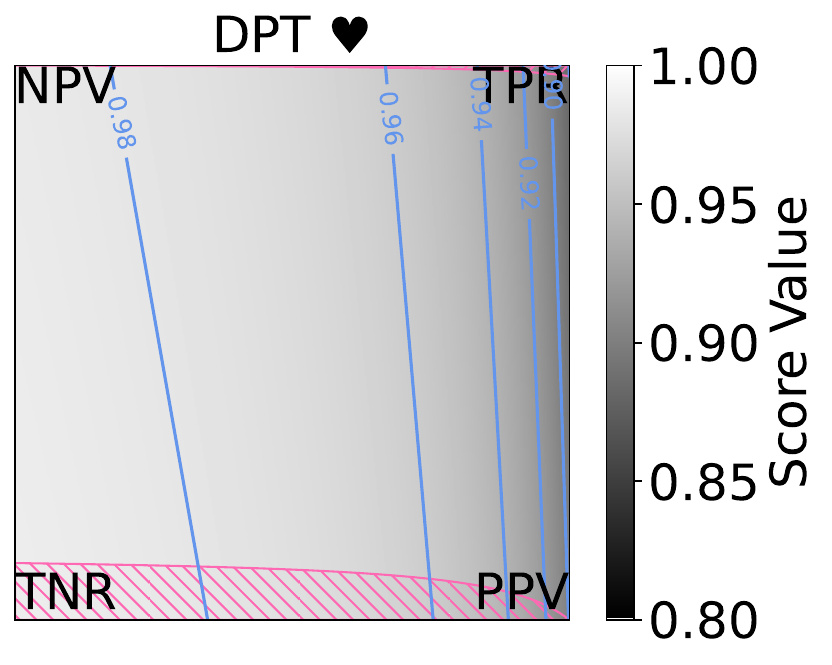}
\includegraphics[scale=0.4]{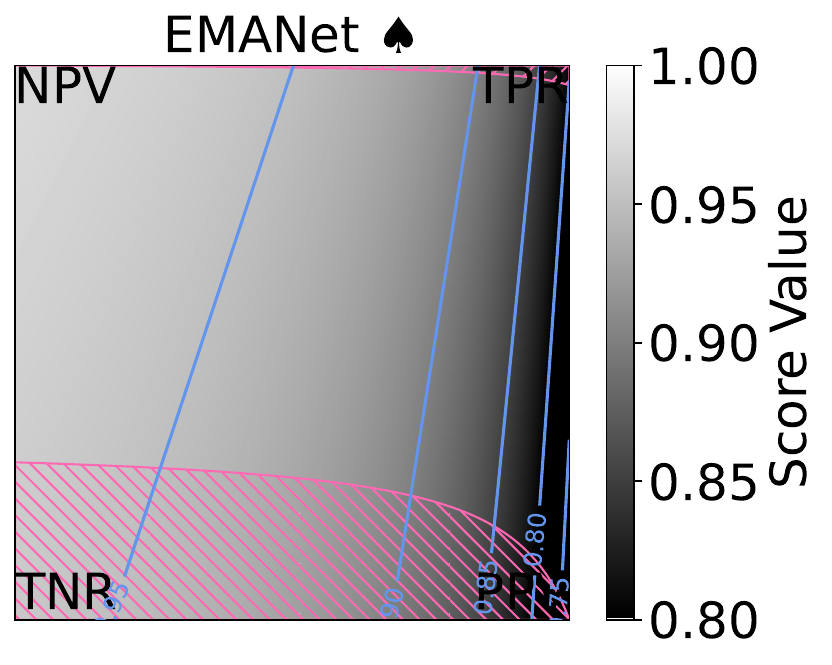}
\includegraphics[scale=0.4]{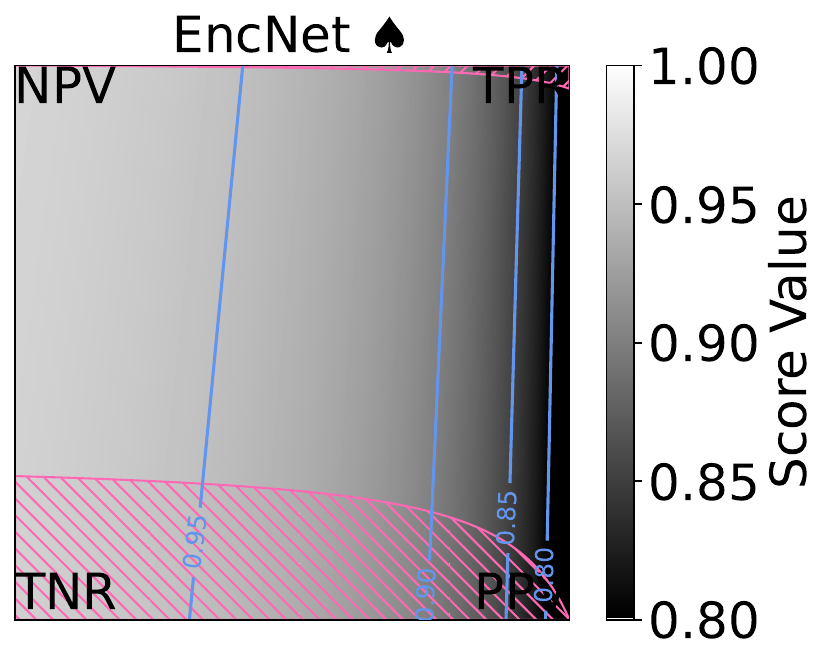}
\includegraphics[scale=0.4]{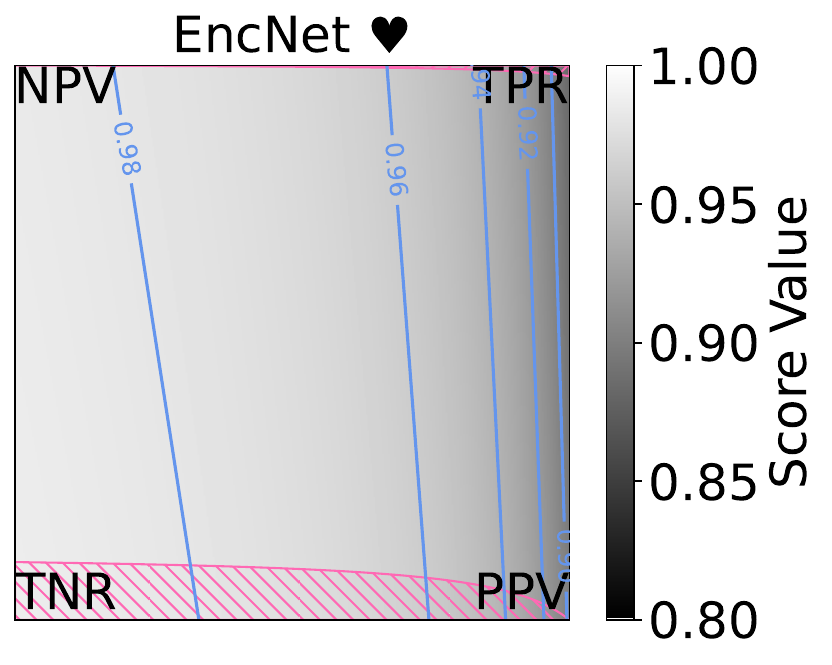}
\includegraphics[scale=0.4]{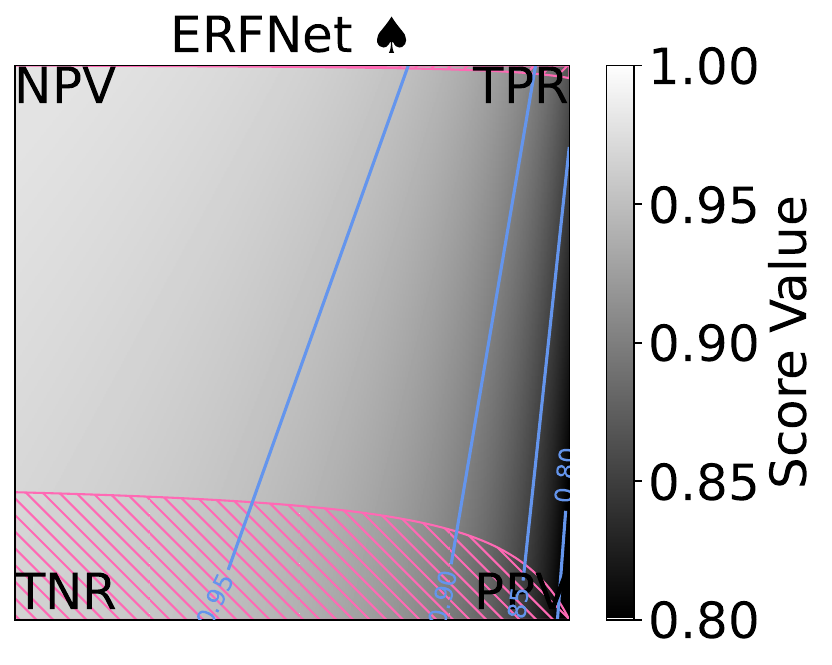}
\includegraphics[scale=0.4]{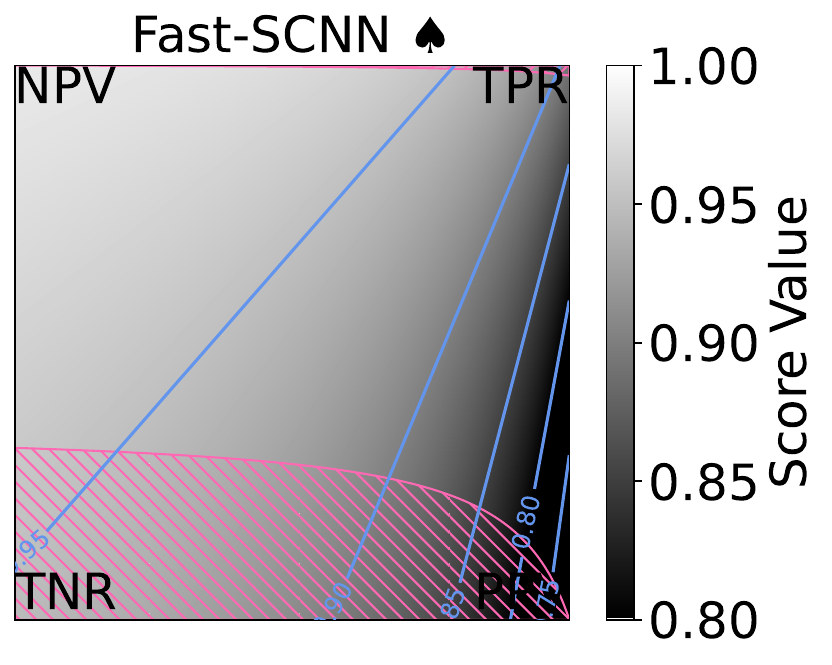}
\includegraphics[scale=0.4]{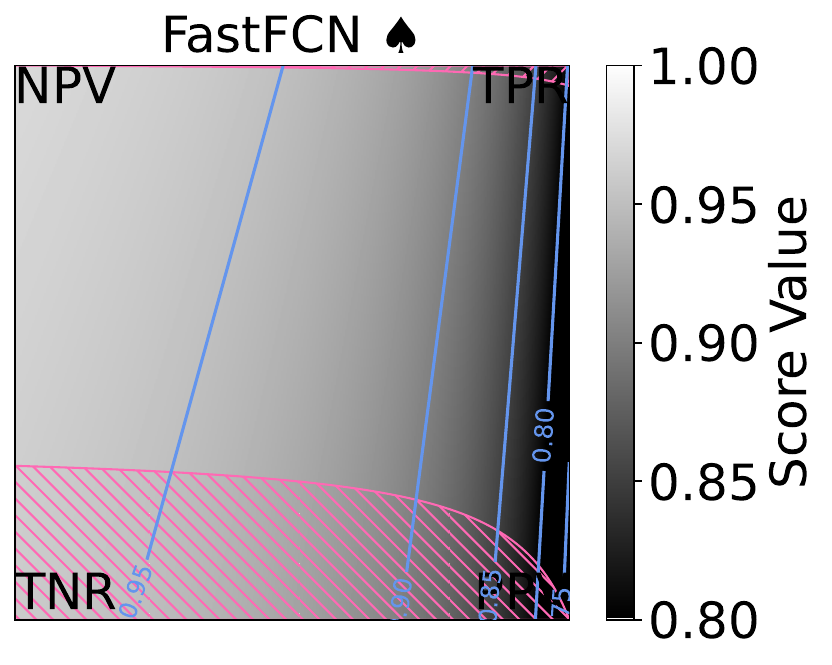}
\includegraphics[scale=0.4]{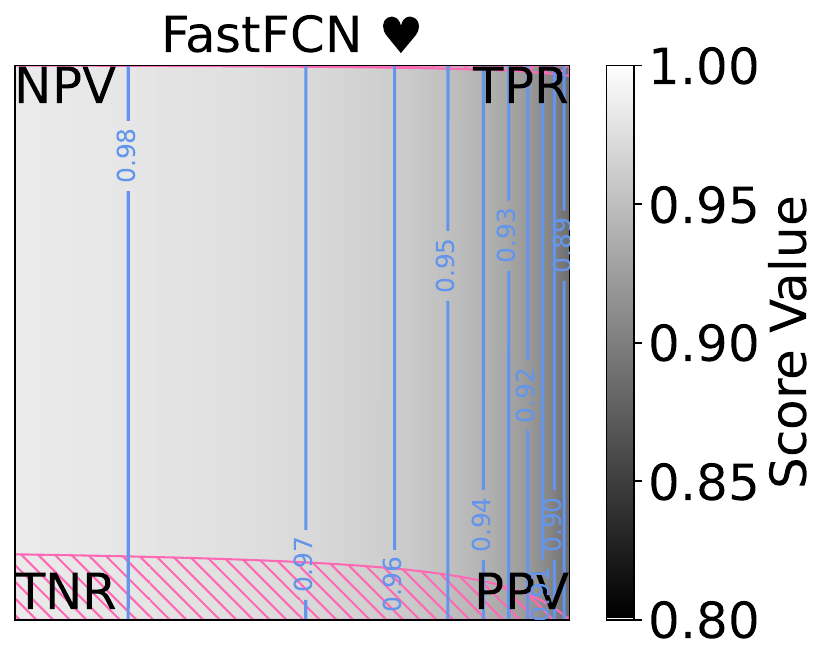}
\includegraphics[scale=0.4]{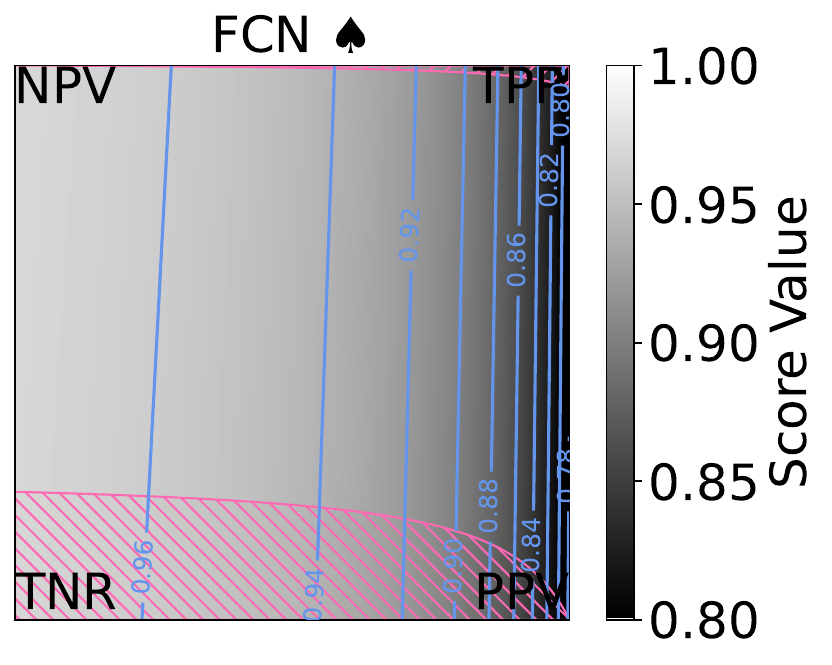}
\includegraphics[scale=0.4]{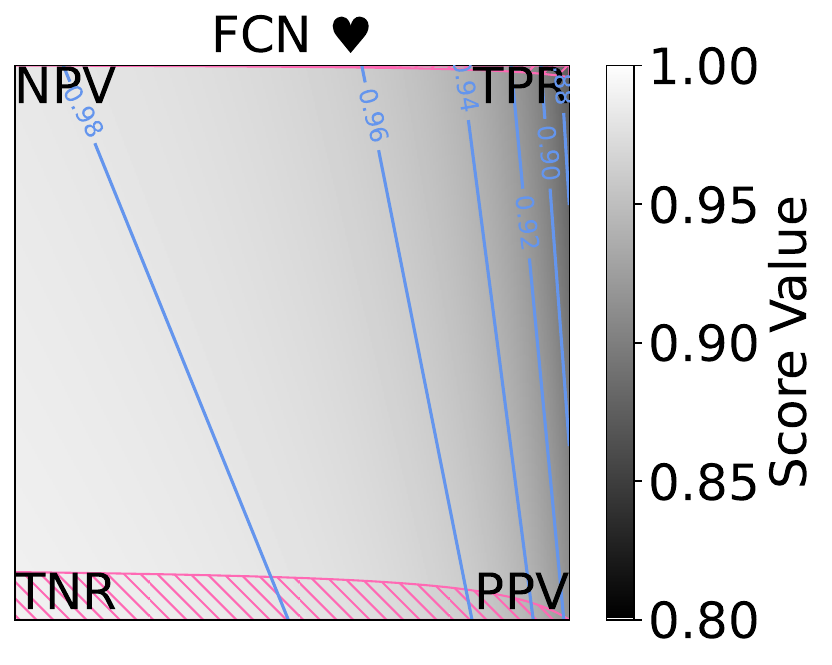}
\includegraphics[scale=0.4]{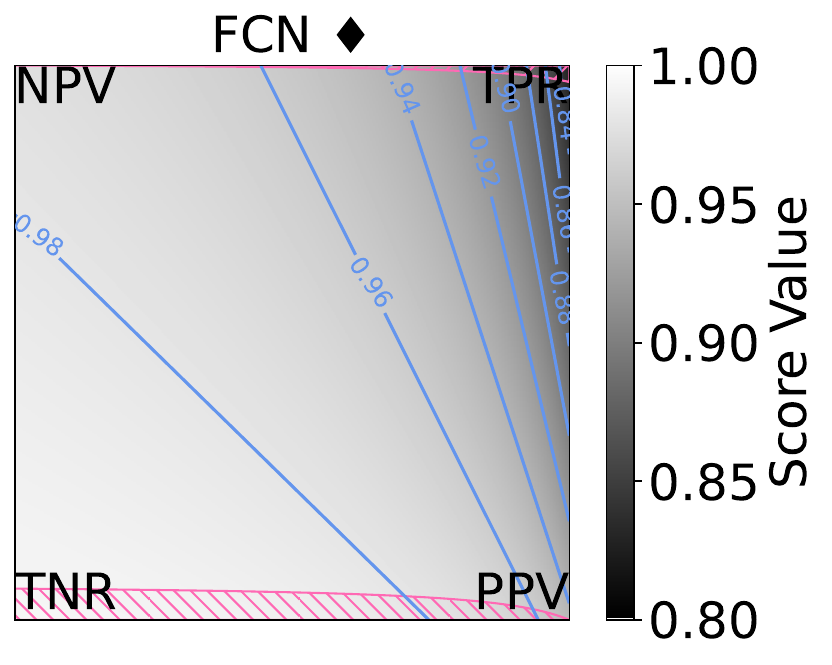}
\includegraphics[scale=0.4]{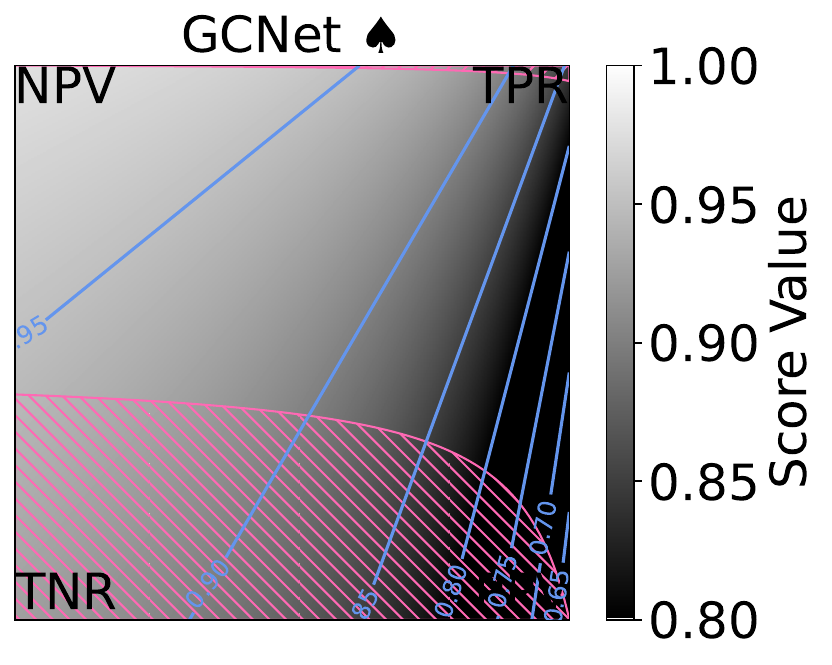}
\includegraphics[scale=0.4]{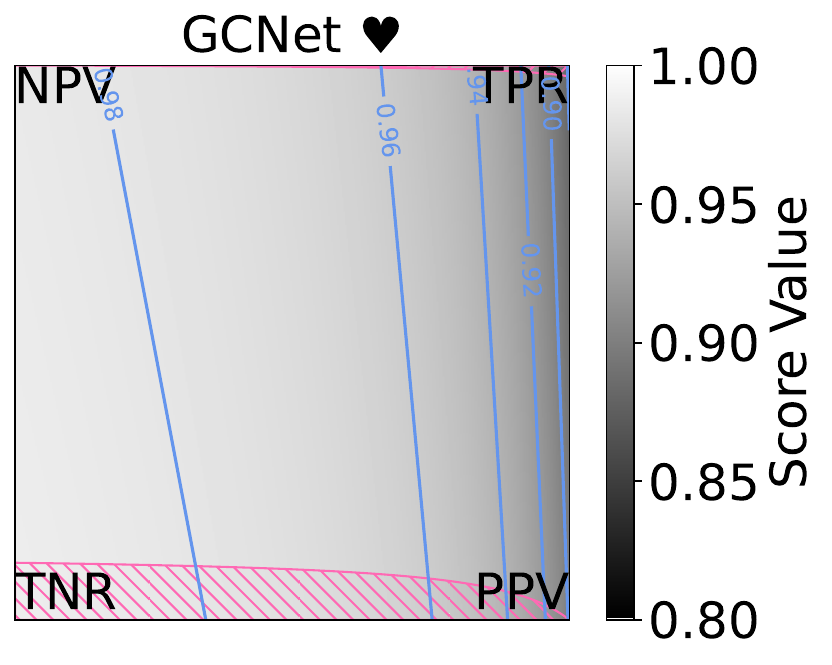}
\includegraphics[scale=0.4]{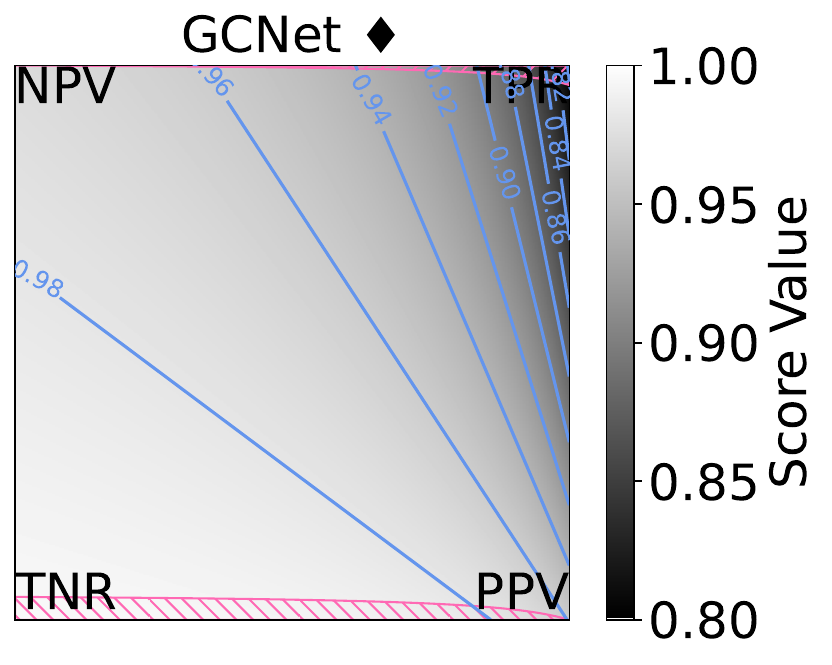}
\includegraphics[scale=0.4]{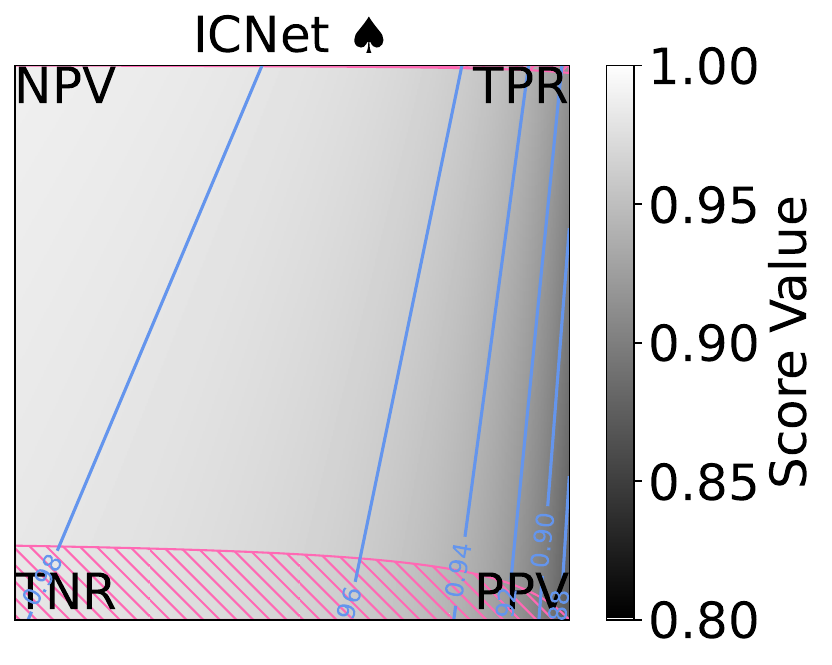}
\includegraphics[scale=0.4]{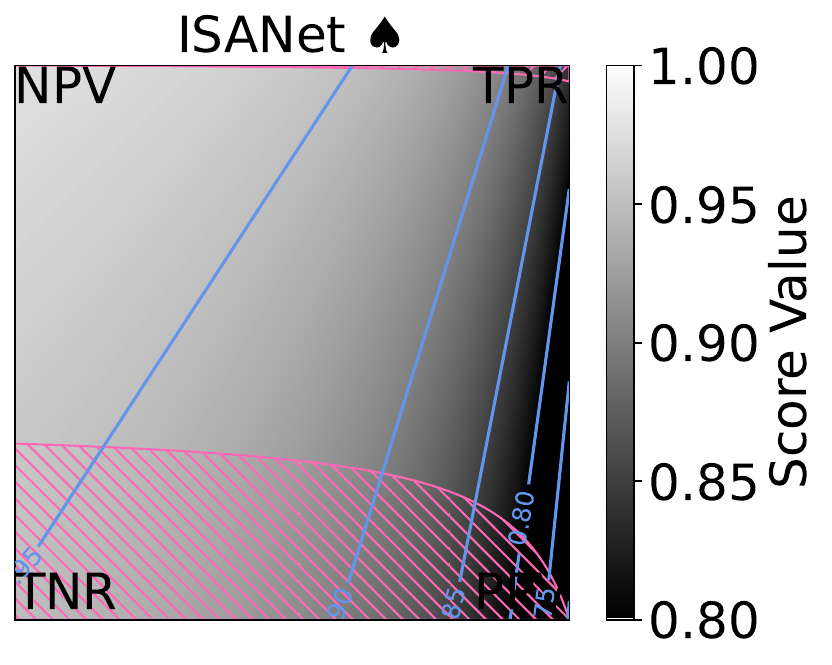}
\includegraphics[scale=0.4]{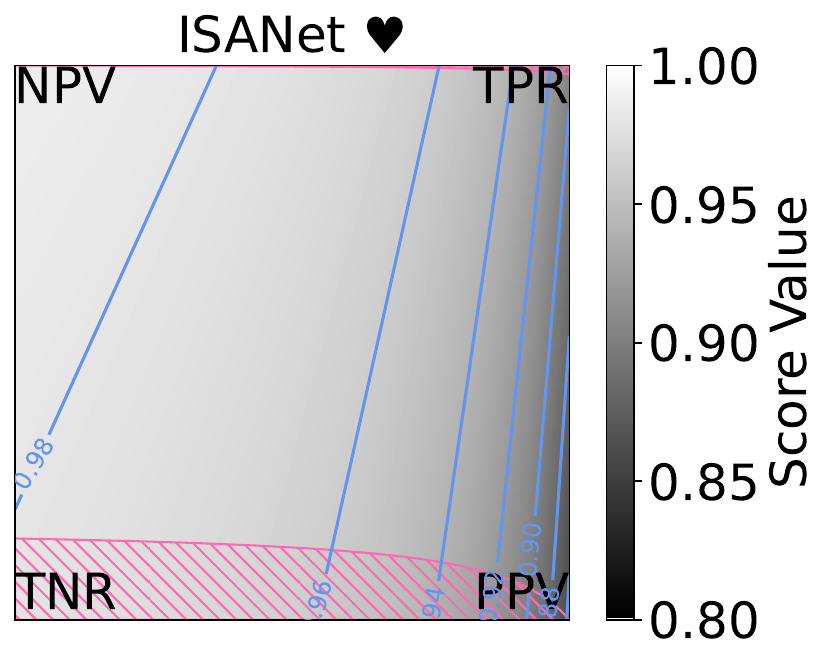}
\includegraphics[scale=0.4]{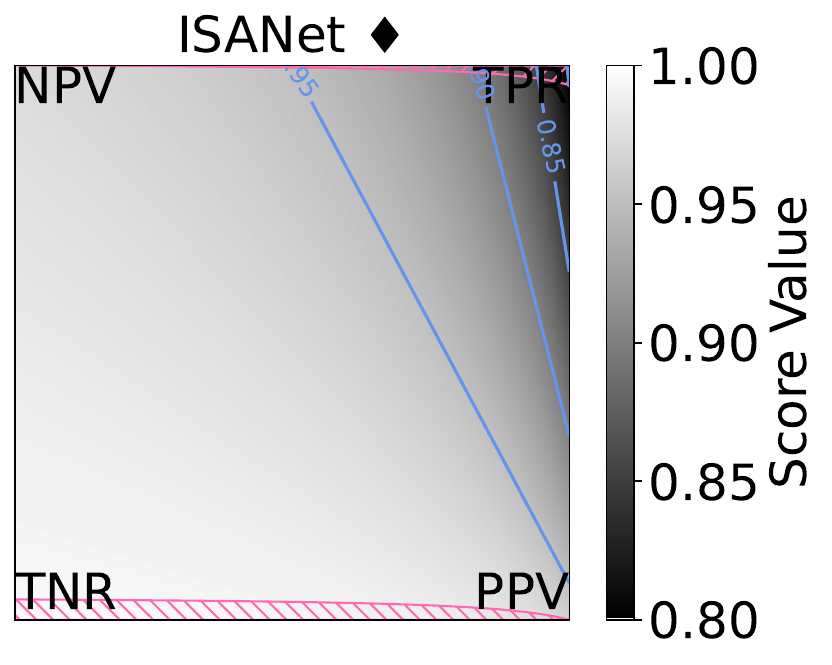}
\includegraphics[scale=0.4]{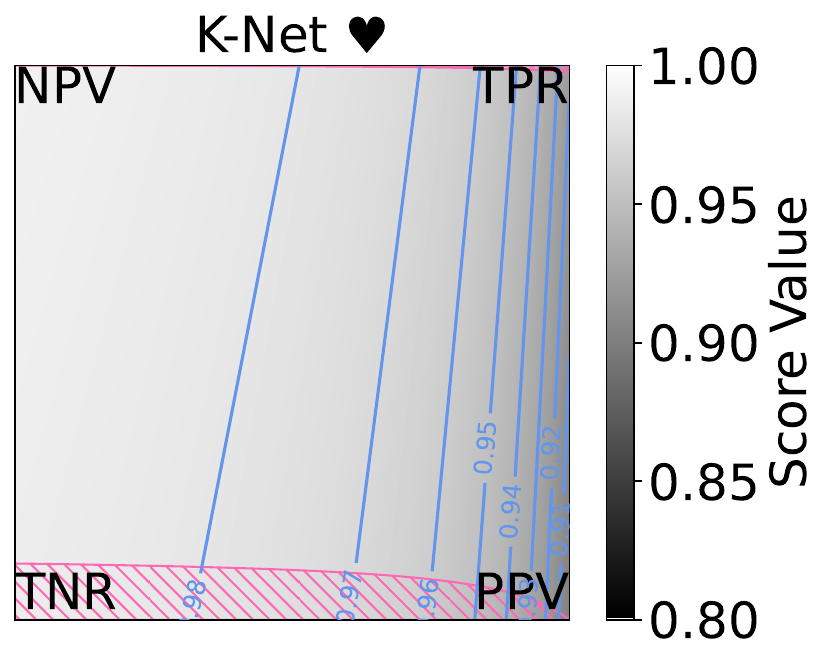}
\includegraphics[scale=0.4]{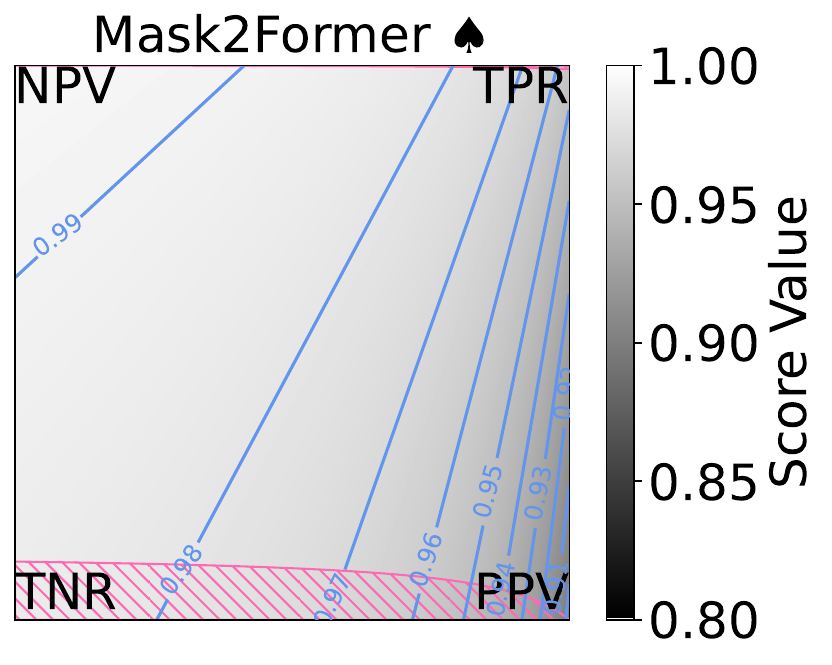}
\includegraphics[scale=0.4]{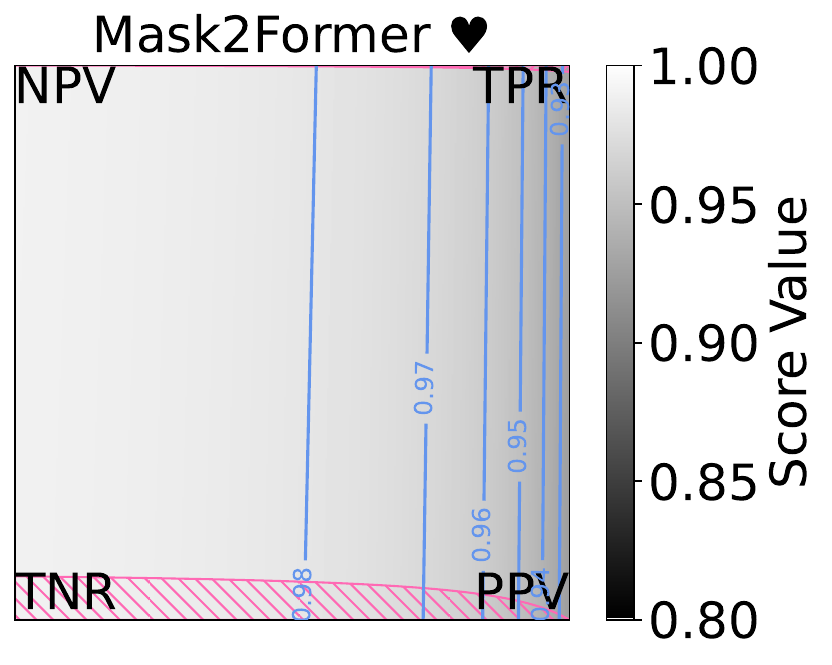}
\includegraphics[scale=0.4]{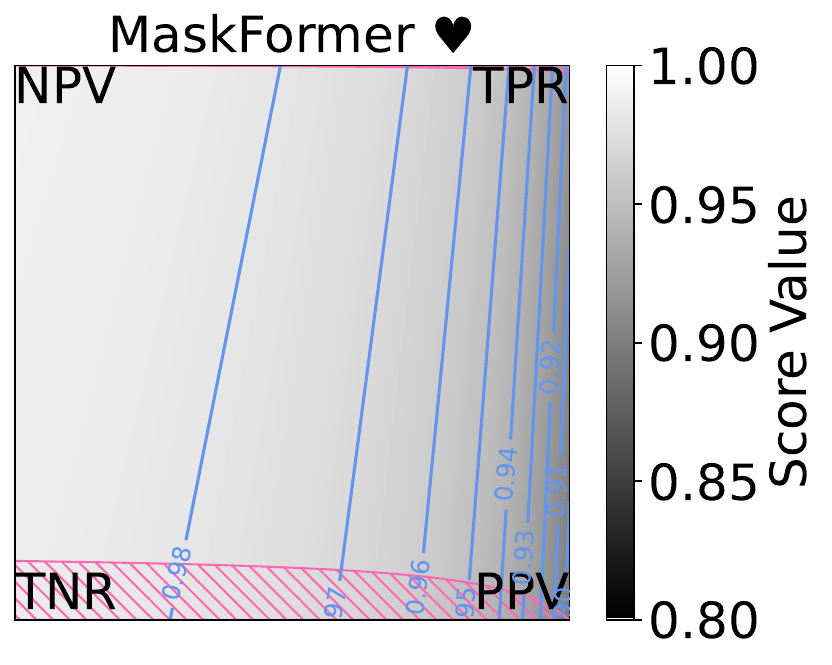}
\includegraphics[scale=0.4]{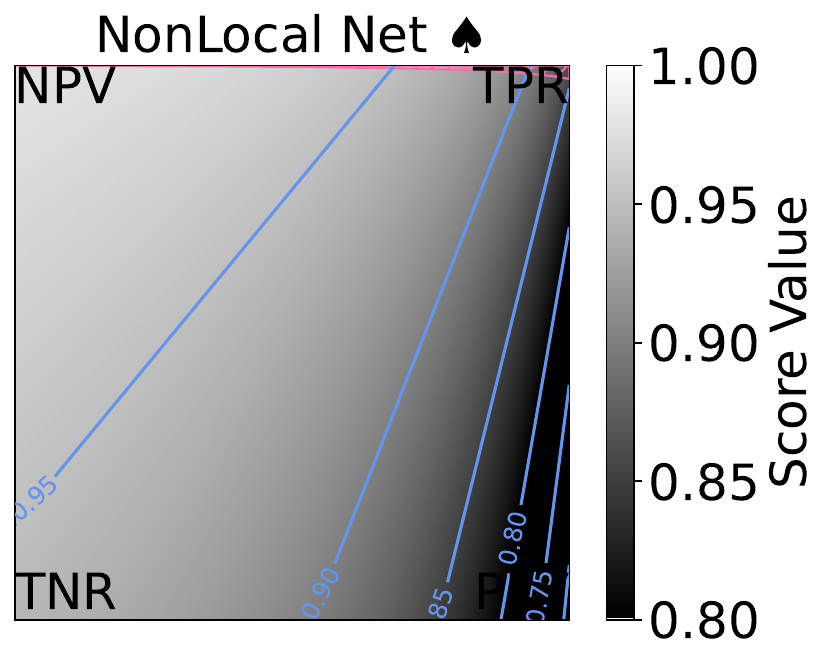}
\includegraphics[scale=0.4]{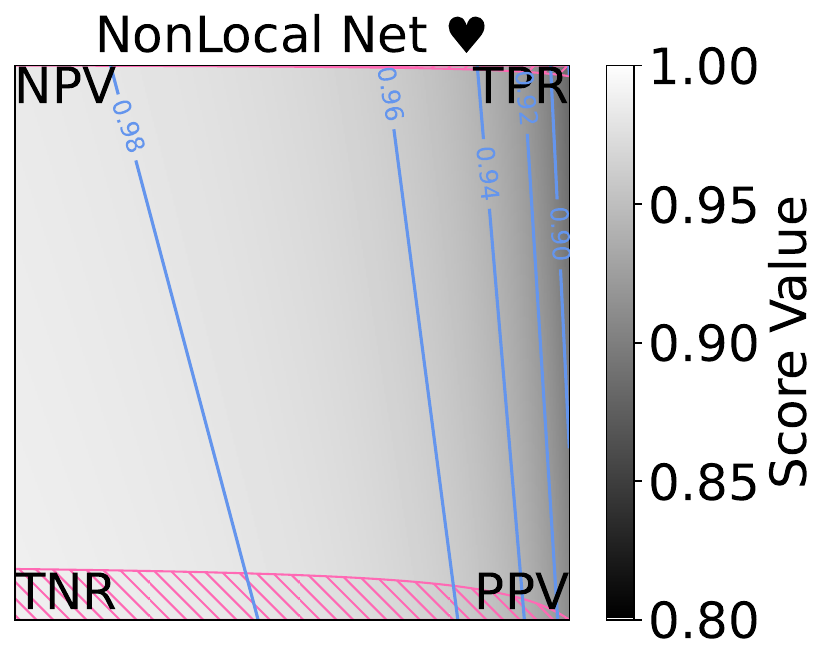}
\includegraphics[scale=0.4]{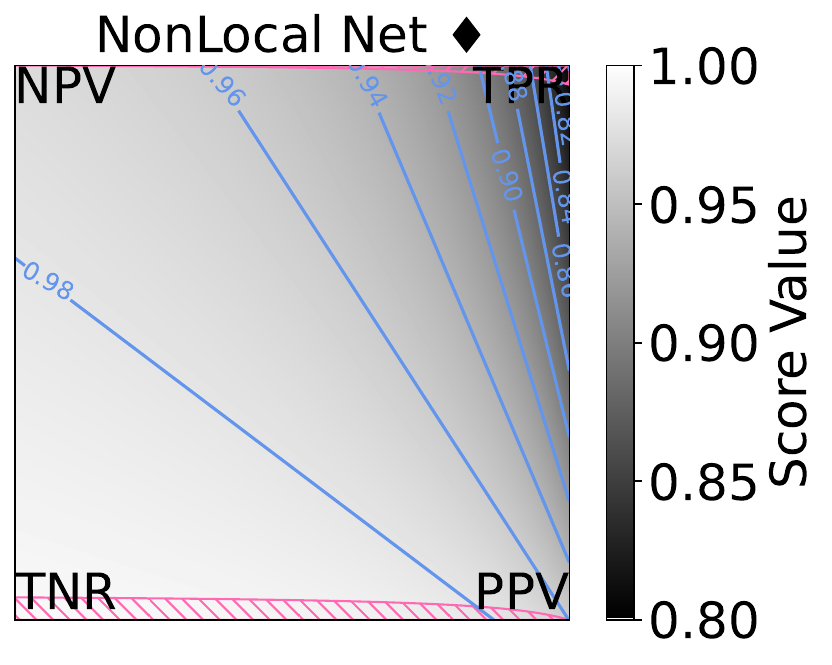}
\includegraphics[scale=0.4]{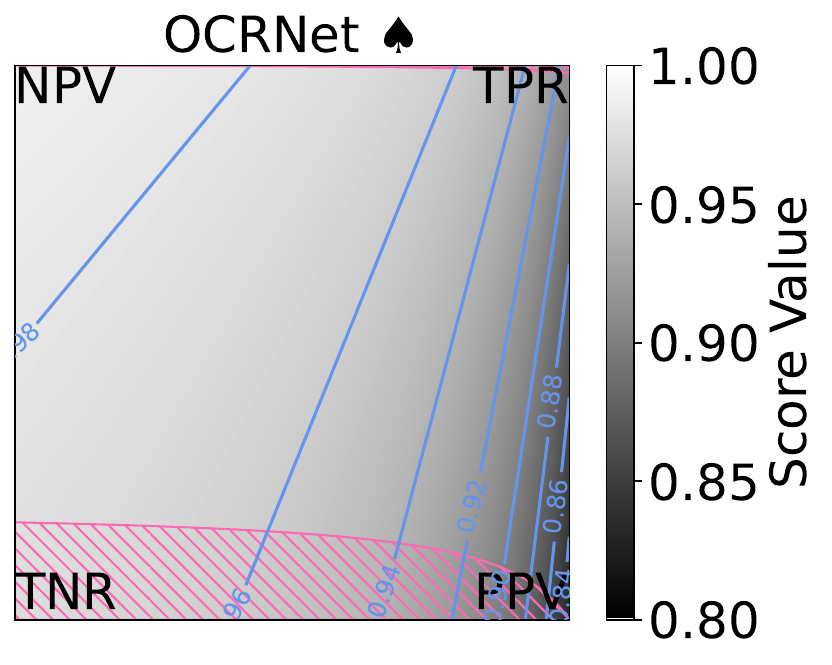}
\includegraphics[scale=0.4]{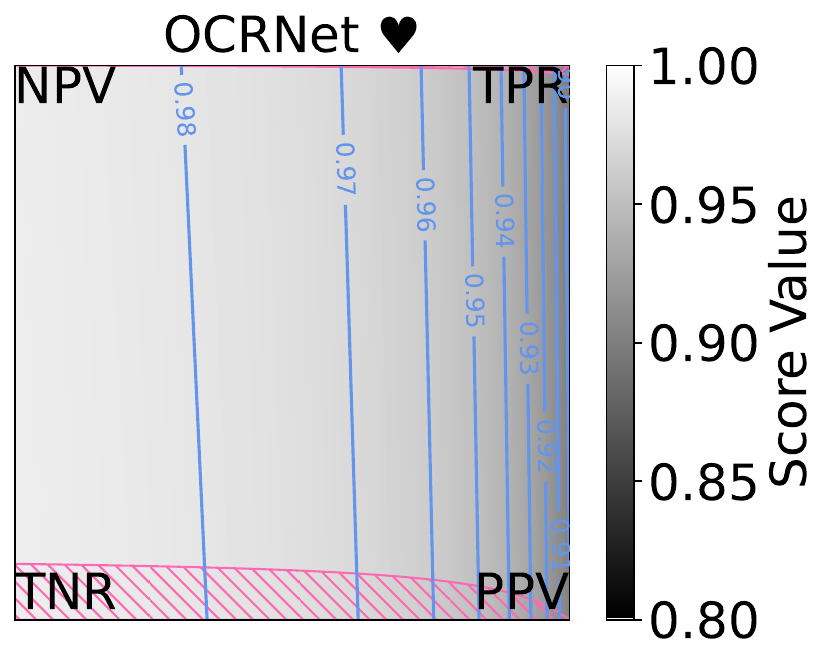}
\includegraphics[scale=0.4]{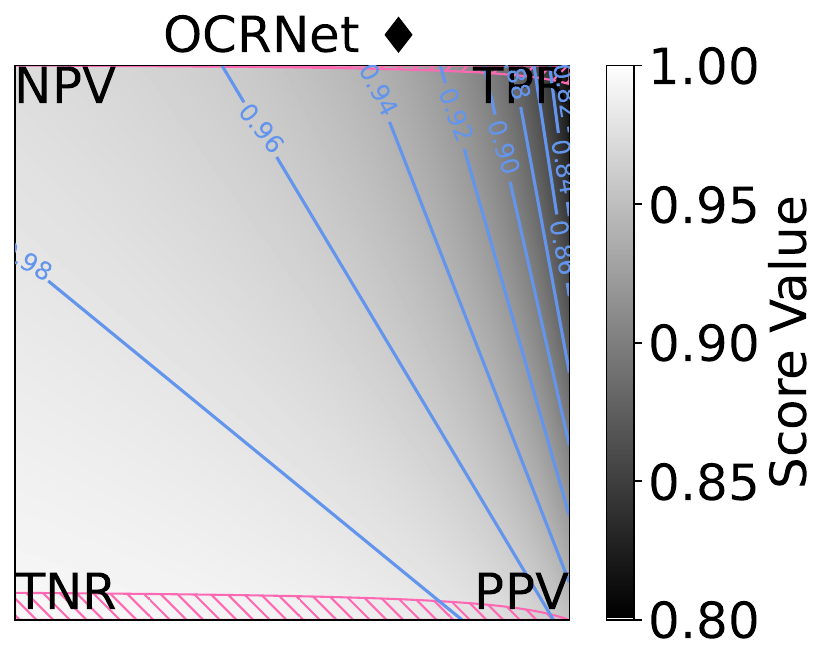}
\includegraphics[scale=0.4]{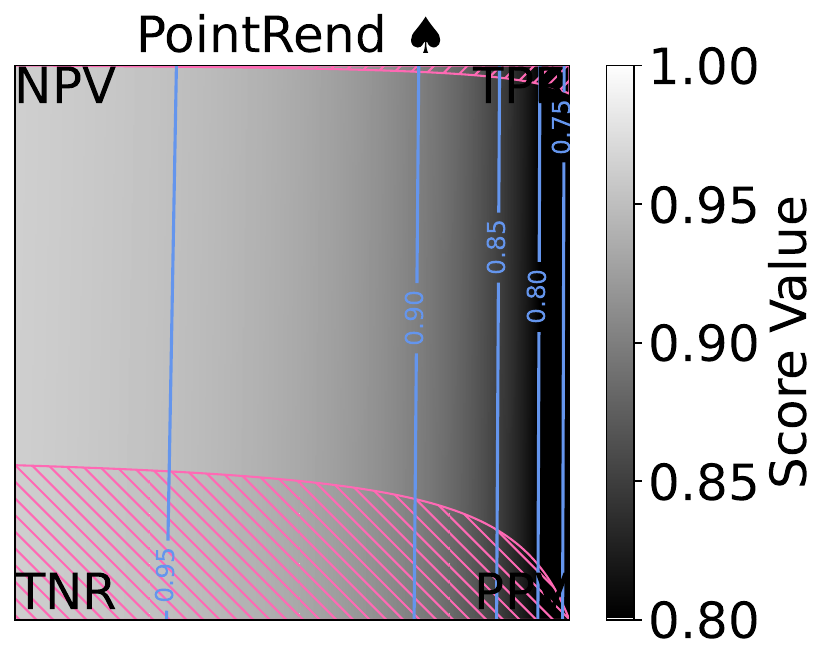}
\includegraphics[scale=0.4]{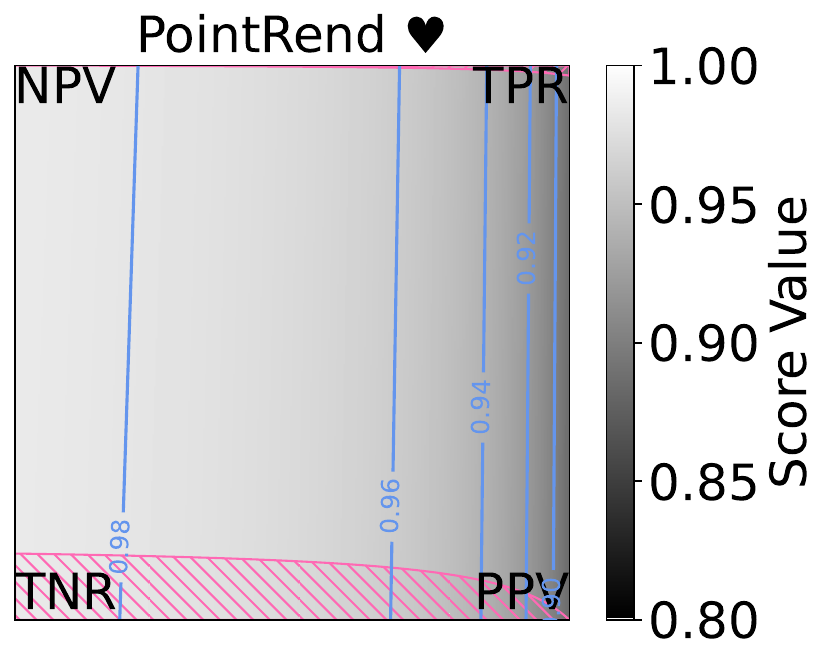}
\includegraphics[scale=0.4]{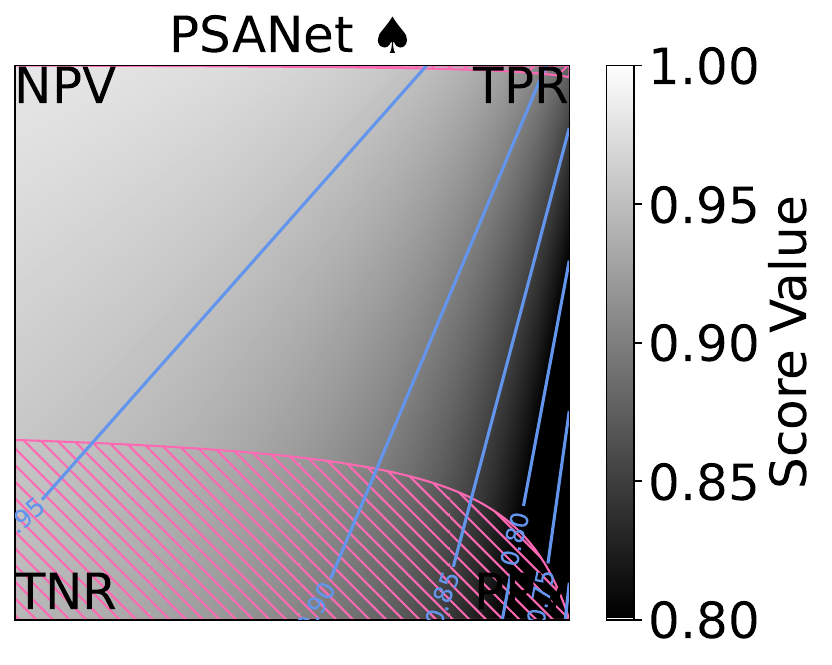}
\includegraphics[scale=0.4]{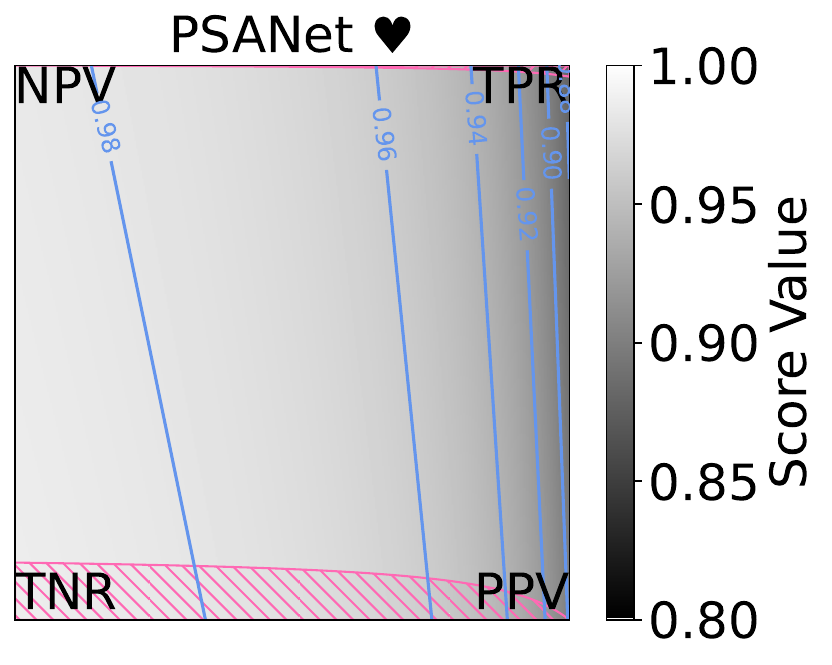}
\includegraphics[scale=0.4]{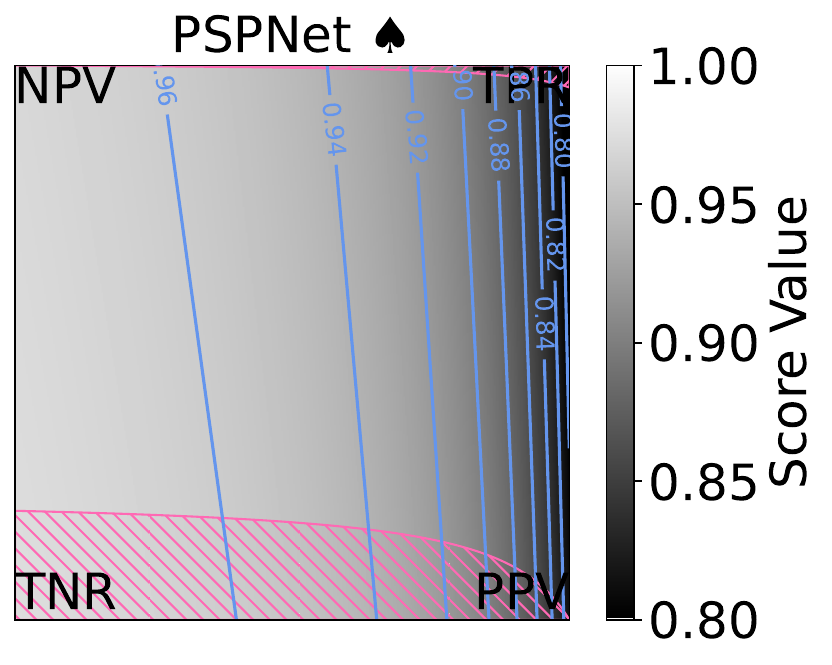}
\includegraphics[scale=0.4]{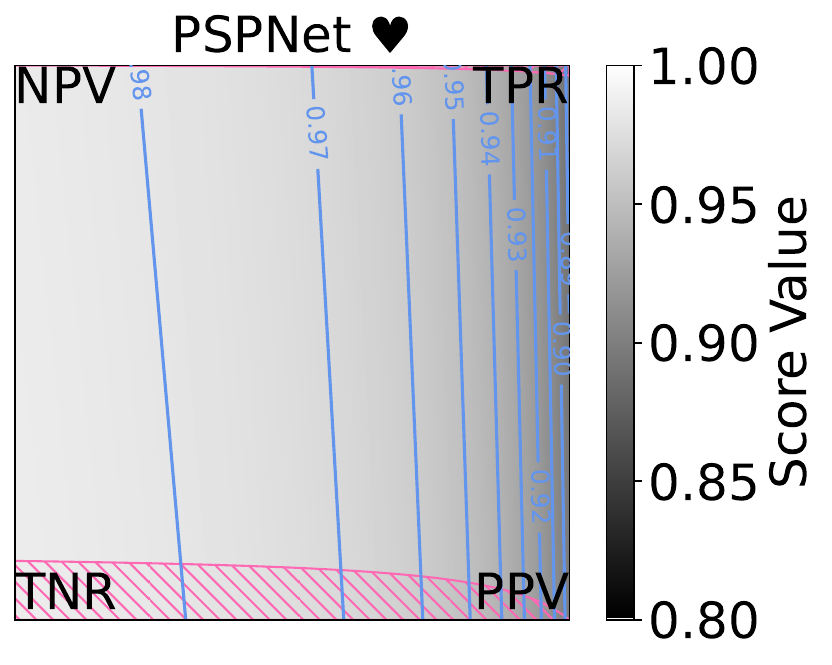}
\includegraphics[scale=0.4]{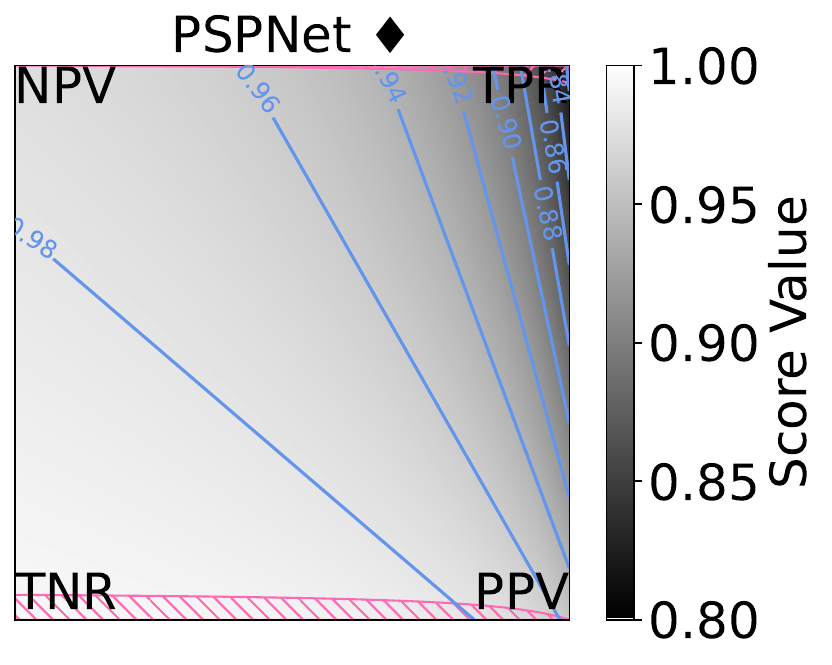}
\includegraphics[scale=0.4]{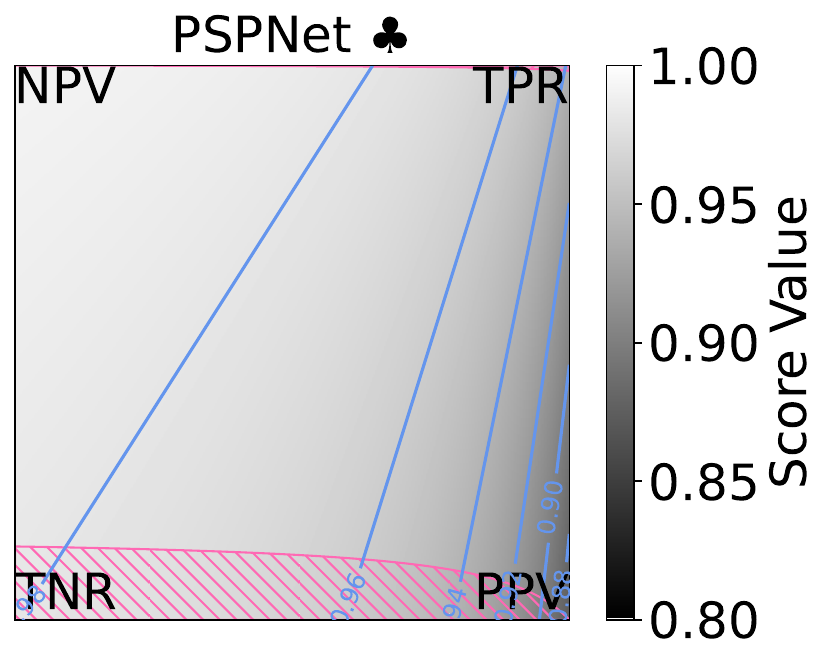}
\includegraphics[scale=0.4]{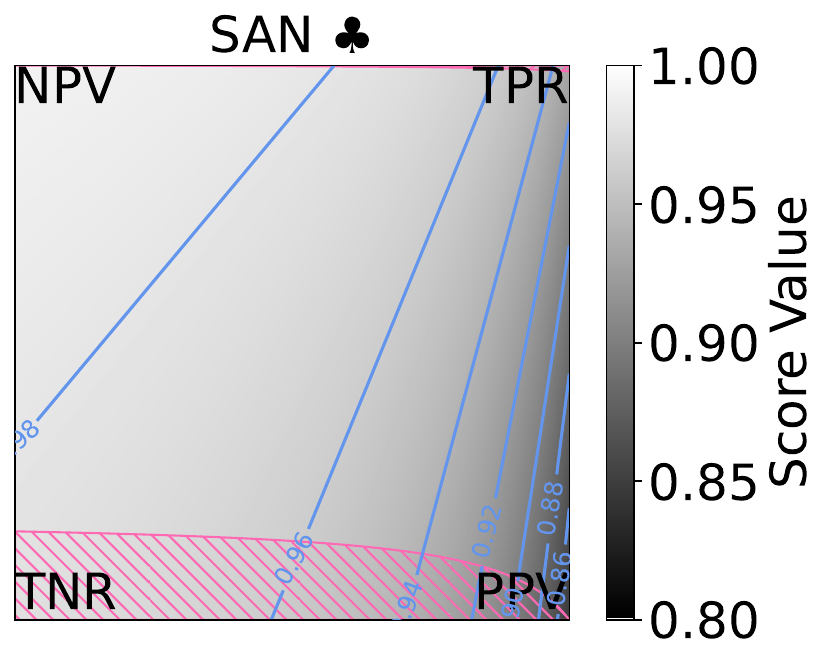}
\includegraphics[scale=0.4]{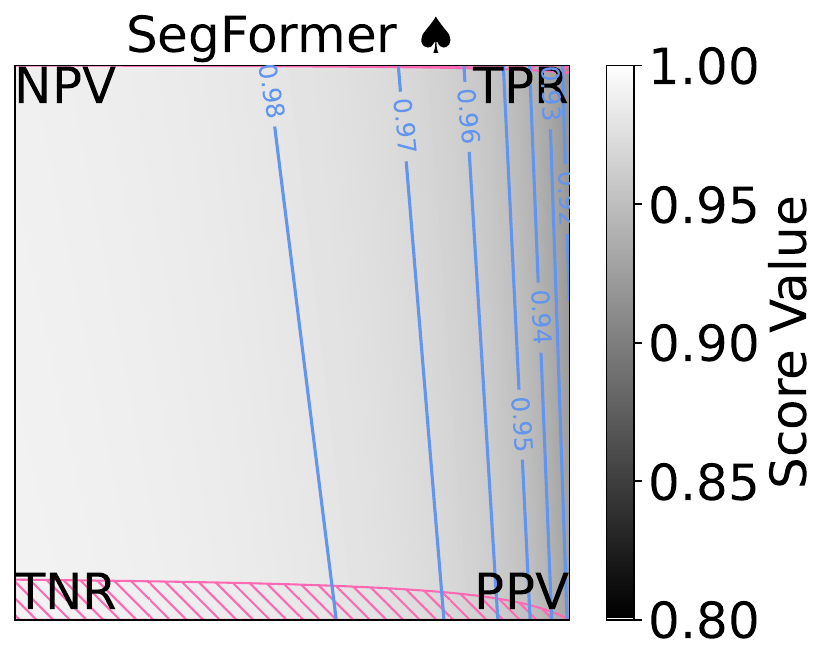}
\includegraphics[scale=0.4]{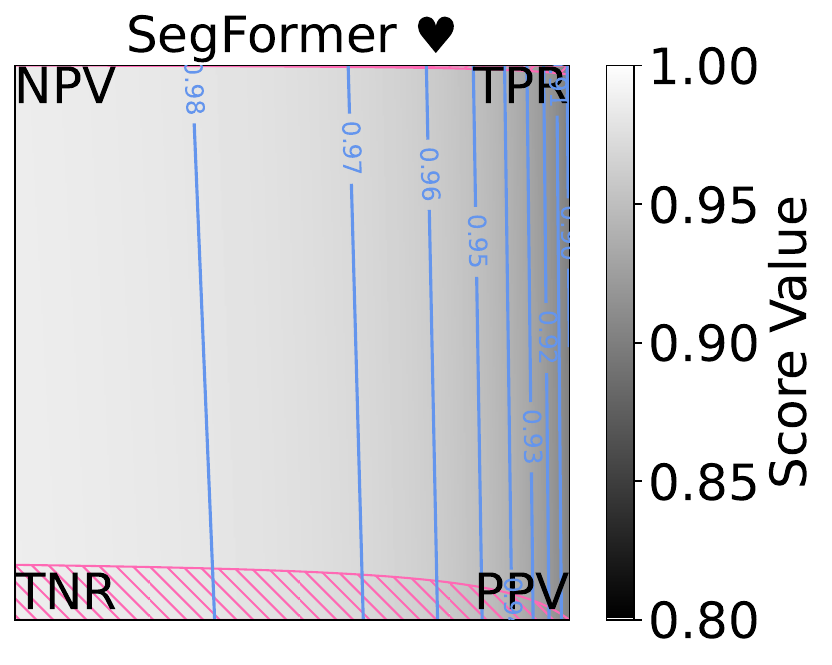}
\includegraphics[scale=0.4]{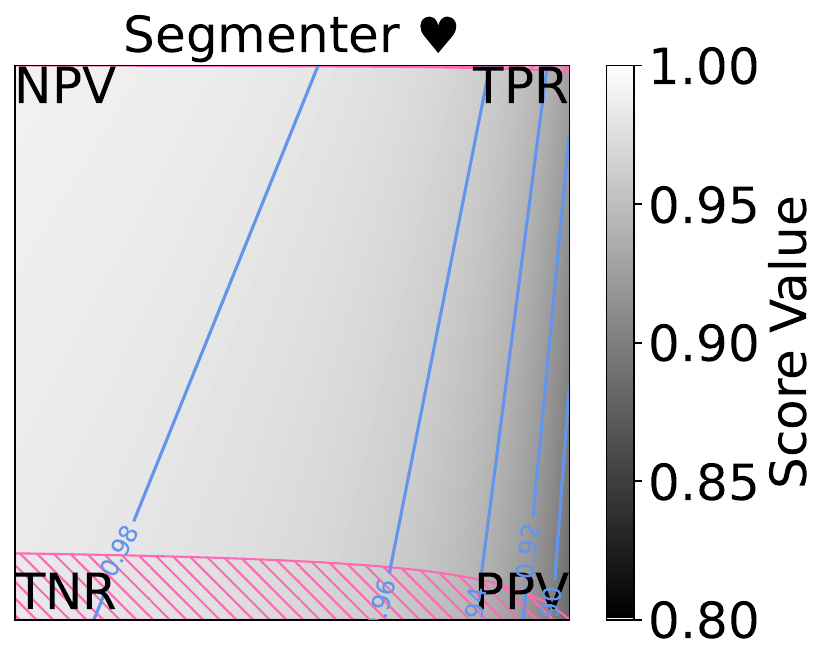}
\includegraphics[scale=0.4]{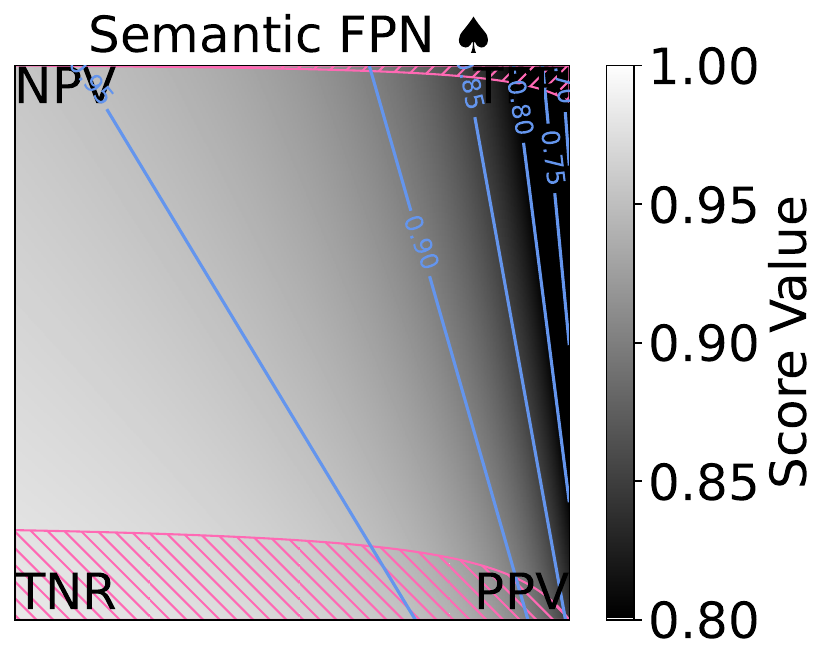}
\includegraphics[scale=0.4]{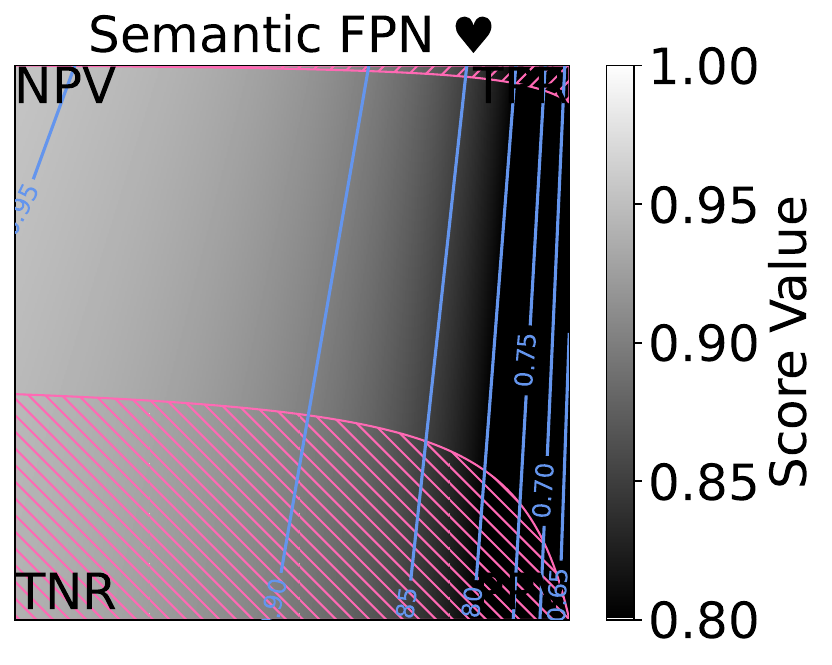}
\includegraphics[scale=0.4]{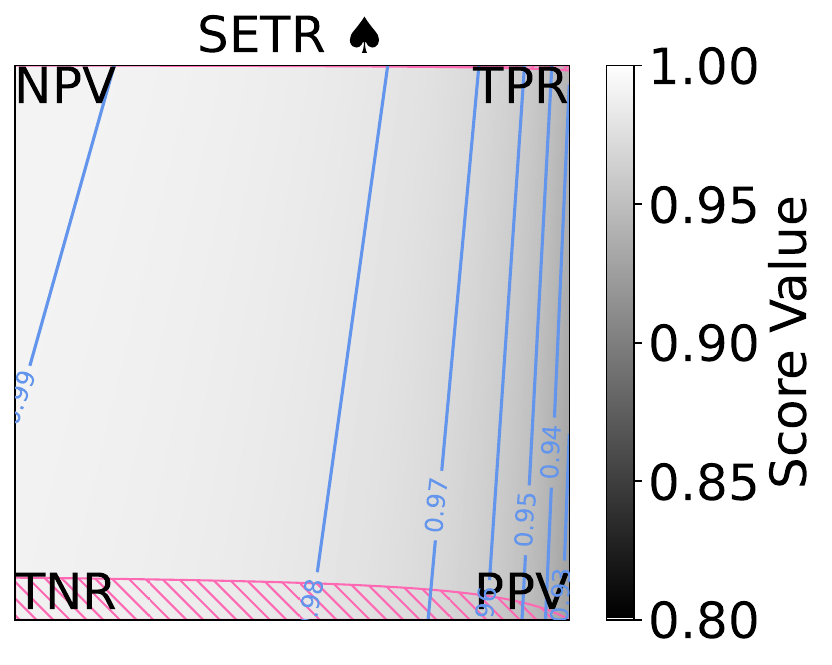}
\includegraphics[scale=0.4]{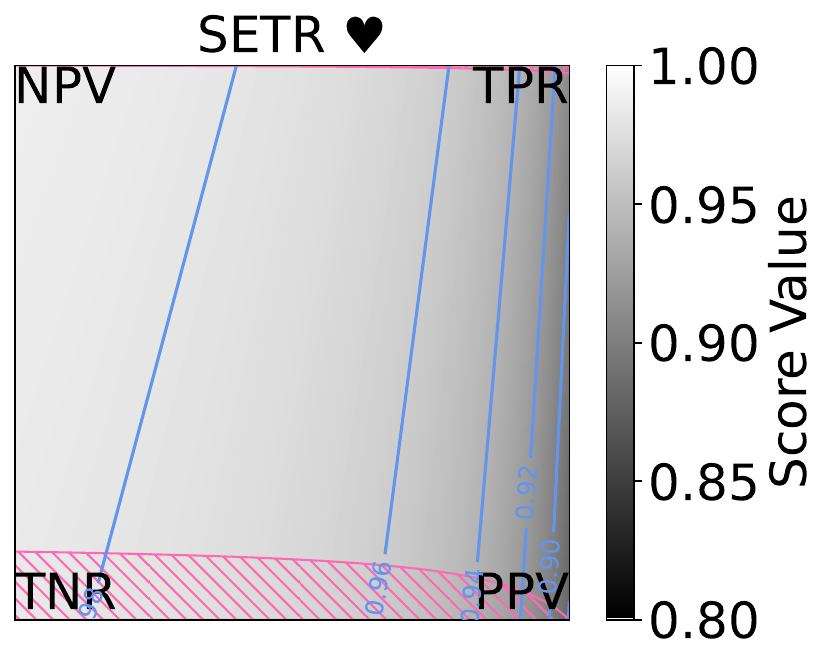}
\includegraphics[scale=0.4]{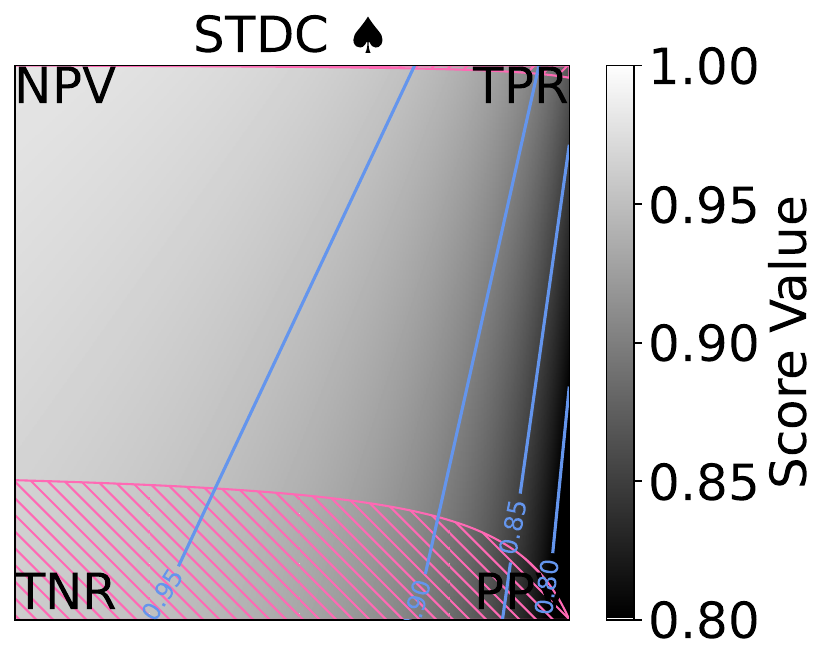}
\includegraphics[scale=0.4]{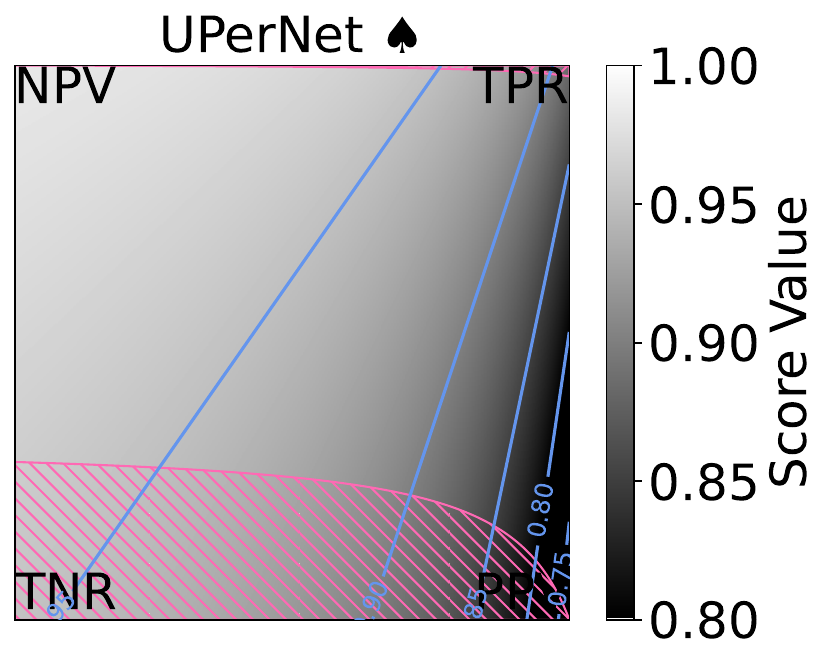}
\includegraphics[scale=0.4]{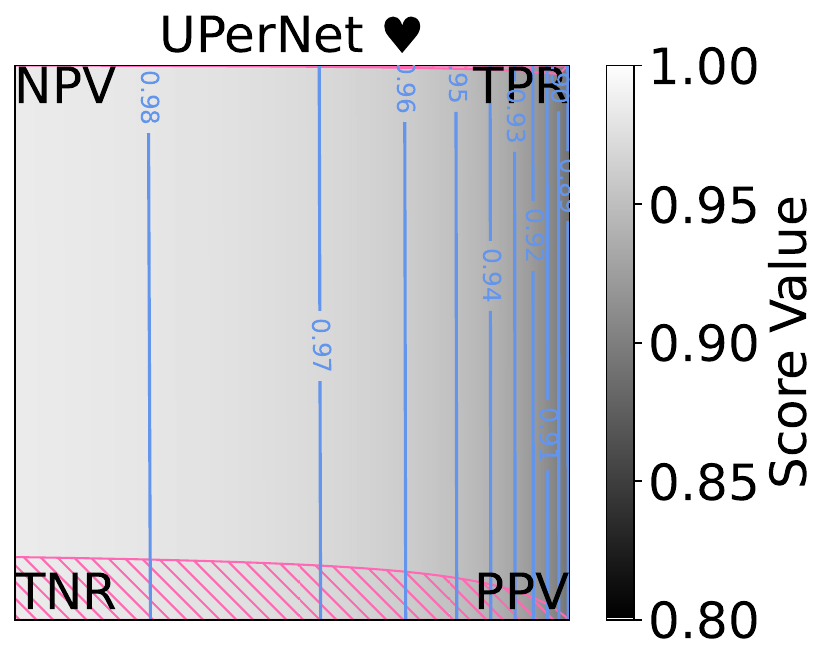}
\includegraphics[scale=0.4]{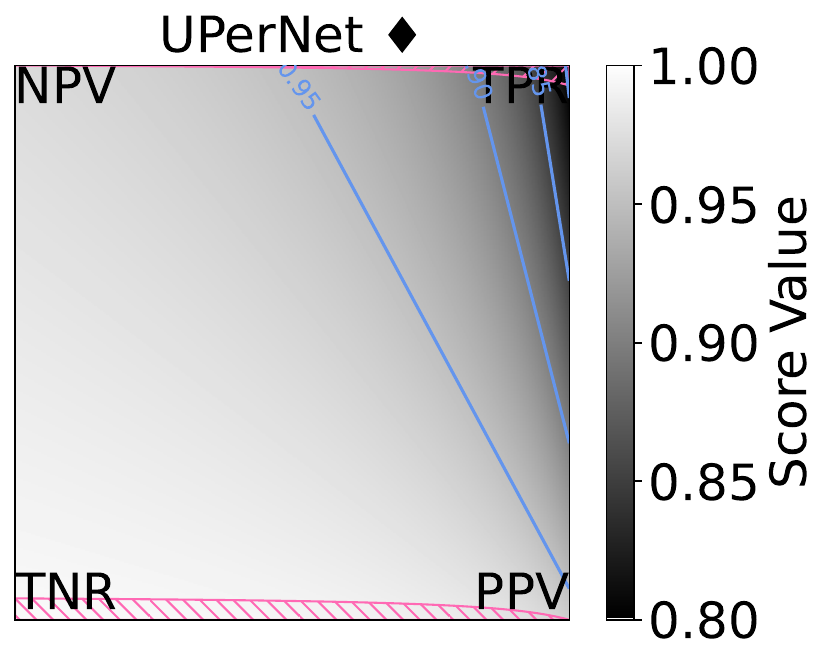}
\par\end{center}

\subsection*{Baseline Value Tile and State-of-the-Art Value Tile: Using the tile to show the 'Baseline' and 'State of the Art'}
\begin{center}
\includegraphics[scale=0.4]{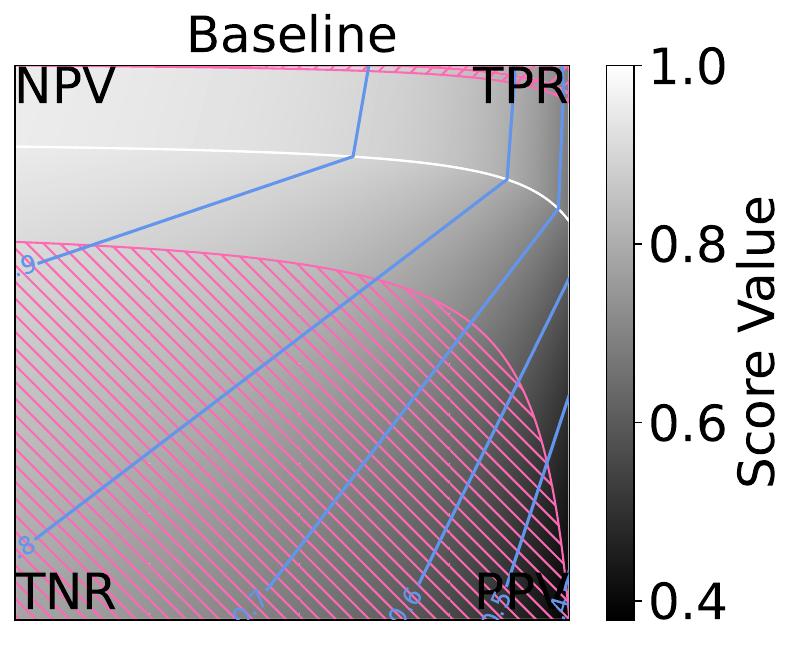}
\includegraphics[scale=0.4]{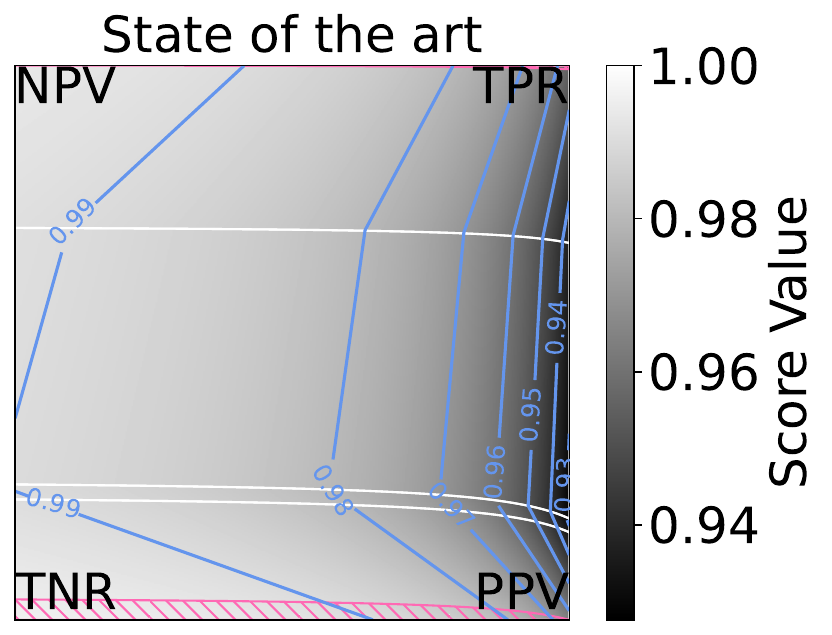}
\par\end{center}

\subsection*{No-Skill Tile and Relative-Skill Tile: Using the tile to show the no-skill values and the relative skill values}
\begin{center}
\includegraphics[scale=0.4]{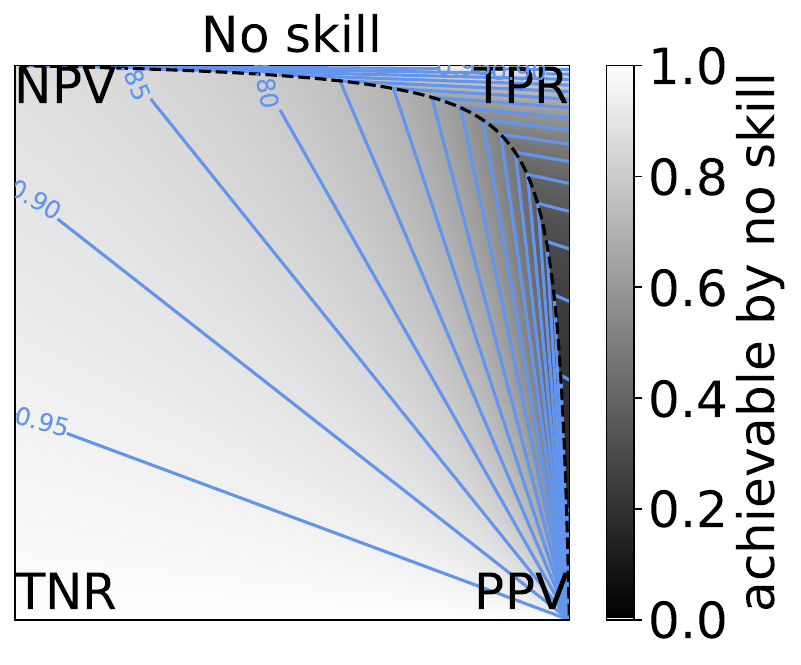}
\includegraphics[scale=0.4]{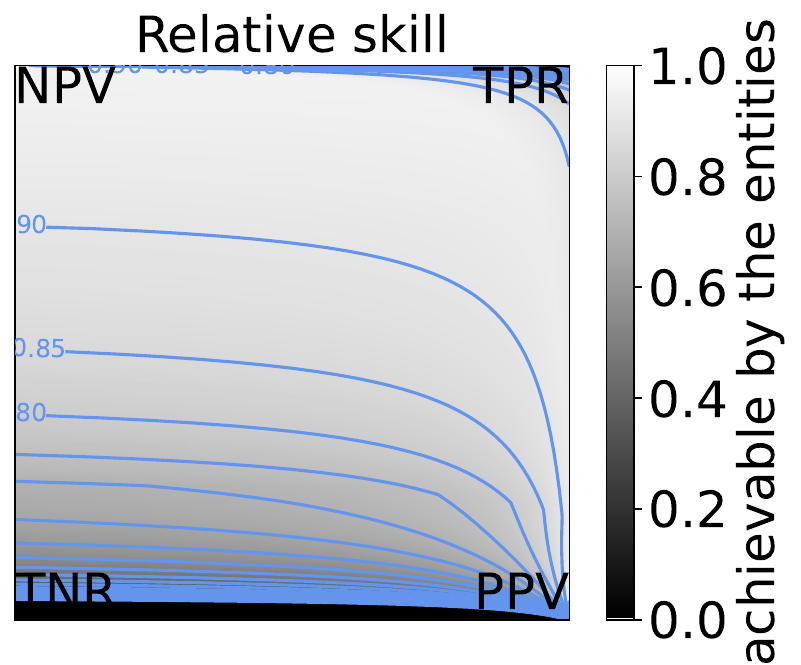}
\par\end{center}

\subsection*{Ranking Tile: Using the tile to show the ranks for each entity}
\begin{center}
\includegraphics[scale=0.4]{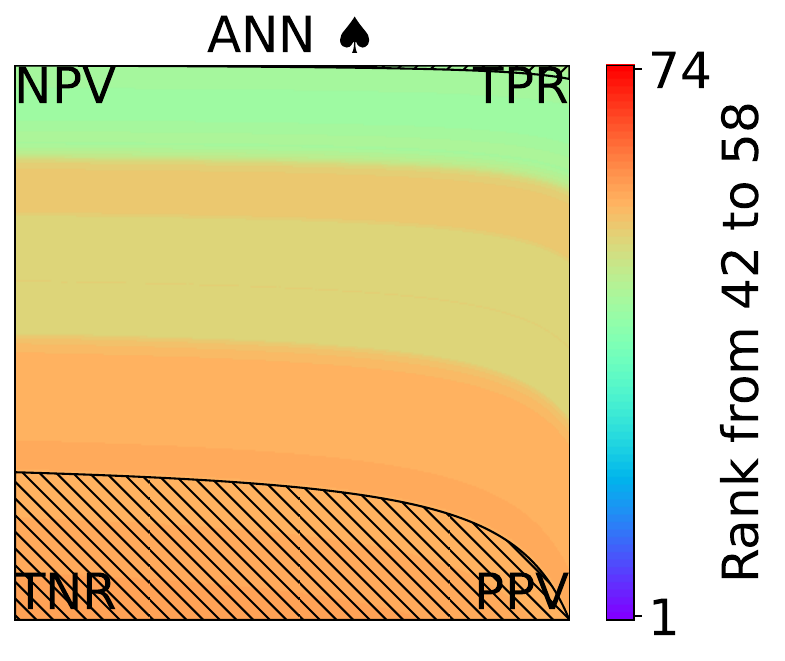}
\includegraphics[scale=0.4]{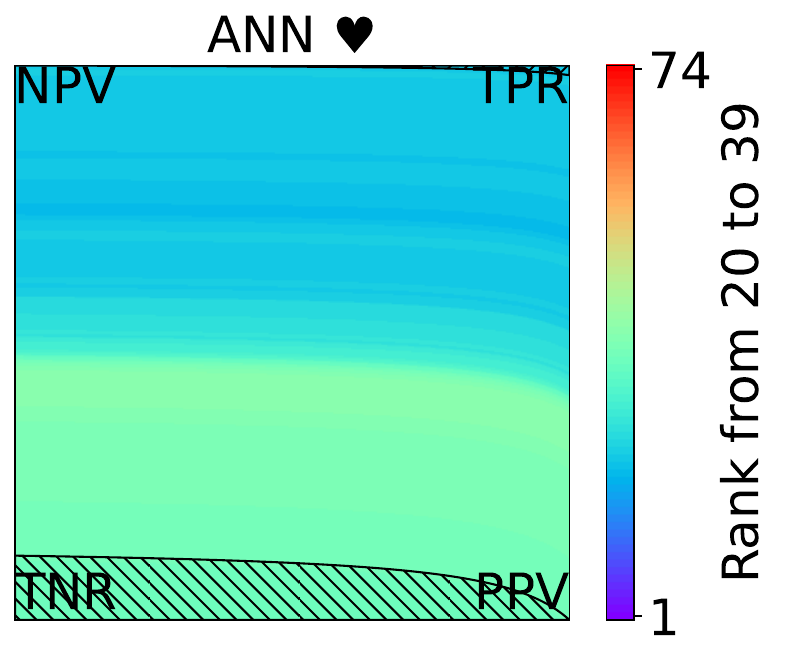}
\includegraphics[scale=0.4]{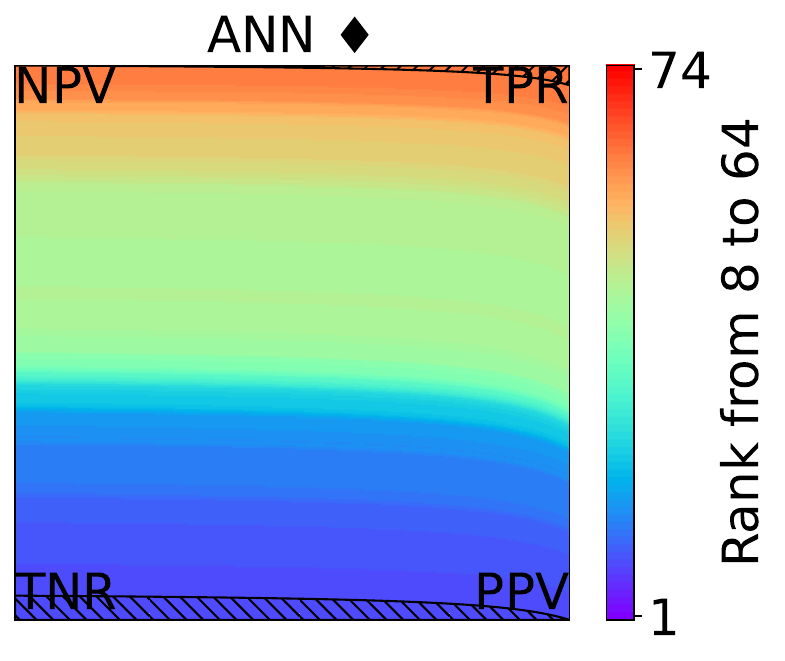}
\includegraphics[scale=0.4]{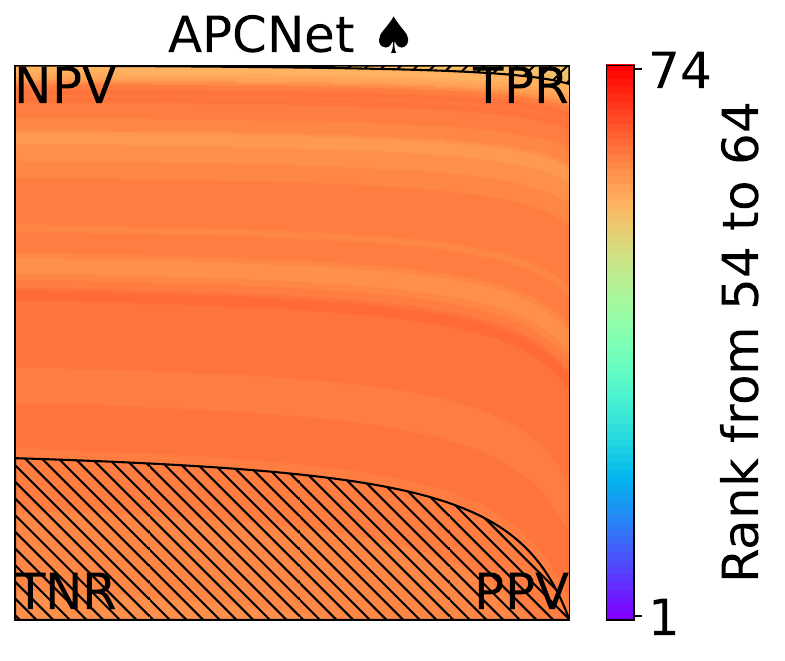}
\includegraphics[scale=0.4]{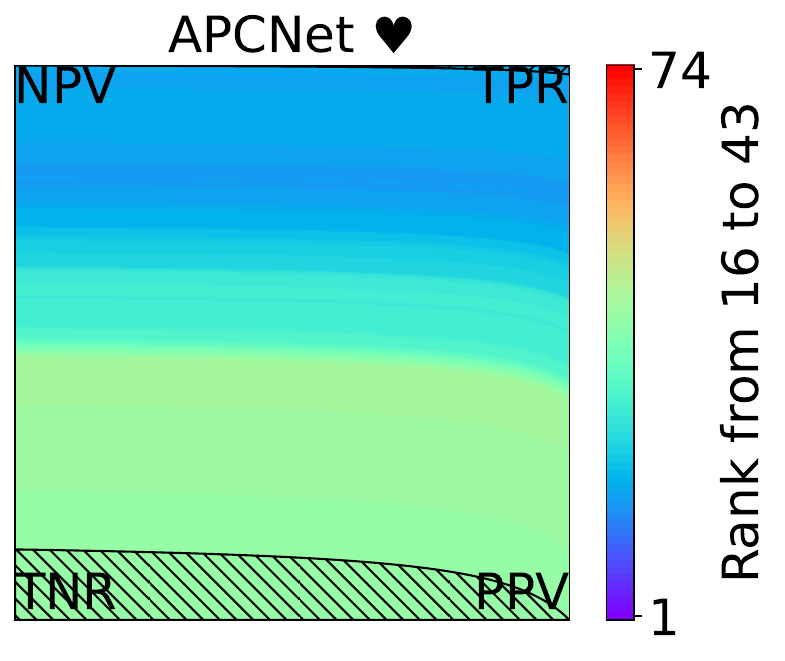}
\includegraphics[scale=0.4]{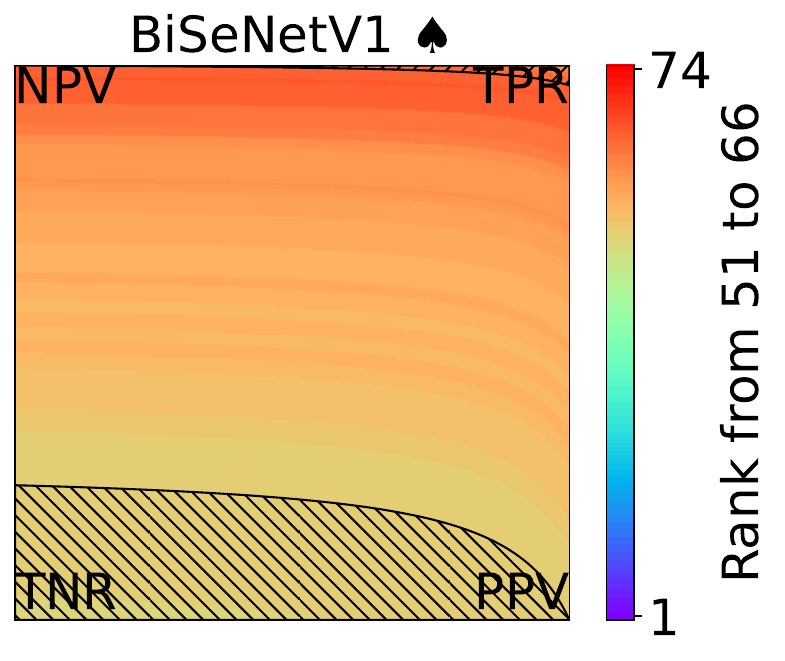}
\includegraphics[scale=0.4]{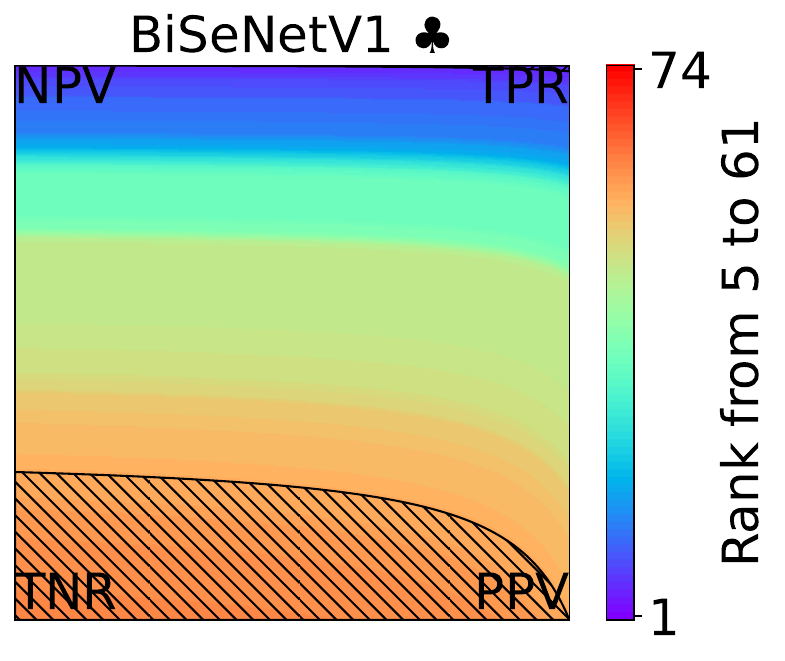}
\includegraphics[scale=0.4]{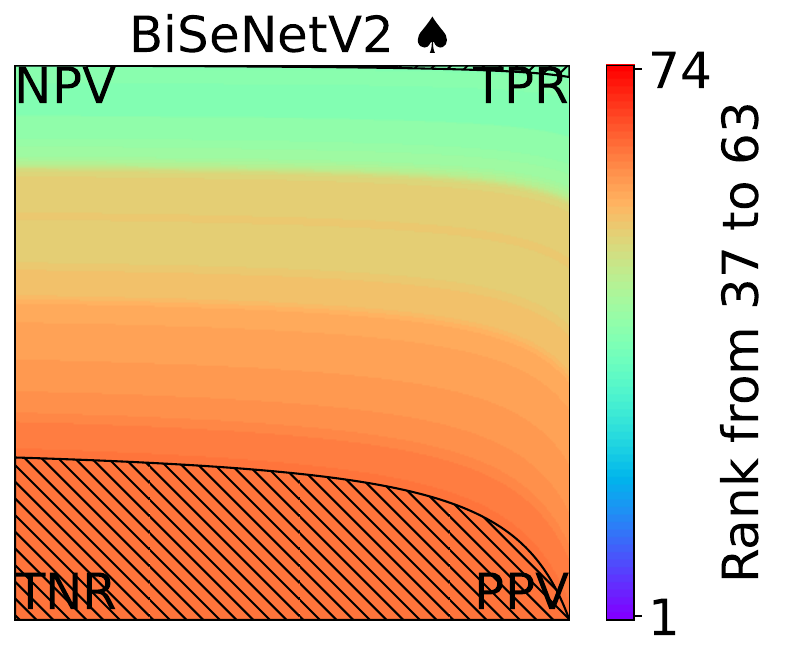}
\includegraphics[scale=0.4]{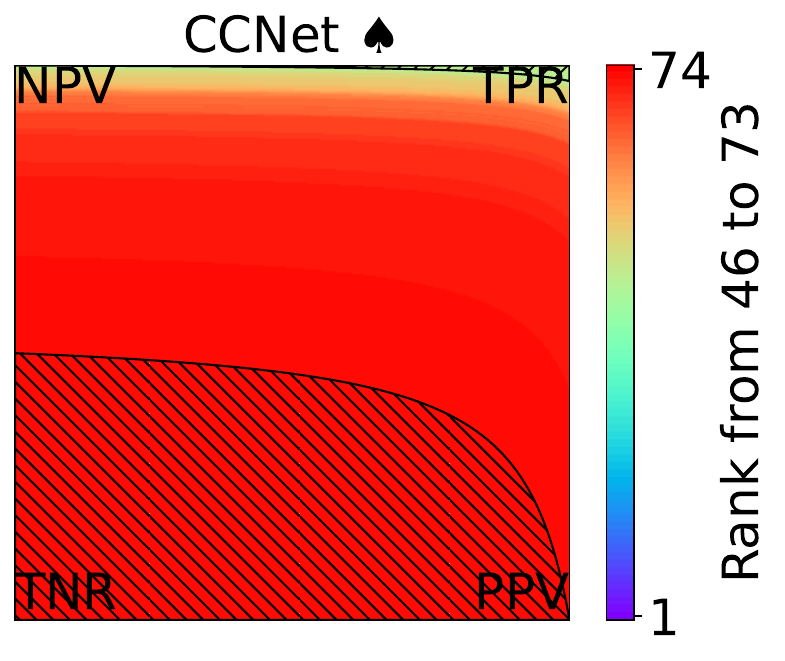}
\includegraphics[scale=0.4]{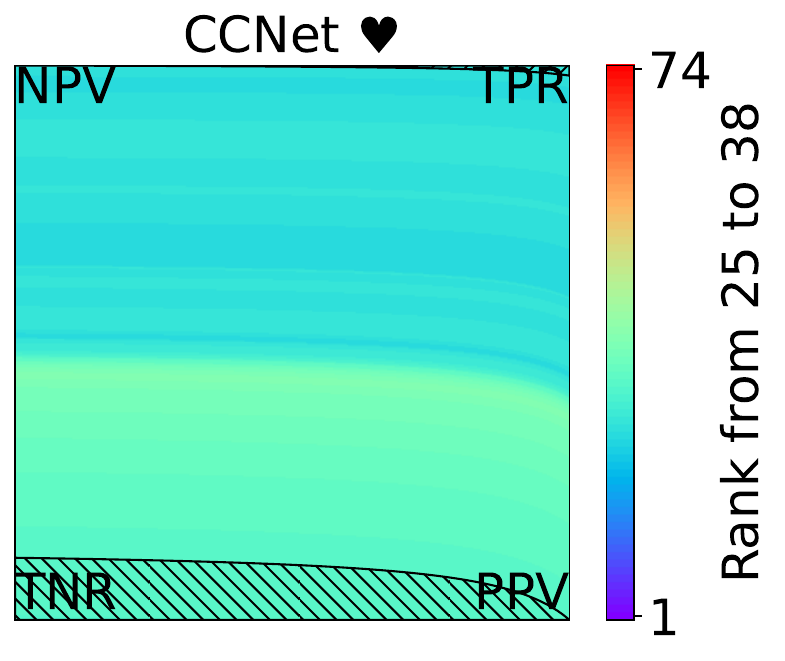}
\includegraphics[scale=0.4]{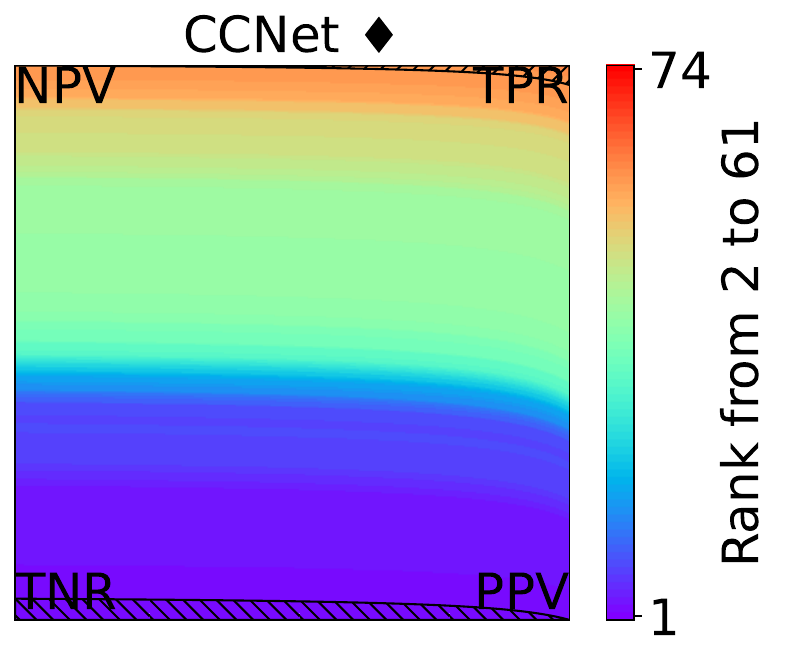}
\includegraphics[scale=0.4]{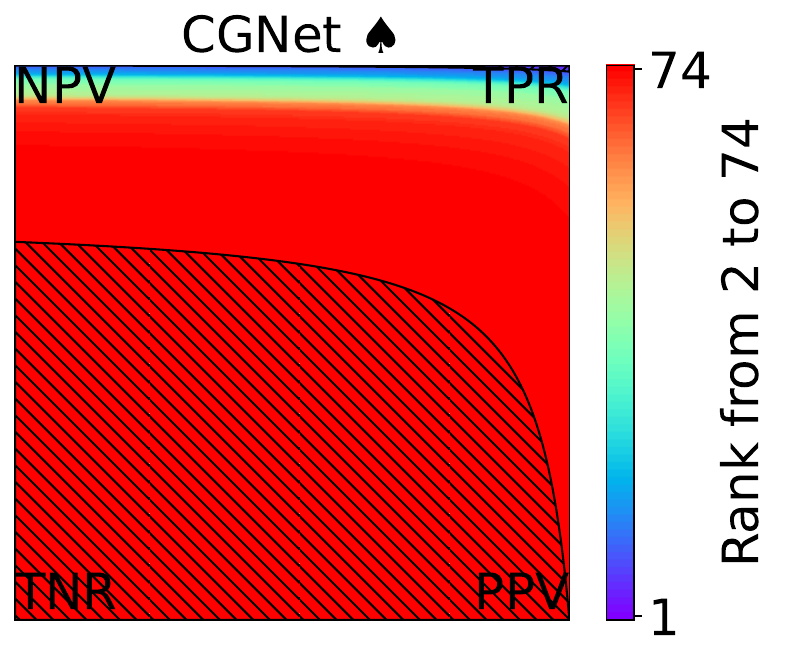}
\includegraphics[scale=0.4]{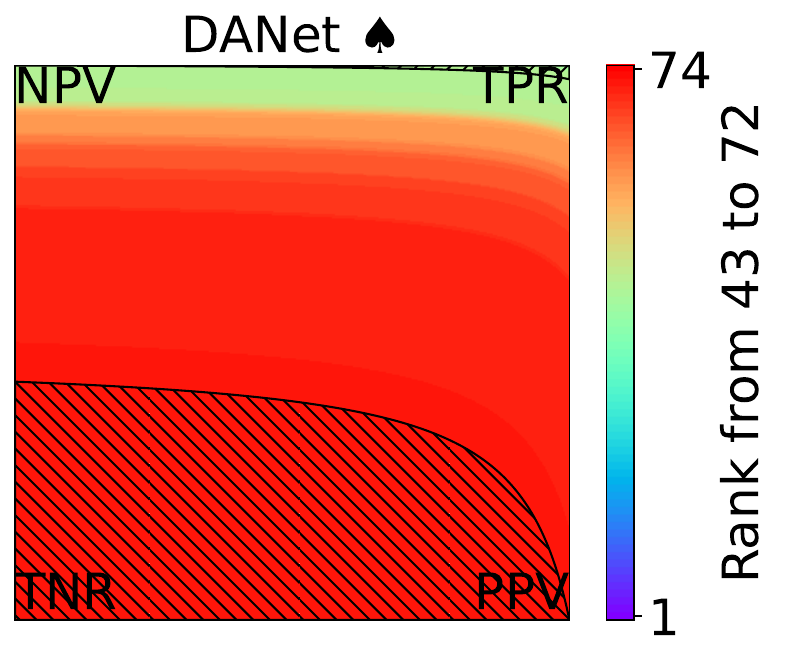}
\includegraphics[scale=0.4]{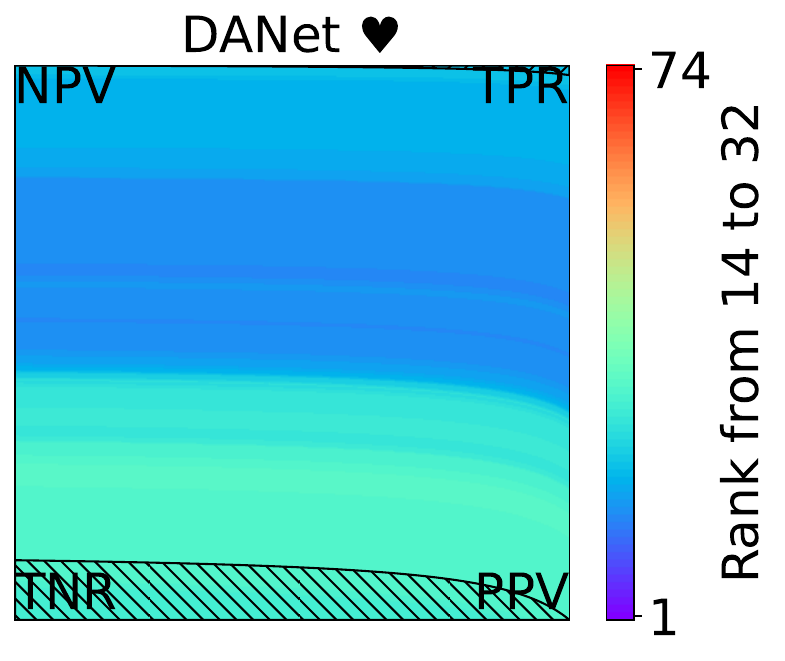}
\includegraphics[scale=0.4]{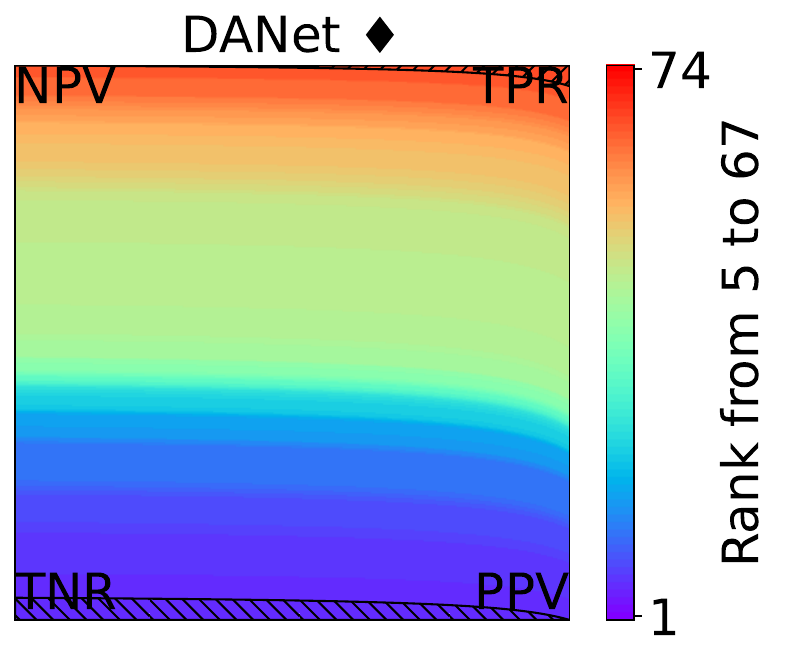}
\includegraphics[scale=0.4]{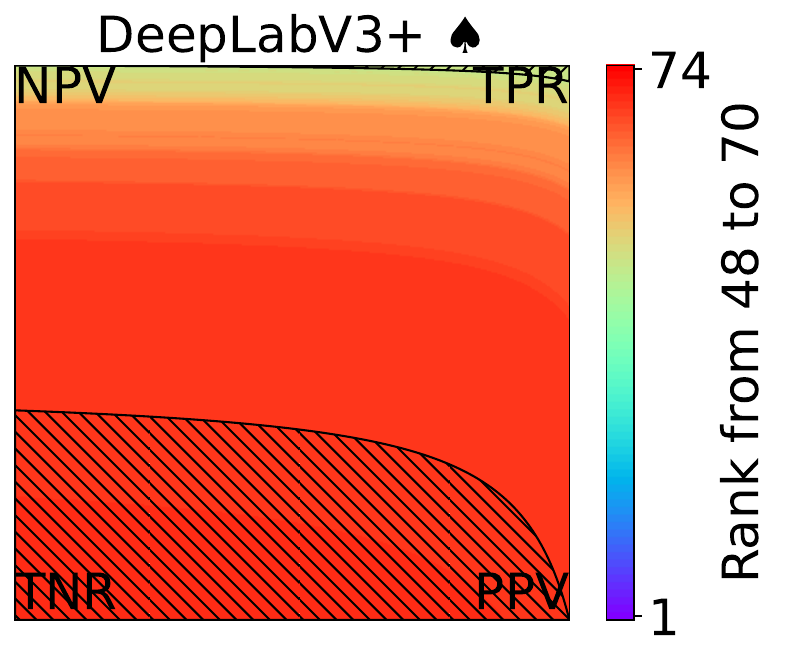}
\includegraphics[scale=0.4]{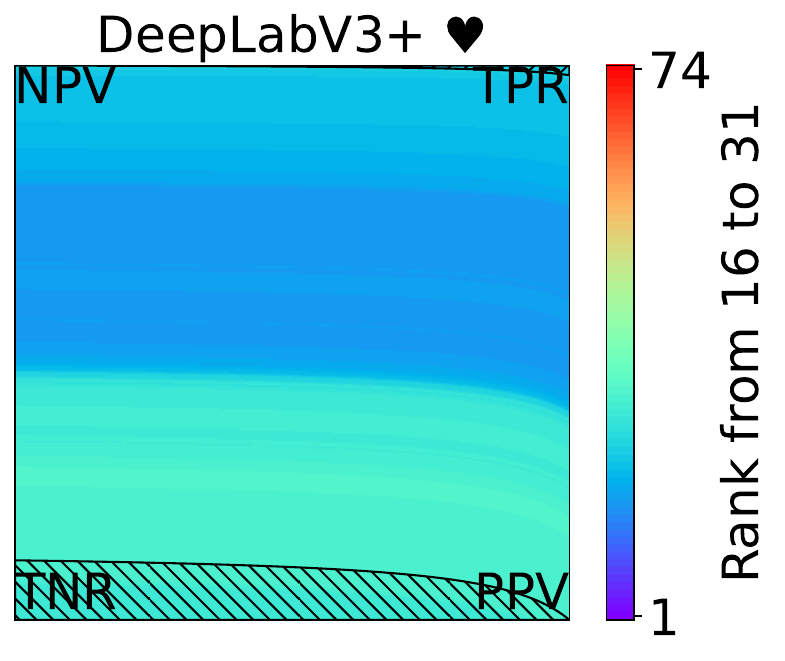}
\includegraphics[scale=0.4]{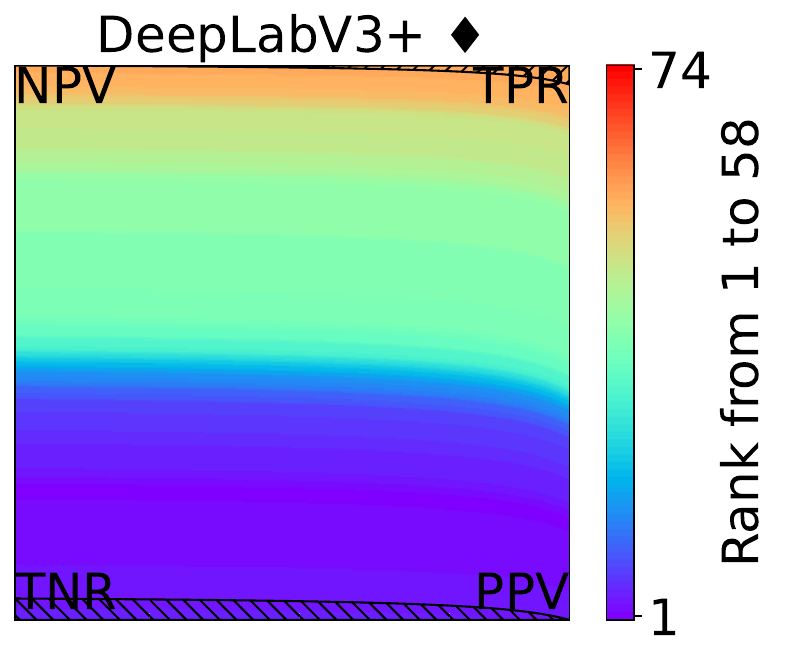}
\includegraphics[scale=0.4]{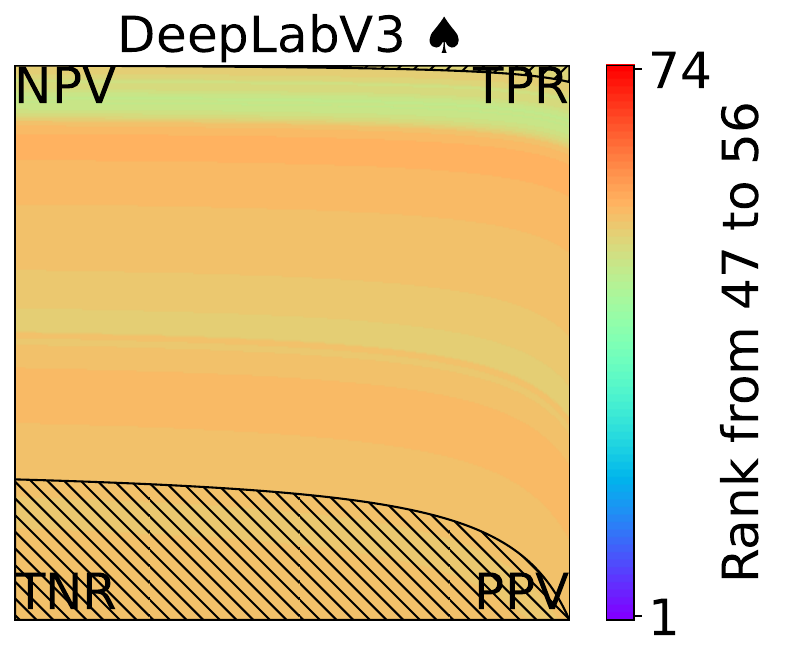}
\includegraphics[scale=0.4]{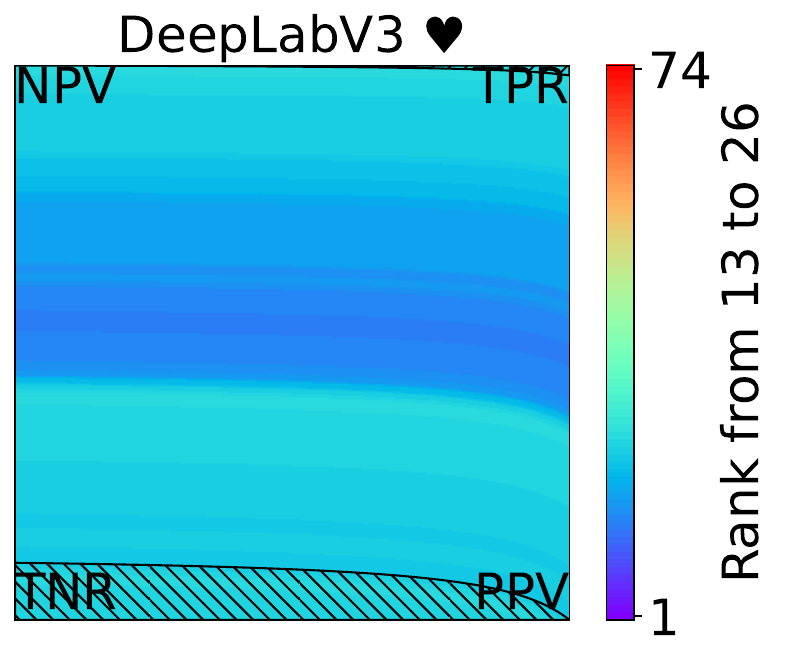}
\includegraphics[scale=0.4]{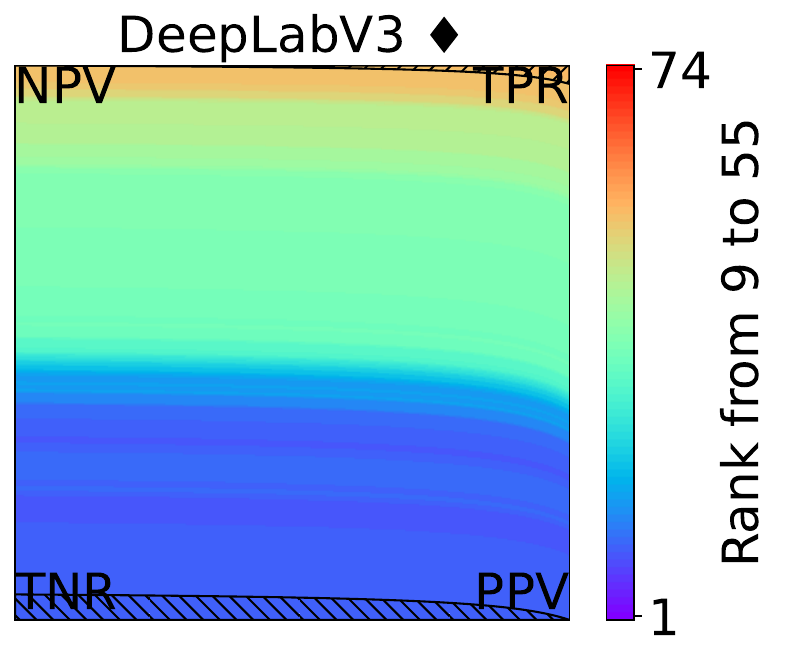}
\includegraphics[scale=0.4]{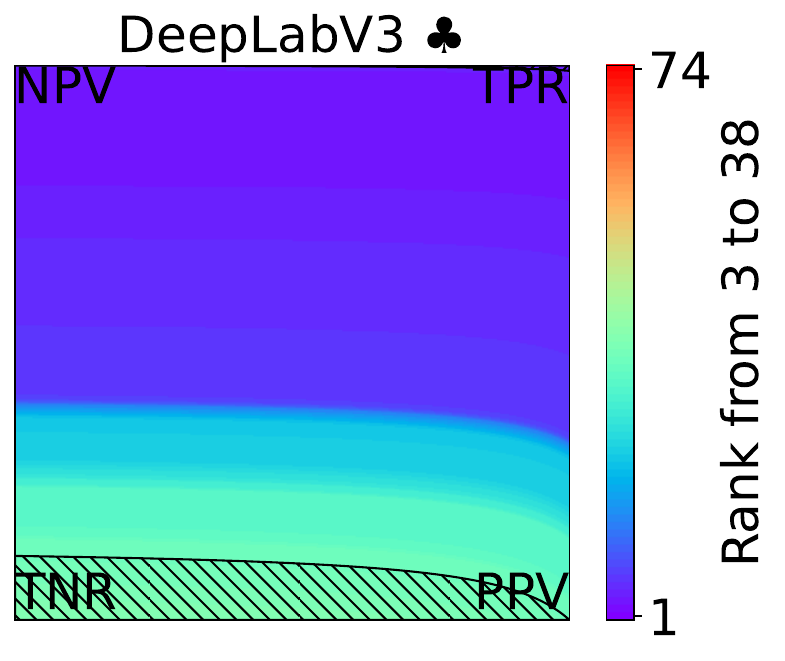}
\includegraphics[scale=0.4]{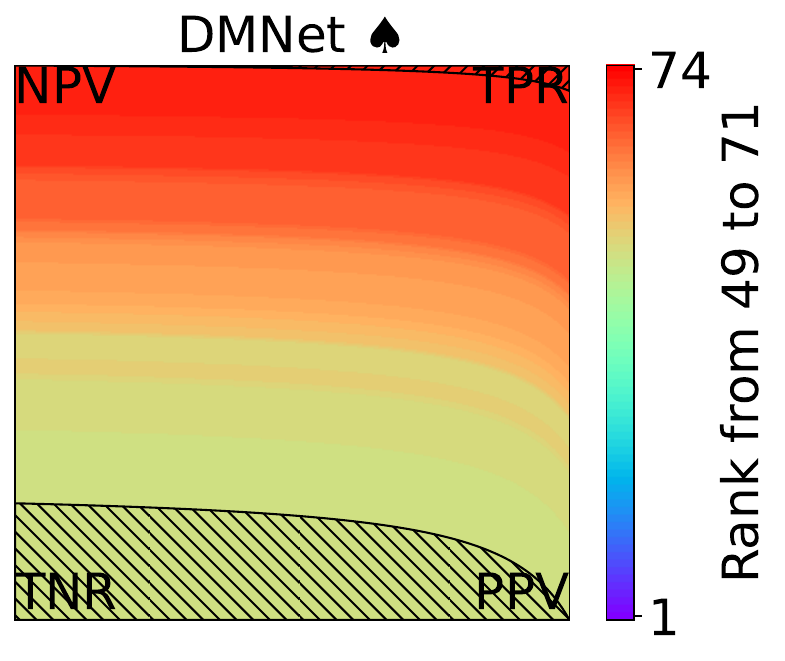}
\includegraphics[scale=0.4]{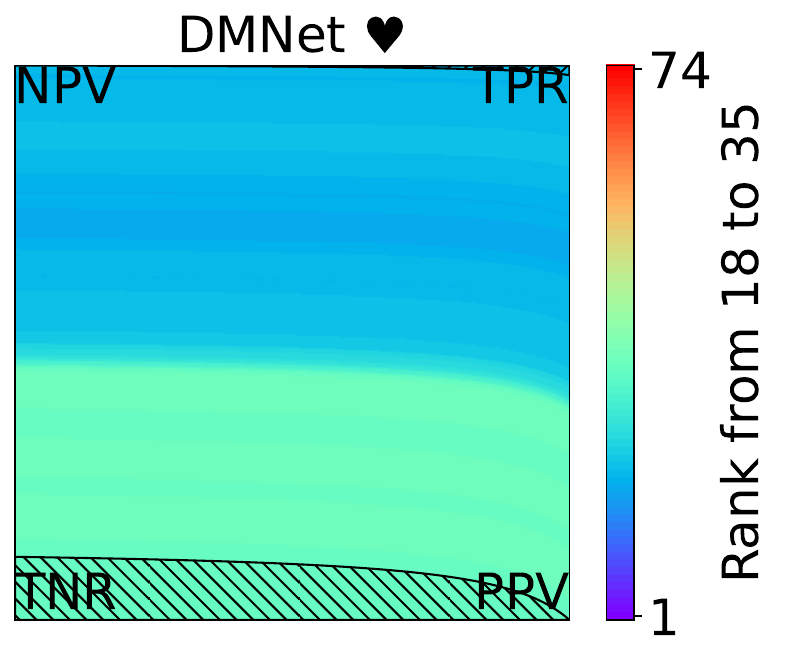}
\includegraphics[scale=0.4]{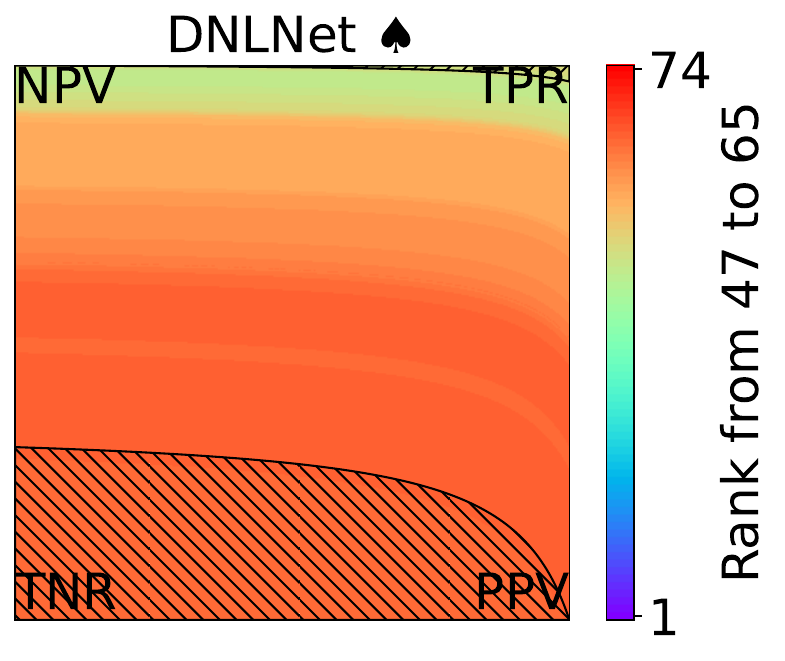}
\includegraphics[scale=0.4]{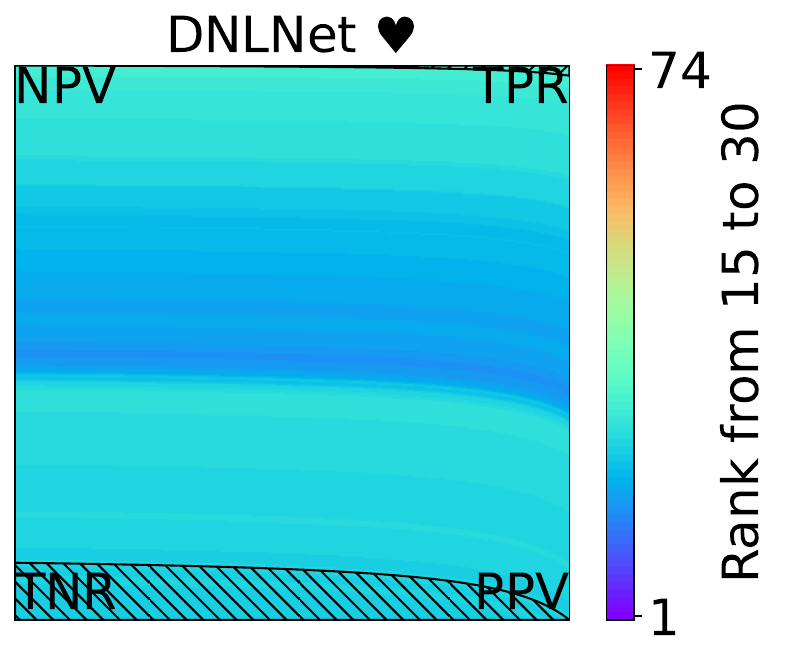}
\includegraphics[scale=0.4]{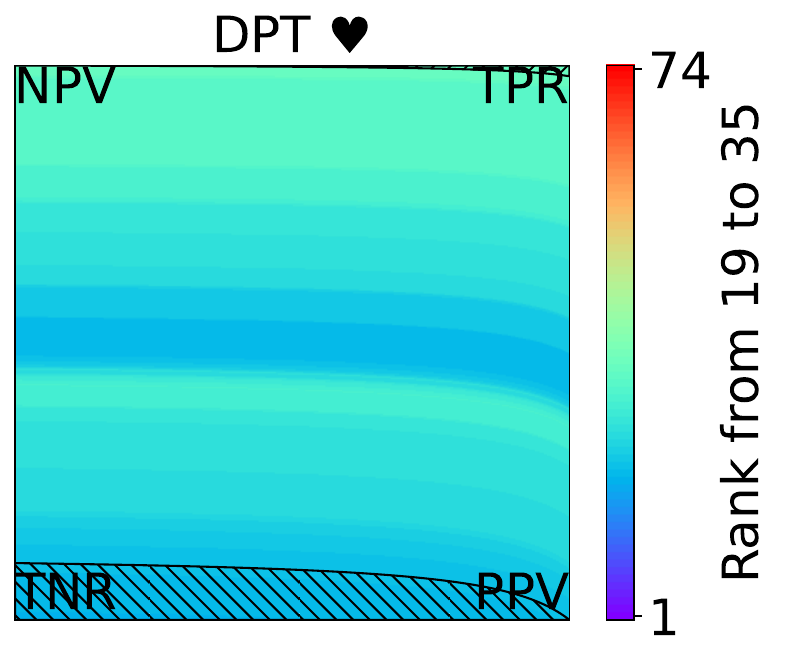}
\includegraphics[scale=0.4]{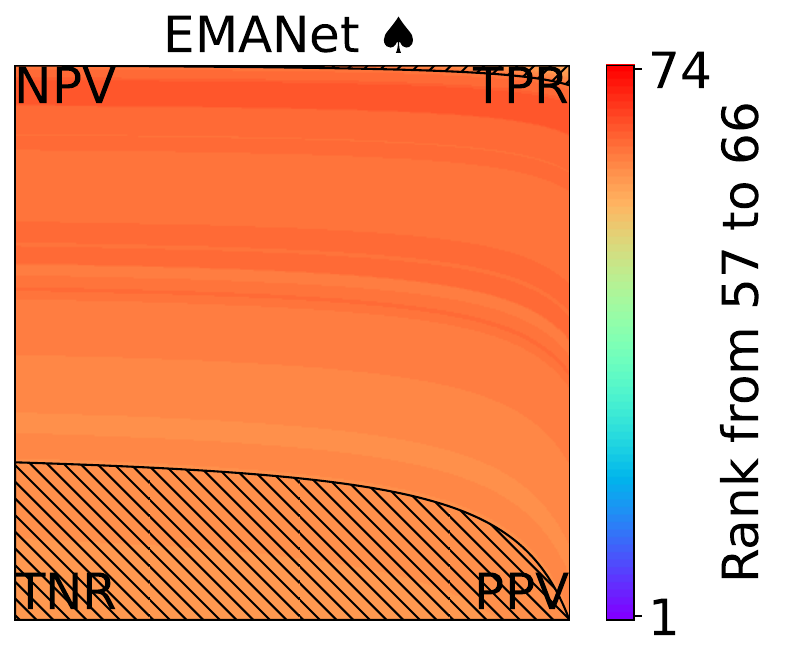}
\includegraphics[scale=0.4]{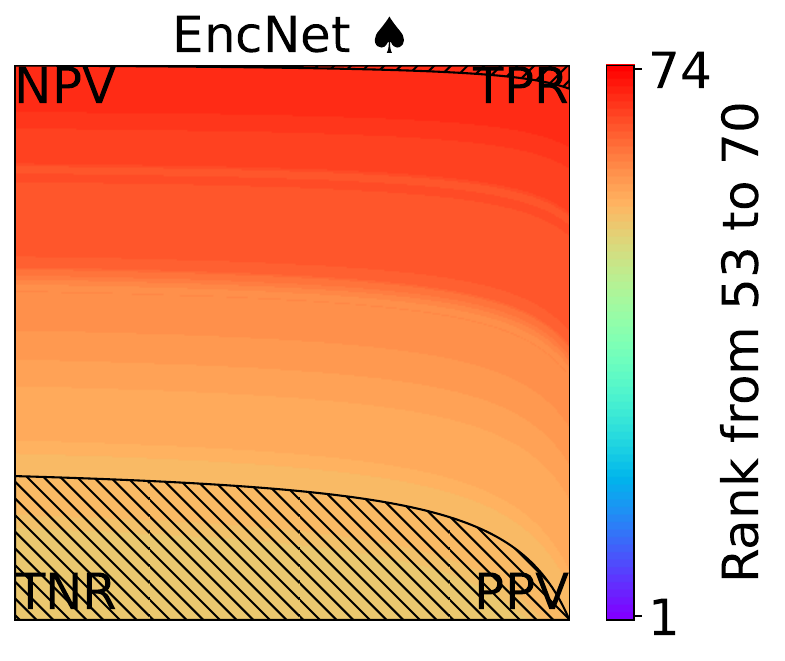}
\includegraphics[scale=0.4]{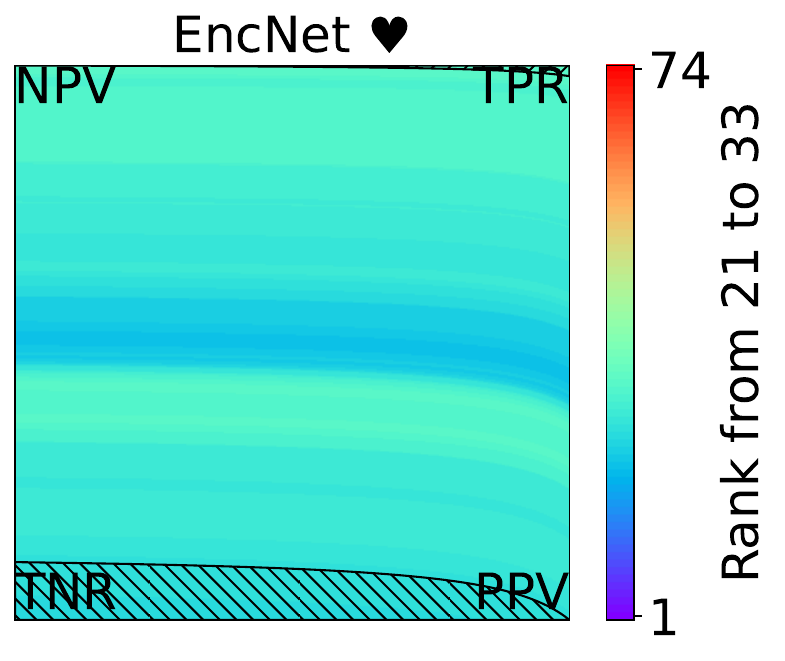}
\includegraphics[scale=0.4]{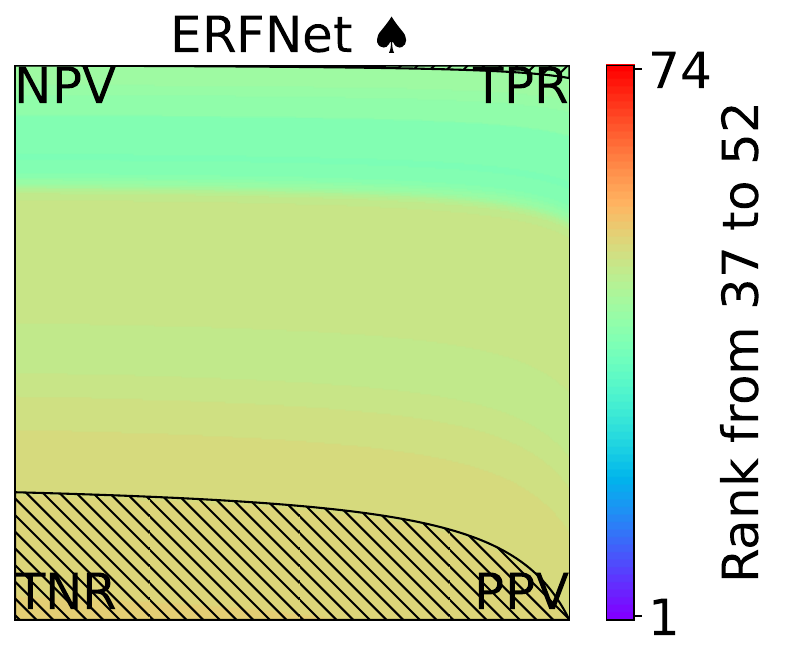}
\includegraphics[scale=0.4]{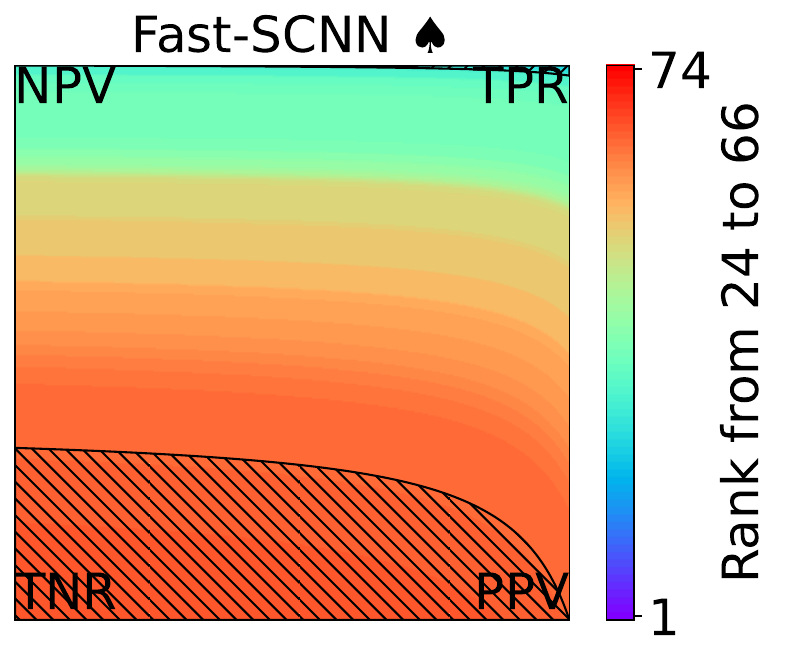}
\includegraphics[scale=0.4]{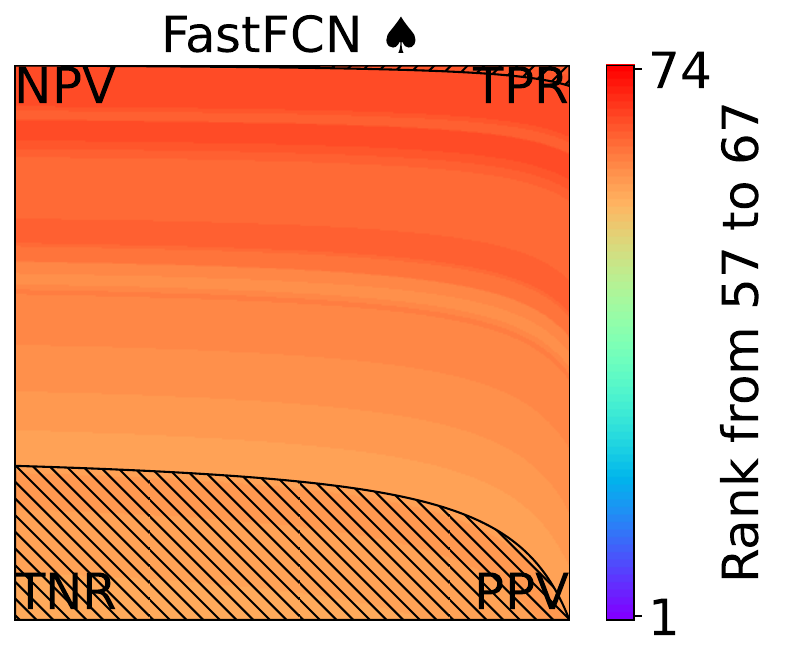}
\includegraphics[scale=0.4]{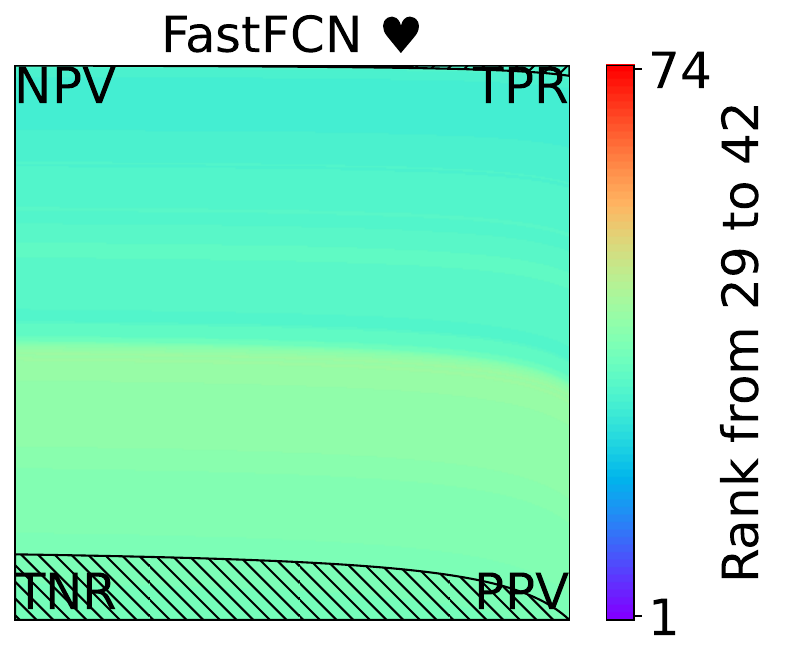}
\includegraphics[scale=0.4]{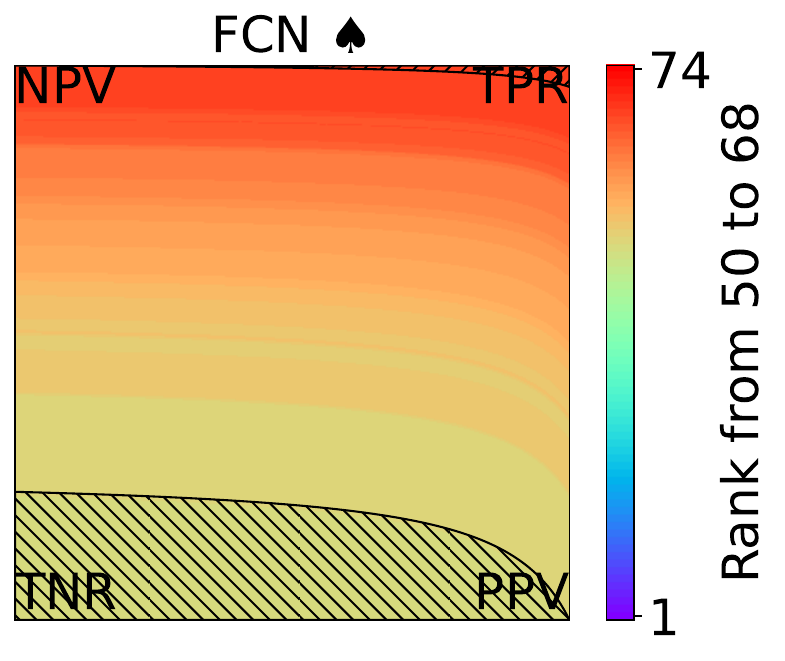}
\includegraphics[scale=0.4]{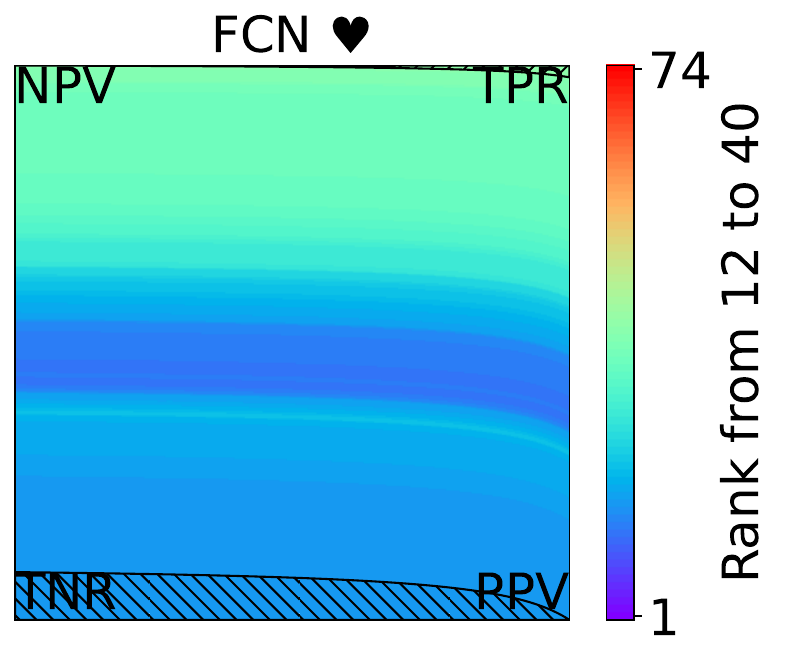}
\includegraphics[scale=0.4]{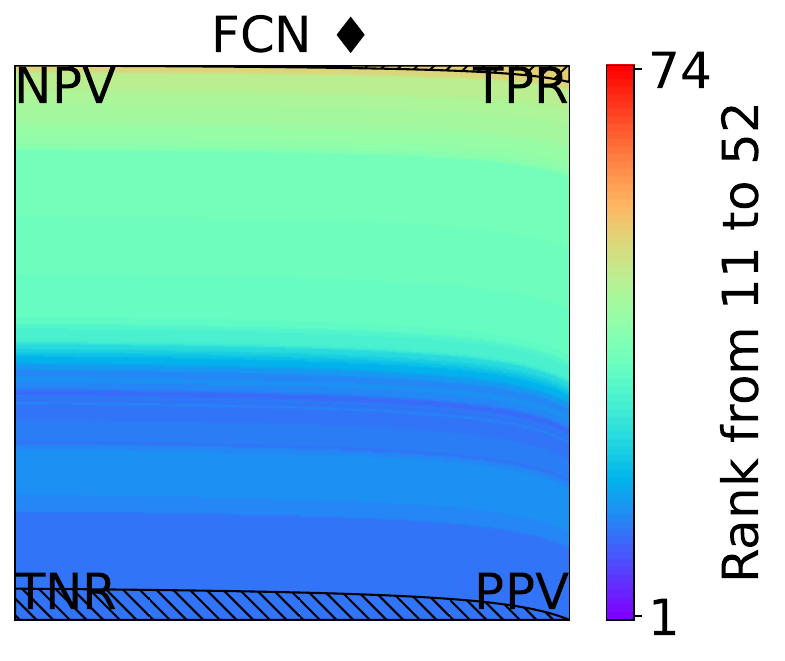}
\includegraphics[scale=0.4]{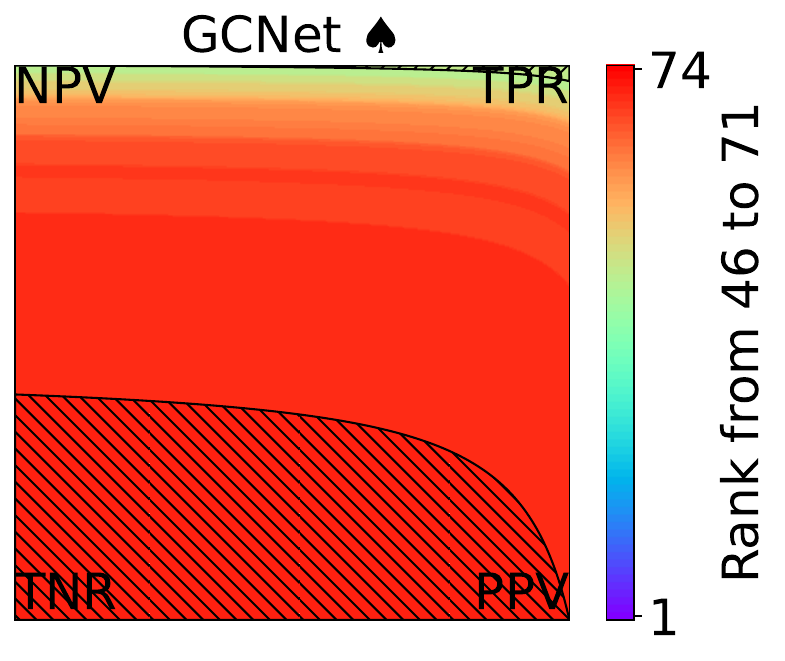}
\includegraphics[scale=0.4]{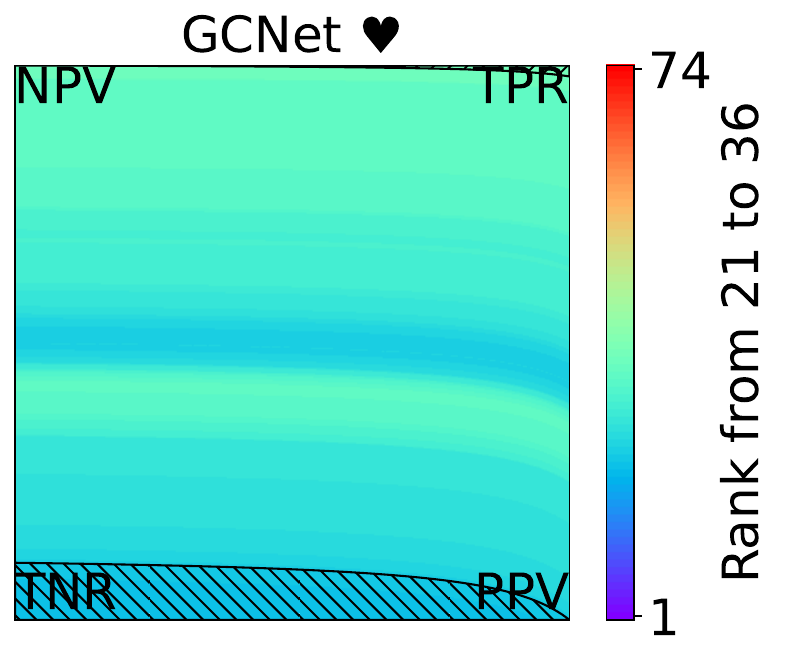}
\includegraphics[scale=0.4]{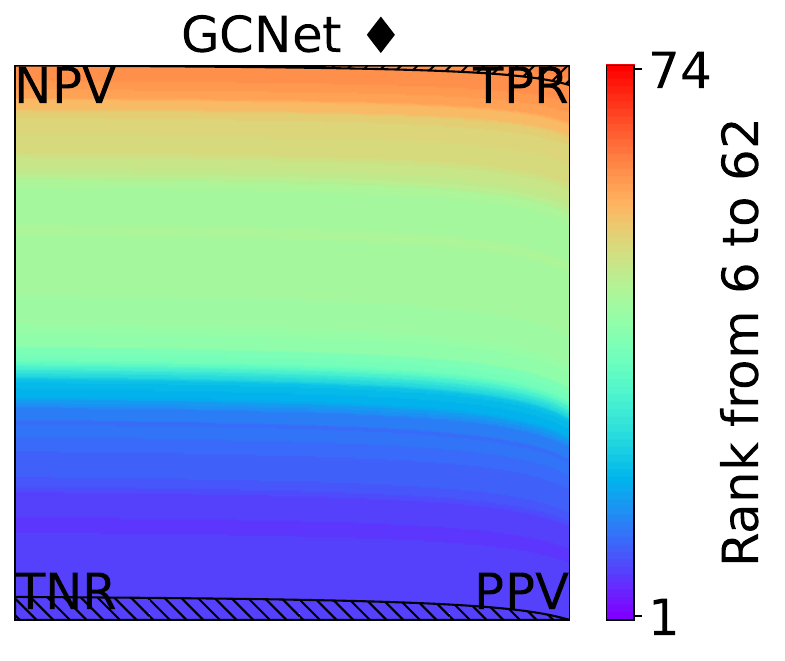}
\includegraphics[scale=0.4]{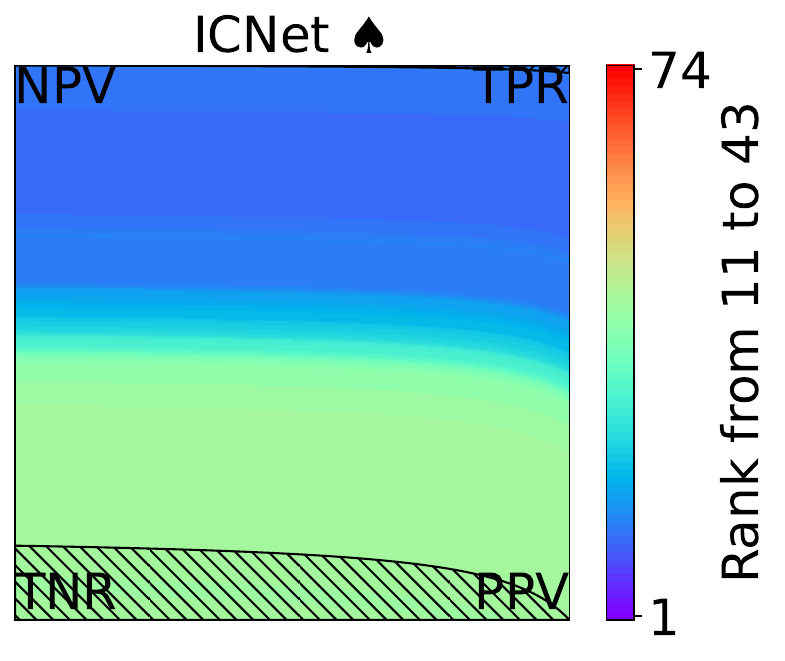}
\includegraphics[scale=0.4]{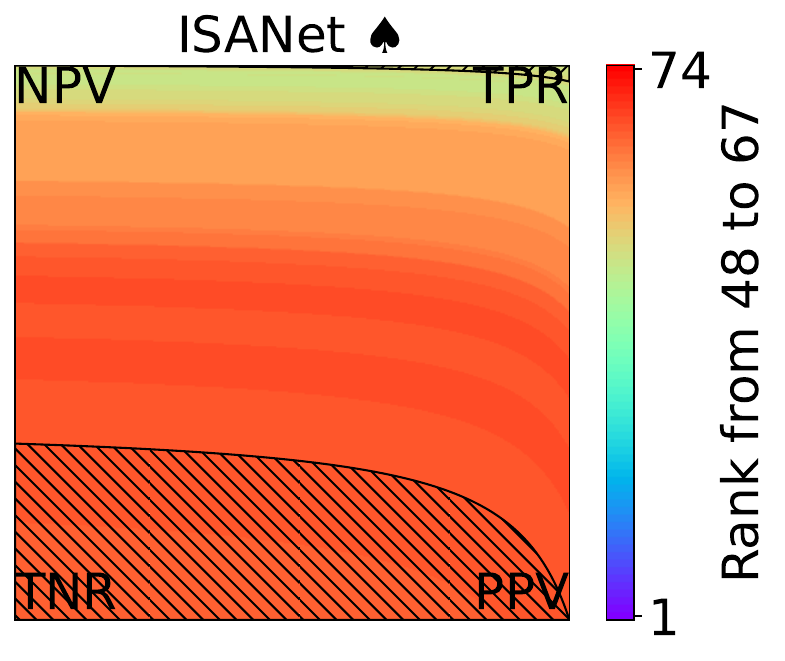}
\includegraphics[scale=0.4]{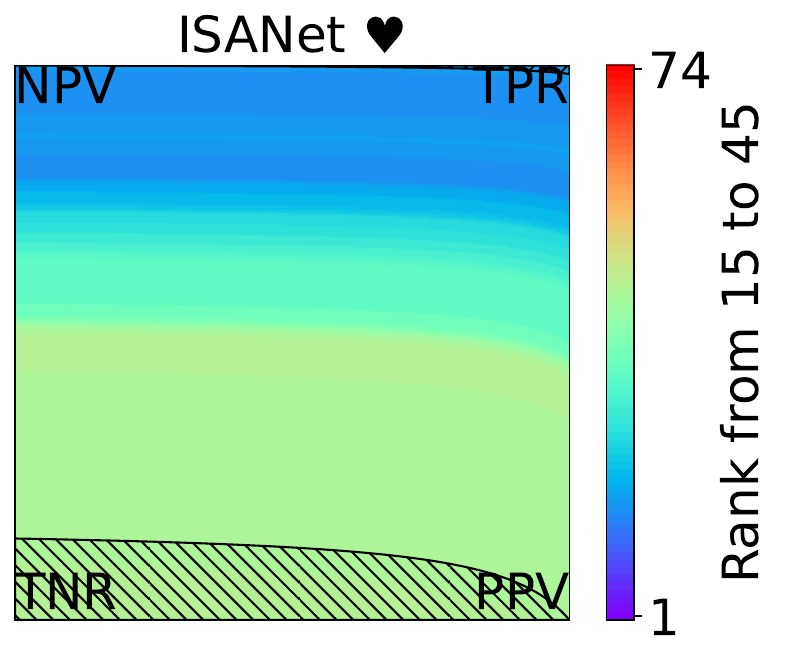}
\includegraphics[scale=0.4]{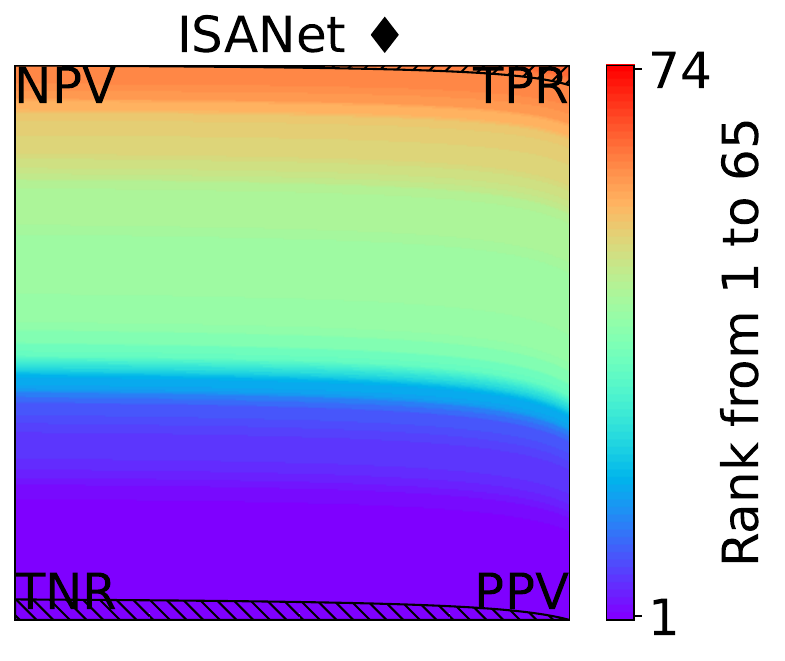}
\includegraphics[scale=0.4]{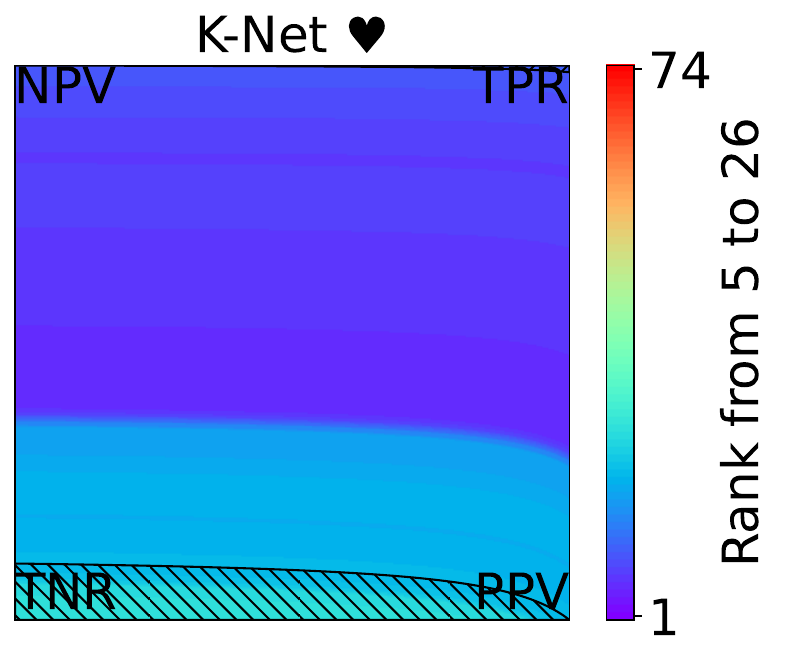}
\includegraphics[scale=0.4]{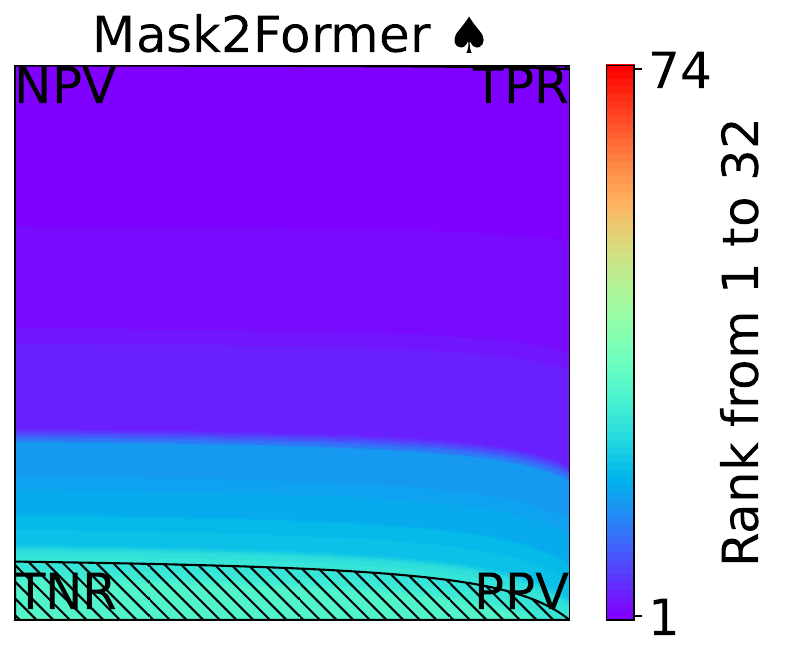}
\includegraphics[scale=0.4]{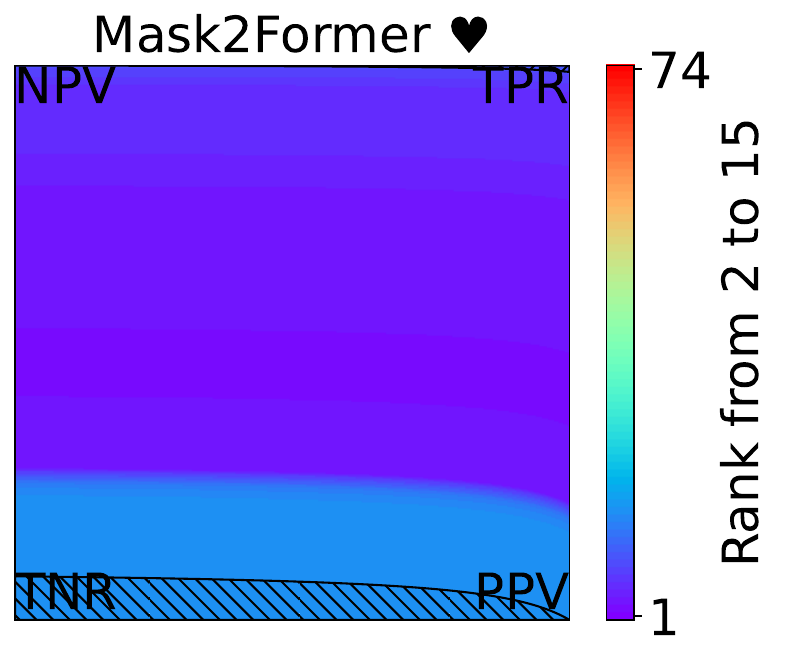}
\includegraphics[scale=0.4]{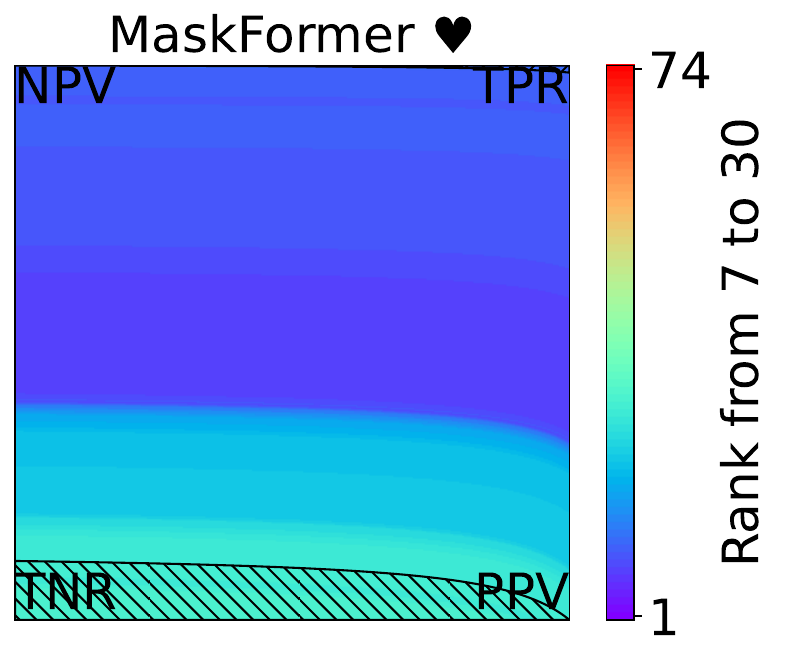}
\includegraphics[scale=0.4]{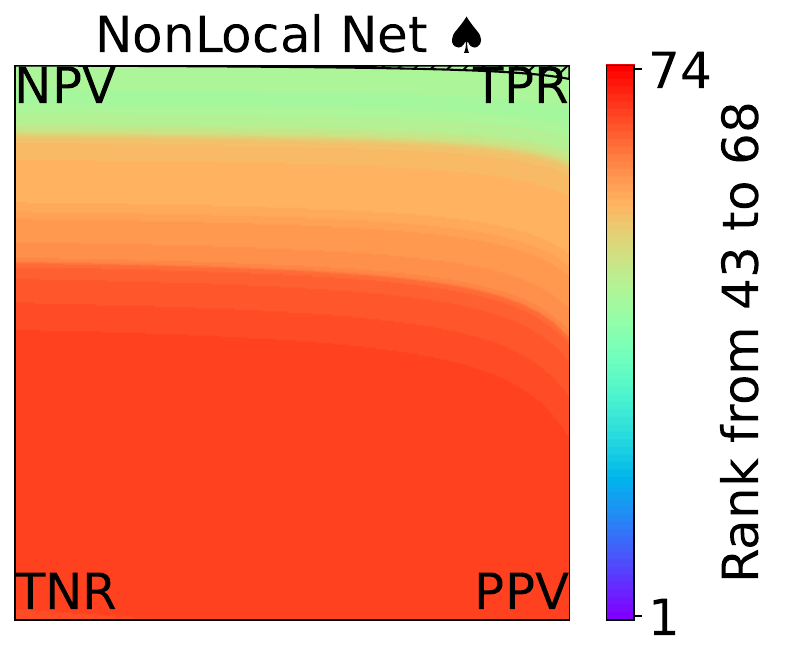}
\includegraphics[scale=0.4]{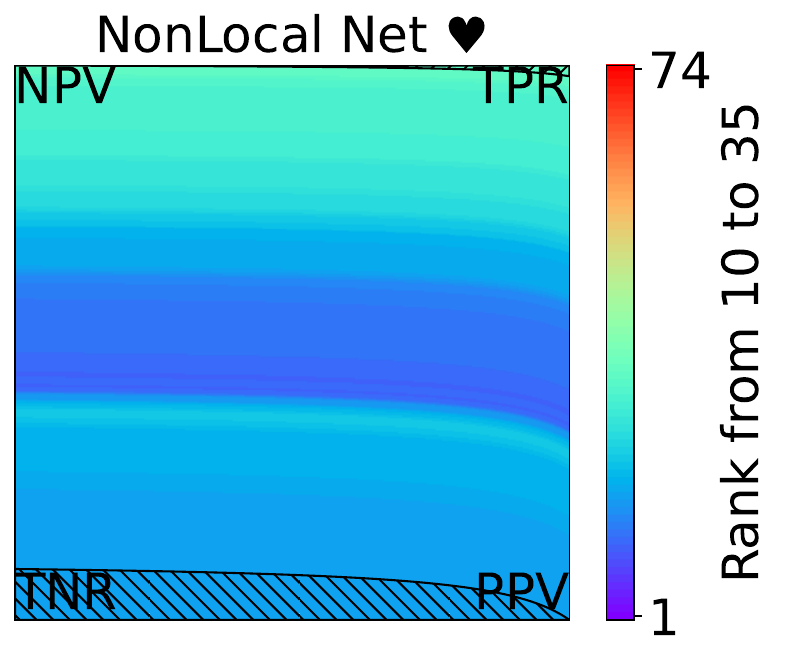}
\includegraphics[scale=0.4]{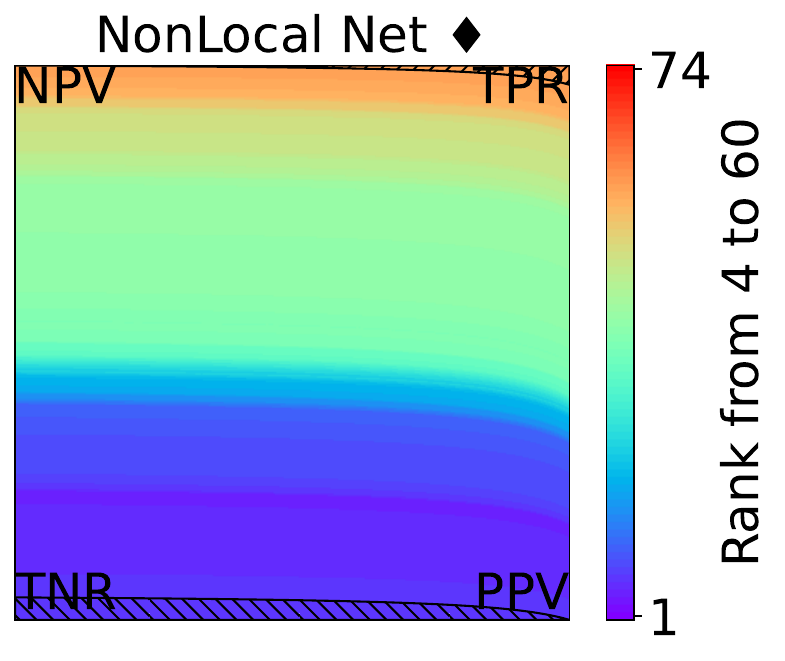}
\includegraphics[scale=0.4]{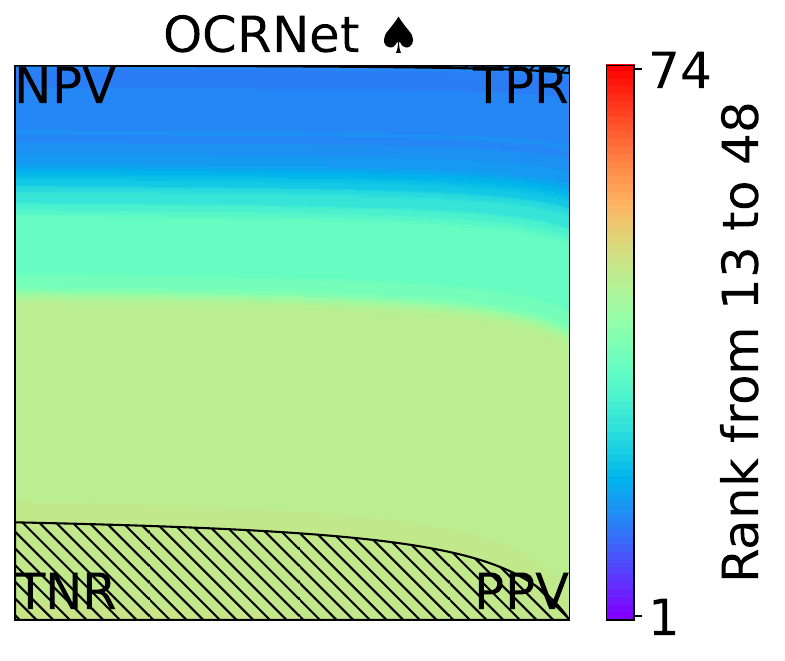}
\includegraphics[scale=0.4]{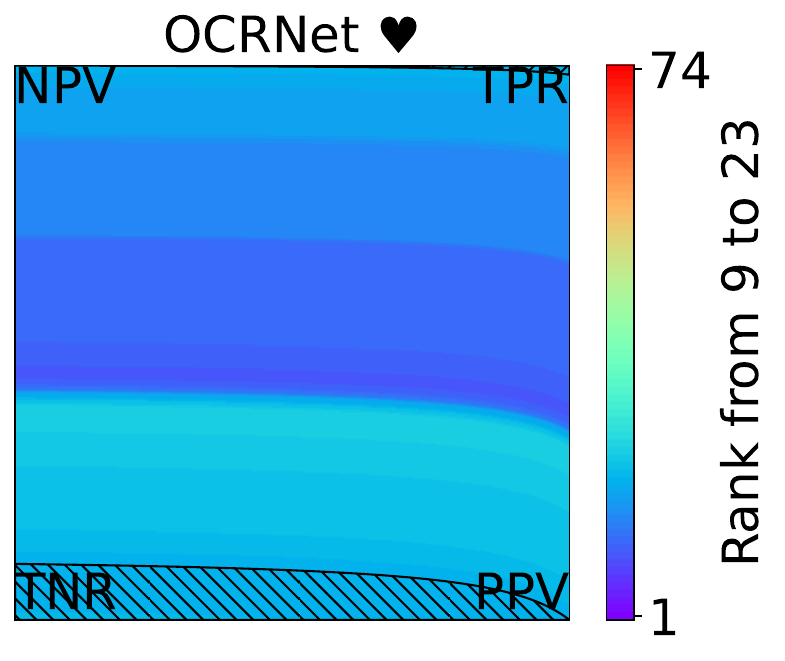}
\includegraphics[scale=0.4]{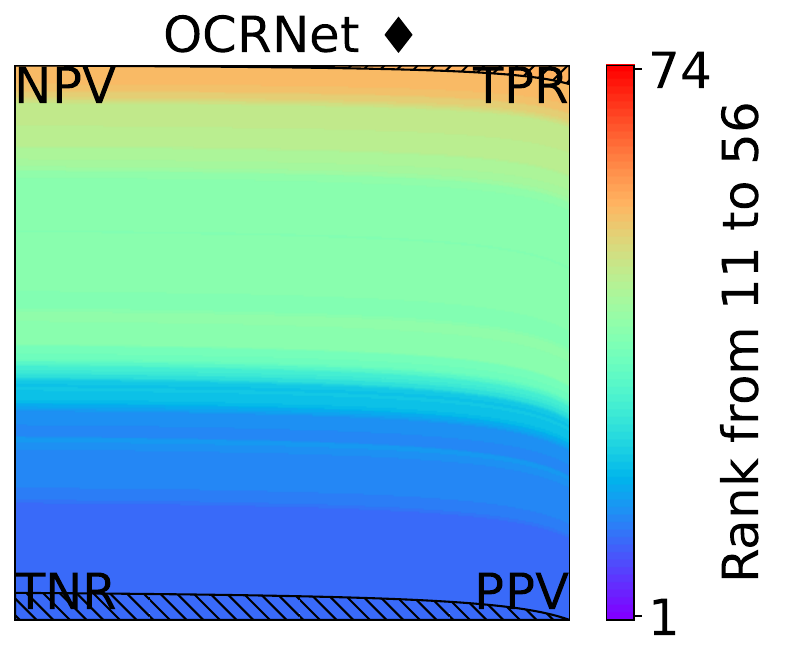}
\includegraphics[scale=0.4]{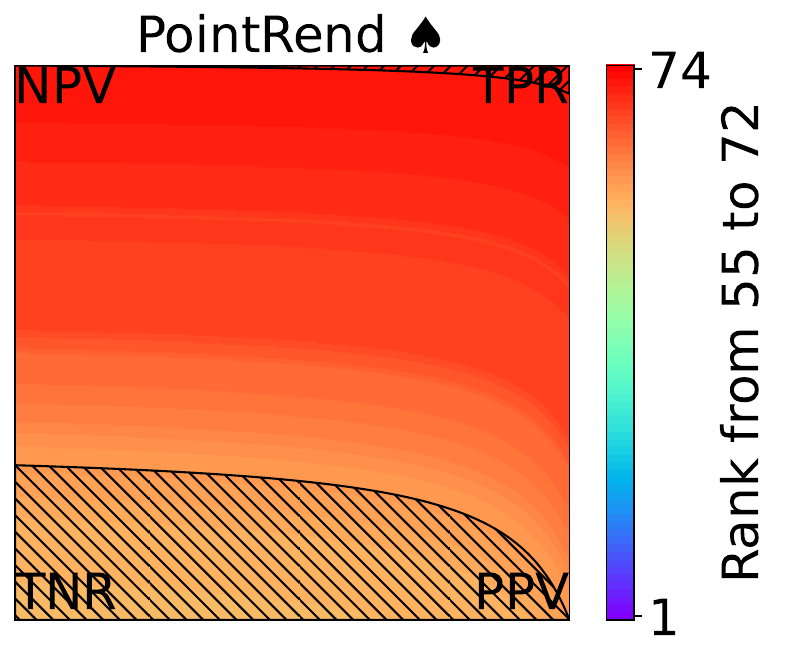}
\includegraphics[scale=0.4]{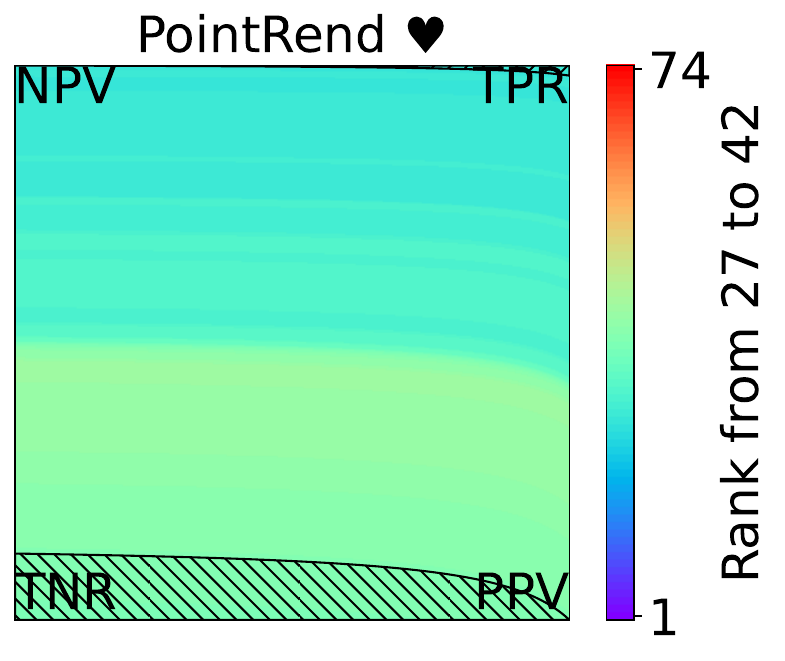}
\includegraphics[scale=0.4]{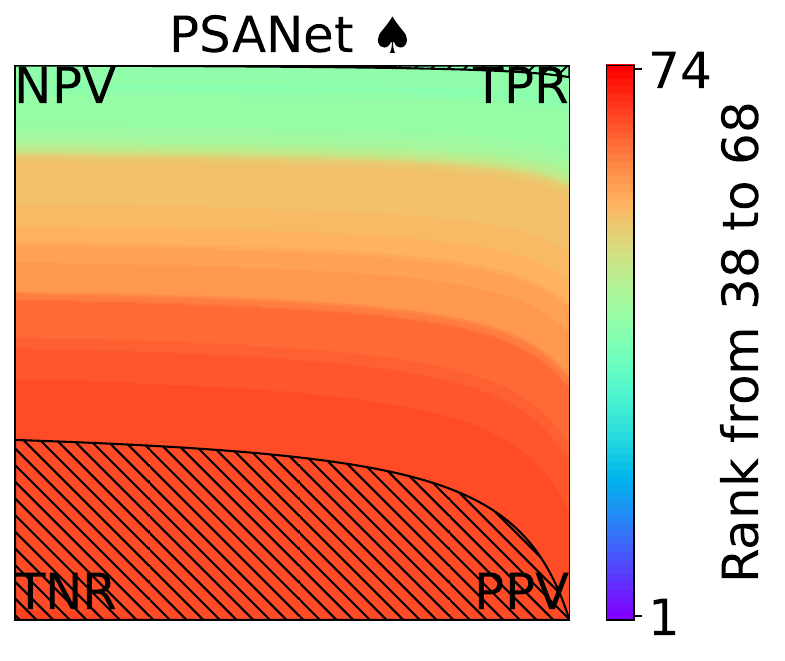}
\includegraphics[scale=0.4]{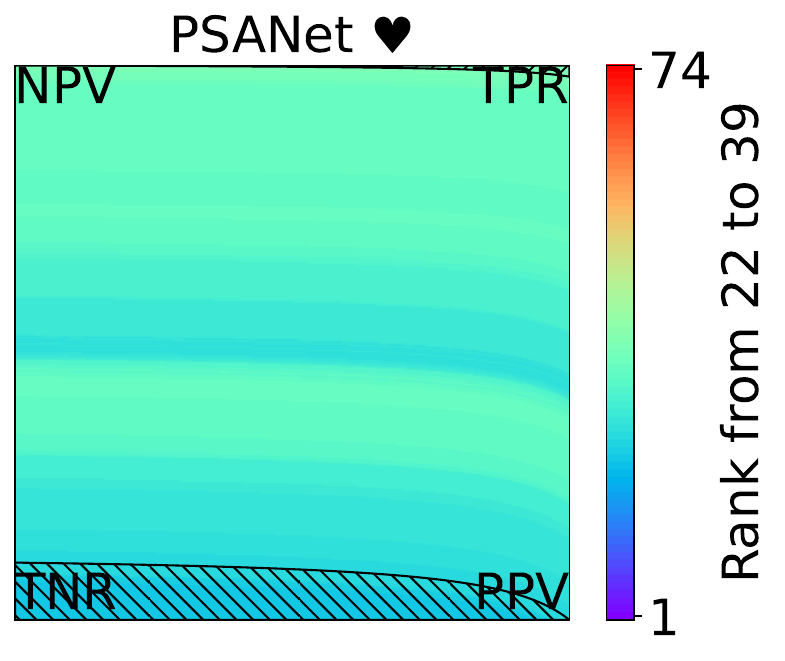}
\includegraphics[scale=0.4]{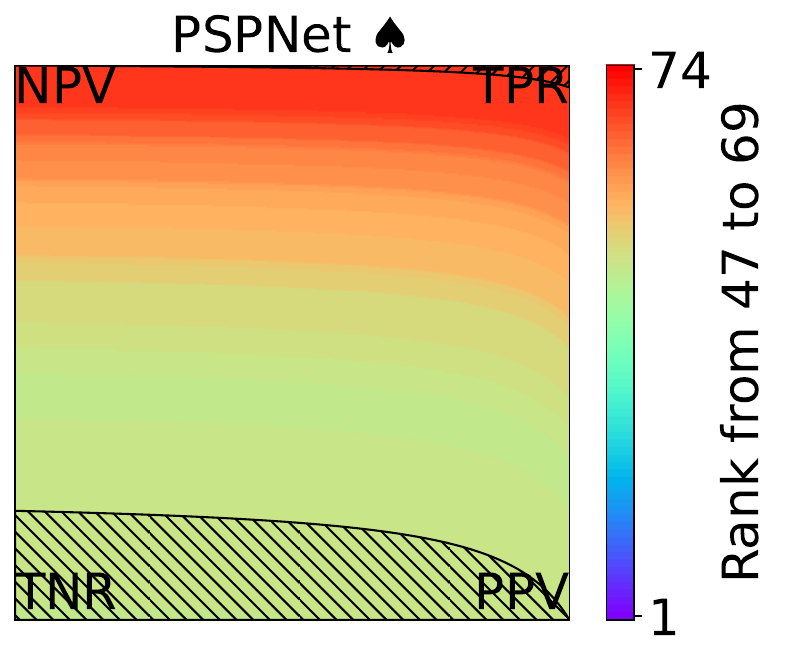}
\includegraphics[scale=0.4]{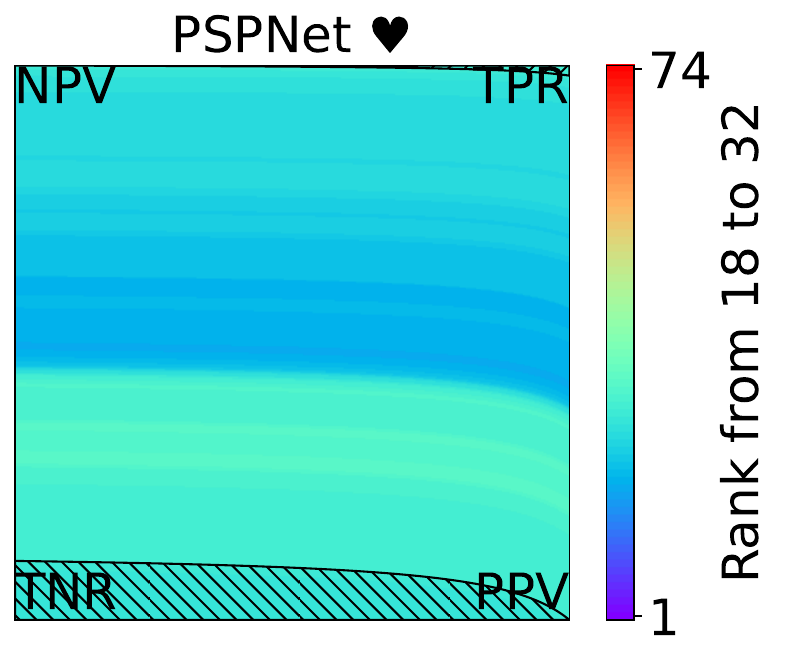}
\includegraphics[scale=0.4]{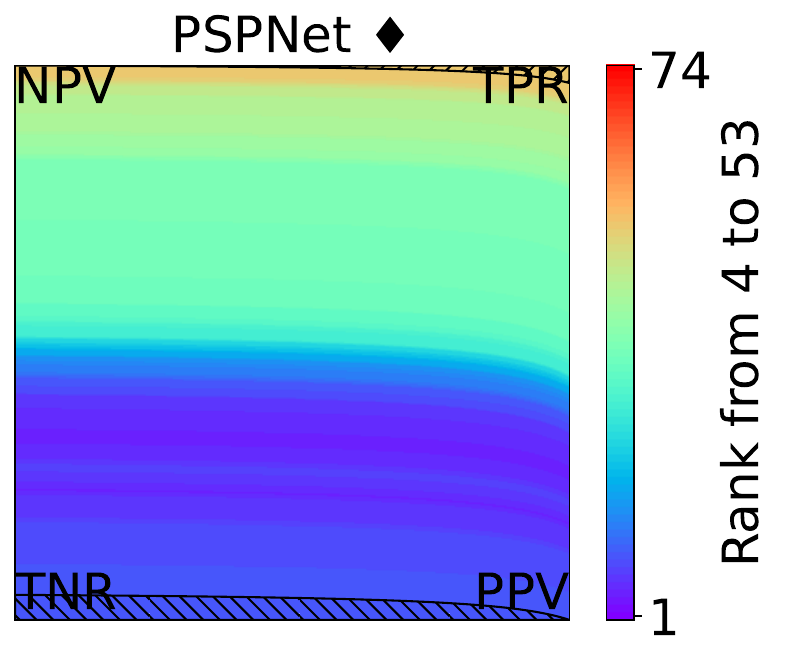}
\includegraphics[scale=0.4]{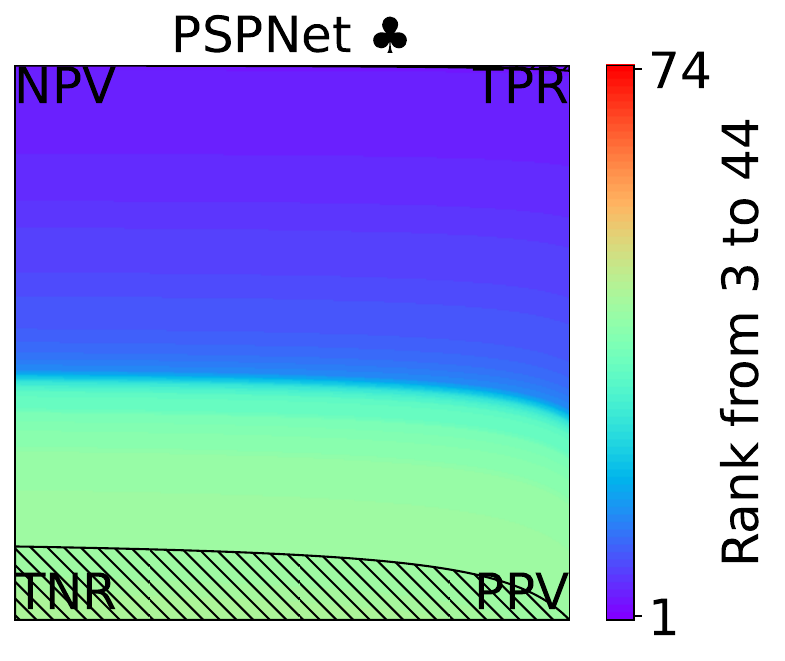}
\includegraphics[scale=0.4]{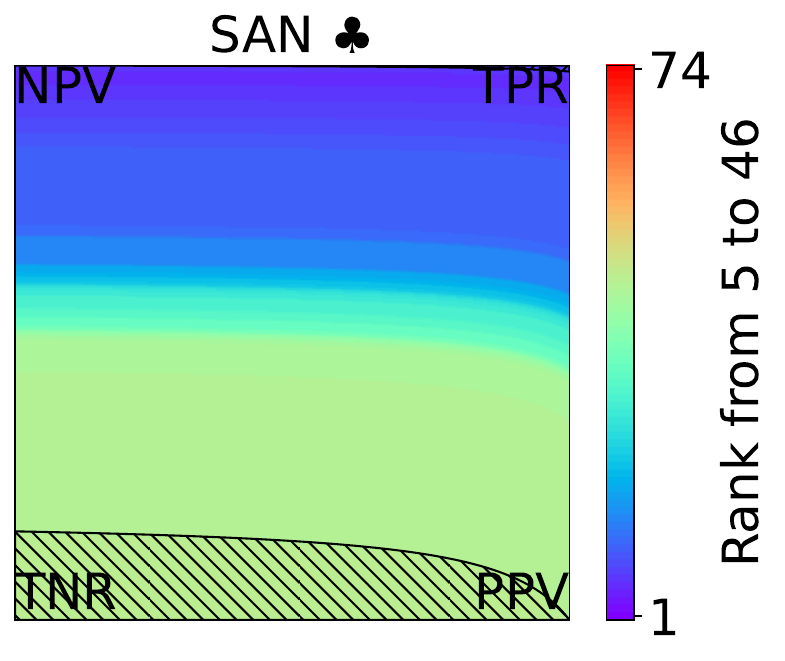}
\includegraphics[scale=0.4]{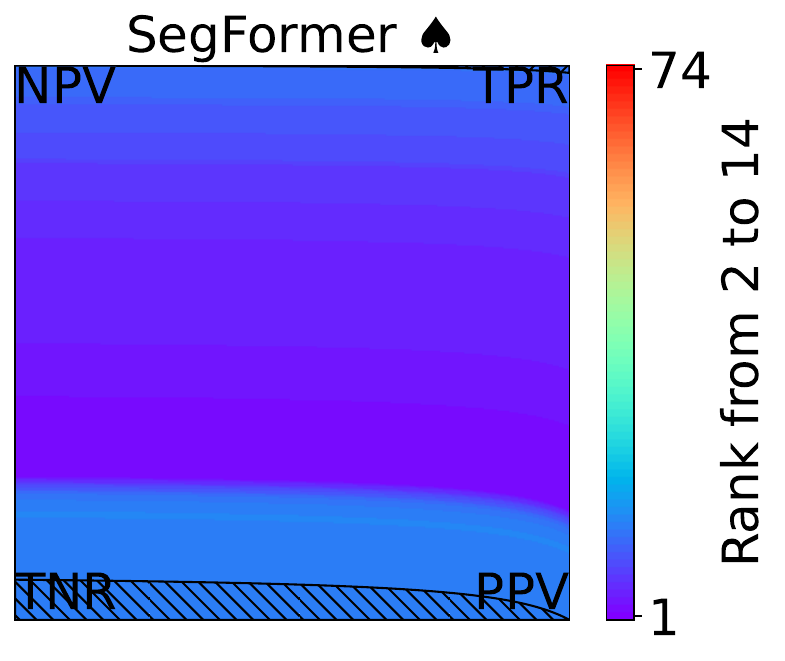}
\includegraphics[scale=0.4]{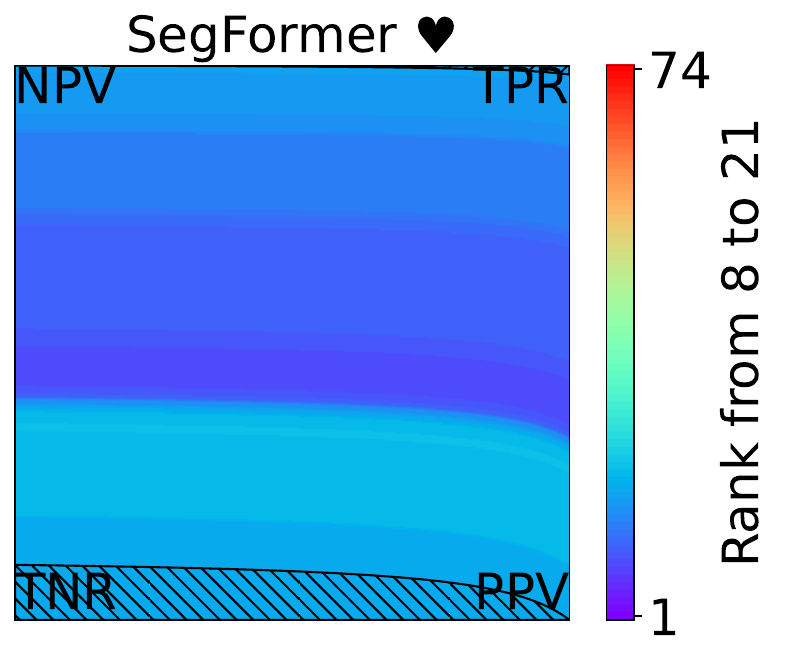}
\includegraphics[scale=0.4]{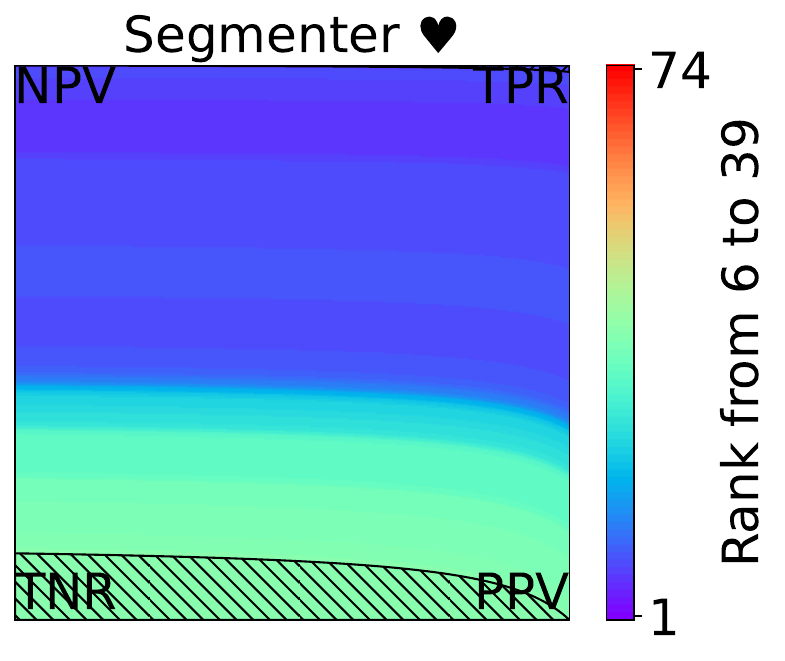}
\includegraphics[scale=0.4]{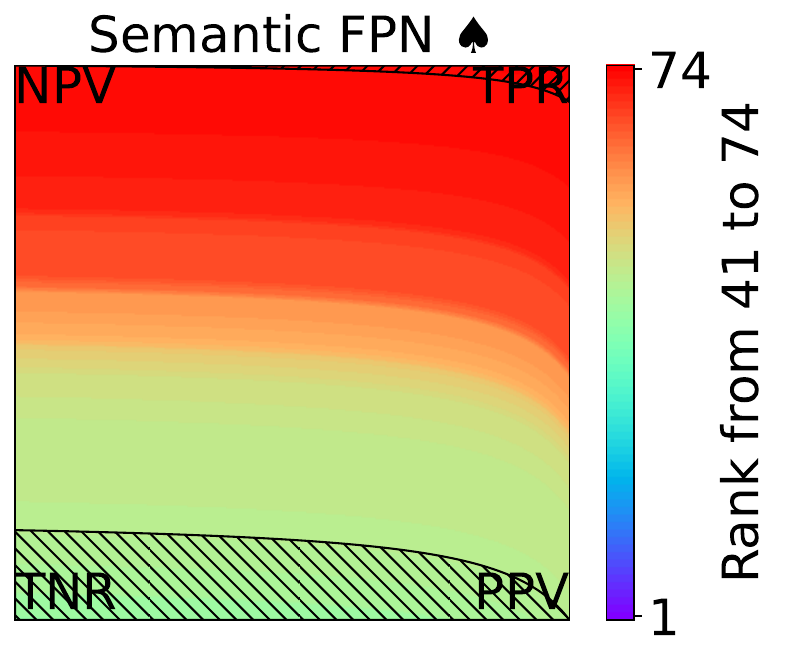}
\includegraphics[scale=0.4]{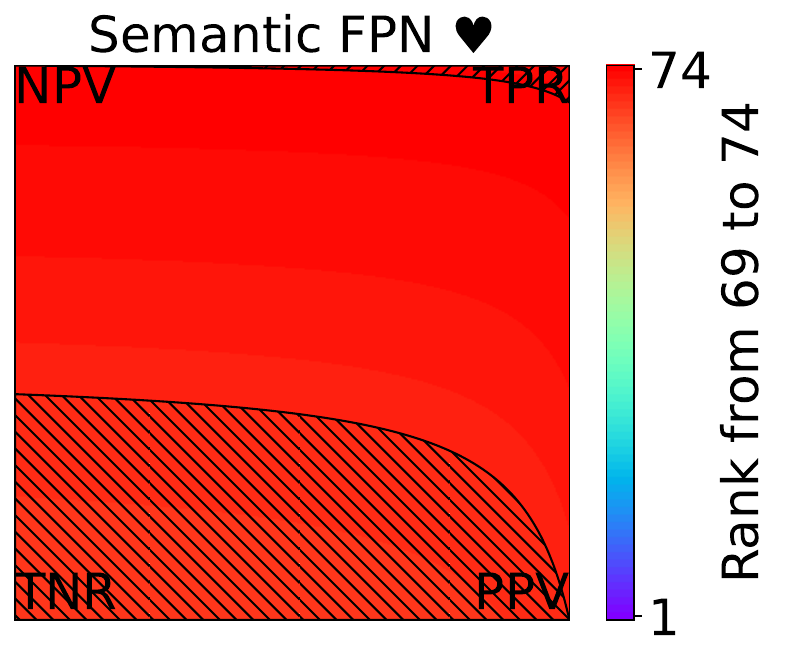}
\includegraphics[scale=0.4]{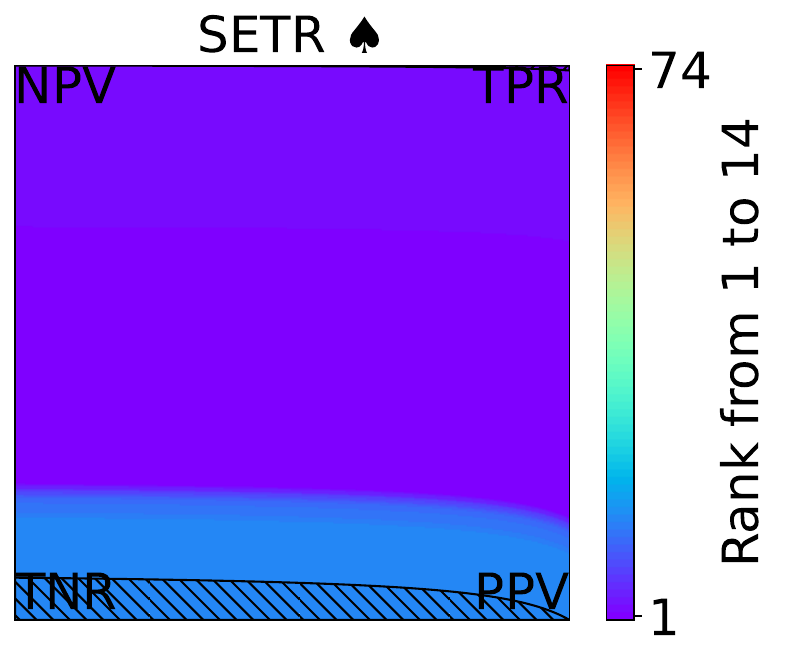}
\includegraphics[scale=0.4]{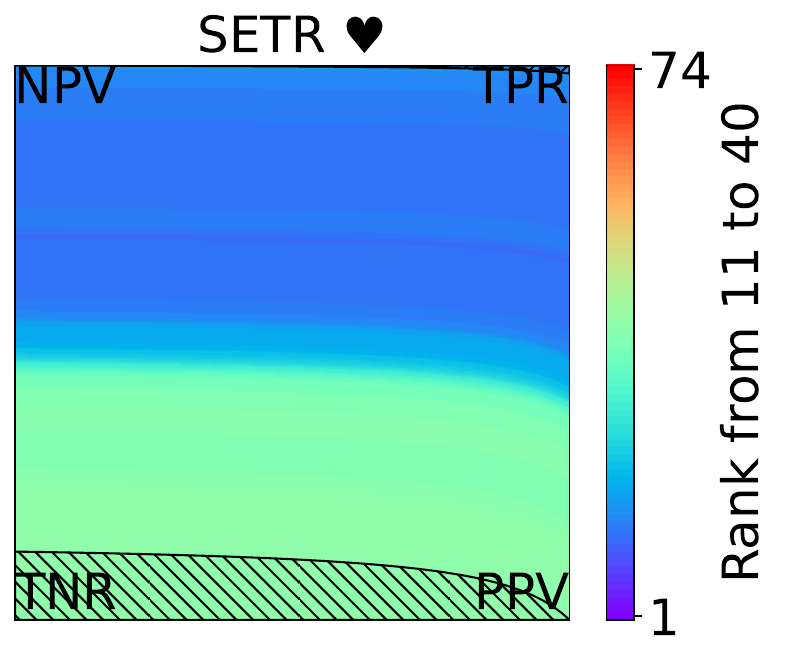}
\includegraphics[scale=0.4]{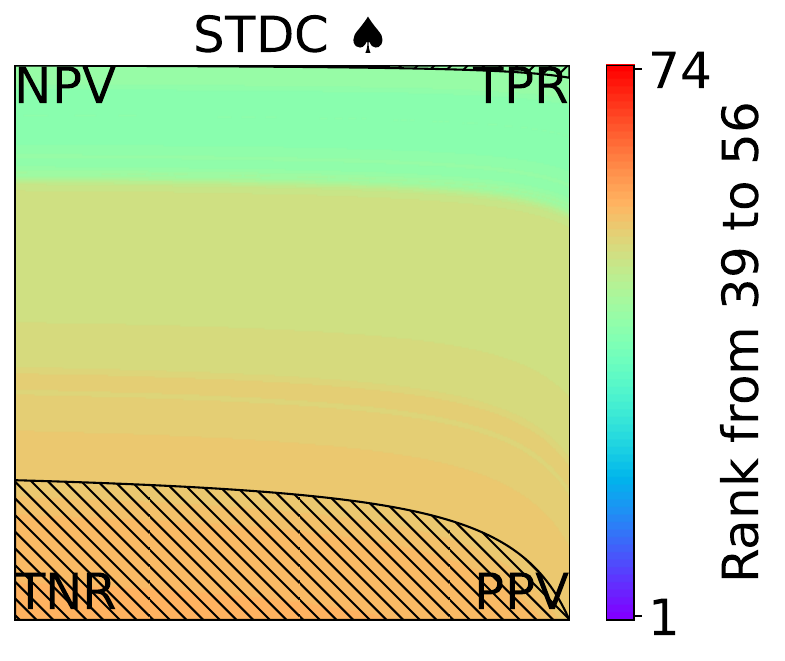}
\includegraphics[scale=0.4]{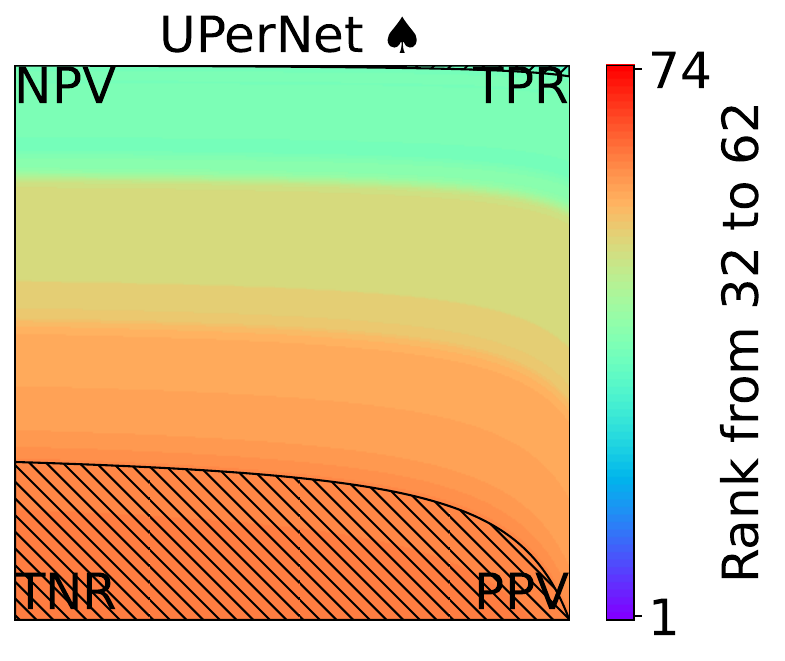}
\includegraphics[scale=0.4]{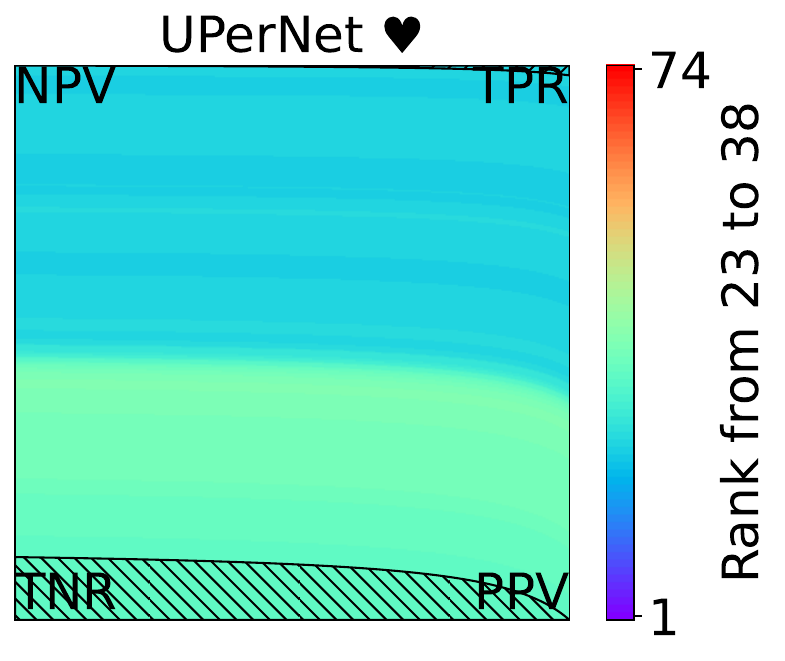}
\includegraphics[scale=0.4]{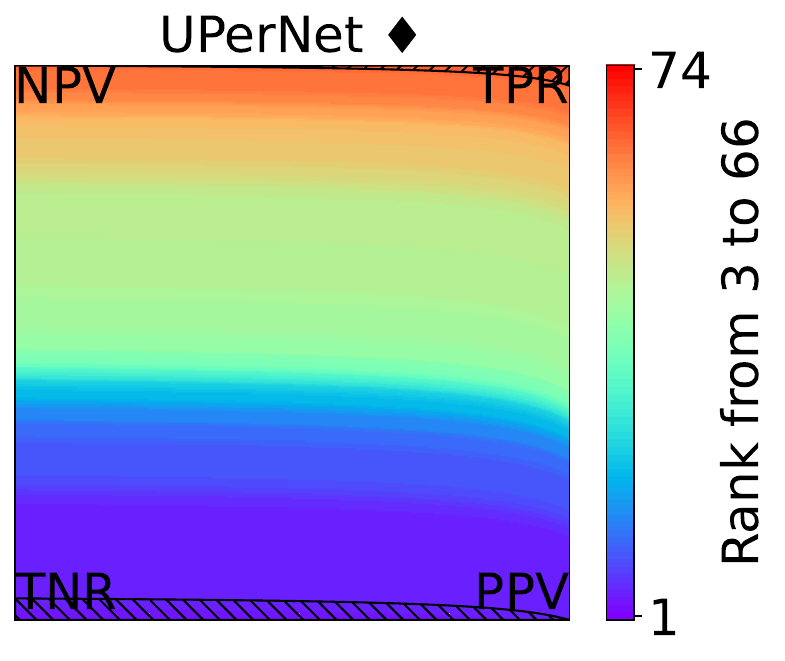}
\par\end{center}
\paragraph{Analysis.}

The following entities mimimize the maximum rank:
\begin{itemize}
    \item  SegFormer \ding{171} (max rank: 14, mean rank: 6.85568)
    \item  SETR \ding{171} (max rank: 14, mean rank: 4.04938)
\end{itemize}

Among them, the following entities have the lowest mean rank:
\begin{itemize}
    \item SETR \ding{171}
\end{itemize}

\subsection*{Entity Tile: Using the tile to show the entities for each rank}
\begin{center}
\includegraphics[scale=0.4]{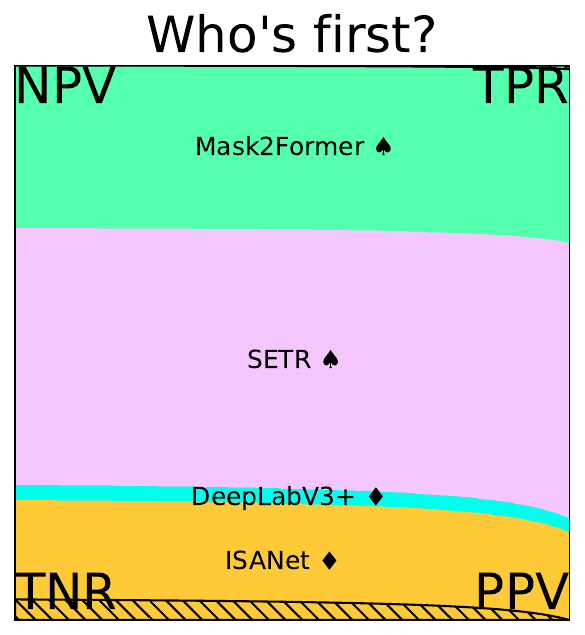}
\includegraphics[scale=0.4]{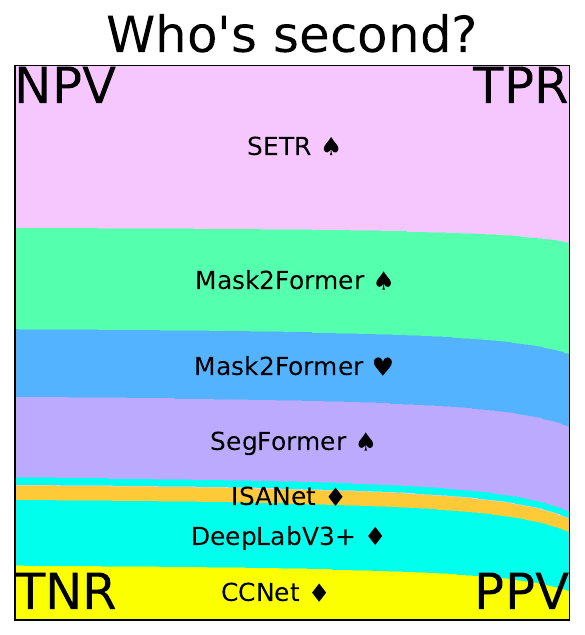}
\includegraphics[scale=0.4]{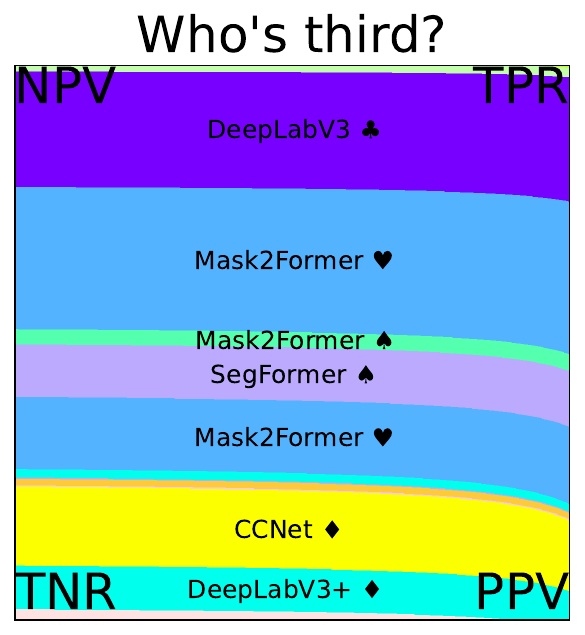}
\includegraphics[scale=0.4]{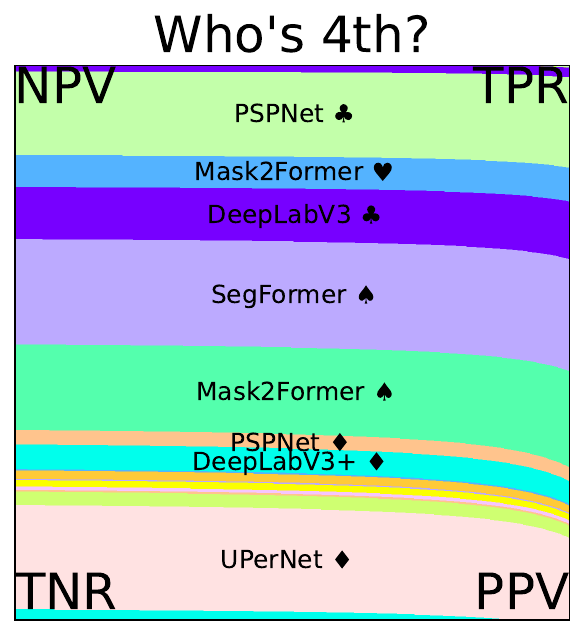}
\includegraphics[scale=0.4]{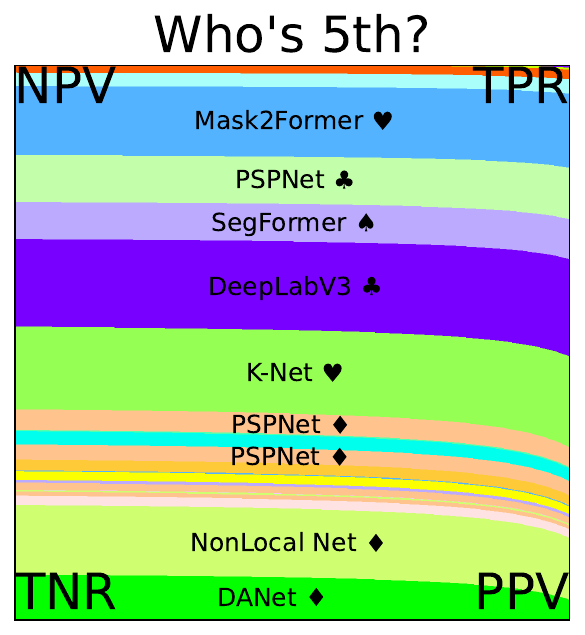}
\includegraphics[scale=0.4]{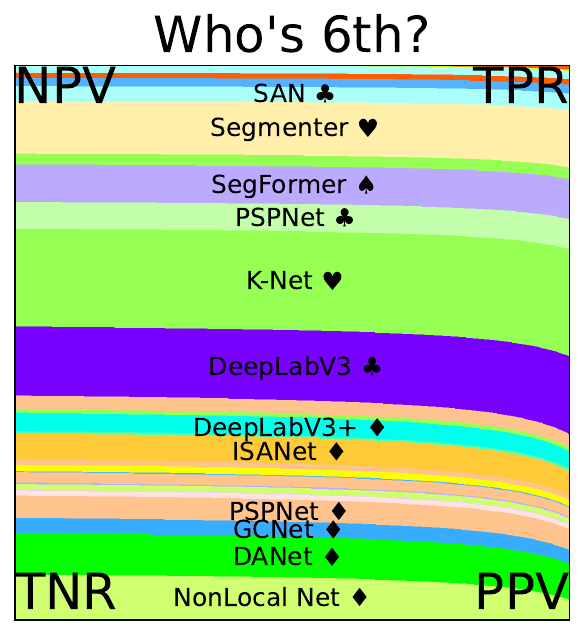}
\includegraphics[scale=0.4]{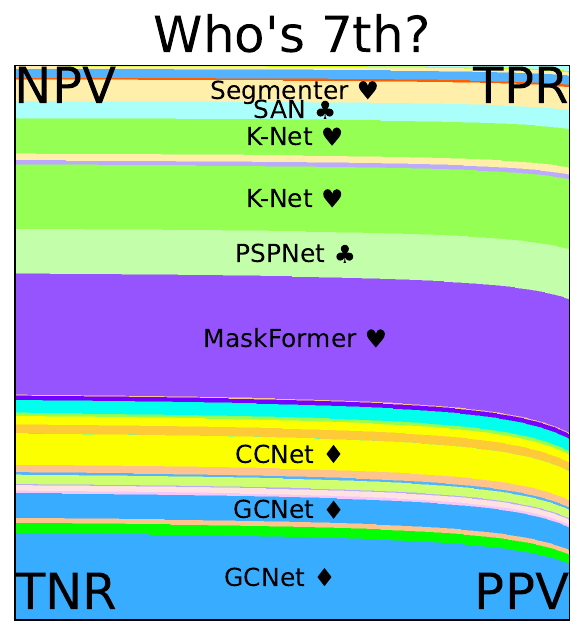}
\includegraphics[scale=0.4]{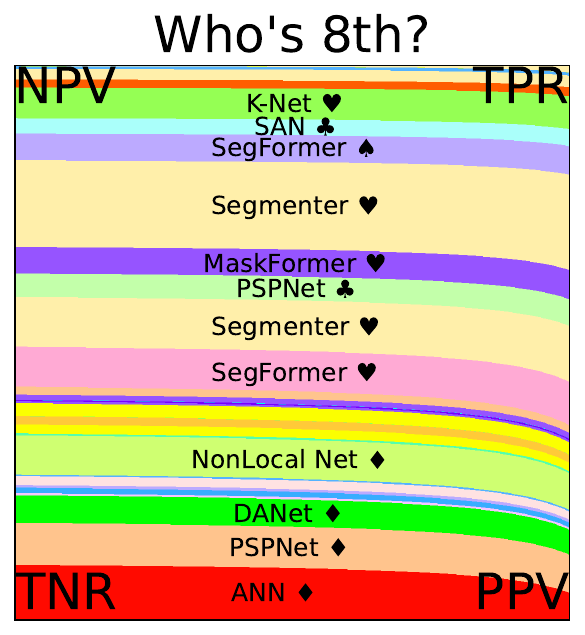}
\includegraphics[scale=0.4]{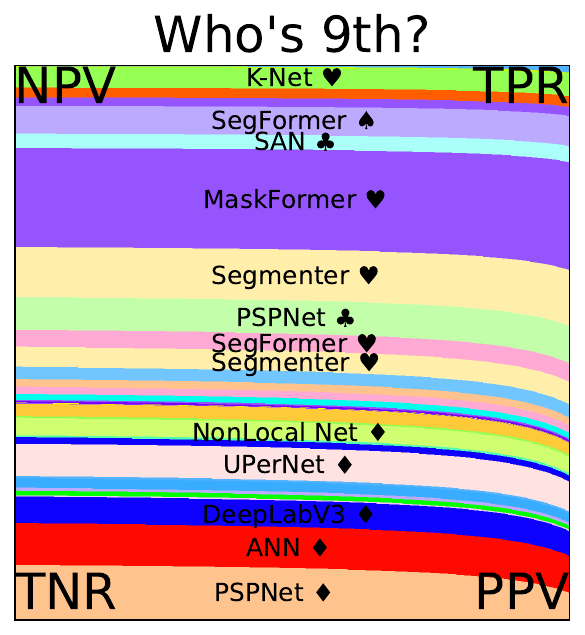}
\includegraphics[scale=0.4]{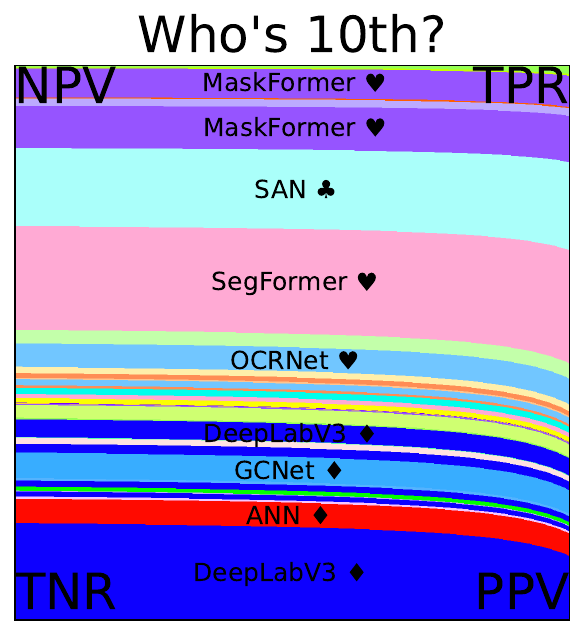}
\includegraphics[scale=0.4]{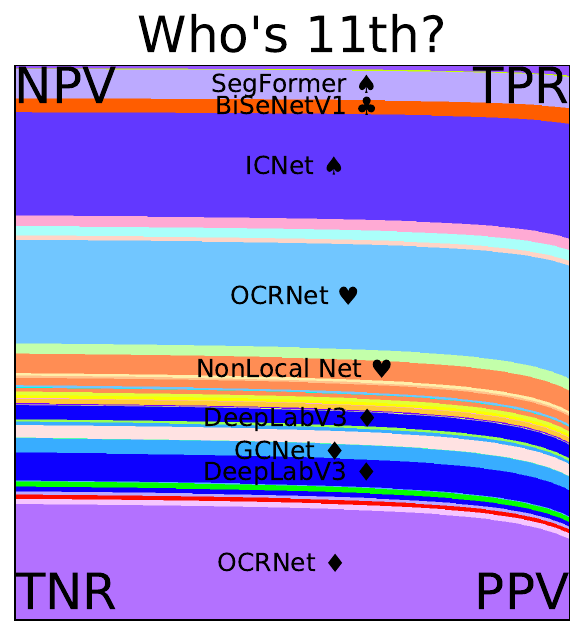}
\includegraphics[scale=0.4]{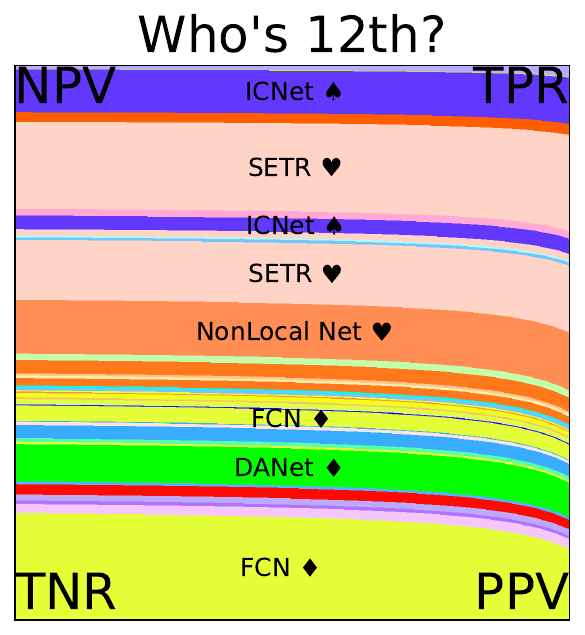}
\includegraphics[scale=0.4]{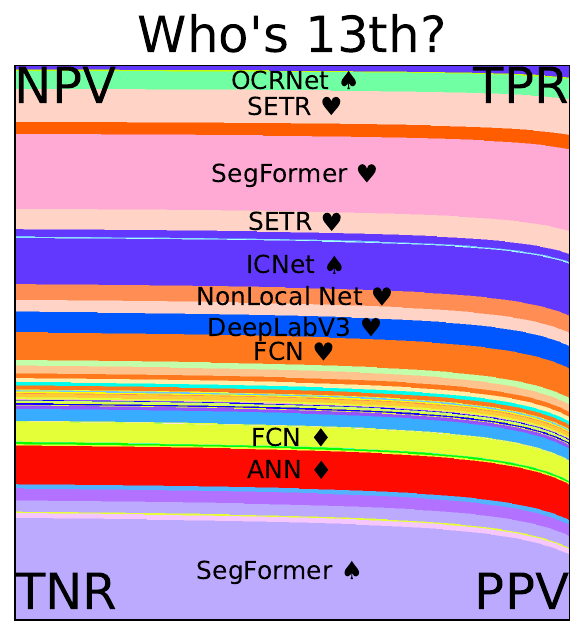}
\includegraphics[scale=0.4]{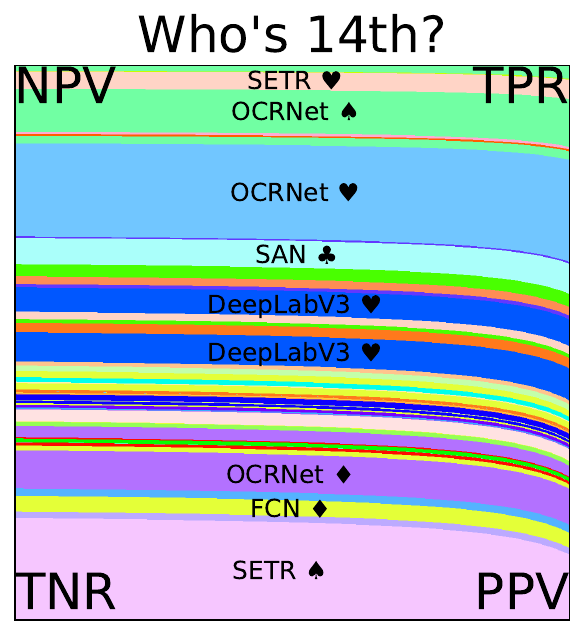}
\includegraphics[scale=0.4]{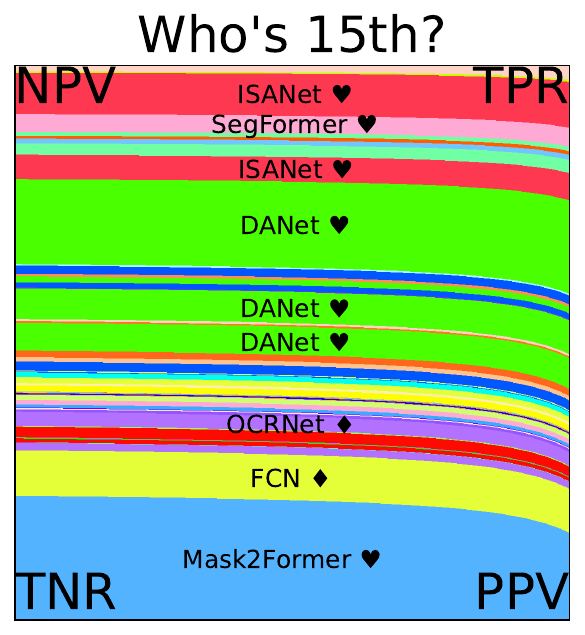}
\includegraphics[scale=0.4]{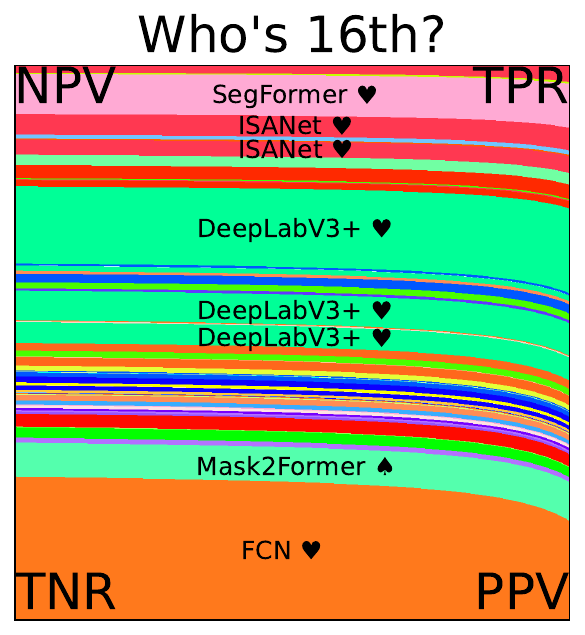}
\includegraphics[scale=0.4]{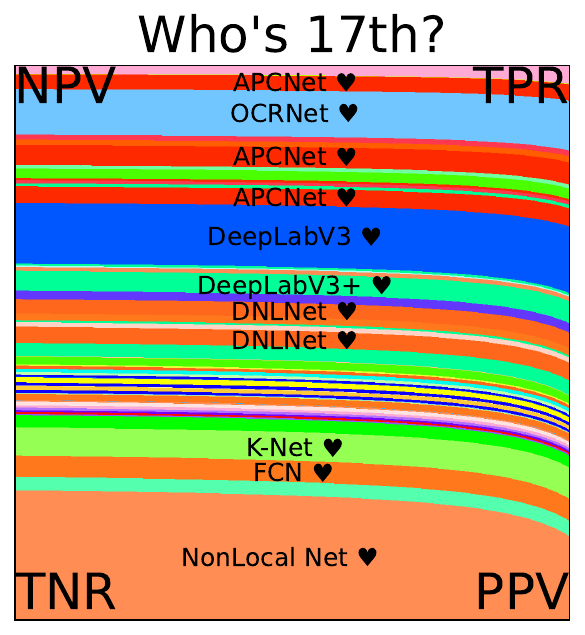}
\includegraphics[scale=0.4]{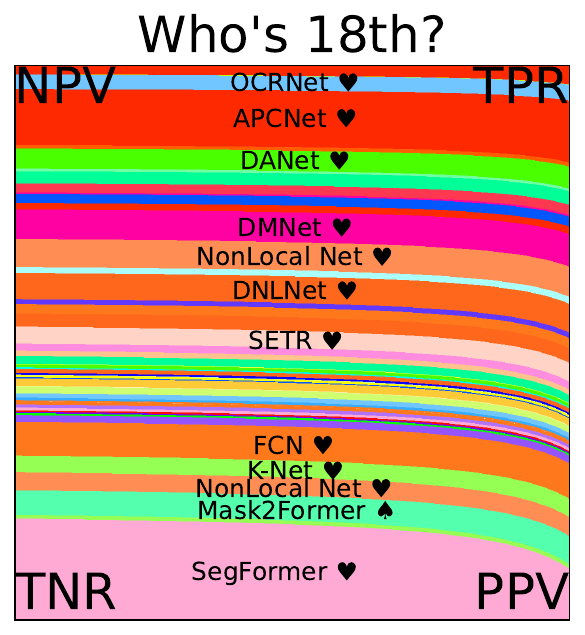}
\includegraphics[scale=0.4]{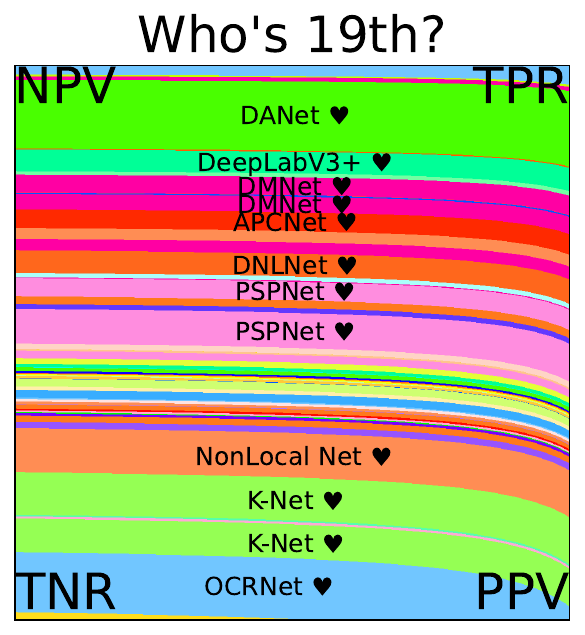}
\includegraphics[scale=0.4]{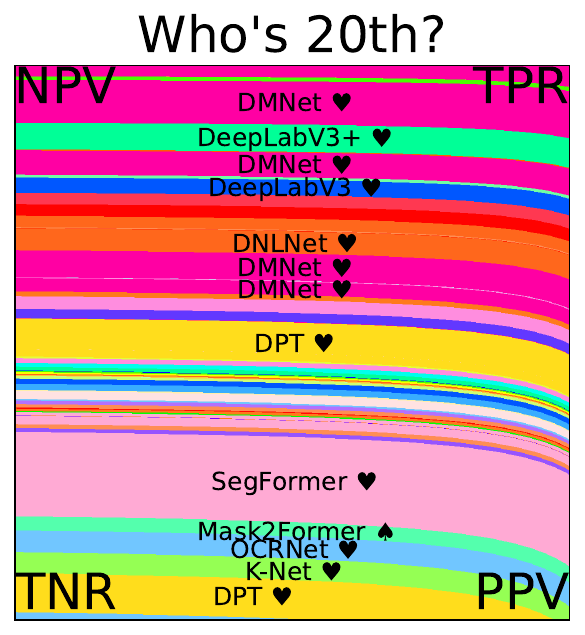}
\includegraphics[scale=0.4]{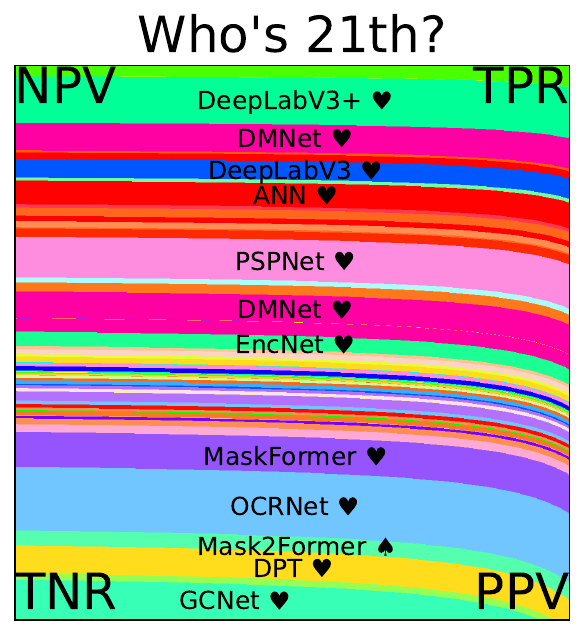}
\includegraphics[scale=0.4]{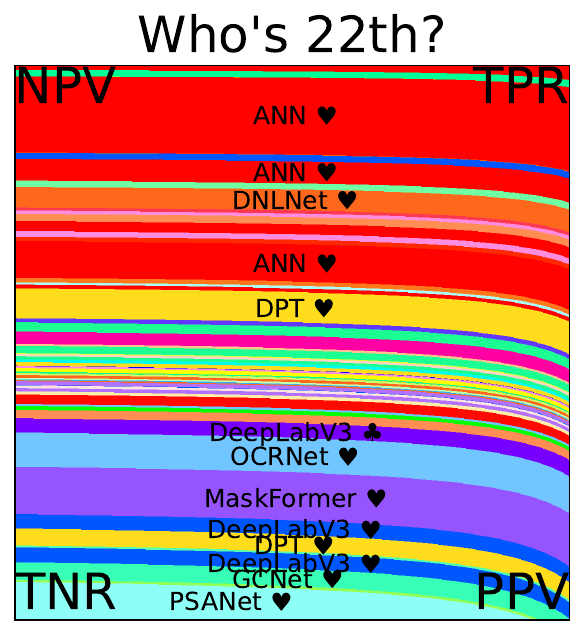}
\includegraphics[scale=0.4]{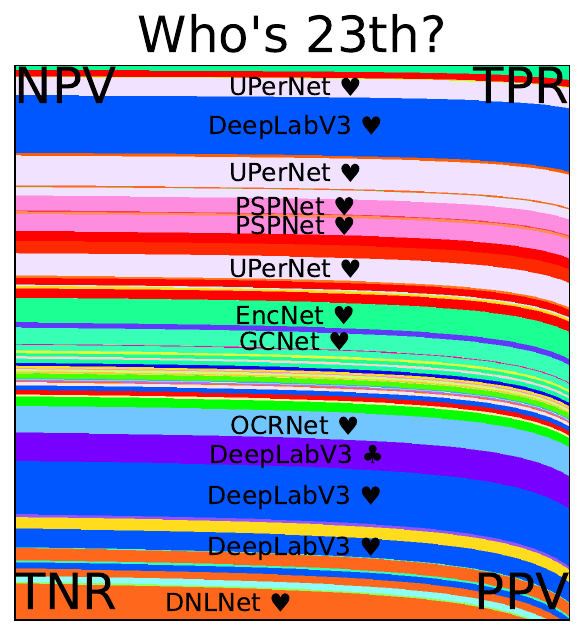}
\includegraphics[scale=0.4]{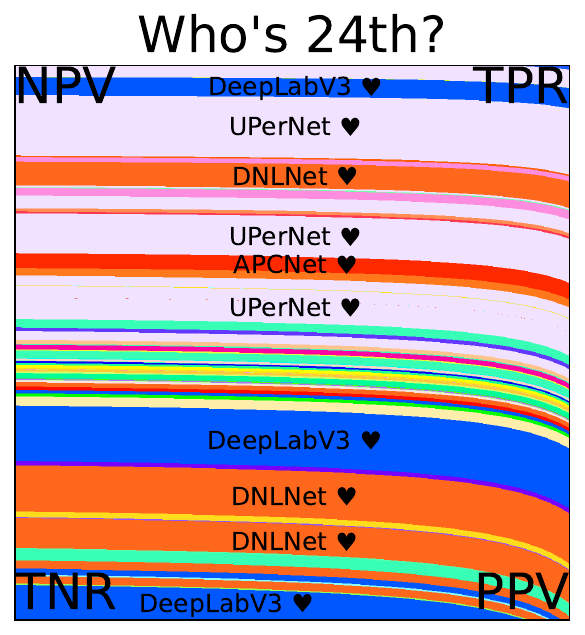}
\includegraphics[scale=0.4]{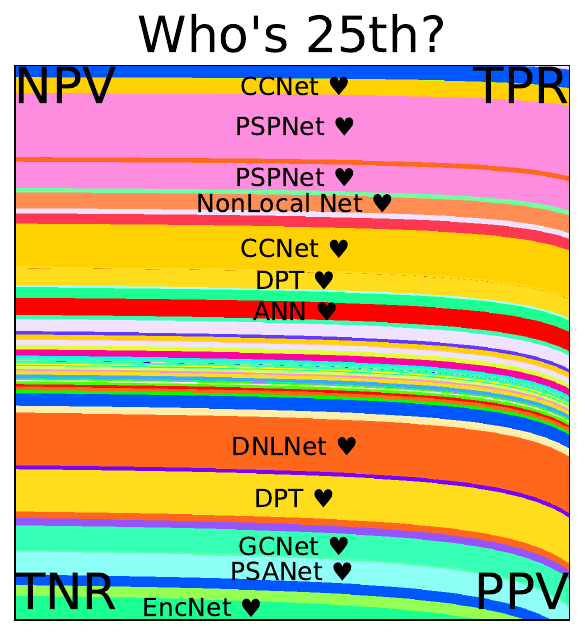}
\includegraphics[scale=0.4]{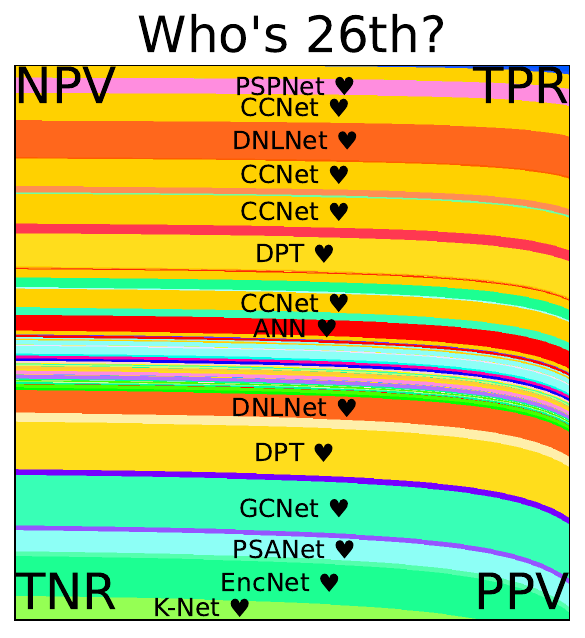}
\includegraphics[scale=0.4]{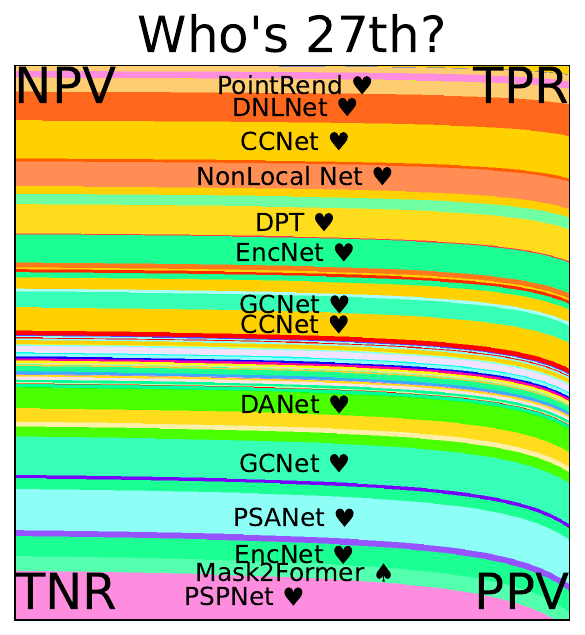}
\includegraphics[scale=0.4]{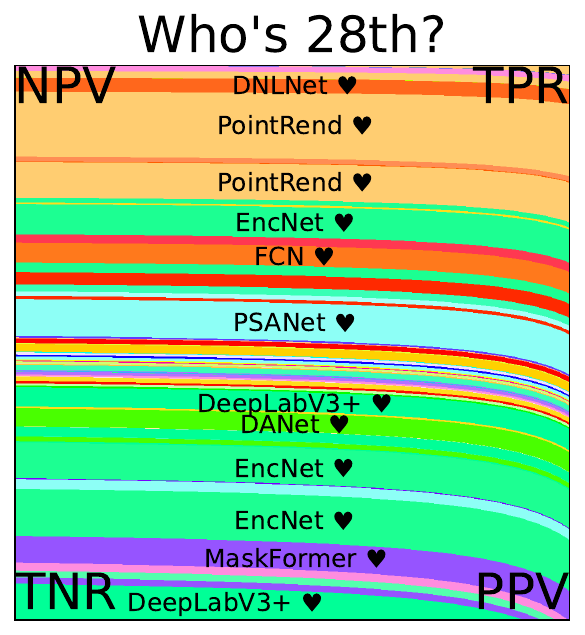}
\includegraphics[scale=0.4]{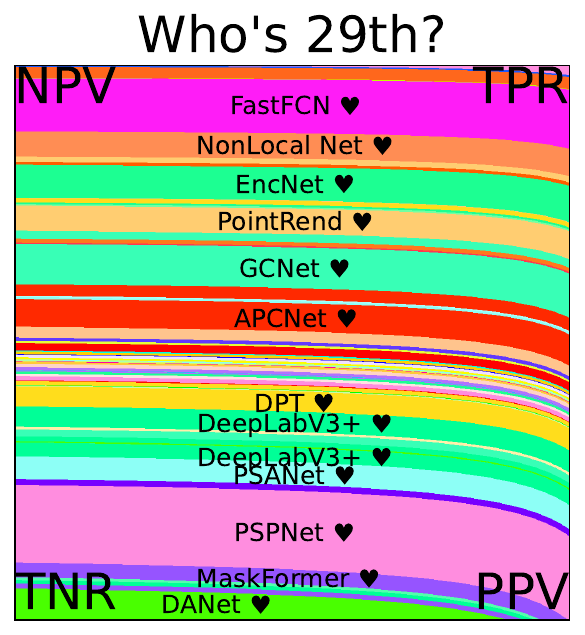}
\includegraphics[scale=0.4]{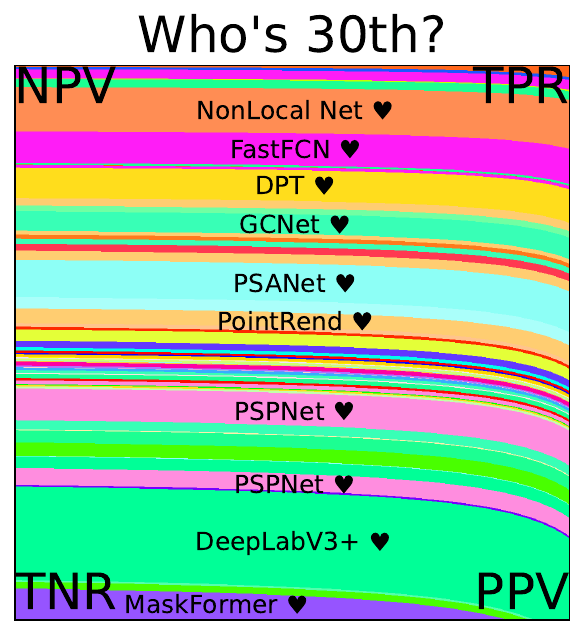}
\includegraphics[scale=0.4]{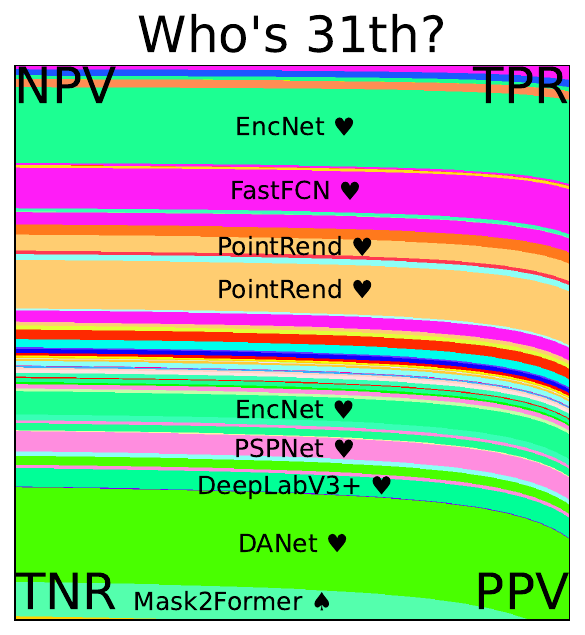}
\includegraphics[scale=0.4]{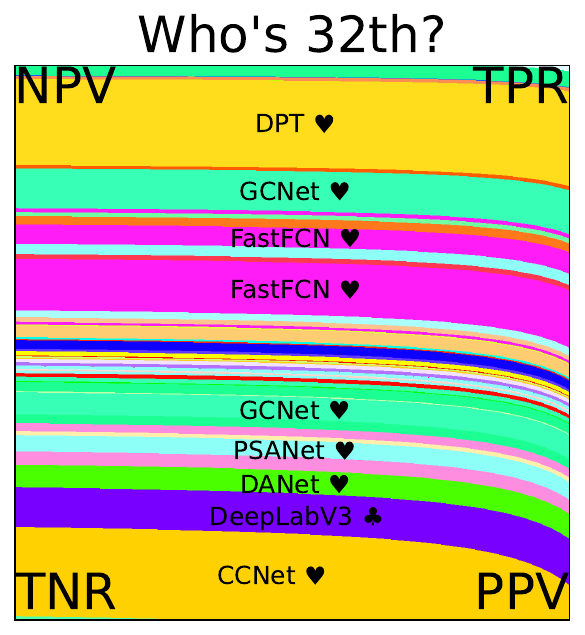}
\includegraphics[scale=0.4]{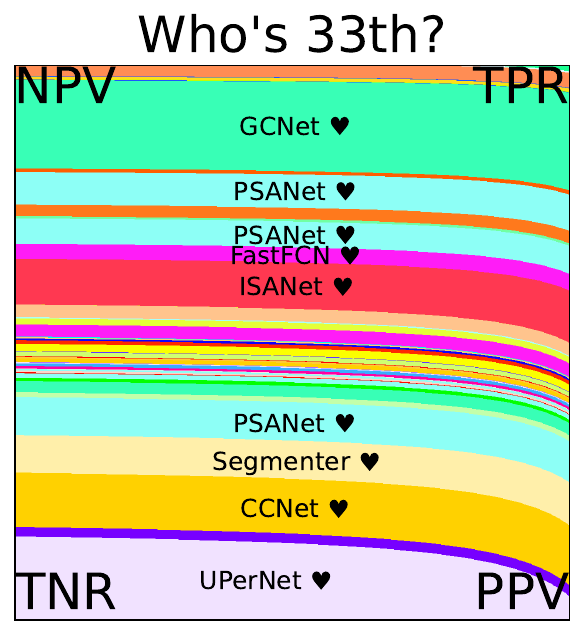}
\includegraphics[scale=0.4]{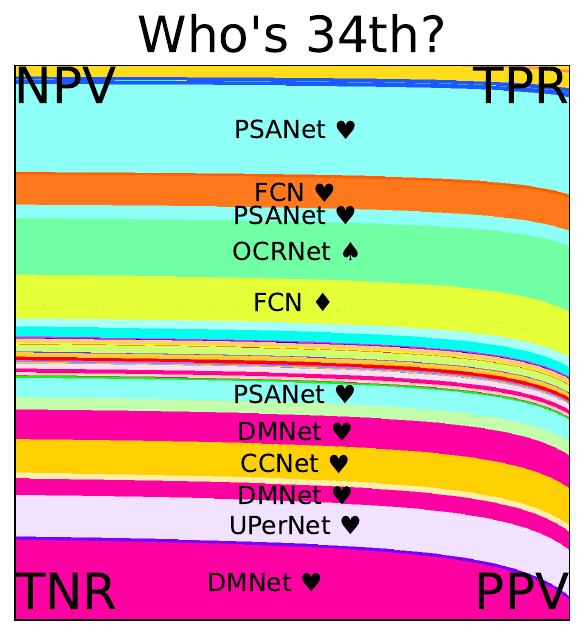}
\includegraphics[scale=0.4]{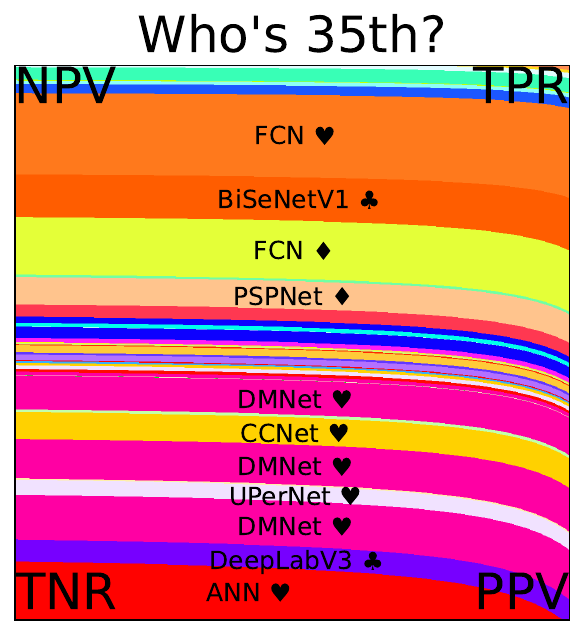}
\includegraphics[scale=0.4]{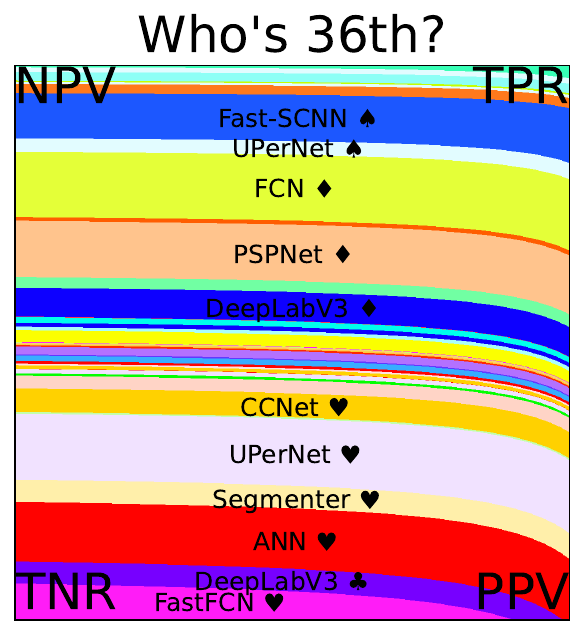}
\includegraphics[scale=0.4]{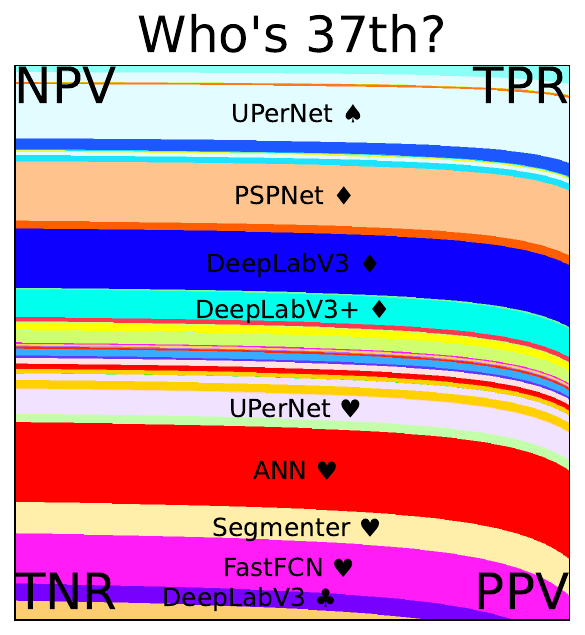}
\includegraphics[scale=0.4]{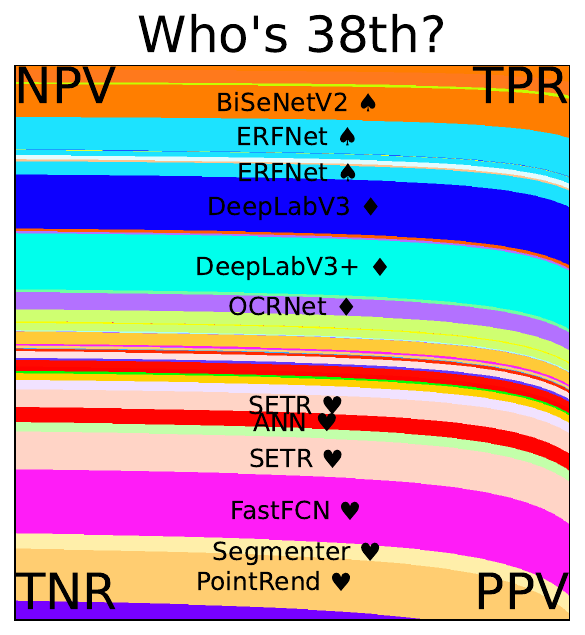}
\includegraphics[scale=0.4]{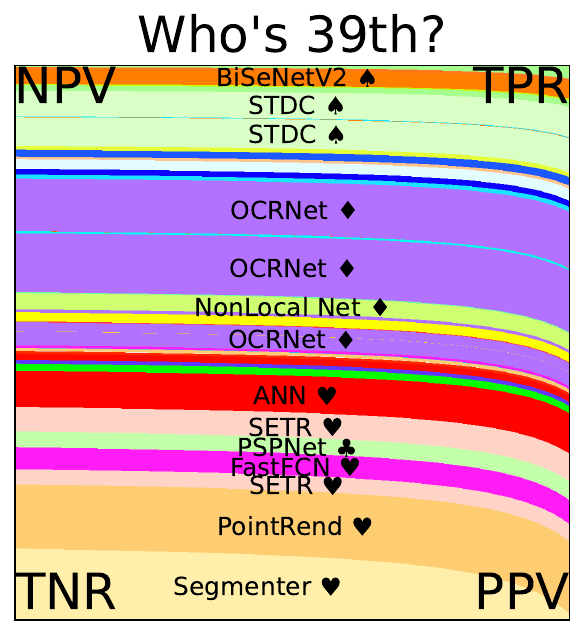}
\includegraphics[scale=0.4]{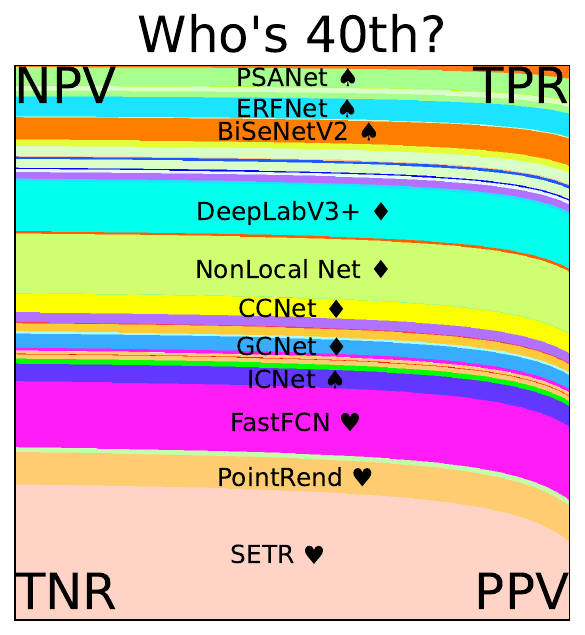}
\includegraphics[scale=0.4]{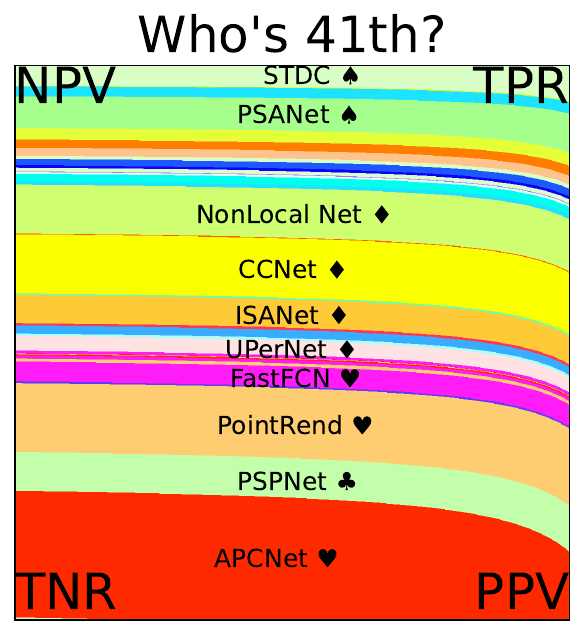}
\includegraphics[scale=0.4]{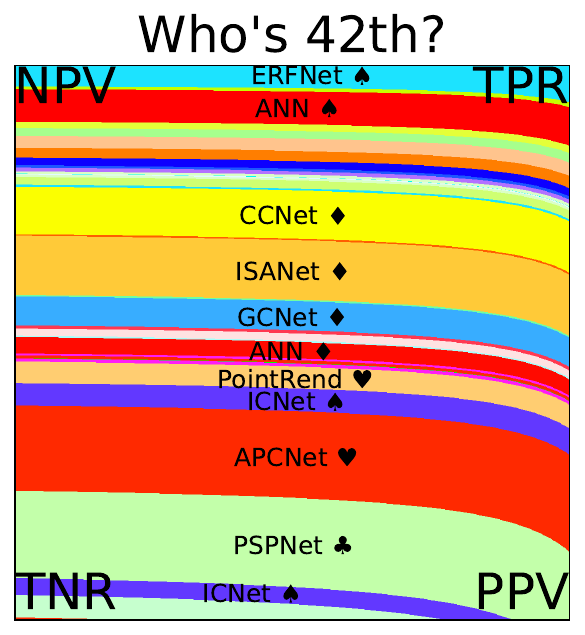}
\includegraphics[scale=0.4]{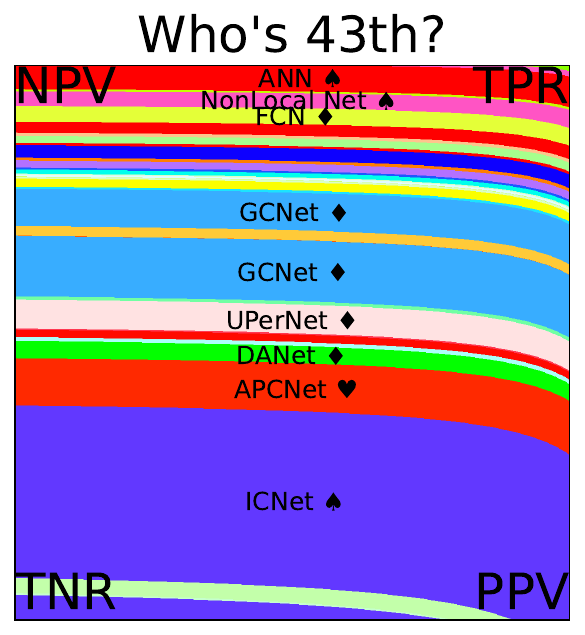}
\includegraphics[scale=0.4]{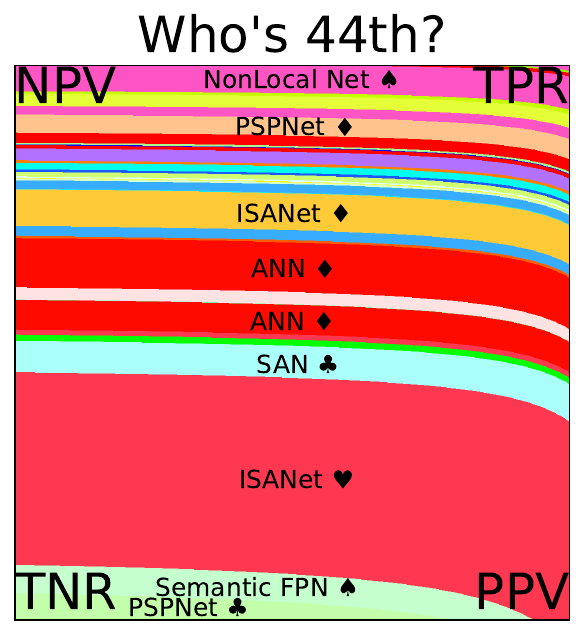}
\includegraphics[scale=0.4]{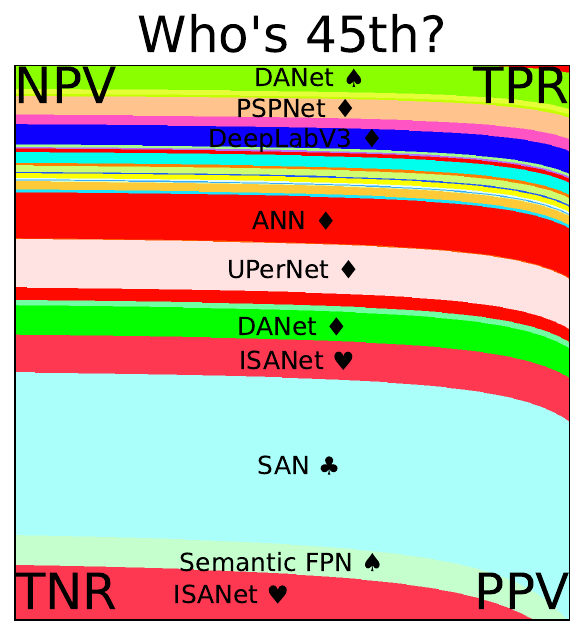}
\includegraphics[scale=0.4]{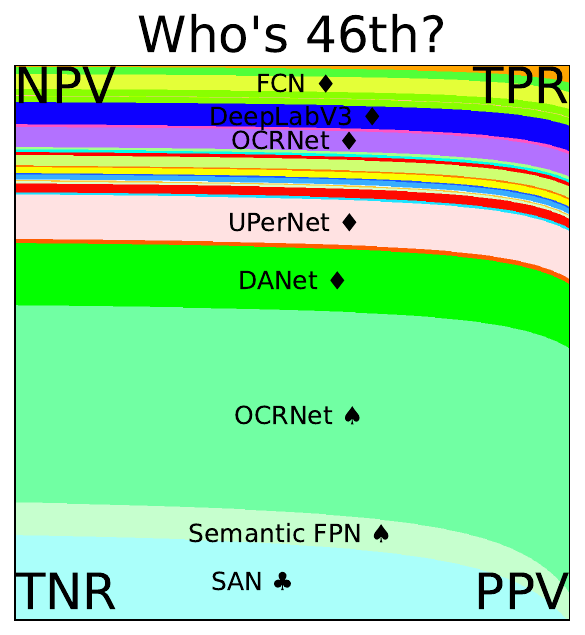}
\includegraphics[scale=0.4]{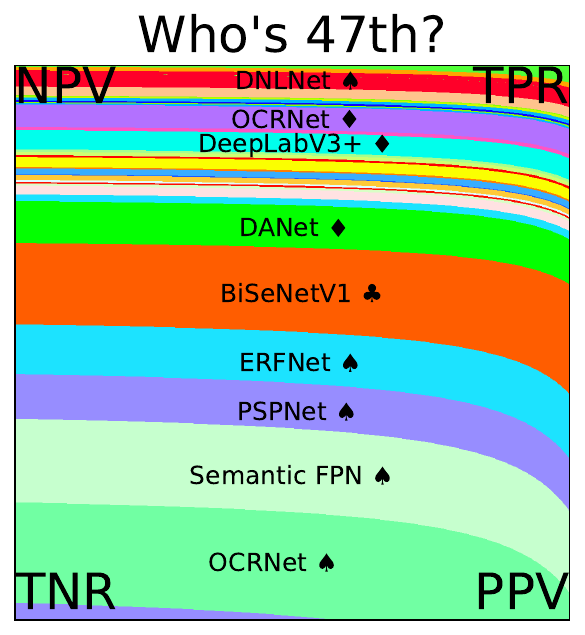}
\includegraphics[scale=0.4]{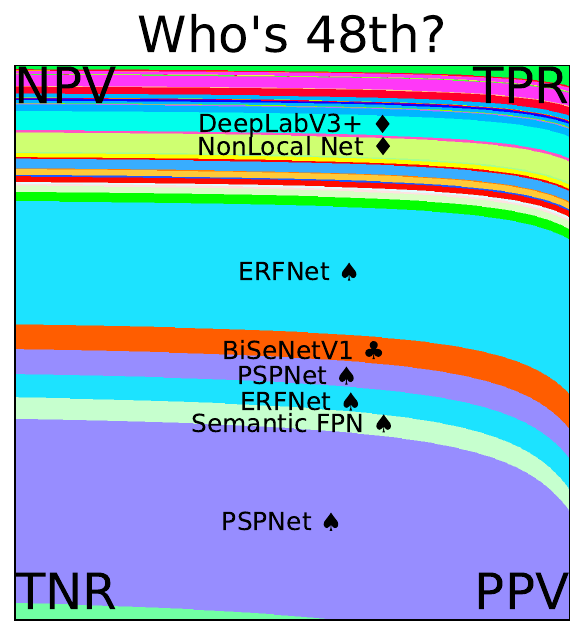}
\includegraphics[scale=0.4]{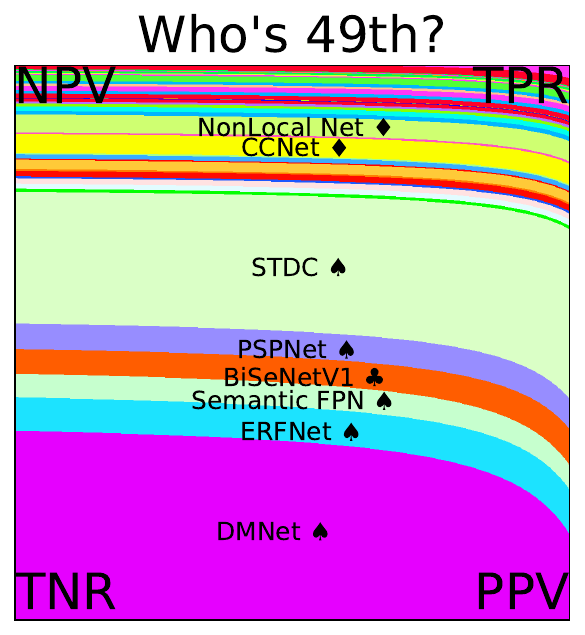}
\includegraphics[scale=0.4]{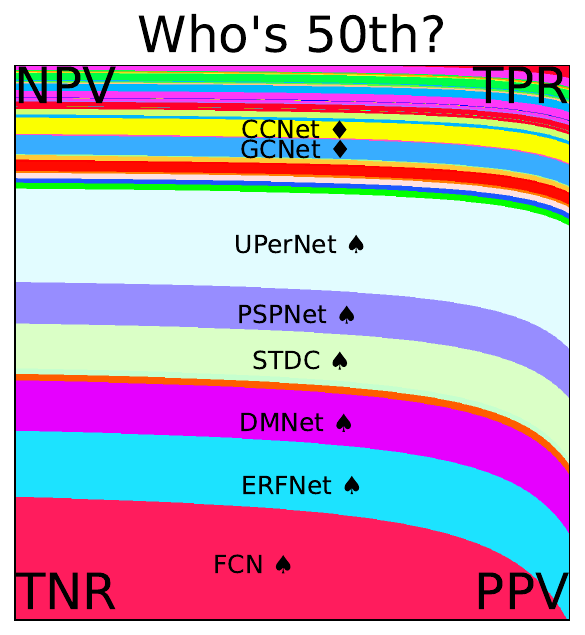}
\includegraphics[scale=0.4]{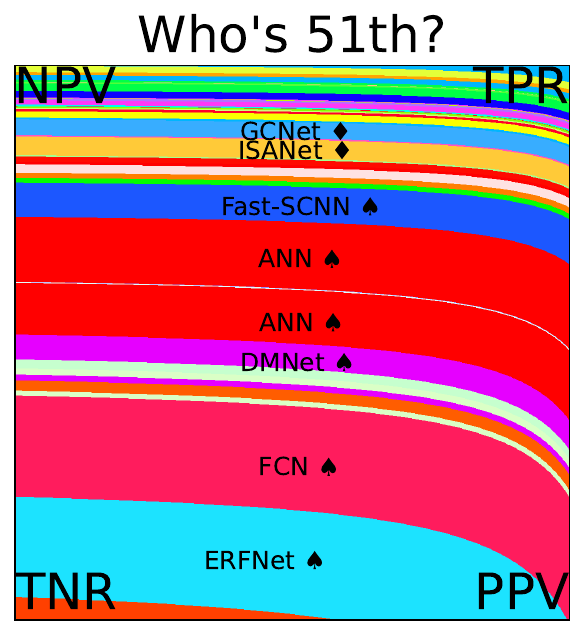}
\includegraphics[scale=0.4]{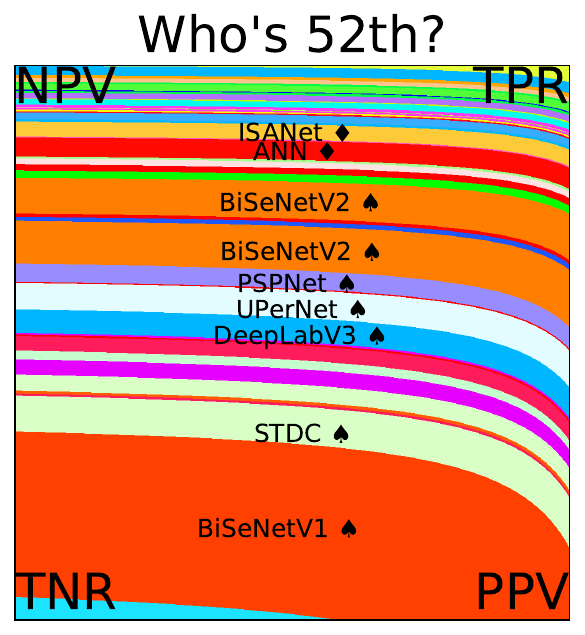}
\includegraphics[scale=0.4]{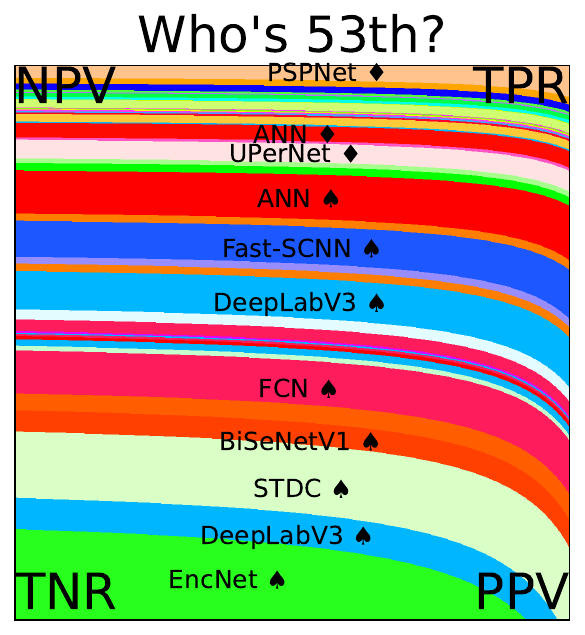}
\includegraphics[scale=0.4]{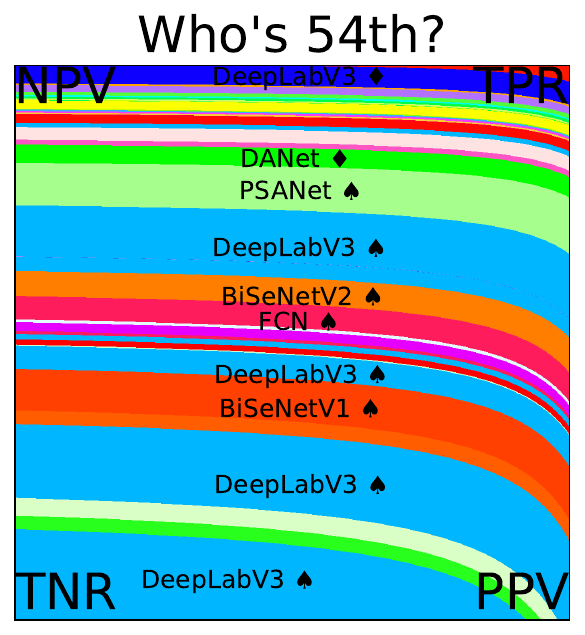}
\includegraphics[scale=0.4]{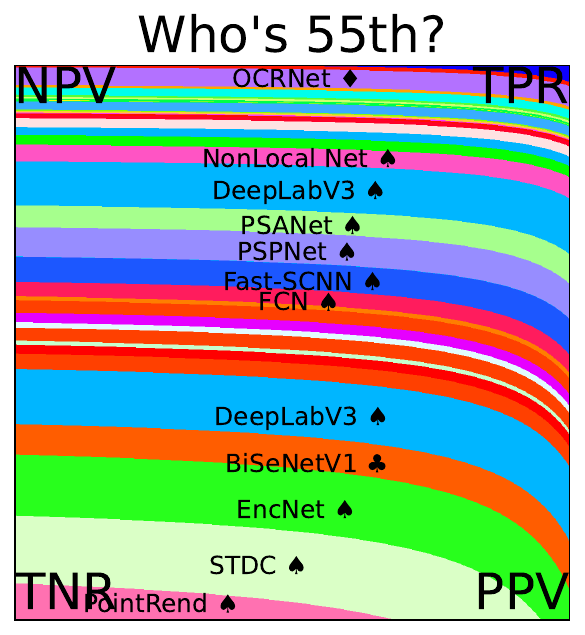}
\includegraphics[scale=0.4]{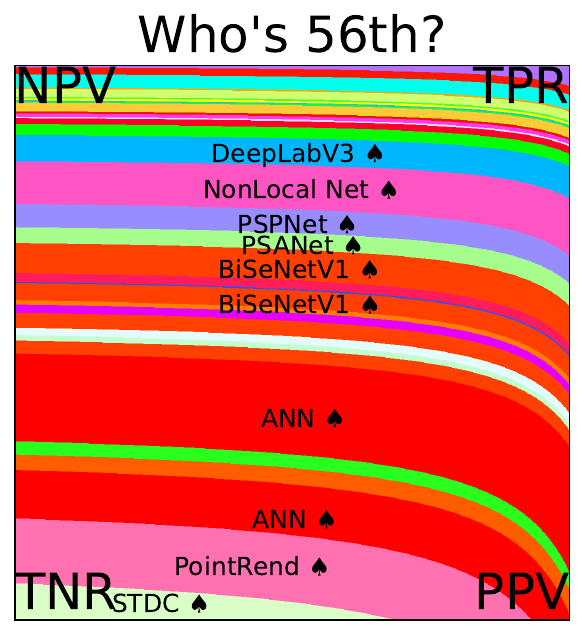}
\includegraphics[scale=0.4]{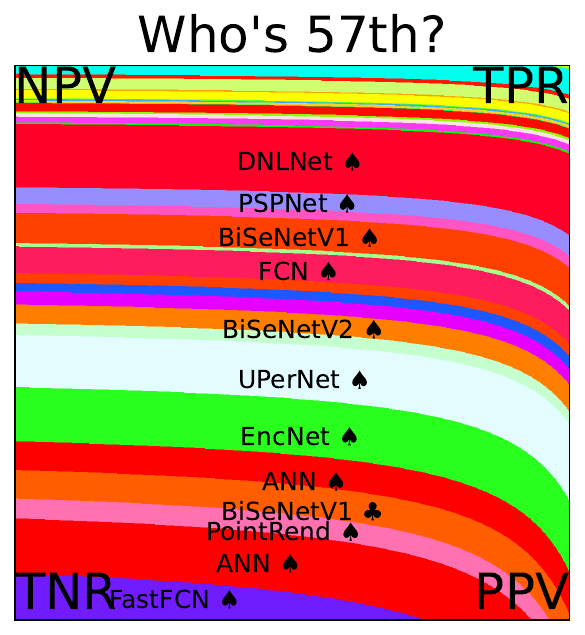}
\includegraphics[scale=0.4]{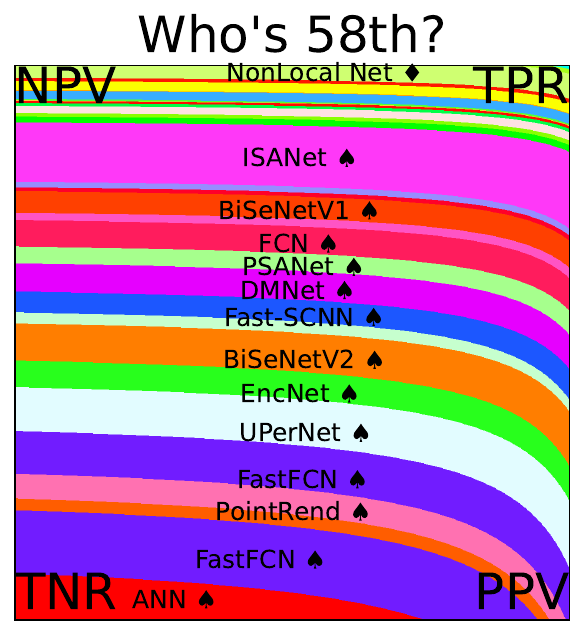}
\includegraphics[scale=0.4]{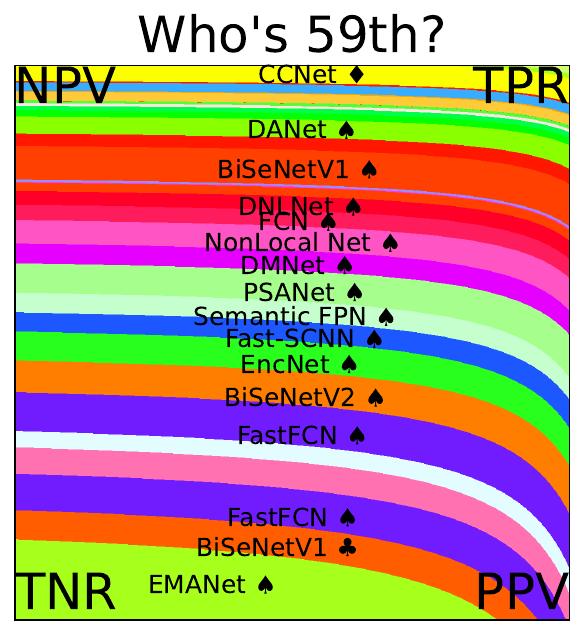}
\includegraphics[scale=0.4]{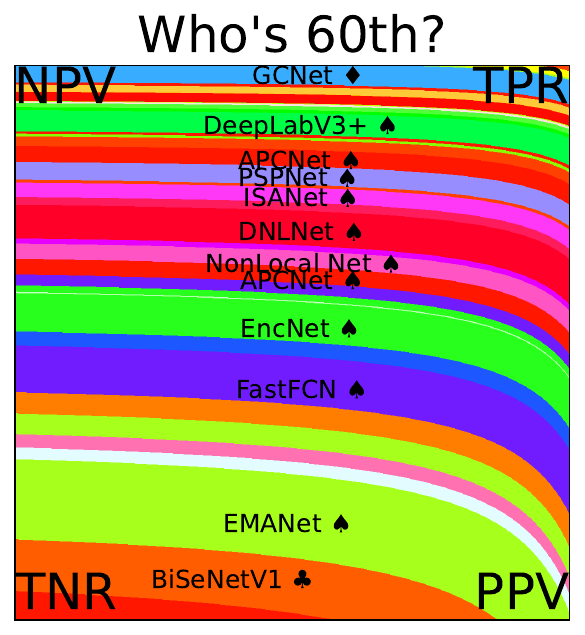}
\includegraphics[scale=0.4]{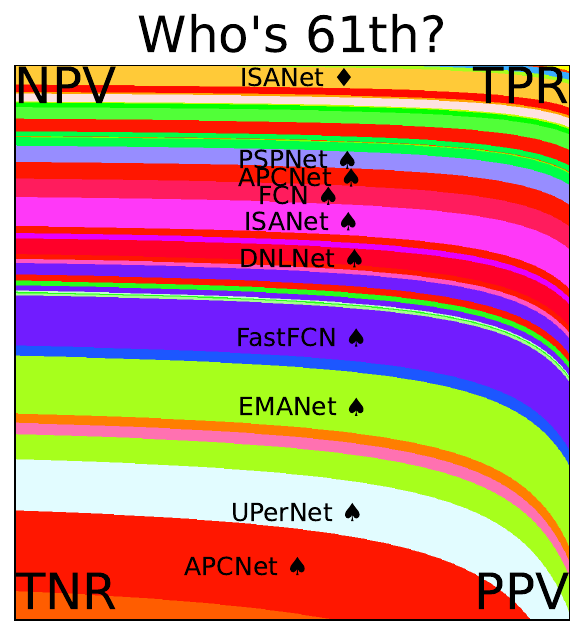}
\includegraphics[scale=0.4]{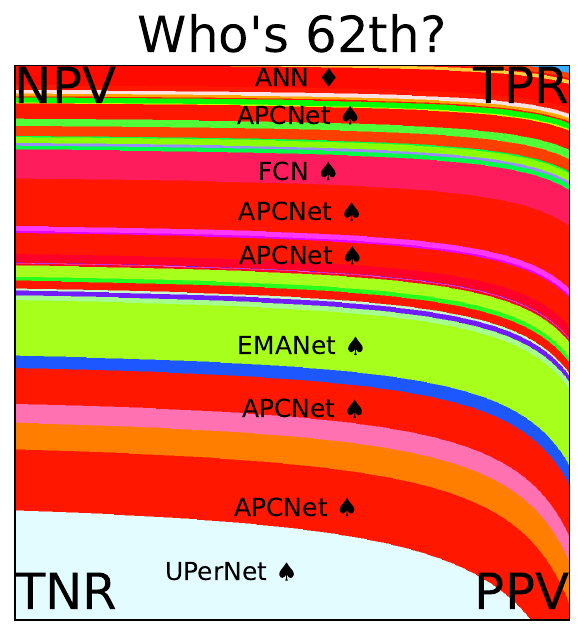}
\includegraphics[scale=0.4]{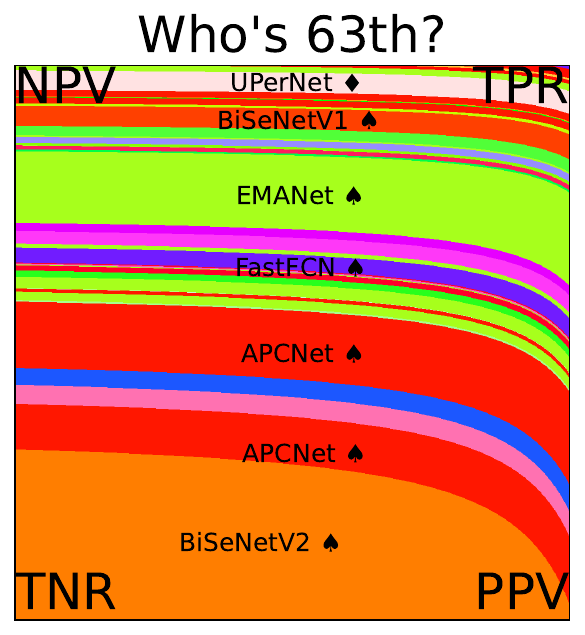}
\includegraphics[scale=0.4]{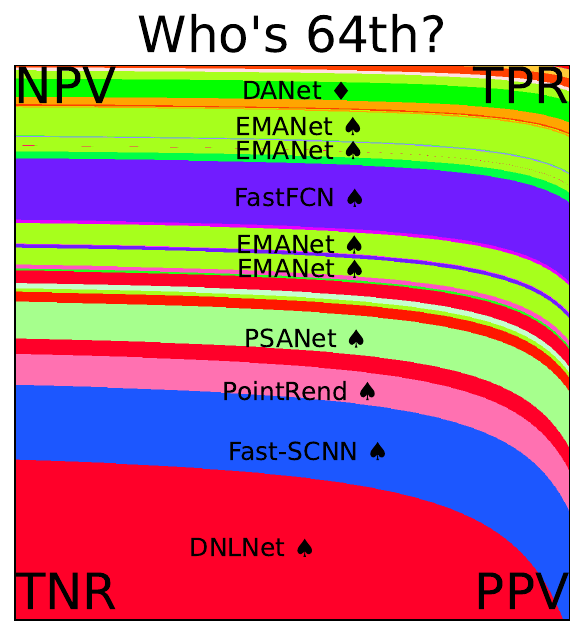}
\includegraphics[scale=0.4]{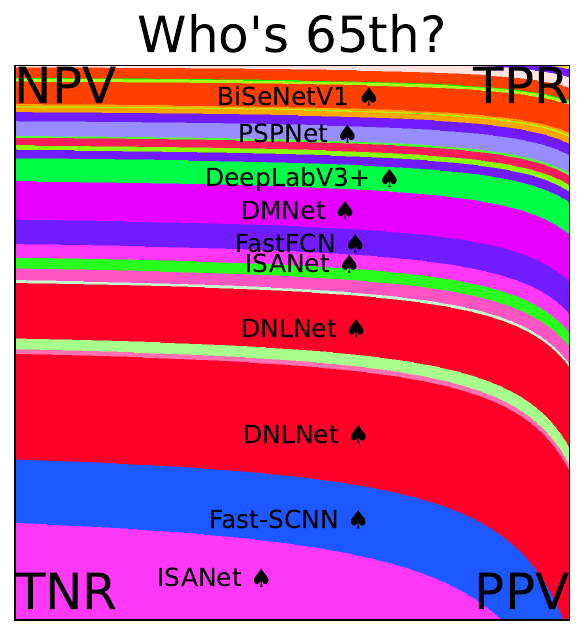}
\includegraphics[scale=0.4]{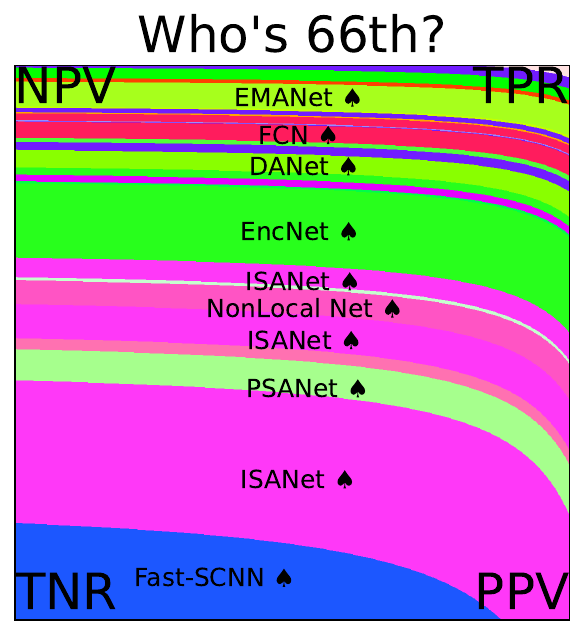}
\includegraphics[scale=0.4]{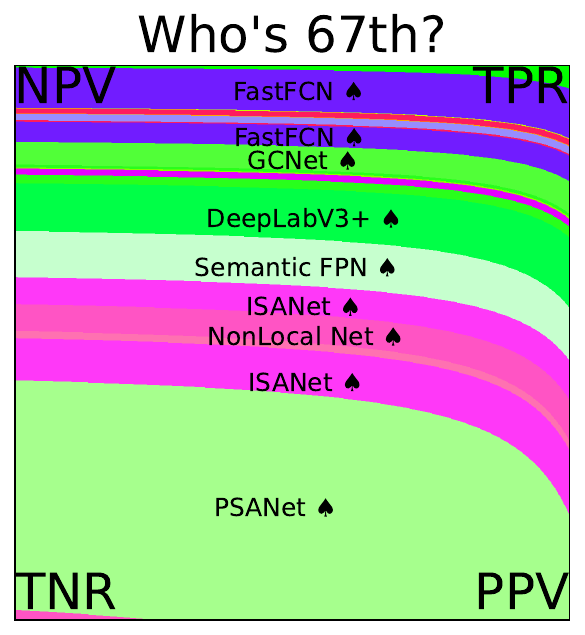}
\includegraphics[scale=0.4]{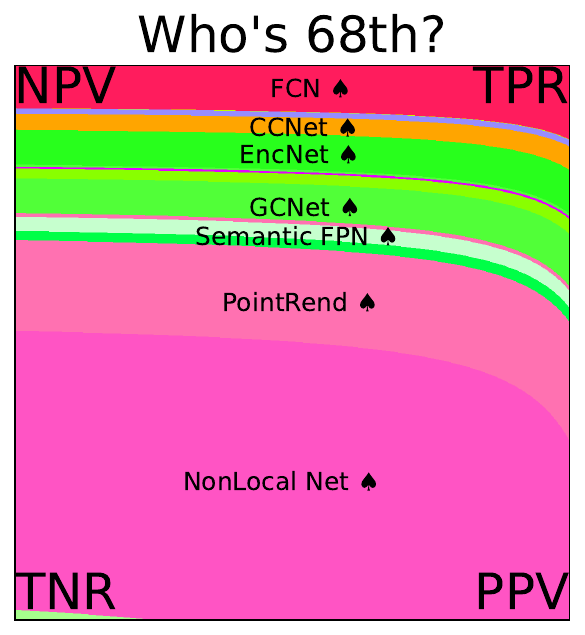}
\includegraphics[scale=0.4]{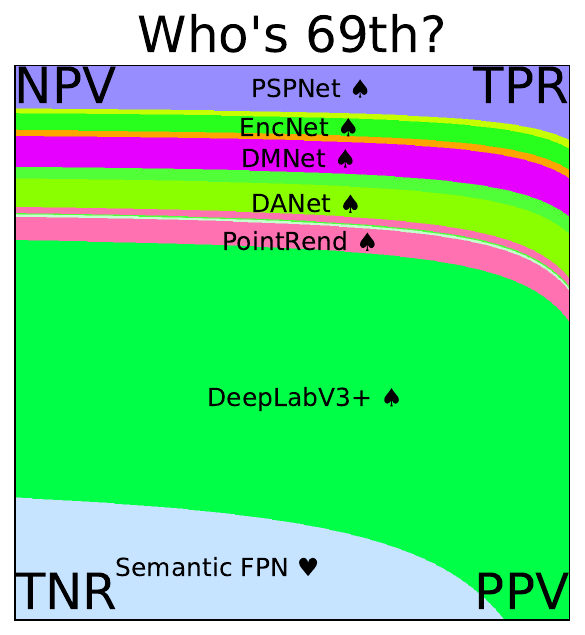}
\includegraphics[scale=0.4]{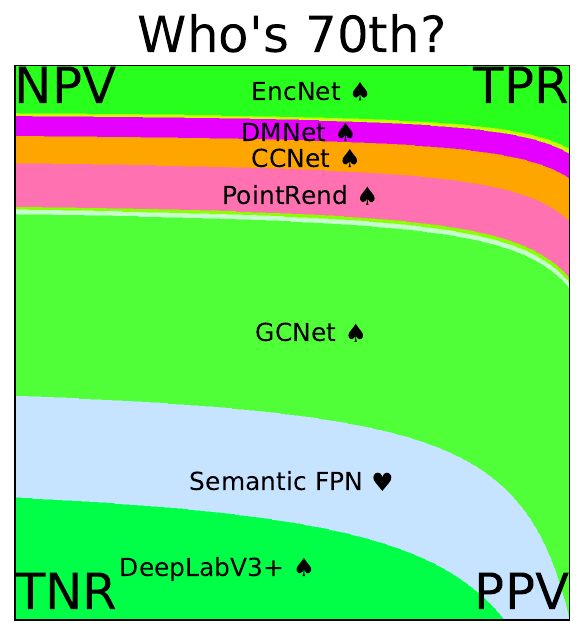}
\includegraphics[scale=0.4]{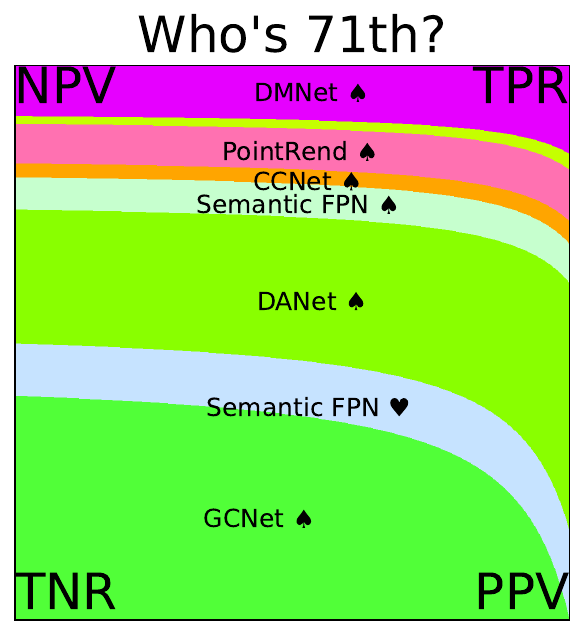}
\includegraphics[scale=0.4]{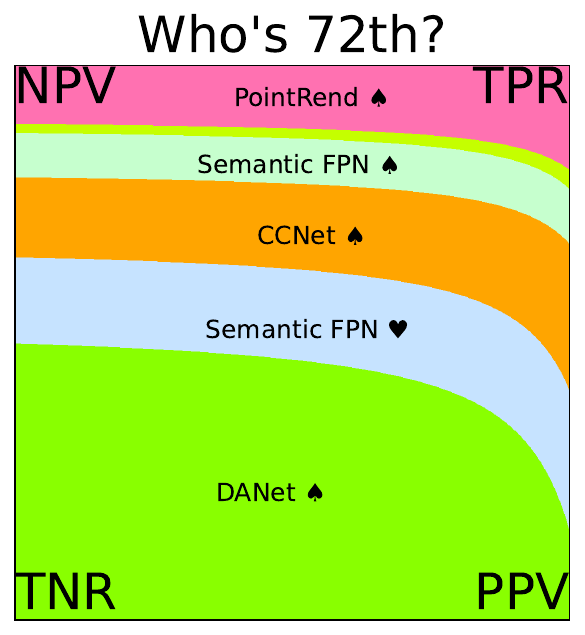}
\includegraphics[scale=0.4]{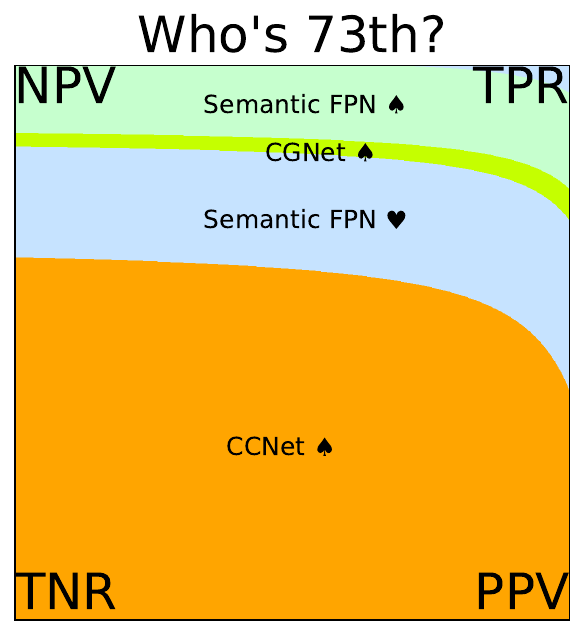}
\includegraphics[scale=0.4]{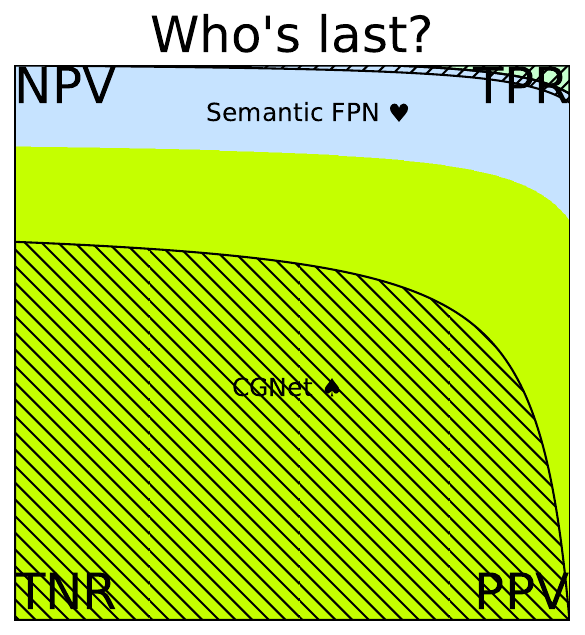}
\par\end{center}
\paragraph{Analysis.}

The following entities are ranked first:
\begin{itemize}
    \item DeepLabV3+ \ding{169} (2.66\% of the tile)
    \item ISANet \ding{169} (20.52\% of the tile)
    \item Mask2Former \ding{171} (29.85\% of the tile)
    \item SETR \ding{171} (46.97\% of the tile)
\end{itemize}

\subsection*{Correlation Tile: Using the tile to show the rank and linear correlations with the mean-IoU}
\begin{center}
\includegraphics[scale=0.35]{images/results/tile_correlation_macro_iou_pearsonr.pdf}
\includegraphics[scale=0.35]{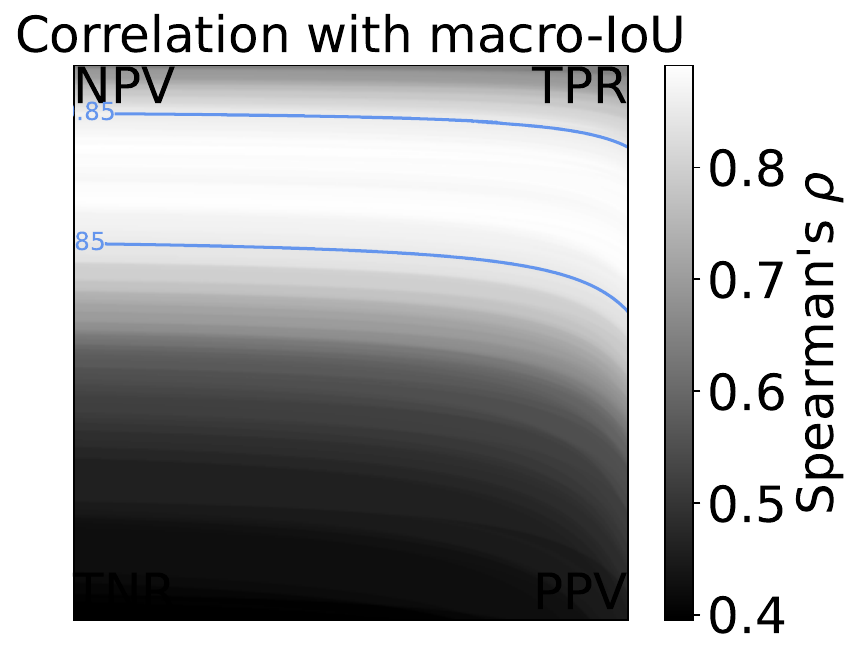}
\includegraphics[scale=0.35]{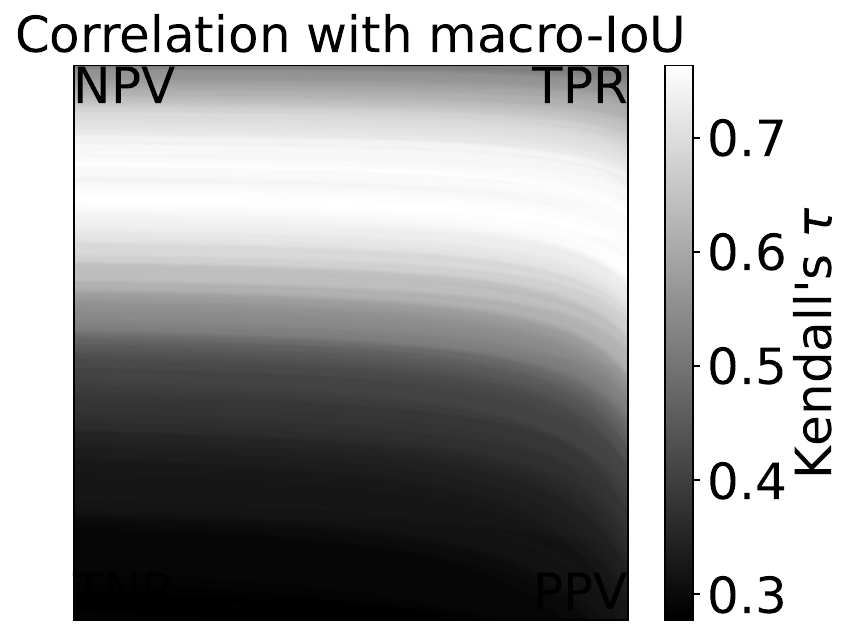}
\par\end{center}
\paragraph{Analysis for the linear correlation with Pearson's $r$.}
\begin{itemize}
    \item In 93.9\% of the zone where $r\ensuremath{\geq}0.85$ in the tile, the best is Mask2Former \ding{171}
    \item In 6.1\% of the zone where $r\ensuremath{\geq}0.85$ in the tile, the best is SETR \ding{171}
\end{itemize}

\paragraph{Analysis for the rank correlation with Spearman's $\rho$.}
\begin{itemize}
    \item In 81.5\% of the zone where $\rho\ensuremath{\geq}0.85$ in the tile, the best is Mask2Former \ding{171}
    \item In 18.5\% of the zone where $\rho\ensuremath{\geq}0.85$ in the tile, the best is SETR \ding{171}
\end{itemize}

\paragraph{Analysis for the rank correlation with Kendall's $\tau$.}
\begin{itemize}
    \item WARNING: There is no zone where $\tau\ensuremath{\geq}0.85$ in the tile! Maybe do you want to change the threshold?
\end{itemize}

\end{document}